\documentclass[9pt]{article} 

\PassOptionsToPackage{numbers,compress}{natbib}
\usepackage[main, final]{neurips_2026}

\usepackage{algorithm}
\usepackage{algpseudocode}
\usepackage[pdftex]{graphicx} 
\usepackage[colorlinks=true, allcolors=blue]{hyperref}     
\usepackage{url}            
\usepackage{wrapfig}
\usepackage{amsmath}
\usepackage{amssymb}
\usepackage{caption} 

\usepackage{makecell}
\usepackage{float}
\usepackage{subcaption}
\usepackage{pifont}        
\usepackage[symbol]{footmisc}
\usepackage{booktabs}
\usepackage[noabbrev,capitalize]{cleveref}
\usepackage{multirow}
\usepackage[inline,shortlabels]{enumitem}
\usepackage{etoolbox}
\usepackage{array}
\usepackage[many]{tcolorbox} 
\usepackage{soul}
\usepackage{appendix}
\usepackage{titletoc}
\usepackage[colorinlistoftodos]{todonotes}

\makeatletter
\patchcmd{\@algocf@start}
  {-1.5em}
  {0pt}
  {}{}
\makeatother

\definecolor{navyblue}{RGB}{0,0,128}

\usepackage{siunitx}
\newcommand{\cmark}{\textcolor{green!70!black}{\ding{51}}}  
\newcommand{\xmark}{\textcolor{red}{\ding{55}}}             

\definecolor{rebuttal_color}{rgb}{0.5, 0.0, 0.5}

\newtcolorbox{boxJ}{
    colback=white,
    colframe=navyblue,
    boxrule=2pt,
    sharp corners,
    fuzzy shadow={0pt}{-2pt}{-0.5pt}{0.5pt}{black!35}
}

\makeatletter
\g@addto@macro{\endtabular}{\rowfont{}}
\makeatother
\newcommand{\rowfonttype}{}
\newcommand{\rowfont}[1]{
\gdef\rowfonttype{#1}#1\ignorespaces%
}
\makeatother

\title{\centering TS-ICL: A Flexible Time-Indexed Foundation Model for Time Series via In-Context Learning}

\usepackage{booktabs}   
\usepackage{tabularx}

\author{\textbf{Etienne Le Naour}\textsuperscript{*, 1} \quad \textbf{Tahar Nabil}\textsuperscript{*, 1} \quad \textbf{Adrien Petralia}\textsuperscript{1} \\
\textsuperscript{*}{Equal contribution}\\
\textsuperscript{1}{EDF R\&D, Palaiseau, France}\\
\vspace{1mm} \\
\texttt{etienne.le-naour@edf.fr} \\
\texttt{tahar.nabil@edf.fr} \\
\texttt{adrien.petralia@gmail.com} \\
}

\begin{document}

\maketitle

\vspace{-0.5cm}

\begin{abstract}

Foundation models mark a profound paradigm shift in time series modeling, with task-specific models being superseded by general-purpose zero-shot models. 
Yet, current approaches primarily focus on forecasting, while real-world time series are often irregularly and partially observed, requiring models that can jointly forecast, impute missing values, and handle degraded sampling conditions. 
To address these challenges, we introduce \texttt{TS-ICL}, a novel probabilistic In-Context Learning encoder--regressor Transformer that unifies forecasting and imputation.
\texttt{TS-ICL} formulates time series tasks as timestamp-aligned regression and naturally incorporates covariates by training on synthetic dependency structures generated from a novel causal data prior.
Empirically, \texttt{TS-ICL} achieves a new state-of-the-art in imputation, while remaining competitive with leading forecasting foundation models across both univariate and covariate-aware benchmarks. It shows particularly strong performance in forecasting with partially observed look-back windows. \textit{GitHub page}: \href{https://github.com/EDF-Lab/ts-icl}{https://github.com/EDF-Lab/ts-icl}.
\end{abstract}
\section{Introduction}


The recent advent of Time Series Foundation Models (TSFMs) drastically changes time series modeling, shifting from task-specific to transferable models that leverage large-scale pretraining and adaptation mechanisms to provide zero-shot inference on unseen data~\cite{auer2025tirex,liu2025moirai,TOTO2025,liu2025sundial}.
TSFMs address limited data regimes and distribution shifts~\citep{wu2026out}, but still face fundamental limitations.
First, many tasks cannot be solved in practice without leveraging external information, requiring covariate-aware inference \citep{taylor2018forecasting}.
Second, real-world observations are often incomplete, with missing values and asynchronous measurements~\citep{clark2004population}.
This challenges standard modeling assumptions and motivates unified frameworks that can jointly handle forecasting and imputation in flexible observation settings.

To address these challenges, (1) recent TSFMs such as \texttt{Chronos-2} \citep{ansari2024chronos2} build on In-Context Learning~\citep{brown2020language} (ICL) to support covariate-informed inference and handle missing values, while remaining highly efficient.
Nevertheless, they do not natively address imputation, thus hindering their practical use.
(2) In a remarkably different approach, \texttt{TabPFN} \citep{hollmann2022tabpfn} and \texttt{TabICLv2} \citep{qu2026tabiclv2} have revolutionized the tabular domain with strong few-shot regression capabilities via Transformer-based ICL and synthetic data priors.
When adapted to time series (e.g., \texttt{TabPFN-TS} \citep{hoo2024tables2time}), these Tabular Foundation Models (TFMs) naturally support covariates and enable both zero-shot imputation \citep{TSFMImputationBenchmark} and forecasting \citep{shchur2025fev}.
Yet, they lack temporal inductive bias and rely on handcrafted time features.
As a result, TFMs lag behind pure TSFMs in forecasting benchmarks, while also incurring high inference cost~\citep{shchur2025fev}.

\begin{table}[h!]
\caption{Capabilities of recent time series foundation models.
Only \texttt{TS-ICL} supports forecasting and imputation, while enabling efficient covariate-aware inference and supporting irregular sampling.}
\begin{center}
\vspace{1ex}
{\centering
\setlength{\tabcolsep}{6pt} 
\scalebox{0.67}{
\begin{tabular}{@{}lccccccc@{}}
\toprule
\multirow{2}{*}{\textbf{Method}} & \textbf{Handles} & \textbf{Handles} & \textbf{Covariate} & \textbf{Probabilistic}  & \textbf{Designed for} & \textbf{Irregular} & \textbf{Fast} \\
  & \textbf{Forecasting} & \textbf{Imputation} & \textbf{Integration} & \textbf{Predictions} & \textbf{Time Series} & \textbf{Sampling}  & \textbf{Inference} \\
\midrule
\texttt{TiRex} \citep{auer2025tirex}, \texttt{Toto} \citep{TOTO2025}, \texttt{TimesFMv2.5} \citep{TimesFM}  & \cmark & \xmark & \xmark & \cmark & \cmark & \xmark & \cmark\\
 \texttt{Chronos-2} \citep{ansari2024chronos2} & \cmark & \xmark & \cmark & \cmark & \cmark &  \xmark & \cmark\\
\texttt{TabPFNv2.5-TS} \citep{hoo2024tables2time}, \texttt{TabICLv2-TS} \citep{qu2026tabiclv2} & \cmark & \cmark & \cmark & \cmark  & \xmark &  \cmark &  \xmark \\
\textbf{\texttt{TS-ICL} (ours)}             & \cmark & \cmark & \cmark & \cmark & \cmark & \cmark & \cmark \\
\bottomrule
\end{tabular}}}
\end{center}
\label{tab:comparison}
\end{table}

Consequently, existing approaches fail to jointly provide
\begin{enumerate*}[(i)]
\item unified forecasting and imputation,
\item covariate-aware inference, and
\item efficient zero-shot performance
\end{enumerate*} (see \cref{tab:comparison}).
In this paper, we tackle this challenge and introduce \texttt{TS-ICL}, a unified probabilistic Transformer foundation model for imputation and forecasting.
\begin{enumerate*}[(i)]
\item \texttt{TS-ICL} casts time series modeling as an in-context regression problem, where observations are represented as timestamp-aligned inputs and encoded into contextual representations that enable forecasting or imputation in a single forward pass.
\item To enable effective covariate-aware inference, a structured synthetic prior over target--covariate relationships is introduced using Directed Acyclic Graphs (DAGs) to define dependency structure, with node-level mechanisms inspired by structural causal models \cite{peters2017elements,hollmann2022tabpfn}.
\item Unlike existing TSFMs, \texttt{TS-ICL} operates directly on timestamped observations rather than fixed grids, allowing flexible handling of missing or irregularly sampled data in practice.
\end{enumerate*}

\paragraph{Contributions.} The main contributions are as follows:
\begin{itemize}[leftmargin=*, itemsep=0.5pt, topsep=0.5pt]
    \item \textbf{A unified and flexible TSFM architecture.} We introduce \texttt{TS-ICL}, a novel probabilistic TSFM that casts time series modeling as a time-indexed in-context regression problem, unifying forecasting and imputation with native support for covariates.
    \item \textbf{Structured synthetic prior.} We design a novel DAG-based causal prior over target--covariate time series, enabling robust zero-shot generalization to unseen dependency structures.
    \item \textbf{State-of-the-art imputation performance.} \texttt{TS-ICL} sets a new state-of-the-art on zero-shot imputation benchmarks, consistently outperforming both task-specific models and TFMs, while being up to $50\times$ faster than TFMs at inference.
    \item \textbf{Competitive forecasting performance.} On forecasting benchmarks, \texttt{TS-ICL} matches state-of-the-art TSFMs while supporting covariate-aware inference, and remains particularly robust to missing observations due to its time-indexed formulation.
\end{itemize}

\section{Related Work}
\label{sec:related-content}

\paragraph{Time series foundation models.} Time Series Foundation Models (TSFMs) are pretrained general-purpose models for time series, typically trained on large mixtures of real and synthetic data and often based on patch-based architectures~\citep{liu2025moirai, auer2025tirex, TOTO2025, TimesFM}. 
While they achieve strong zero-shot forecasting performance on standard benchmarks~\citep{aksu2024gift, qiao2026sTIME}, they exhibit several limitations in practical settings: they are typically not designed for covariate-aware inference, and primarily focus on forecasting without addressing imputation. 
\texttt{Chronos-2}~\citep{ansari2024chronos2} partially mitigates the covariate limitation by enabling inference-time conditioning on exogenous time series, leading to improved performance when informative covariates are available. 
However, its patch-based formulation assumes regularly sampled inputs and does not natively support imputation, which is critical in many real-world time series applications~\cite{schulz1997spectrum,clark2004population,cao2018brits, du2023saits}.

\paragraph{Tabular foundation models for time series.} Tabular Foundation Models (TFMs) such as \texttt{TabPFN} \citep{hollmann2022tabpfn} and \texttt{TabICLv2} \citep{qu2026tabiclv2} leverage in-context learning over synthetic task distributions to achieve strong few-shot regression performance. 
Extensions to time series \citep{hoo2024tables2time} demonstrate competitive results for both forecasting \citep{shchur2025fev} and imputation~\citep{TSFMImputationBenchmark}, while naturally supporting covariates at inference time. 
However, these approaches typically rely on handcrafted temporal features, such as Fourier features operating at predefined frequencies,
and incur higher inference costs compared to dedicated TSFMs~\citep{shchur2025fev}, limiting their scalability in practice.

\paragraph{Supervised imputation models.} Classical supervised imputation methods such as \texttt{BRITS}~\citep{cao2018brits} and \texttt{SAITS}~\citep{du2023saits} achieve strong performance in in-domain settings but require task-specific training and generalize poorly across domains. 
Recent large-scale evaluations~\citep{TSFMImputationBenchmark} suggest that TFM-based approaches can significantly outperform these methods in realistic scenarios.

\paragraph{Time-indexed and Neural Field models.} Continuous-time modeling of time series (also referred to as time-indexed modeling~\citep{DeepTime}) has been explored through Neural Ordinary Differential Equations~\citep{chen2018neural} (ODE) and latent ODE frameworks~\citep{latent_ode2019}. 
More recently, Neural Field-based representations~\citep{naour2023time, serrano2024aroma} encode irregular observations into continuous latent functions without requiring explicit temporal alignment. 
These approaches provide a principled foundation for flexible time representations, but are not designed for zero-shot inference in forecasting or imputation settings.

Despite these advances, existing approaches remain fragmented across forecasting, imputation, and representation learning. 
No framework jointly provides efficient zero-shot inference, covariate-aware modeling, and a unified treatment of forecasting and imputation within a single architecture.
\section{\texttt{TS-ICL} Architecture}
\label{sec:model-univar}


This section introduces the proposed \texttt{TS-ICL} architecture, designed for efficient probabilistic zero-shot time series forecasting and imputation, while maintaining flexibility in handling covariates and irregularly sampled observations. A high-level overview of the architecture and its main components is provided, while detailed descriptions of each module are deferred to Appendix \ref{appendix-archi}.

\paragraph{Problem setting.}
Let $\boldsymbol{x} = (x_t)_{t \in \mathcal{T}}$ denote a time series defined over a (possibly irregular) set of timestamps $\mathcal{T}$. The index set $\mathcal{T}$ is partitioned into two disjoint subsets: (i) the context timestamps $\mathcal{T}^{\mathrm{ctxt}}$, corresponding to observed values, and (ii) the target timestamps $\mathcal{T}^{\mathrm{tgt}}$, corresponding to values to be predicted. Accordingly, $\boldsymbol{x}^{\mathrm{ctxt}} = (x_t)_{t \in \mathcal{T}^{\mathrm{ctxt}}}$ denotes the observed context values, and $\boldsymbol{x}^{\mathrm{tgt}} = (x_t)_{t \in \mathcal{T}^{\mathrm{tgt}}}$ the target values.


Additionally, an optional set of exogenous covariate time series is considered:
\[
\boldsymbol{X}^{\mathrm{covar}} = \left\{ \boldsymbol{x}^{\mathrm{covar}}_{c} = (x^{\mathrm{covar}}_{c,t})_{t \in \mathcal{T}^{\mathrm{covar}}_{c}} \right\}_{c=1}^{C-1},
\]

where $C$ refers to the channel dimension, with one channel corresponding to the time series of interest and the remaining $C-1$ channels corresponding to exogenous covariates.

Each covariate is defined over its own set of timestamps $\mathcal{T}^{\mathrm{covar}}_{c} \subseteq \mathcal{T}$. Covariates may be observed over arbitrary subsets of timestamps, including only the context (e.g. the look-back window in forecasting) or both context and target (e.g. look-back and horizon windows), and may be sparsely observed. The \texttt{TS-ICL} framework can handle such heterogeneous availability without requiring any imputation preprocessing.

The objective, common to both forecasting and imputation, is to infer the target values conditioned on the observed context and, when available, the covariates:
\[
p\!\left(\boldsymbol{x}^{\mathrm{tgt}} \mid \boldsymbol{x}^{\mathrm{ctxt}}, \boldsymbol{X}^{\mathrm{covar}}\right).
\]
This formulation unifies forecasting and imputation as conditional inference problems. For clarity, the next section omits batch and timestamp indices whenever no ambiguity arises.

\subsection{\texttt{TS-ICL} Overview}

\begin{figure}[t!]
    \centering
    \includegraphics[width=1.00\linewidth]{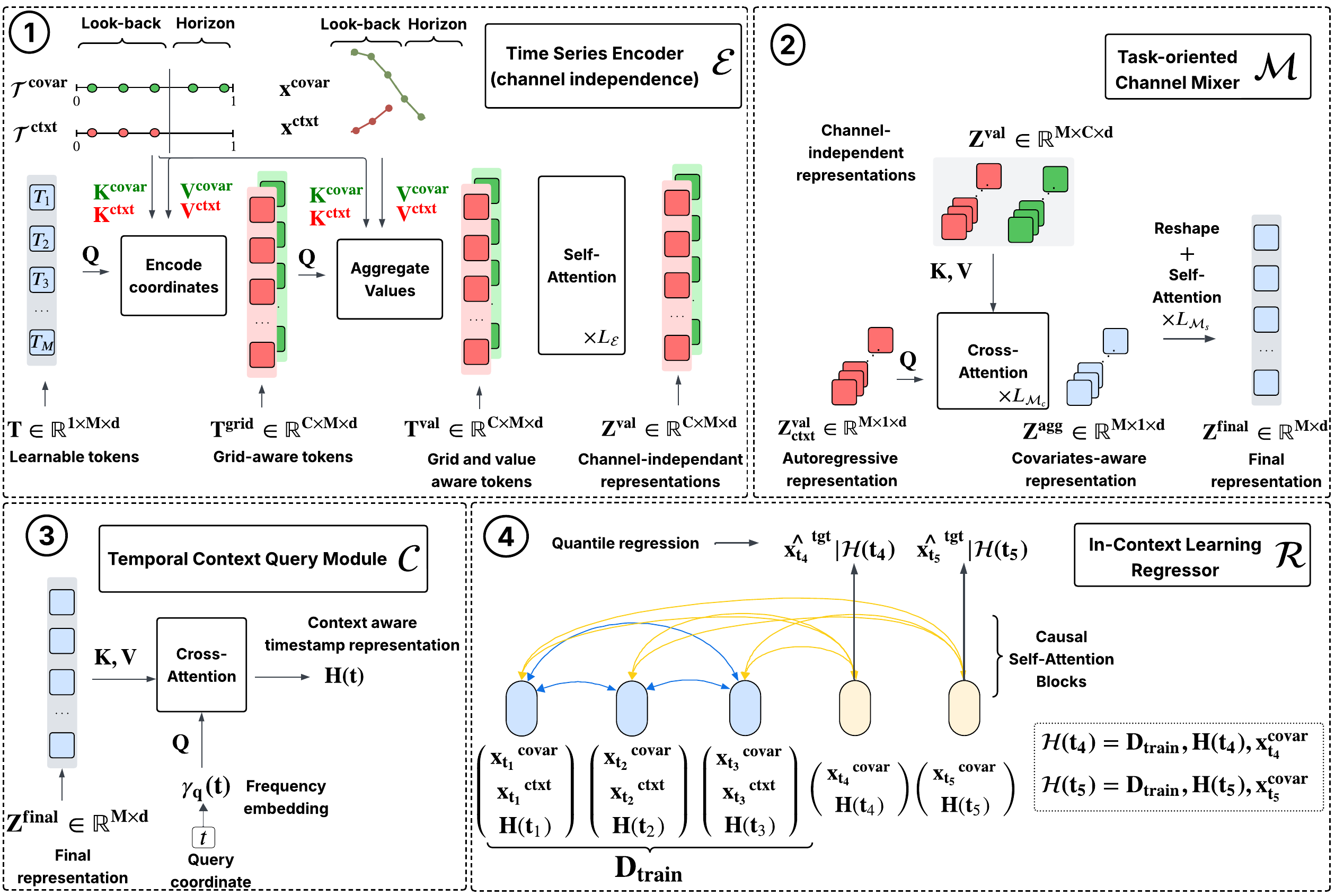}
    \caption{The \texttt{TS-ICL} pipeline. From temporal encoding to in-context regression. The diagram illustrates the four-module transformation for forecasting with one covariate observed on the horizon.}
    \label{fig:TS-ICL-archi}
\end{figure}

The key idea behind \texttt{TS-ICL} is to reformulate time series prediction as an in-context regression problem over learned temporal representations. Thus, the architecture is composed of four successive modules that transform raw observations into global and local context-aware representations used for prediction. The pipeline first encodes each time series, then aggregates information across covariates, and finally produces timestamp-specific representations that enable in-context regression. The overall architecture is illustrated in \cref{fig:TS-ICL-archi}.

\paragraph{(i) Time Series Encoder $\mathcal{E}$.}
This module extracts representations from the context time series and optional $C-1$ covariates in a channel-independent manner. It follows a Perceiver encoder design \citep{jaegle2021perceiver, serrano2024aroma} based on a fixed set of $M$ learnable tokens that sequentially cross-attend to timestamp–value pairs. This allows compressing an arbitrary-length input into a compact latent sequence.
\begin{equation*}
(\boldsymbol{x}^{\mathrm{ctxt}}, \mathcal{T}^{\mathrm{ctxt}}, \boldsymbol{X}^{\mathrm{covar}}, \mathcal{T}^{\mathrm{covar}}) \xrightarrow{\mathcal{E}} \boldsymbol{Z}^{\mathrm{val}} \in \mathbb{R}^{C \times M \times d}.
\end{equation*}

\paragraph{(ii) Channel Mixer $\mathcal{M}$.}
This module aggregates information across the $C$ channels to produce a unified representation. For each of the $M$ latent positions, the token corresponding to the time series of interest queries the tokens of the $C-1$ covariates via cross-attention. This mechanism collapses the channel dimension by selectively integrating exogenous information into the main series' representation. A subsequent stack of self-attention layers models global dependencies among the resulting $M$ tokens, yielding a compact task-oriented representation. This step is critical to capture inter-channel dependencies, which are not modeled by the channel-independent encoder.
\begin{equation*}
\boldsymbol{Z}^{\mathrm{val}} \in \mathbb{R}^{C \times M \times d} \xrightarrow{\mathcal{M}} \boldsymbol{Z}^{\mathrm{final}} \in \mathbb{R}^{M \times d}.
\end{equation*}

\paragraph{(iii) Temporal Context Query Module $\mathcal{C}$.}
This modules aims at bridging the gap between the discrete latent tokens $\boldsymbol{Z}^{\mathrm{final}}$ and the continuous time domain.
Any timestamp $t \in \mathcal{T}$ is first embedded using a frequency-based positional encoding \citep{mildenhall2021nerf}.
This local encoding then cross-attends to the time series representation $\boldsymbol{Z}^{\mathrm{final}}$, providing a single local and global context-aware representation of arbitrary timestamps.
This design enables querying at arbitrary timestamps, supporting irregular sampling and unifying forecasting and imputation.
\begin{equation*}
(t, \boldsymbol{Z}^{\mathrm{final}}) \xrightarrow{\mathcal{C}} \boldsymbol{H}(t) \in \mathbb{R}^{d}.
\end{equation*}


\paragraph{(iv) In-Context Learning Regressor $\mathcal{R}$.} Given the representations $\boldsymbol{H}(t)$, prediction is formulated as an in-context regression problem \cite{garg2022what,vanoswald2023transformers}, where the observed context defines a training set:
\[
\mathcal{D}_{\mathrm{train}} = \{ \boldsymbol{H}(t), \boldsymbol{x}^{\mathrm{ctxt}}_t, \boldsymbol{X}_t^{\mathrm{covar}} \}_{t \in \mathcal{T}^{\mathrm{ctxt}}},
\]
used to infer target values at unseen target timestamps $t^{\mathrm{tgt}}$. The regressor is implemented as a Transformer that performs causal self-attention over the training (context) set, effectively learning to map representations to values in an in-context manner. When available, covariates are incorporated in the training set via cross-attention, allowing the model to condition on exogenous time series under arbitrary availability patterns (e.g., context-only, context and target, or sparse observations). \texttt{TS-ICL} outputs a dense set of quantile estimates of the target distribution, trained using a smoothed pinball loss \citep{steinwart2011estimating}.
\begin{equation*}
\boldsymbol{H}(t^{\mathrm{tgt}}), \boldsymbol{X}^{\mathrm{covar}}_{t^{\mathrm{tgt}}} \xrightarrow{\mathcal{R}} p\!\left(\boldsymbol{x}^{\mathrm{tgt}} \mid \boldsymbol{H}(t^{\mathrm{tgt}}), \boldsymbol{X}_{t^{\mathrm{tgt}}}^{\mathrm{covar}}, \mathcal{D}_{\mathrm{train}}\right).
\end{equation*}

This four-step formulation unifies time series representation learning and in-context regression, enabling \texttt{TS-ICL} to perform forecasting and imputation while flexibly operating over timestamped observations and optionally observed covariates. Additional architectural details and hyperparameters are provided in Appendix~\ref{appendix-archi} and Appendix~\ref{appendix:hyperparameters}, respectively.

\section{Data Prior and Training Procedure}
\label{sec:data-prior}

\texttt{TS-ICL} is trained on a structured data prior combining real-world and synthetic time series, spanning both univariate signals and multivariate covariate--target structures. This prior is designed to jointly capture temporal dynamics and inter-variable dependencies within a unified training distribution. Concretely, training samples consist of either univariate time series or multivariate problems where target--covariate relationships are generated via structured transformations of base signals.

\paragraph{Univariate time series.}
In the univariate setting, a large and highly heterogeneous pretraining prior is constructed by combining real-world and synthetic time series.
This mixture is designed to expose \texttt{TS-ICL} to a broad spectrum of temporal dynamics. For \textit{real-world data}, the prior leverages a collection of 31 datasets spanning multiple domains, as detailed in Appendix~\ref{subsec:univar-real-ts}.
These datasets cover diverse temporal phenomena, including trends, seasonal patterns, regime shifts, and varying levels of non-stationarity. The training distribution is further augmented with \textit{synthetic data} sampled from the \texttt{TempoPFN} univariate time series generator \citep{moroshan2025tempopfn}, which induces controlled but diverse stochastic processes.
Overall, this yields a large-scale pretraining distribution over univariate time series, comprising 40 datasets and approximately 2M time series with lengths ranging from $\sim100$ to $\sim600$k time steps.

\begin{wrapfigure}[23]{R}{0.50\linewidth}
\vspace{-1.2em}
    \centering
    \includegraphics[width=\linewidth]{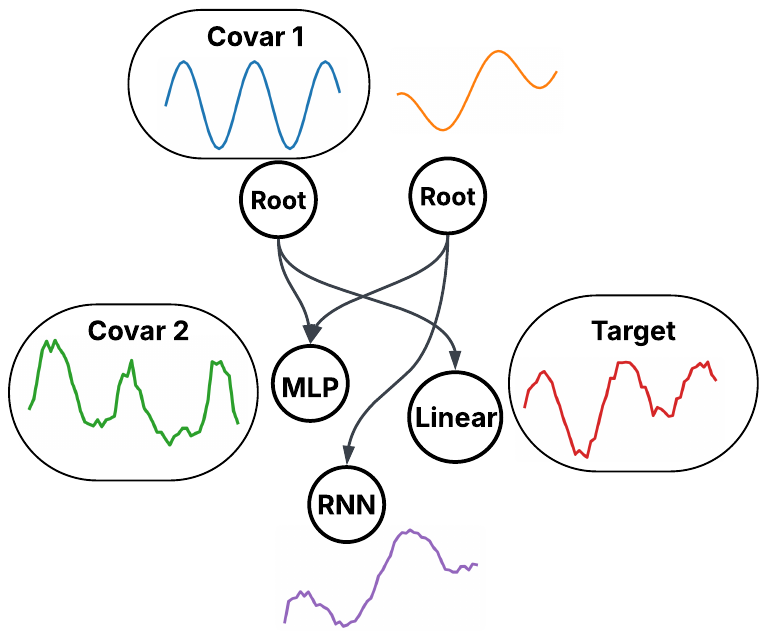}
    \caption{Synthetic target--covariate generation. Multivariate structures are constructed from a sampled DAG, where base signals are transformed via linear and non-linear SCMs. One node is selected as the target, while others serve as informative or redundant covariates.}
    \label{fig:dag-scheme}
\end{wrapfigure}

\paragraph{Covariates generator.}
To enable \texttt{TS-ICL} to learn under structured covariates, synthetic multivariate time series are constructed from base univariate signals (either real or generated as described above). Specifically, \begin{enumerate*}[(i)]
    \item A Directed Acyclic Graph (DAG) is generated over univariate signals where nodes correspond to time series and edges encode causal dependencies. 
    Followingé \cite{qu2026tabiclv2}, each non-root node is produced by applying a transformation sampled from a Structural Causal Model (SCM) registry, including both linear and non-linear operators (e.g., linear mappings, MLP, RNN). 
    This yields heterogeneous dependency structures while preserving temporal coherence.
    \item Given the resulting graph, one node is selected as the prediction target and a subset of the remaining nodes is sampled as covariates, producing multivariate problems with varying numbers of covariates and controllable dependency strength.
\end{enumerate*} This design ensures that covariates can be causally informative, redundant, or entirely independent of the target, preventing reliance on spurious correlations. \cref{fig:dag-scheme} illustrates the synthetic target--covariate generation process. Additional details on the data prior are provided in Appendix \ref{subsec:dag-explanations}.

\paragraph{Whole procedure.} Overall, each training sample is constructed by first sampling base time series (real or synthetic), which may either be used directly or serve as building blocks for multivariate structures via the DAG-based generator. This yields a unified training distribution over univariate and multivariate time series. The training task is then defined dynamically. $\bullet$ For imputation, observations are masked either point-wise or in contiguous segments with randomly sampled masking ratios. $\bullet$ For forecasting, a future horizon of random length is masked. 
The full data generation pipeline is summarized in \cref{alg:data_prior} in Appendix \ref{subsec:whole-training-procedure}, while hyperparameters and training details are reported in Appendix \ref{appendix:hyperparameters}.
\section{Experiments} 
\label{sec:experiements}

\texttt{TS-ICL} is evaluated on zero-shot imputation (\cref{sec:imputation-expe}) and forecasting (\cref{sec:forecasting-expe}) under two settings: \begin{enumerate*}[(i)] \item \textit{univariate time series} and \item \textit{time series with known covariates} available at inference time. \end{enumerate*} 
All experiments use a 27M-parameter version of \texttt{TS-ICL} (see Appendix \ref{appendix:hyperparameters} for architectural details). 
Imputation experiments rely on \texttt{fm-impute-bench}~\citep{TSFMImputationBenchmark}, while forecasting is evaluated on \texttt{fev-bench}~\citep{shchur2025fev}, following recent protocols for time series foundation models. 
Ablation studies on \texttt{TS-ICL} components are provided in Appendix \ref{appendix:ablations} and additional results on the \texttt{TIME} benchmark~\citep{qiao2026sTIME} are presented in Appendices \ref{extended-imputation-expes} and \ref{extended-forecasting-expes}.
\subsection{Imputation Experiments}
\label{sec:imputation-expe}

The zero-shot imputation capability of \texttt{TS-ICL} is evaluated on \texttt{fm-impute-bench}~\citep{TSFMImputationBenchmark}, covering both \textit{univariate} settings and scenarios with \textit{covariates} available at inference time. 
The benchmark spans diverse missingness patterns, sequence lengths, and application domains. 

\paragraph{Setting.} \begin{enumerate*}[(i)] 
\item In the \textit{univariate setting}, \texttt{fm-impute-bench} comprises 33 datasets across multiple domains (e.g., energy, climate, etc.) with varying lengths and frequencies (see \cref{tab:tmlr-bench-datasets}, Appendix \ref{sec:tmlr-benchmark-imputation}). Each sample corresponds to a four-week window. \texttt{TS-ICL} is evaluated under four masking scenarios, namely: 
$\bullet$ 50\% or $\bullet$ 70\% pointwise missingness, and $\bullet$ two or $\bullet$ four disjoint one-day gaps. This results in 132 tasks and about 1.3M windows to impute.
\item The same four masking scenarios are applied to the \textit{known-covariates} setting on six datasets providing informative covariates (see \cref{tab:dataset-covar-tmlr-bench}, Appendix \ref{sec:tmlr-benchmark-imputation}). This results in 24 tasks and approximately 1K windows to impute.
\end{enumerate*}

\paragraph{Baselines.} \begin{enumerate*}[(i)] 
    \item \textit{Univariate comparisons} include Tabular Foundation Models (TFMs) adapted to time series: \texttt{TabPFNv2.5-TS}~\citep{hoo2024tables2time}, \texttt{TabICLv2-TS}~\citep{qu2026tabiclv2}, along with standard local methods: \texttt{Linear Interpolation}, \texttt{Seasonal Naive}, and \texttt{Last Observation Carried Forward (LOCF)}. 
    In addition, supervised imputation models \texttt{SAITS}~\citep{du2023saits} and \texttt{BRITS}~\citep{cao2018brits}, trained per dataset, serve as strong task-specific baselines.
    \item In the \textit{known-covariates} setting, the same foundation models are considered, together with ridge regression on covariates to quantify the predictive signal of exogenous variables.
    Variants of foundation models without covariates are also included for comparison.
\end{enumerate*}
Following \texttt{fm-impute-bench}, pointwise performance is evaluated with Normalized Mean Absolute Error (NMAE) and probabilistic performance with Continuous Ranked Probability Score (CRPS).
Metric definitions are provided in Appendix \ref{sec:scores-metrics}.
\cref{fig:imputations-score-tmlr} illustrates the trade-off between pointwise and probabilistic performance, while \cref{tab:imputation-tmlr-inference-cost} reports the median inference time.

\begin{figure}[h!]
    \centering
    \begin{subfigure}[b]{0.495\textwidth}
        \centering
        \includegraphics[width=\linewidth]{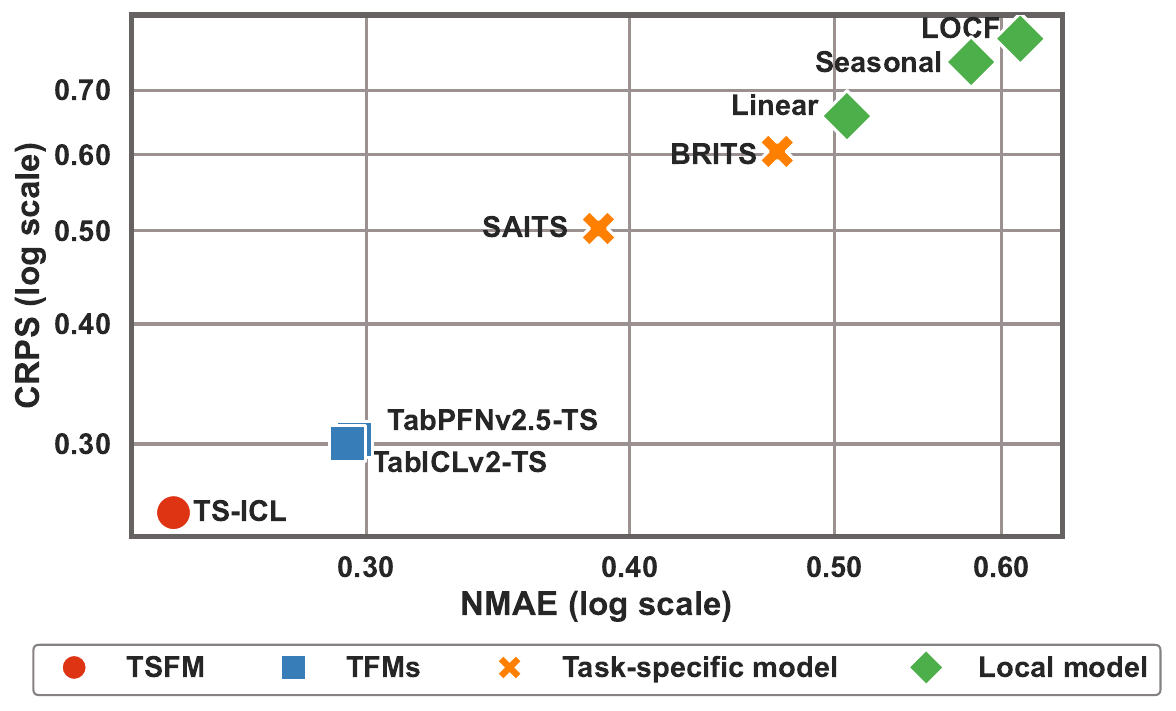}
        \caption{\textit{Univariate} imputation across 132 tasks.}
        \label{fig:Univariate-imputation-score-tmlr}
    \end{subfigure}
    \hfill
    \begin{subfigure}[b]{0.495\textwidth}
        \centering
        \includegraphics[width=\linewidth]{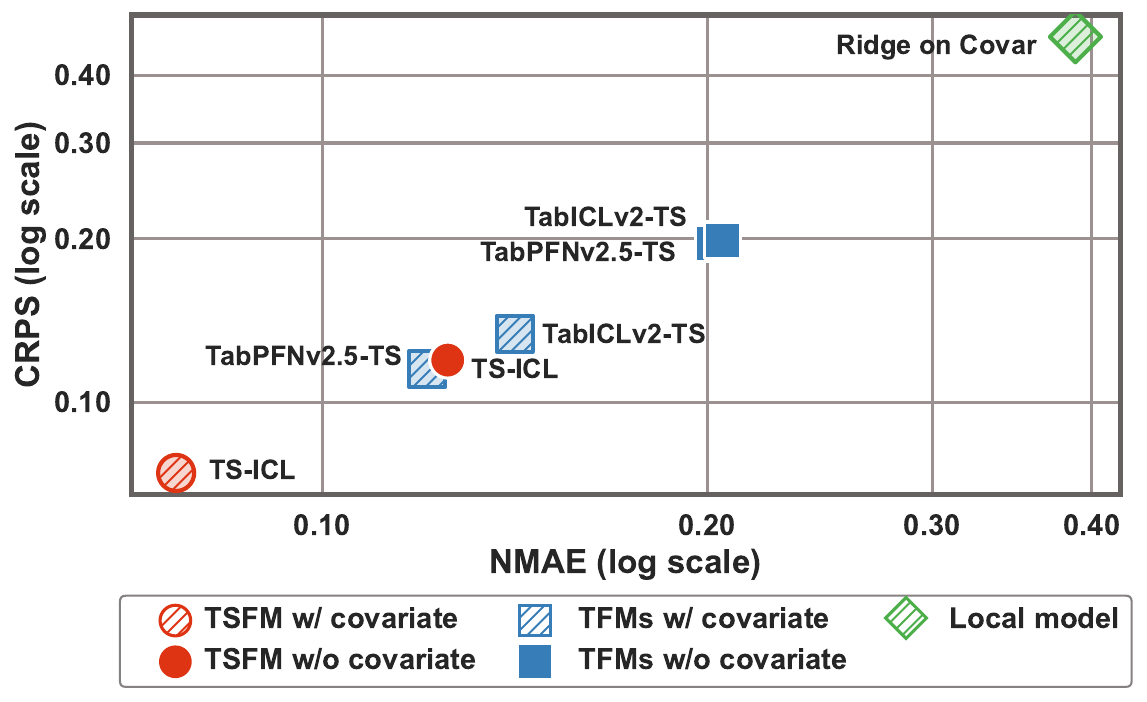}
        \caption{Imputation with \textit{known covariates} across 24 tasks.}
        \label{fig:covariates-imputation-score-tmlr}
    \end{subfigure}
    \caption{NMAE-CRPS (lower is better) on the \texttt{fm-impute} benchmark. Each point corresponds to a method, averaged across tasks.}
    \label{fig:imputations-score-tmlr}
\end{figure}

\paragraph{Results.}
\texttt{TS-ICL} sets a new state-of-the-art for zero-shot imputation, improving both pointwise and probabilistic scores over TFMs while being up to two orders of magnitude faster at inference. 

\begin{figure}[t!]
    \centering
    \begin{subfigure}[b]{0.495\textwidth}
        \centering
        \includegraphics[width=\linewidth]{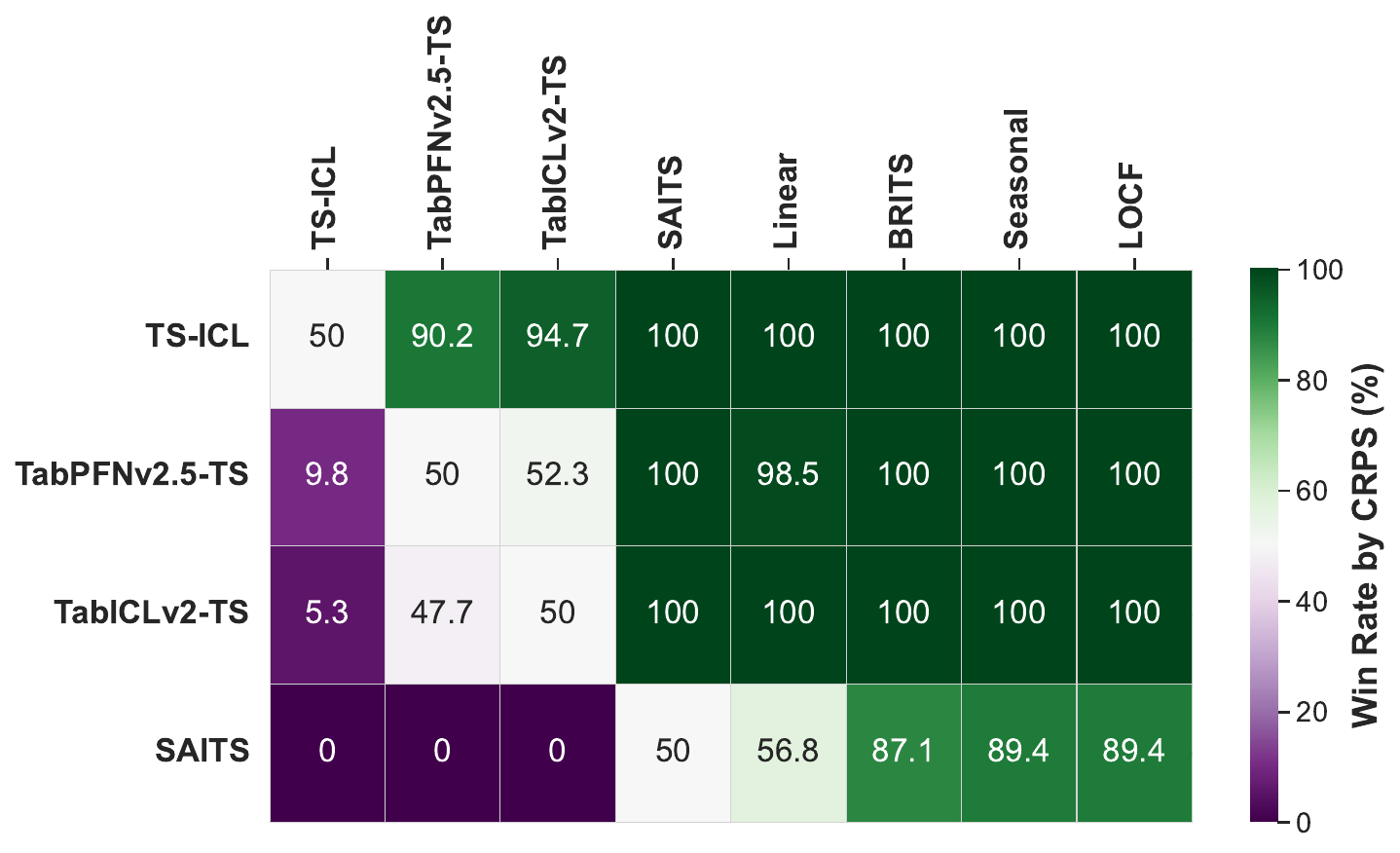}
        \caption{\textit{Univariate} imputation across 132 tasks.}
        \label{fig:Univariate-imputation-wins-tmlr}
    \end{subfigure}
    \hfill
    \begin{subfigure}[b]{0.495\textwidth}
        \centering
        \includegraphics[width=\linewidth]{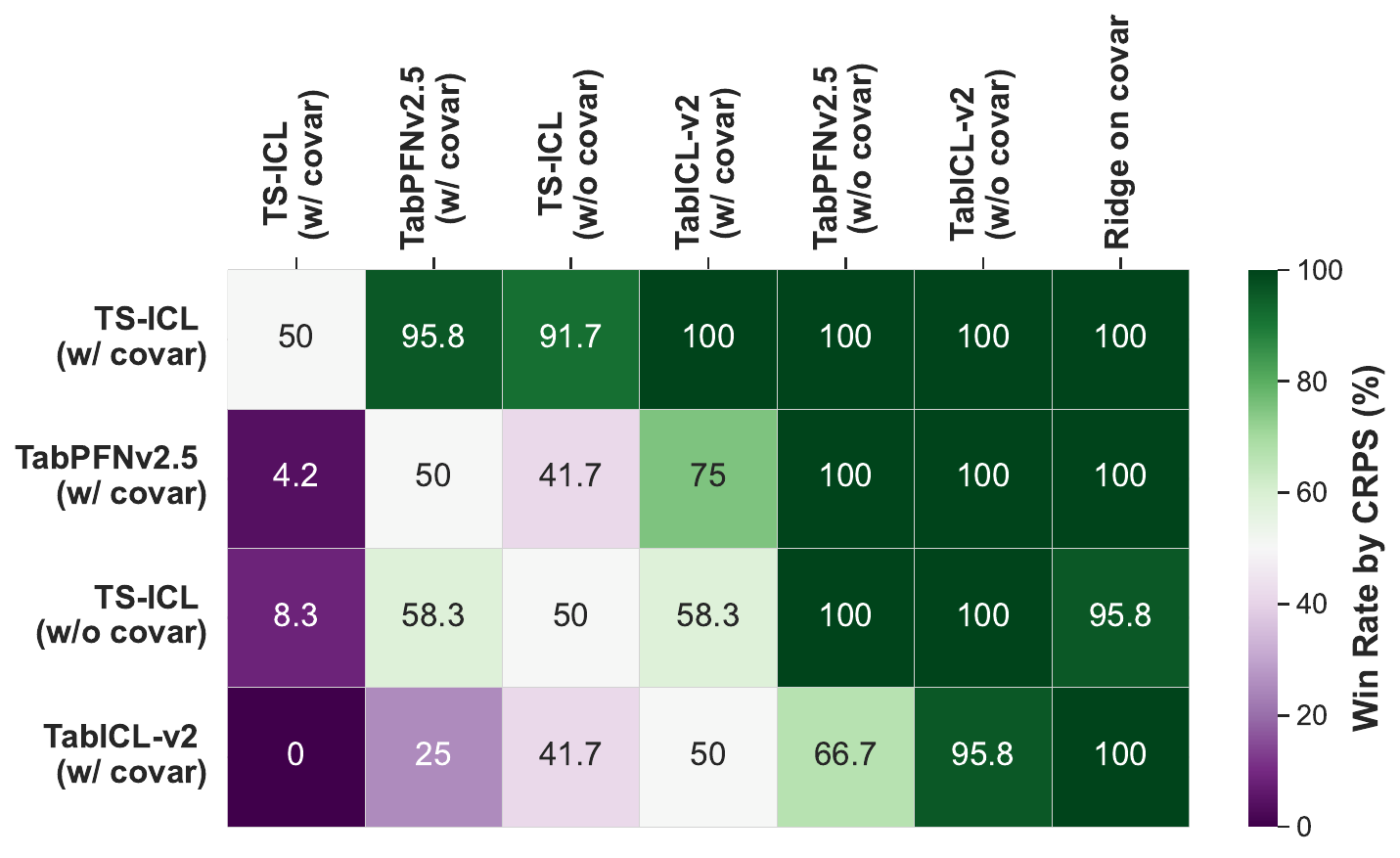}
        \caption{Imputation with \textit{known covariates} across 24 tasks.}
        \label{fig:covariates-imputation-wins-tmlr}
    \end{subfigure}
    \caption{Pairwise win rates of the top-4 models on the \texttt{fm-impute} benchmark. Each entry indicates the fraction of tasks where a method outperforms another according to the CRPS.}
    \label{fig:imputations-wins-tmlr}
    \vspace{-0.3cm}
\end{figure}

\begin{enumerate}[(i),
leftmargin=0pt,
labelsep=0.5em,
itemindent=1.5em]
\item \textit{Univariate setting.} As shown in \cref{fig:Univariate-imputation-score-tmlr}, \texttt{TS-ICL} achieves lower NMAE and CRPS than competing TFMs, improving over \texttt{TabICLv2-TS} by 17\% and 15\%, respectively, while being $\sim\!50\times$ faster at inference (\cref{tab:imputation-tmlr-inference-cost}). \texttt{TabPFNv2.5-TS} and \texttt{TabICLv2-TS} perform similarly and outperform task-specific and local baselines by a wide margin. 
Pairwise win rates (\cref{fig:Univariate-imputation-wins-tmlr}) further show that \texttt{TS-ICL} dominates on the vast majority of tasks, indicating robustness across tasks.

\item \textit{Covariate-aware setting.} Results in \cref{fig:covariates-imputation-score-tmlr} show similar trends when covariates are available. 
\texttt{TS-ICL} improves over \texttt{TabPFNv2.5-TS} by 36\% in NMAE and 35\% in CRPS, and gains 39\% (NMAE) and 38\% (CRPS) over its variant without covariates. 
While all TFMs benefit from covariates, they remain below \texttt{TS-ICL}. Ridge regression indicates limited predictive power of covariates alone, and pairwise comparisons (\cref{fig:covariates-imputation-wins-tmlr}) demonstrate the clear dominance of \texttt{TS-ICL}.
\end{enumerate}

\begin{table}[h!]
\caption{Median imputation inference time on \texttt{fm-impute-bench univar} with a H100 GPU.}
\centering
\scalebox{0.63}{
\begin{tabular}{lcccccccc}
\toprule
 & {TSFM} & \multicolumn{2}{c}{Tabular Foundation models} & \multicolumn{2}{c}{Task-Specific Models} & \multicolumn{3}{c}{Local models} \\
\cmidrule(r){2-2} \cmidrule(r){3-4} \cmidrule(r){5-6} \cmidrule(r){7-9} 
 & \texttt{TS-ICL} & \texttt{TabPFNv2.5-TS} & \texttt{TabICLv2-TS} & \texttt{SAITS} & \texttt{BRITS}
 & \shortstack{\texttt{Linear} \\ \texttt{interpolation}} 
 & \shortstack{\texttt{Seasonal} \\ \texttt{Naive}} 
 & \texttt{LOCF} \\
\midrule
\shortstack{Inference time \\ (s per window)} & $6.51 \times 10^{-3}$ & $2.80 \times 10^{-1}$ & $3.07 \times 10^{-1}$ & $1.33 \times 10^{-2}$ & $4.52 \times 10^{-1}$ & $1.61 \times 10^{-4}$ & $7.54 \times 10^{-4}$ & $5.58 \times 10^{-4}$ \\
\bottomrule
\end{tabular}}
\label{tab:imputation-tmlr-inference-cost}
\end{table}

Overall, \texttt{TS-ICL} outperforms state-of-the-art TFMs for zero-shot imputation in both \textit{univariate} and \textit{covariate-informed} settings, while being two orders of magnitude faster at inference.
Additional qualitative results in Appendix~\ref{extended-imputation-expes}, including \cref{fig:tmlr-plots-1,fig:tmlr-plots-2}, as well as results on the \texttt{TIME} benchmark~\citep{qiao2026sTIME}, further support these findings.



\subsection{Forecasting Experiments}
\label{sec:forecasting-expe}

The zero-shot forecasting capability of \texttt{TS-ICL} is evaluated on \texttt{fev-bench} \citep{shchur2025fev}, a comprehensive benchmark covering diverse datasets, horizons, and sampling frequencies, with controlled evaluations in both \textit{univariate} and \textit{known-covariates} settings. 
An additional univariate setting with \textit{missing values in the look-back window} is also considered across the entire \texttt{fev-bench} benchmark.

\paragraph{Setting.} Evaluation considers two forecasting regimes. \begin{enumerate*}[(i)]
    \item In the \textit{univariate setting}, the benchmark comprises 100 tasks and $\sim$235k forecasting windows (see \cref{tab:fevbench-full-datasets}, Appendix~\ref{sec:fevbench-appendix}). 
    Models rely solely on past observations, with no access to covariates or cross-series information.
    \item In the \textit{known-covariate setting}, we evaluate all methods on the same 100 tasks. 
    Among these, 30 datasets include meaningful exogenous time series, denoted as "known dynamics" covariates in \cref{tab:fevbench-full-datasets}. 
    For these datasets, methods that support covariates are evaluated both with and without covariate inputs, while methods that do not are kept unchanged. 
    This protocol enables a controlled assessment of the effect of covariate information while preserving comparability across tasks.
\end{enumerate*}

\begin{figure}[h!] 
\centering
\begin{subfigure}[b]{0.49\textwidth}
    \centering
    \includegraphics[width=\linewidth]{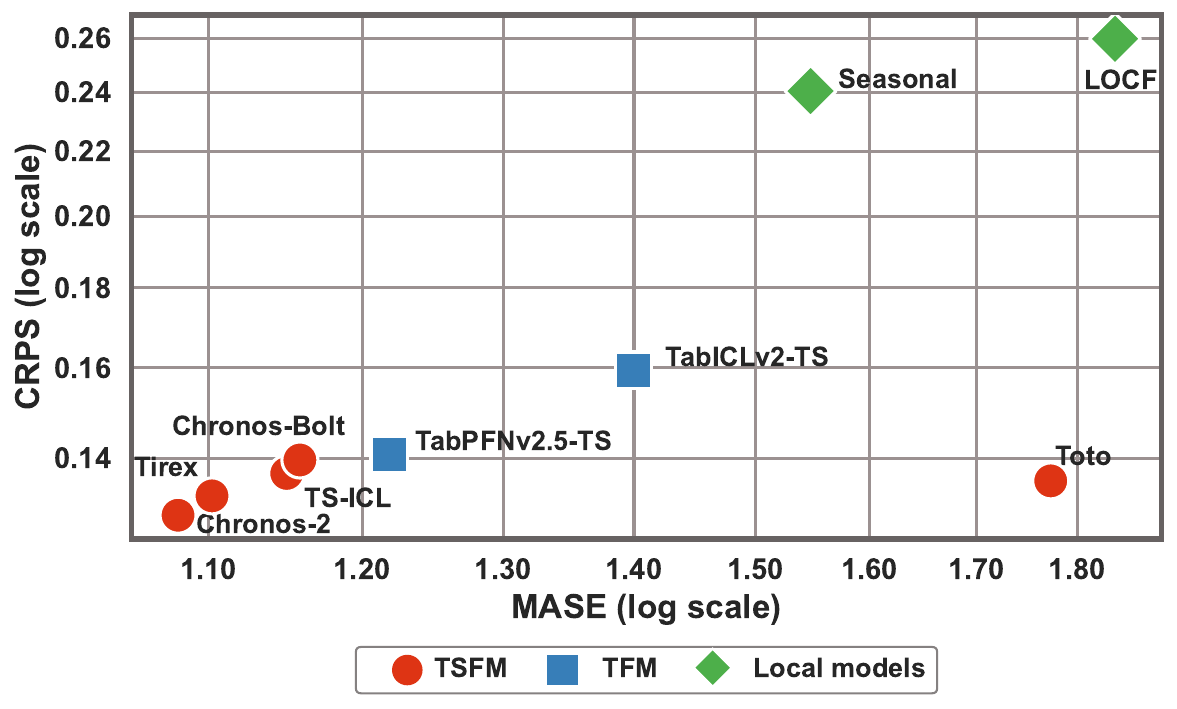}
    \caption{\textit{Univariate} forecasting (100 tasks).}
    \label{fig:Univariate-forecasting-score-fevbench}
\end{subfigure}
\hfill
\begin{subfigure}[b]{0.49\textwidth}
    \centering
    \includegraphics[width=\linewidth]{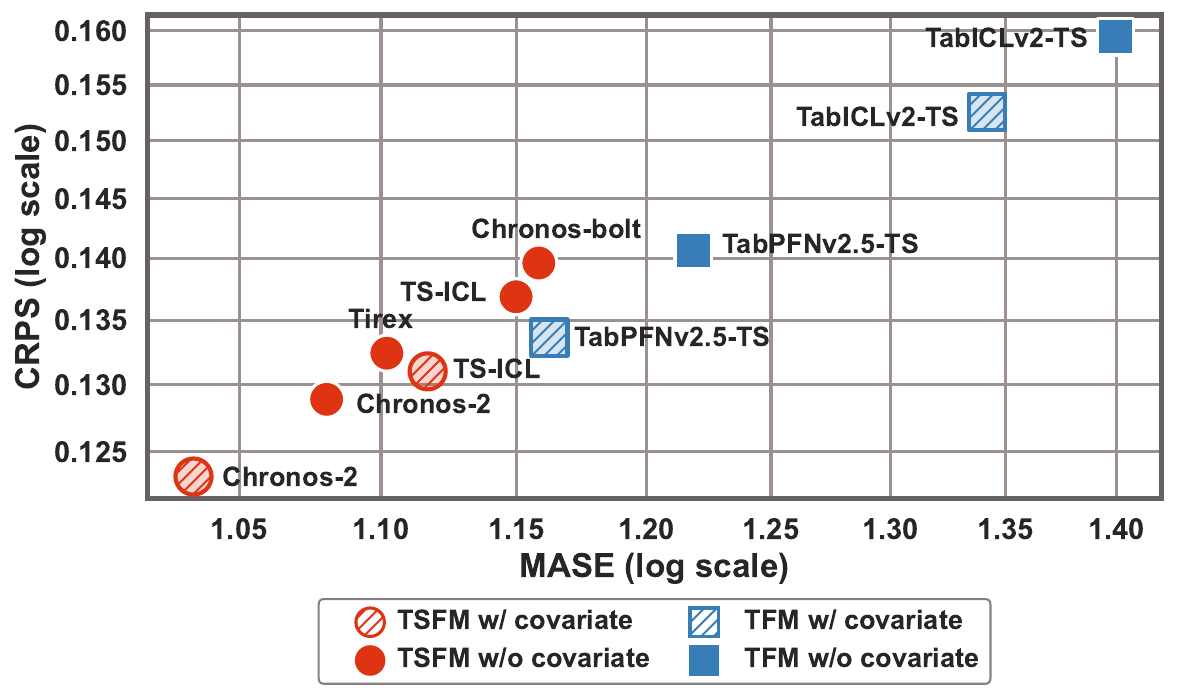}
    \caption{Forecasting on 100 tasks (30 \textit{covariate-aware}).}
    \label{fig:covariate-forecasting-score-fevbench}
\end{subfigure}
    \begin{subfigure}[b]{0.49\textwidth}
        \centering
        \includegraphics[width=\linewidth]{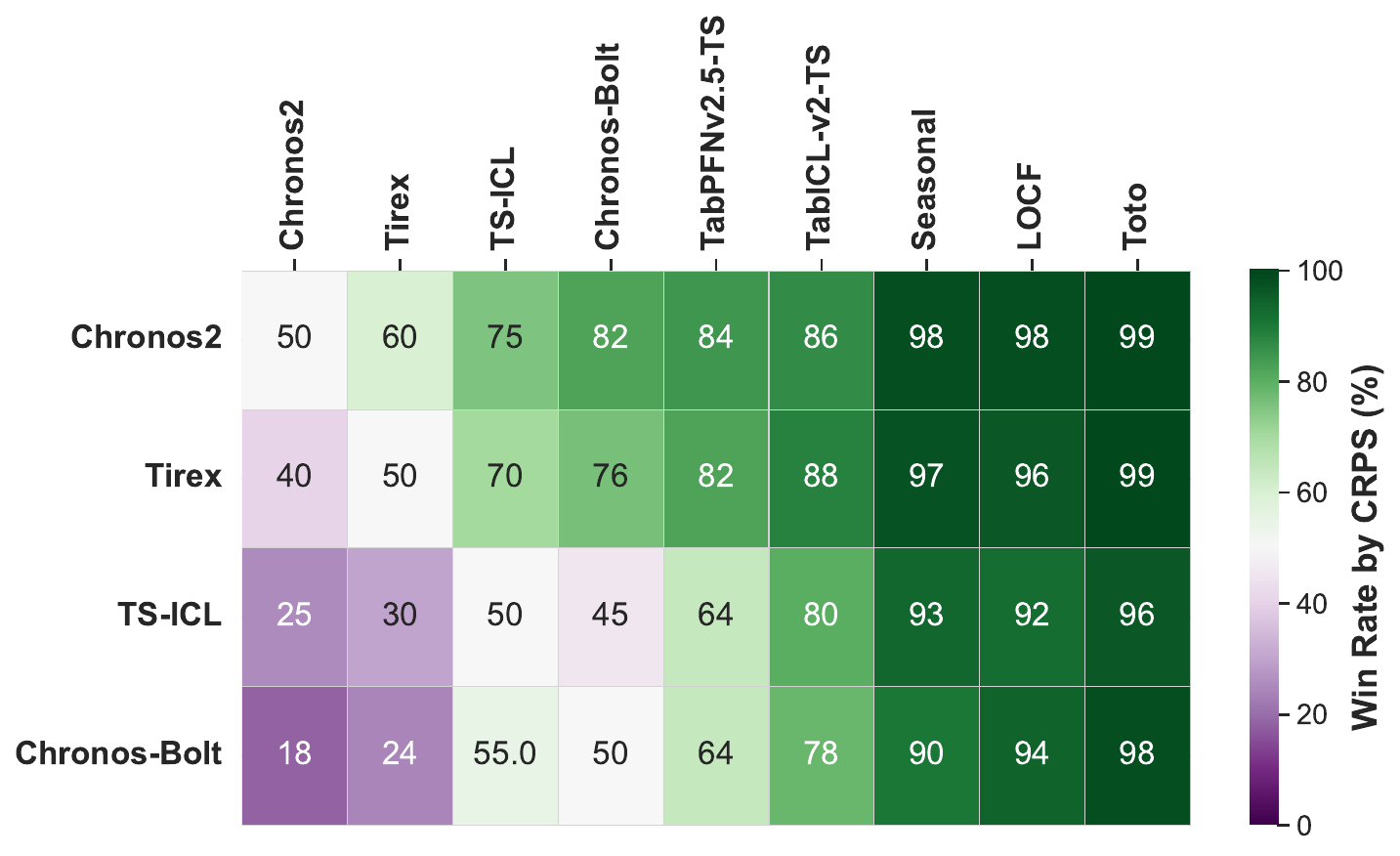}
        \caption{\texttt{Fev-bench}. \textit{Univariate} forecasting (100 tasks)}
        \label{fig:Univariate-forecasting-fev-wins}
    \end{subfigure}
    \hfill
    \begin{subfigure}[b]{0.49\textwidth}
        \centering
        \includegraphics[width=\linewidth]{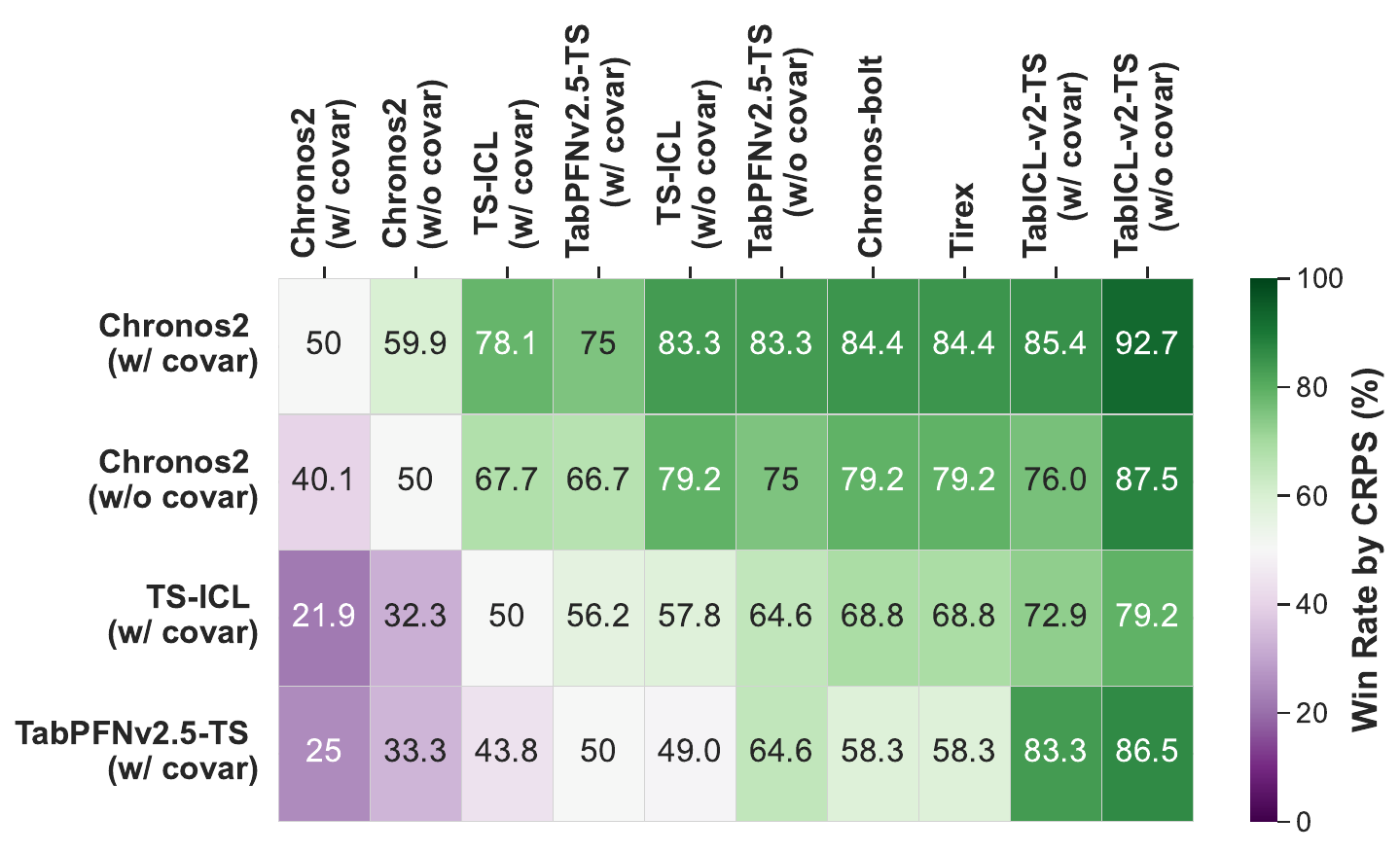}
        \caption{Forecasting on 100 tasks (30 \textit{covariate-aware}).}
        \label{fig:covariates-forecasting-fev-wins}
    \end{subfigure}
\caption{
\texttt{Fev-bench} forecasting benchmark.
(a)-(b) MASE-CRPS trade-off (lower is better).
Points correspond to per-method scores averaged across tasks.
(c)-(d) Pairwise win rates. Each entry indicates the fraction of tasks where a method outperforms another according to the CRPS.
}
\label{fig:forecasting-scores-fevbench}
\end{figure}

\paragraph{Baselines.}
\texttt{TS-ICL} is compared against a broad set of baselines covering foundation models and local methods. 
\begin{enumerate*}[(i)]
\item \textit{Univariate comparisons} include state-of-the-art TSFMs (\texttt{Chronos-2} \citep{ansari2024chronos2}, \texttt{TiRex} \citep{auer2025tirex}, \texttt{Chronos-bolt} and \texttt{Toto} \citep{TOTO2025}), TFMs adapted to time series (\texttt{TabPFNv2.5-TS}, \texttt{TabICLv2-TS}) and local baselines (\texttt{Seasonal Naive}, \texttt{LOCF}). 
Supervised forecasting models are omitted, as prior large-scale studies indicate that TSFMs generally outperform them on established benchmarks \citep{aksu2024gift}. \texttt{TimesFM 2.5} \citep{TimesFM} and \texttt{Moirai2} \citep{liu2025moirai} are excluded from evaluations due to substantial data leakage with \texttt{fev-bench}. 
\item In the \textit{known-covariate} setting, we evaluate both TSFMs and TFMs under the unified protocol described above. 
Methods supporting covariates (\texttt{TS-ICL}, \texttt{Chronos-2}, \texttt{TabPFNv2.5-TS}, \texttt{TabICLv2-TS}) are evaluated with and without covariates on the 30 relevant datasets, while covariate-agnostic TSFMs are included unchanged to quantify the benefit of incorporating exogenous information.
\cref{fig:forecasting-scores-fevbench} reports the results following the \texttt{fev-bench} protocol, using Mean Absolute Scaled Error (MASE) and CRPS (metric definitions provided in Appendix \ref{sec:scores-metrics}) while \cref{tab:forecasting-time-inference-fevbench} reports the median inference time.
\end{enumerate*}

%


\paragraph{Results.} \texttt{TS-ICL} achieves strong zero-shot performance on \texttt{fev-bench}, remaining competitive with leading TSFMs while consistently outperforming TFMs and local baselines on both point and probabilistic metrics.

\begin{figure}[h!]
\centering
\includegraphics[width=0.9\linewidth]{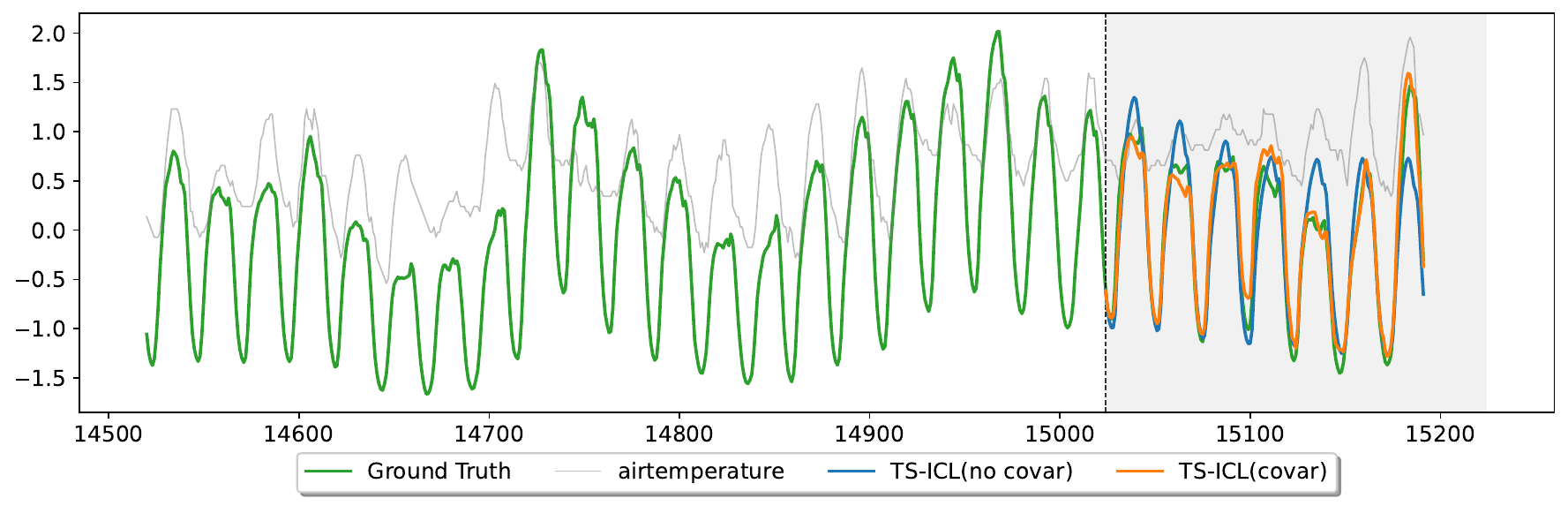}
\caption{\texttt{TS-ICL} forecast on an horizon of length 168, with one additional covariate, on \emph{GFC17}.}
\label{fig:tsicl-quali}
\end{figure}

\begin{enumerate}[(i), 
    leftmargin=0pt, 
    labelsep=0.5em, 
    itemindent=1.5em]
\item In the \textit{univariate setting}, \texttt{TS-ICL} ranks among the top-performing methods (\cref{fig:Univariate-forecasting-score-fevbench}), remaining within $\sim$6\% of \texttt{Chronos-2} and $\sim$3\% of \texttt{TiRex}, while outperforming all other baselines, including TFMs and local methods. Pairwise comparisons (\cref{fig:Univariate-forecasting-fev-wins}) confirm consistent majority wins across tasks against tabular and local approaches. Finally, it offers a strong accuracy--efficiency trade-off (\cref{tab:forecasting-time-inference-fevbench}), with inference time on the order of $10^{-2}$ seconds per window, around $4\times$ slower than \texttt{Chronos2} but still about $40\times$ faster than TFMs.

\item In the \textit{known-covariate setting}, performance improves with exogenous information (\cref{fig:covariate-forecasting-score-fevbench}). \texttt{Chronos2} remains the strongest overall method, while \texttt{TS-ICL} benefits from covariates, as illustrated qualitatively in \cref{fig:tsicl-quali}, and consistently outperforms TFMs under identical inputs.
It also improves its relative ranking among TSFMs, surpassing \texttt{TiRex} in CRPS and achieving $\sim$70\% pairwise win rates (\cref{fig:covariates-forecasting-fev-wins}), indicating stable gains on \texttt{fev-bench} when leveraging covariates.
\end{enumerate}

\begin{table}[h!] 
\centering
\caption{Median forecasting inference time on \texttt{fev-bench} (\emph{univariate setting}) with a H100 GPU.}
\label{tab:forecasting-time-inference-fevbench}
\scalebox{0.64}{ 
\begin{tabular}{lcccccccc}
\toprule
 & \multicolumn{4}{c}{Time Series Foundation Models} & \multicolumn{2}{c}{Tabular Foundation Models} & \multicolumn{2}{c}{Local Methods} \\
\cmidrule(r){2-5} \cmidrule(r){6-7} \cmidrule(r){8-9}
 & \texttt{TS-ICL} & \texttt{Chronos-2} & \texttt{TiRex} & \texttt{TOTO} 
 & \texttt{TabPFNv2.5-TS} & \texttt{TabICLv2-TS} 
 & \texttt{S-Naive} & \texttt{LOCF} \\
\midrule
\shortstack{Median Inference \\ time (s / window)} 
& $1.54 \times 10^{-2}$ & $3.53 \times 10^{-3}$ & $5.20 \times 10^{-2}$ & $1.28 \times 10^{-1}$ & $4.33 \times 10^{-1}$ & $3.76 \times 10^{-1}$ & $3.09 \times 10^{-4}$ & $1.73 \times 10^{-4}$ \\
\bottomrule
\end{tabular}}
\end{table}

\paragraph{Forecasting with missing values.} Beyond controlled evaluation settings on \texttt{fev-bench}, evaluating \texttt{TS-ICL} robustness in realistic scenarios with partially missing historical observations is crucial. \cref{tab:forecasting-with-missing} compares \texttt{TS-ICL} and \texttt{Chronos-2} forecasting performance under increasing levels of look-back window missingness (30\%–90\%) across the entire \texttt{fev-bench} benchmark. While both models degrade as the context becomes sparser, \texttt{TS-ICL} consistently outperforms \texttt{Chronos-2}, with significant gaps at all missingness levels. As a result, \texttt{TS-ICL} remains substantially more robust for forecasting under missing historical observations.

\begin{table}[h!] 
\centering 
\caption{
Zero-shot forecasting MASE under increasing missingness in the look-back window for \texttt{TS-ICL} and \texttt{Chronos-2} on \texttt{fev-bench} (100 univariate tasks, look-back = 4092, arithmetic mean).
Relative degradation (\%) is reported in parentheses.
\texttt{Seasonal Naive} serves as a simple baseline for evaluating forecasting performance under degraded conditions. Best results are in \textbf{bold}.
}
\label{tab:forecasting-with-missing} 
\scalebox{0.80}{ 
\begin{tabular}{lccccc} 
\toprule 
 & 0 \% missing & 30 \% missing & 50 \% missing & 70 \% missing & 90 \% missing \\ 
\midrule 
\texttt{Chronos-2}  
& \textbf{1.62 (0\%)} & 2.16 (-33\%) & 2.44 (-50\%) & 2.54 (-56\%) & 3.97 (-144\%) \\ 
\texttt{TS-ICL}  
& 1.70 (0\%) & \textbf{1.77 (-4\%)} & \textbf{1.89} (-11\%) & \textbf{2.16} (-27\%) & 3.63 (-113\%) \\ 
\midrule
\shortstack{\texttt{Seasonal}}{\texttt{(0\% missing)}} 
& 2.48 & 2.48 & 2.48 & 2.48 & \textbf{2.48} \\ 
\bottomrule 
\end{tabular} 
} 
\end{table}


Overall, \texttt{TS-ICL} is thus a competitive non patch-based TSFM for zero-shot forecasting.
It remains close to state-of-the art TSFMs such as \texttt{Chronos-2} and \texttt{TiRex} across standard benchmarks, while effectively leveraging covariates when available.
Its most notable advantage lies in its robustness to missing history, where it consistently outperforms \texttt{Chronos-2}, highlighting the sensitivity of patch-based models to incomplete context.
In contrast, \texttt{TS-ICL} benefits from its time-indexed formulation to tackle realistic forecasting settings.
Additional analyses and forecasting plots are provided in Appendix~\ref{sec:fevbench-appendix}, while Appendix~\ref{sec:time-forecasting-appendix} reports leakage-free comparisons against 12 TSFMs on the \texttt{TIME} benchmark \citep{qiao2026sTIME} across 98 zero-shot tasks.

\section{Conclusion}
\label{sec:conclusion}

\texttt{TS-ICL} is a flexible probabilistic time series foundation model based on an in-context regression formulation, unifying forecasting and imputation within a continuous-time framework.
It sets new state-of-the-art performance on zero-shot imputation, while remaining competitive with leading TSFMs on forecasting benchmarks and serving as a strong alternative to patch-based models. 
A key advantage of \texttt{TS-ICL} lies in its robustness to incomplete historical observations: it consistently outperforms \texttt{Chronos-2} under partially observed look-back windows, highlighting the benefits of its time-indexed formulation in realistic forecasting settings. 
It also enables efficient covariate-aware inference, further improving performance when exogenous information is available.
Despite these strengths, \texttt{TS-ICL} exhibits higher inference cost than highly optimized models such as \texttt{Chronos-2} (up to $4\times$ slower despite having $4\times$ fewer parameters), mainly due to its pointwise regression formulation, which increases computational cost during both training and inference. 
These limitations may be mitigated through architectural optimizations such as caching or mixed-precision training. 
In addition, further increasing the diversity of the training data prior may improve the zero-shot generalization of \texttt{TS-ICL}, as observed in tabular foundation models \citep{hollmann2025accurate, qu2026tabiclv2}. 
Finally, the flexibility of the \texttt{TS-ICL} encoder--regressor framework makes it a natural foundation for extending beyond forecasting and imputation to tasks such as zero-shot anomaly detection and time series classification.

\clearpage
\section*{Acknowledgements}

We would like to thank Ghislain Agoua, whose contributions during the early stages of this work helped lay the foundations for this project. We would also like to express our sincere gratitude to Louis Serrano for the insightful discussions on encoder architectures and for open-sourcing the AROMA implementation. We are particularly grateful to Marc Héry for his invaluable help in developing the package accompanying this work. We are also grateful to the TabICL team for the stimulating exchanges on synthetic priors and, more broadly, on foundation models. Their openness and dedication to open-source research have been invaluable to this work.

We further thank our colleagues at EDF for their careful review of the manuscript and for the constructive feedback they provided throughout the project. Finally, we acknowledge the broader time series research community for openly sharing datasets, code, and tools, without which this work would not have been possible.

\bibliography{mybibliography}

\begin{thebibliography}{45}
\providecommand{\natexlab}[1]{#1}
\providecommand{\url}[1]{\texttt{#1}}
\expandafter\ifx\csname urlstyle\endcsname\relax
  \providecommand{\doi}[1]{doi: #1}\else
  \providecommand{\doi}{doi: \begingroup \urlstyle{rm}\Url}\fi

\bibitem[Aksu et~al.(2024)Aksu, Woo, Liu, Liu, Liu, Savarese, Xiong, and Sahoo]{aksu2024gift}
Taha Aksu, Gerald Woo, Juncheng Liu, Xu~Liu, Chenghao Liu, Silvio Savarese, Caiming Xiong, and Doyen Sahoo.
\newblock {GIFT}-eval: A benchmark for general time series forecasting model evaluation.
\newblock In \emph{NeurIPS Workshop on Time Series in the Age of Large Models}, 2024.
\newblock URL \url{https://openreview.net/forum?id=Z2cMOOANFX}.

\bibitem[Ansari et~al.(2024)Ansari, Stella, T{\"{u}}rkmen, Zhang, Mercado, Shen, Shchur, Rangapuram, Pineda{-}Arango, Kapoor, Zschiegner, Maddix, Wang, Mahoney, Torkkola, Wilson, Bohlke{-}Schneider, and Wang]{Chronosv1}
Abdul~Fatir Ansari, Lorenzo Stella, Ali~Caner T{\"{u}}rkmen, Xiyuan Zhang, Pedro Mercado, Huibin Shen, Oleksandr Shchur, Syama~Sundar Rangapuram, Sebastian Pineda{-}Arango, Shubham Kapoor, Jasper Zschiegner, Danielle~C. Maddix, Hao Wang, Michael~W. Mahoney, Kari Torkkola, Andrew~Gordon Wilson, Michael Bohlke{-}Schneider, and Bernie Wang.
\newblock Chronos: Learning the language of time series.
\newblock \emph{Trans. Mach. Learn. Res.}, 2024.

\bibitem[Ansari et~al.(2025)Ansari, Shchur, K{\"u}ken, Auer, Han, Mercado, Rangapuram, Shen, Stella, Zhang, et~al.]{ansari2024chronos2}
Abdul~Fatir Ansari, Oleksandr Shchur, Jaris K{\"u}ken, Andreas Auer, Boran Han, Pedro Mercado, Syama~Sundar Rangapuram, Huibin Shen, Lorenzo Stella, Xiyuan Zhang, et~al.
\newblock Chronos-2: From univariate to universal forecasting.
\newblock \emph{arXiv preprint arXiv:2510.15821}, 2025.

\bibitem[Auer et~al.(2025)Auer, Podest, Klotz, B{\"o}ck, Klambauer, and Hochreiter]{auer2025tirex}
Andreas Auer, Patrick Podest, Daniel Klotz, Sebastian B{\"o}ck, G{\"u}nter Klambauer, and Sepp Hochreiter.
\newblock Tirex: Zero-shot forecasting across long and short horizons with enhanced in-context learning.
\newblock In \emph{The Thirty-ninth Annual Conference on Neural Information Processing Systems}, 2025.
\newblock URL \url{https://openreview.net/forum?id=v7UqniC9pF}.

\bibitem[Brown et~al.(2020)Brown, Mann, Ryder, Subbiah, Kaplan, Dhariwal, Neelakantan, Shyam, Sastry, Askell, et~al.]{brown2020language}
Tom Brown, Benjamin Mann, Nick Ryder, Melanie Subbiah, Jared~D Kaplan, Prafulla Dhariwal, Arvind Neelakantan, Pranav Shyam, Girish Sastry, Amanda Askell, et~al.
\newblock Language models are few-shot learners.
\newblock \emph{Advances in Neural Information Processing Systems}, 33:\penalty0 1877--1901, 2020.

\bibitem[Cao et~al.(2018)Cao, Wang, Li, Zhou, Li, and Li]{cao2018brits}
Wei Cao, Dong Wang, Jian Li, Hao Zhou, Yitan Li, and Lei Li.
\newblock {BRITS}: bidirectional recurrent imputation for time series.
\newblock In \emph{Advances in Neural Information Processing Systems}, volume~31, 2018.

\bibitem[Chen et~al.(2025)Chen, Shen, Li, Wang, Sun, and Liu]{chen2024visionts}
Mouxiang Chen, Lefei Shen, Zhuo Li, Xiaoyun~Joy Wang, Jianling Sun, and Chenghao Liu.
\newblock Vision{TS}: Visual masked autoencoders are free-lunch zero-shot time series forecasters.
\newblock In \emph{Forty-second International Conference on Machine Learning}, 2025.
\newblock URL \url{https://openreview.net/forum?id=5DSj3MfWrB}.

\bibitem[Chen et~al.(2018)Chen, Rubanova, Bettencourt, and Duvenaud]{chen2018neural}
Ricky T.~Q. Chen, Yulia Rubanova, Jesse Bettencourt, and David Duvenaud.
\newblock Neural ordinary differential equations.
\newblock In \emph{Advances in Neural Information Processing Systems}, volume~31, 2018.

\bibitem[Clark \& Bj{\o}rnstad(2004)Clark and Bj{\o}rnstad]{clark2004population}
James~S Clark and Ottar~N Bj{\o}rnstad.
\newblock Population time series: process variability, observation errors, missing values, lags, and hidden states.
\newblock \emph{Ecology}, 85\penalty0 (11):\penalty0 3140--3150, 2004.

\bibitem[Cohen et~al.(2025)Cohen, Khwaja, Doubli, Lemaachi, Lettieri, Masson, Miccinilli, Ram{\'e}, Ren, Rostamizadeh, du~Terrail, Toon, Wang, Xie, Xu, Zhukova, Asker, Talwalkar, and Abou-Amal]{TOTO2025}
Ben Cohen, Emaad Khwaja, Youssef Doubli, Salahidine Lemaachi, Chris Lettieri, Charles Masson, Hugo Miccinilli, Elise Ram{\'e}, Qiqi Ren, Afshin Rostamizadeh, Jean~Ogier du~Terrail, Anna-Monica Toon, Kan Wang, Stephan Xie, Zongzhe Xu, Viktoriya Zhukova, David Asker, Ameet Talwalkar, and Othmane Abou-Amal.
\newblock This time is different: An observability perspective on time series foundation models.
\newblock In \emph{The Thirty-ninth Annual Conference on Neural Information Processing Systems}, 2025.
\newblock URL \url{https://openreview.net/forum?id=1jDAYXfcS2}.

\bibitem[Das et~al.(2024)Das, Kong, Sen, and Zhou]{TimesFM}
Abhimanyu Das, Weihao Kong, Rajat Sen, and Yichen Zhou.
\newblock A decoder-only foundation model for time-series forecasting.
\newblock In \emph{Forty-first International Conference on Machine Learning, {ICML} 2024, Vienna, Austria, July 21-27, 2024}, Proceedings of Machine Learning Research, 2024.

\bibitem[Dooley et~al.(2023)Dooley, Khurana, Mohapatra, Naidu, and White]{dooley2023forecastpfn}
Samuel Dooley, Gurnoor~Singh Khurana, Chirag Mohapatra, Siddartha~V Naidu, and Colin White.
\newblock {ForecastPFN}: Synthetically-trained zero-shot forecasting.
\newblock In \emph{Advances in Neural Information Processing Systems}, volume~36, pp.\  2403--2426, 2023.

\bibitem[Du et~al.(2023{\natexlab{a}})Du, C{\^o}t{\'e}, and Liu]{du2023saits}
Wenjie Du, David C{\^o}t{\'e}, and Yan Liu.
\newblock {SAITS}: Self-attention-based imputation for time series.
\newblock \emph{Expert Systems with Applications}, 219:\penalty0 119619, 2023{\natexlab{a}}.
\newblock \doi{https://doi.org/10.1016/j.eswa.2023.119619}.

\bibitem[Du et~al.(2023{\natexlab{b}})Du, Yang, Qian, Wang, and Wen]{du2023pypots}
Wenjie Du, Yiyuan Yang, Linglong Qian, Jun Wang, and Qingsong Wen.
\newblock {PyPOTS: A Python Toolkit for Machine Learning on Partially-Observed Time Series}.
\newblock \emph{arXiv:2305.18811}, 2023{\natexlab{b}}.

\bibitem[Garg et~al.(2022)Garg, Tsipras, Liang, and Valiant]{garg2022what}
Shivam Garg, Dimitris Tsipras, Percy~S Liang, and Gregory Valiant.
\newblock What can transformers learn in-context? a case study of simple function classes.
\newblock In S.~Koyejo, S.~Mohamed, A.~Agarwal, D.~Belgrave, K.~Cho, and A.~Oh (eds.), \emph{Advances in Neural Information Processing Systems}, volume~35, pp.\  30583--30598. Curran Associates, Inc., 2022.

\bibitem[Gneiting et~al.(2007)Gneiting, Balabdaoui, and Raftery]{gneiting2007probabilistic}
Tilmann Gneiting, Fadoua Balabdaoui, and Adrian~E Raftery.
\newblock Probabilistic forecasts, calibration and sharpness.
\newblock \emph{Journal of the Royal Statistical Society Series B: Statistical Methodology}, 69\penalty0 (2):\penalty0 243--268, 2007.

\bibitem[Hollmann et~al.(2022)Hollmann, M{\"u}ller, Eggensperger, and Hutter]{hollmann2022tabpfn}
Noah Hollmann, Samuel M{\"u}ller, Katharina Eggensperger, and Frank Hutter.
\newblock {TabPFN}: A transformer that solves small tabular classification problems in a second.
\newblock In \emph{The Eleventh International Conference on Learning Representations}, 2022.

\bibitem[Hollmann et~al.(2025)Hollmann, M{\"u}ller, Purucker, Krishnakumar, K{\"o}rfer, Hoo, Schirrmeister, and Hutter]{hollmann2025accurate}
Noah Hollmann, Samuel M{\"u}ller, Lennart Purucker, Arjun Krishnakumar, Max K{\"o}rfer, Shi~Bin Hoo, Robin~Tibor Schirrmeister, and Frank Hutter.
\newblock Accurate predictions on small data with a tabular foundation model.
\newblock \emph{Nature}, 637\penalty0 (8045):\penalty0 319--326, 2025.

\bibitem[Hoo et~al.(2025)Hoo, M{\"u}ller, Salinas, and Hutter]{hoo2024tables2time}
Shi~Bin Hoo, Samuel M{\"u}ller, David Salinas, and Frank Hutter.
\newblock From tables to time: How {TabPFN-v2} outperforms specialized time series forecasting models.
\newblock \emph{arXiv preprint arXiv:2501.02945}, 2025.

\bibitem[Hyndman \& Athanasopoulos(2018)Hyndman and Athanasopoulos]{hyndman2018forecasting}
Rob~J Hyndman and George Athanasopoulos.
\newblock \emph{Forecasting: principles and practice}.
\newblock OTexts, 2018.

\bibitem[Jaegle et~al.(2021)Jaegle, Gimeno, Brock, Vinyals, Zisserman, and Carreira]{jaegle2021perceiver}
Andrew Jaegle, Felix Gimeno, Andy Brock, Oriol Vinyals, Andrew Zisserman, and Joao Carreira.
\newblock Perceiver: General perception with iterative attention.
\newblock In \emph{International conference on machine learning}, pp.\  4651--4664. PMLR, 2021.

\bibitem[Jordan et~al.(2024)Jordan, Jin, Boza, You, Cesista, Newhouse, and Bernstein]{jordan2024muon}
Keller Jordan, Yuchen Jin, Vlado Boza, Jiacheng You, Franz Cesista, Laker Newhouse, and Jeremy Bernstein.
\newblock Muon: An optimizer for hidden layers in neural networks, 2024.
\newblock URL \url{https://kellerjordan.github.io/posts/muon/}.

\bibitem[Kingma \& Ba(2015)Kingma and Ba]{kingma2015adam}
Diederik~P Kingma and Jimmy~Lei Ba.
\newblock {Adam: A Method for Stochastic Optimization}.
\newblock In \emph{International Conference on Learning Representations}, 2015.

\bibitem[Koenker \& Hallock(2001)Koenker and Hallock]{koenker2001quantile}
Roger Koenker and Kevin~F Hallock.
\newblock {Quantile Regression}.
\newblock \emph{Journal of Economic Perspectives}, 15\penalty0 (4):\penalty0 143--156, 2001.
\newblock \doi{10.1257/jep.15.4.143}.

\bibitem[Le~Naour et~al.(2024)Le~Naour, Serrano, Migus, Yin, Agoua, Baskiotis, Gallinari, and Guigue]{naour2023time}
Etienne Le~Naour, Louis Serrano, L{\'e}on Migus, Yuan Yin, Ghislain Agoua, Nicolas Baskiotis, Patrick Gallinari, and Vincent Guigue.
\newblock Time series continuous modeling for imputation and forecasting with implicit neural representations.
\newblock \emph{Transactions on Machine Learning Research}, 2024.
\newblock ISSN 2835-8856.
\newblock URL \url{https://openreview.net/forum?id=P1vzXDklar}.

\bibitem[Le~Naour et~al.(2026)Le~Naour, Nabil, Petralia, and Agoua]{TSFMImputationBenchmark}
Etienne Le~Naour, Tahar Nabil, Adrien Petralia, and Ghislain Agoua.
\newblock Are time-indexed foundation models the future of time series imputation?
\newblock \emph{Transactions on Machine Learning Research}, 2026.
\newblock ISSN 2835-8856.
\newblock URL \url{https://openreview.net/forum?id=cTk56KpsP5}.

\bibitem[Liu et~al.(2025{\natexlab{a}})Liu, Aksu, Liu, Liu, Yan, Pham, Savarese, Sahoo, Xiong, and Li]{liu2025moirai}
Chenghao Liu, Taha Aksu, Juncheng Liu, Xu~Liu, Hanshu Yan, Quang Pham, Silvio Savarese, Doyen Sahoo, Caiming Xiong, and Junnan Li.
\newblock Moirai 2.0: When less is more for time series forecasting.
\newblock \emph{arXiv preprint arXiv:2511.11698}, 2025{\natexlab{a}}.

\bibitem[Liu et~al.(2025{\natexlab{b}})Liu, Qin, Shi, Chen, Yang, Huang, Wang, and Long]{liu2025sundial}
Yong Liu, Guo Qin, Zhiyuan Shi, Zhi Chen, Caiyin Yang, Xiangdong Huang, Jianmin Wang, and Mingsheng Long.
\newblock Sundial: A family of highly capable time series foundation models.
\newblock In \emph{Forty-second International Conference on Machine Learning}, 2025{\natexlab{b}}.
\newblock URL \url{https://openreview.net/forum?id=LO7ciRpjI5}.

\bibitem[Mildenhall et~al.(2021)Mildenhall, Srinivasan, Tancik, Barron, Ramamoorthi, and Ng]{mildenhall2021nerf}
Ben Mildenhall, Pratul~P Srinivasan, Matthew Tancik, Jonathan~T Barron, Ravi Ramamoorthi, and Ren Ng.
\newblock Nerf: Representing scenes as neural radiance fields for view synthesis.
\newblock \emph{Communications of the ACM}, 65\penalty0 (1):\penalty0 99--106, 2021.

\bibitem[Moroshan et~al.(2025)Moroshan, Siems, Zela, Carstensen, and Hutter]{moroshan2025tempopfn}
Vladyslav Moroshan, Julien Siems, Arber Zela, Timur Carstensen, and Frank Hutter.
\newblock Tempo{PFN}: Towards synthetic pre-training of linear {RNN}s for zero-shot time series forecasting.
\newblock In \emph{EurIPS 2025 Workshop: AI for Tabular Data}, 2025.
\newblock URL \url{https://openreview.net/forum?id=Iqex1gfnvc}.

\bibitem[Nie et~al.(2023)Nie, Nguyen, Sinthong, and Kalagnanam]{PatchTST}
Yuqi Nie, Nam~H. Nguyen, Phanwadee Sinthong, and Jayant Kalagnanam.
\newblock A time series is worth 64 words: Long-term forecasting with transformers.
\newblock In \emph{International Conference on Learning Representations, ICLR}, 2023.

\bibitem[Peters et~al.(2017)Peters, Janzing, and Scholkopf]{peters2017elements}
Jonas Peters, Dominik Janzing, and Bernhard Scholkopf.
\newblock \emph{Elements of causal inference: foundations and learning algorithms}.
\newblock MIT press, 2017.

\bibitem[Qiao et~al.(2026)Qiao, Pan, Wang, Zhukova, Liu, Jiang, Wen, Long, Jin, and Liu]{qiao2026sTIME}
Zhongzheng Qiao, Sheng Pan, Anni Wang, Viktoriya Zhukova, Yong Liu, Xudong Jiang, Qingsong Wen, Mingsheng Long, Ming Jin, and Chenghao Liu.
\newblock It's {TIME}: Towards the next generation of time series forecasting benchmarks.
\newblock \emph{arXiv preprint arXiv:2602.12147}, 2026.

\bibitem[Qu et~al.(2026)Qu, Holzm{\"u}ller, Varoquaux, and Morvan]{qu2026tabiclv2}
Jingang Qu, David Holzm{\"u}ller, Ga{\"e}l Varoquaux, and Marine~Le Morvan.
\newblock {TabICLv2}: A better, faster, scalable, and open tabular foundation model.
\newblock \emph{arXiv preprint arXiv:2602.11139}, 2026.

\bibitem[Rubanova et~al.(2019)Rubanova, Chen, and Duvenaud]{latent_ode2019}
Yulia Rubanova, Ricky T.~Q. Chen, and David Duvenaud.
\newblock Latent odes for irregularly-sampled time series.
\newblock In \emph{Proceedings of the 33rd International Conference on Neural Information Processing Systems}, Red Hook, NY, USA, 2019. Curran Associates Inc.

\bibitem[Schulz \& Stattegger(1997)Schulz and Stattegger]{schulz1997spectrum}
Michael Schulz and Karl Stattegger.
\newblock Spectrum: Spectral analysis of unevenly spaced paleoclimatic time series.
\newblock \emph{Computers \& Geosciences}, 23\penalty0 (9):\penalty0 929--945, 1997.

\bibitem[Serrano et~al.(2024)Serrano, Wang, Le~Naour, Vittaut, and Gallinari]{serrano2024aroma}
Louis Serrano, Thomas~X Wang, Etienne Le~Naour, Jean-No{\"e}l Vittaut, and Patrick Gallinari.
\newblock Aroma: Preserving spatial structure for latent pde modeling with local neural fields.
\newblock \emph{Advances in Neural Information Processing Systems}, 37:\penalty0 13489--13521, 2024.

\bibitem[Shchur et~al.(2025)Shchur, Ansari, Turkmen, Stella, Erickson, Guerron, Bohlke-Schneider, and Wang]{shchur2025fev}
Oleksandr Shchur, Abdul~Fatir Ansari, Caner Turkmen, Lorenzo Stella, Nick Erickson, Pablo Guerron, Michael Bohlke-Schneider, and Yuyang Wang.
\newblock fev-bench: A realistic benchmark for time series forecasting.
\newblock \emph{arXiv preprint arXiv:2509.26468}, 2025.

\bibitem[Steinwart \& Christmann(2011)Steinwart and Christmann]{steinwart2011estimating}
Ingo Steinwart and Andreas Christmann.
\newblock Estimating conditional quantiles with the help of the pinball loss.
\newblock \emph{Bernoulli}, 17\penalty0 (1), 2011.
\newblock \doi{10.3150/10-BEJ267}.

\bibitem[Taylor \& Letham(2018)Taylor and Letham]{taylor2018forecasting}
Sean~J Taylor and Benjamin Letham.
\newblock Forecasting at scale.
\newblock \emph{The American Statistician}, 72\penalty0 (1):\penalty0 37--45, 2018.

\bibitem[Von~Oswald et~al.(2023)Von~Oswald, Niklasson, Randazzo, Sacramento, Mordvintsev, Zhmoginov, and Vladymyrov]{vanoswald2023transformers}
Johannes Von~Oswald, Eyvind Niklasson, Ettore Randazzo, Jo\~{a}o Sacramento, Alexander Mordvintsev, Andrey Zhmoginov, and Max Vladymyrov.
\newblock Transformers learn in-context by gradient descent.
\newblock In \emph{Proceedings of the 40th International Conference on Machine Learning}, 2023.

\bibitem[Woo et~al.(2023)Woo, Liu, Sahoo, Kumar, and Hoi]{DeepTime}
Gerald Woo, Chenghao Liu, Doyen Sahoo, Akshat Kumar, and Steven Hoi.
\newblock Learning deep time-index models for time series forecasting.
\newblock In \emph{International Conference on Machine Learning}, pp.\  37217--37237. PMLR, 2023.

\bibitem[Woo et~al.(2024)Woo, Liu, Kumar, Xiong, Savarese, and Sahoo]{woo2024unified}
Gerald Woo, Chenghao Liu, Akshat Kumar, Caiming Xiong, Silvio Savarese, and Doyen Sahoo.
\newblock Unified training of universal time series forecasting transformers.
\newblock In \emph{Forty-first International Conference on Machine Learning}, 2024.

\bibitem[Wu et~al.(2026)Wu, Teng, Li, Zhang, Duan, and Li]{wu2026out}
Xin Wu, Fei Teng, Xingwang Li, Ji~Zhang, Qiang Duan, and Tianrui Li.
\newblock Out-of-distribution generalization in time series: A survey.
\newblock \emph{Information Fusion}, pp.\  104336, 2026.

\bibitem[Xie et~al.(2026)Xie, Feofanov, Zhang, Palpanas, and Redko]{xie2025cauker}
Shifeng Xie, Vasilii Feofanov, Jianfeng Zhang, Themis Palpanas, and Ievgen Redko.
\newblock Cauker: Classification time series foundation models can be pretrained on synthetic data.
\newblock In \emph{The Fourteenth International Conference on Learning Representations}, 2026.
\newblock URL \url{https://openreview.net/forum?id=xBW2FIfswU}.

\end{thebibliography}
\bibliographystyle{tmlr}

\clearpage
\appendix
\appendixpage
\addappheadtotoc
\startcontents[sections]
\printcontents[sections]{l}{1}{\setcounter{tocdepth}{2}}

\clearpage

\section{\texttt{TS-ICL} Architecture Details}
\label{appendix-archi}

This section provides a detailed breakdown of the \texttt{TS-ICL} framework introduced in \cref{sec:model-univar}.
Casting both imputation and forecasting tasks as in-context regression problems over learned temporal representation, \texttt{TS-ICL} consists in four successive modules that transform raw observations into global and local context-aware representations used for prediction.


\subsection{The Time Series Encoder $\mathcal{E}$}
\label{sec:encoder-details}

\begin{figure}[h!]
    \centering
    \includegraphics[width=0.89\linewidth]{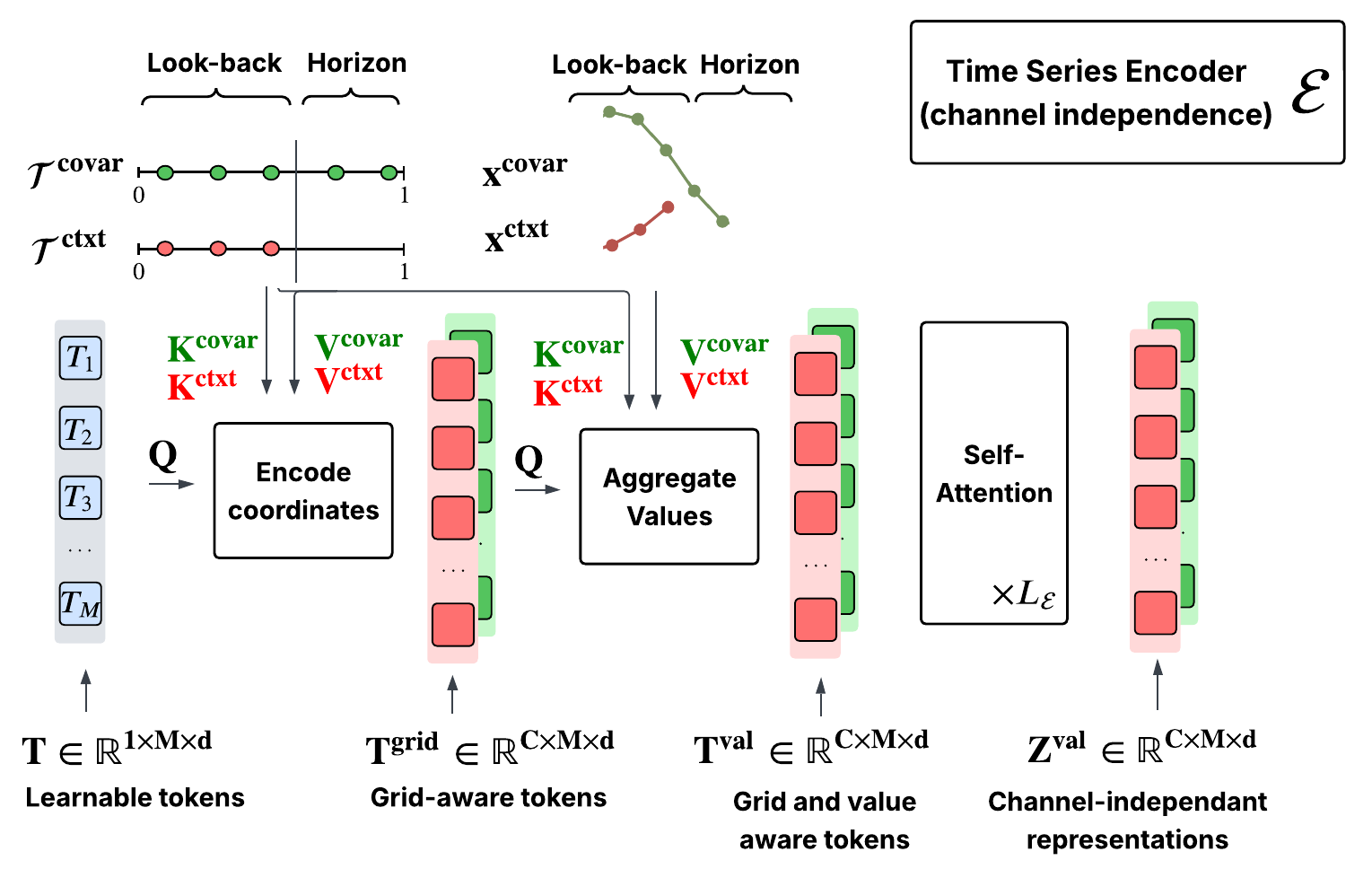}
    \caption{
    Overview of the Time Series Encoder $\mathcal{E}$.
    Forecasting task shown for illustration.}
    \label{fig:encoder}
\end{figure}

\paragraph{Encoder Overview.}

\begin{figure}[h!]
    \centering
    \includegraphics[width=0.75\linewidth]{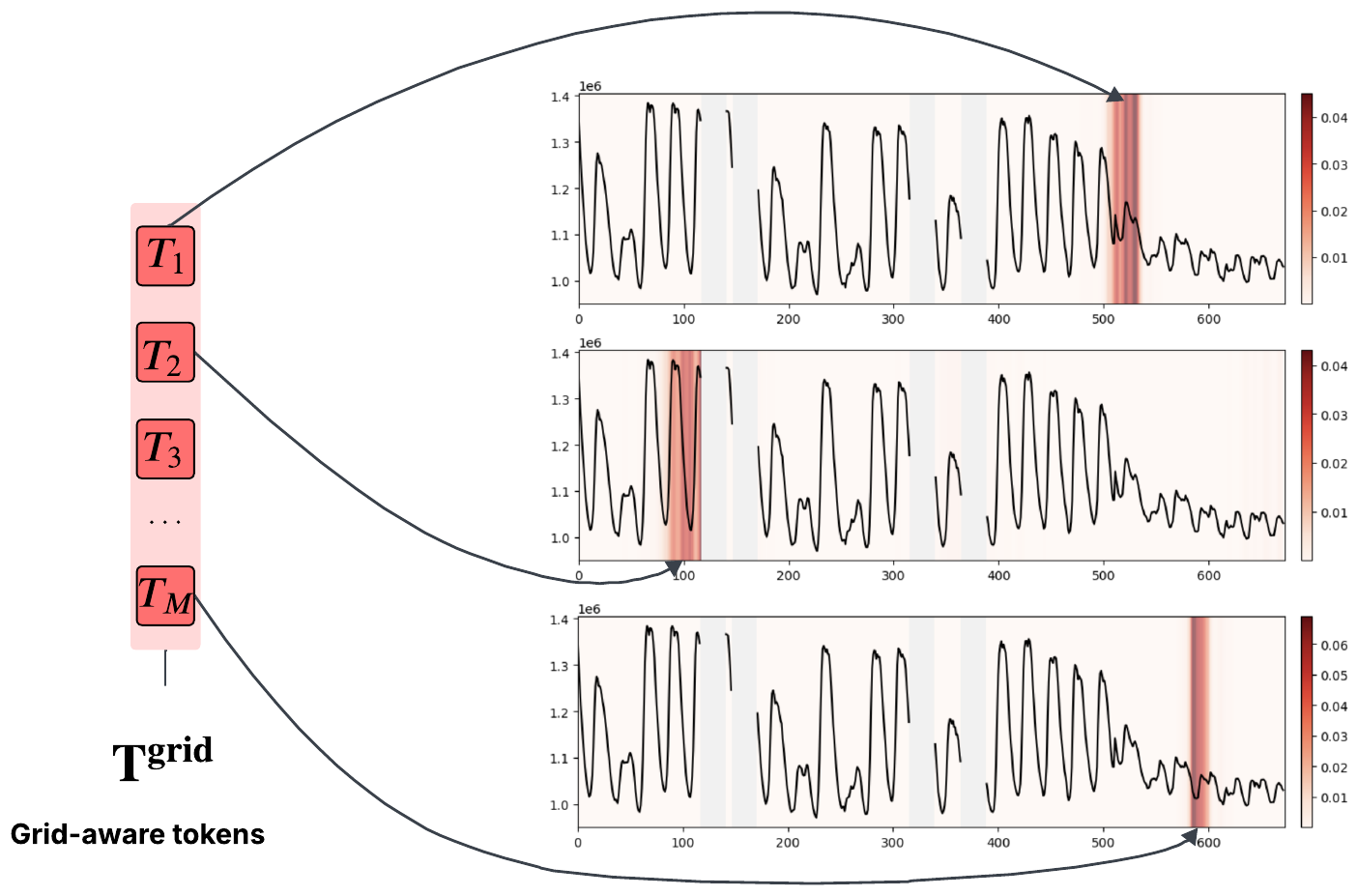}
    \caption{Temporal interpretation of tokens via cross-attention between $\mathbf{T}^{grid}$ and $\gamma(t)$ for each $t \in \mathcal{T}^{obs}$. Cross-attention maps for three distinct tokens are shown for a given attention head.}
    \label{fig:token-vizu}
\end{figure}

The encoder $\mathcal{E}$ maps the observed context $(\mathcal{T}^{\mathrm{ctxt}}, \boldsymbol{x}^{\mathrm{ctxt}})$ and $C-1$ optional covariates $(\mathcal{T}^{\mathrm{covar}}, \boldsymbol{X}^{\mathrm{covar}})$, where $C\geq1$, jointly into a channel-independent latent representation $\boldsymbol{Z}^{\mathrm{val}} \in \mathbb{R}^{C \times M \times d}$.
Shown in \Cref{fig:encoder}, the module operates through the following steps:

\begin{enumerate}[(i)]

\item \textbf{Temporal Encoding.}  
The timestamps from both the context $\mathcal{T}^{\mathrm{ctxt}}$ and covariates $\mathcal{T}^{\mathrm{covar}}$ are independently mapped into higher-dimensional representations using Fourier features \citep{mildenhall2021nerf} followed by a linear projection:
\[
\mathcal{T} 
\xrightarrow{\text{Fourier + Linear}} 
\gamma(\mathcal{T}) \in \mathbb{R}^{T \times d}.
\]

\item \textbf{Coordinate Encoding.}  
A set of $M$ learnable latent tokens $\boldsymbol{T} \in \mathbb{R}^{1 \times M \times d}$ serves as a query ($Q$) and attends to the temporal embeddings of all channels (context and covariates) through cross-attention:
\[
\boldsymbol{T}^{\mathrm{grid}} 
=
\mathrm{CrossAttn}\!\left(
Q=\boldsymbol{T},\;
K=V=\gamma(\mathcal{T})
\right)
\in \mathbb{R}^{C \times M \times d}.
\]
This step produces \textit{grid-aware tokens} that capture the geometric structure of the sampling grid for each channel.

\item \textbf{Value Aggregation.}  
The observed values $(\boldsymbol{x}^{\mathrm{ctxt}}, \boldsymbol{X}^{\mathrm{covar}})$ are first projected into the latent dimension (\textit{value lifting}).
The grid-aware tokens $\boldsymbol{T}^{\mathrm{grid}}$ then attend to these value embeddings (see \cref{fig:token-vizu} for a temporal interpretation):
\[
\boldsymbol{T}^{\mathrm{val}}
=
\mathrm{CrossAttn}\!\left(
Q=\boldsymbol{T}^{\mathrm{grid}},\;
K=\gamma(\mathcal{T}),\;
V=\mathrm{lift}(\boldsymbol{X})
\right)
\in \mathbb{R}^{C \times M \times d}.
\]
This step integrates the specific observed values into the latent representations, resulting in \textit{grid and value aware tokens}.

\item \textbf{Latent Refinement.}  
Finally, $L_{\mathcal{E}}$ layers of channel-independent self-attention are applied to refine the representations:
\[
\boldsymbol{Z}^{\mathrm{val}}
=
\mathrm{Transformer}_L(\boldsymbol{T}^{\mathrm{val}})
\in \mathbb{R}^{C \times M \times d}.
\]
This produces the final latent representations $\boldsymbol{Z}^{\mathrm{val}}$, where each channel has been compressed into $M$ informative tokens.

\end{enumerate}

\subsection{The Channel Mixer $\mathcal{M}$}
\label{sec:channel-mixer-details}

\begin{figure}[h!]
    \centering
    \includegraphics[width=0.99\linewidth]{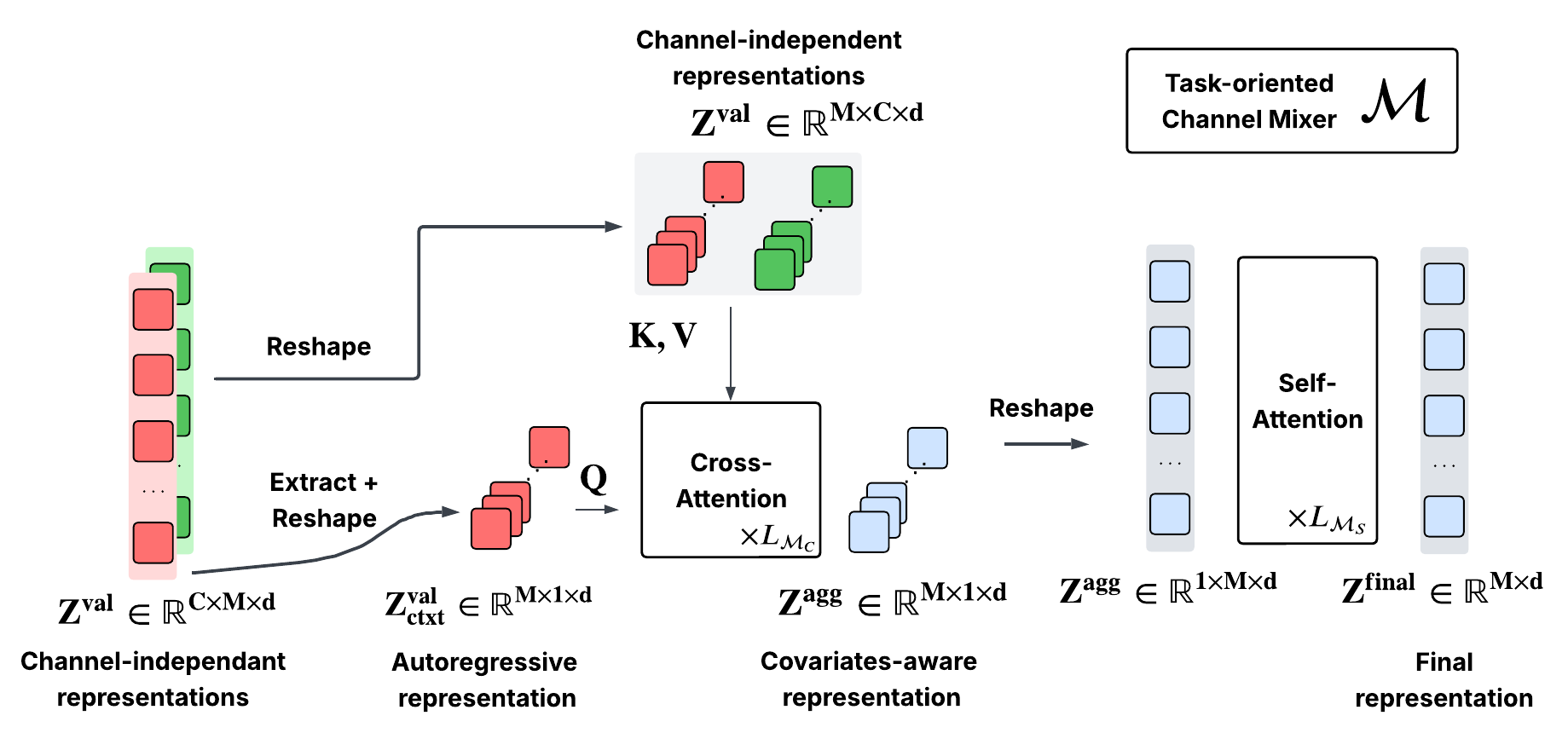}
    \caption{Overview of the Channel Mixer $\mathcal{M}$.
    }
    \label{fig:mixer}
\end{figure}

\paragraph{Module $\mathcal{M}$ Overview.}
The Channel Mixer $\mathcal{M}$ is designed to aggregate information across multiple channels (time series of interest and covariates representations) by conditioning the channel-independent features on the target series context. It transforms a set of independent representations $\boldsymbol{Z}^{\mathrm{val}}$ into a unified, covariate-aware representation $\boldsymbol{Z}^{\mathrm{final}}$.

Shown in \Cref{fig:mixer}, the module follows a three-step process:

\begin{enumerate}[(i)]
\item \textbf{Cross-Channel Attention.} 
The autoregressive context representation $\boldsymbol{Z}_{\mathrm{ctxt}}^{\mathrm{val}} \in \mathbb{R}^{M \times 1 \times d}$, which represents the specific temporal dynamics of the target time series, acts as a query ($Q$). It attends to the channel-independent representations $\boldsymbol{Z}^{\mathrm{val}} \in \mathbb{R}^{M \times C \times d}$ (where $C$ is the number of channels), which serve as keys ($K$) and values ($V$):
\[
\boldsymbol{Z}^{\mathrm{agg}} 
= 
\mathrm{CrossAttn}_L\!\left(
Q=\boldsymbol{Z}_{\mathrm{ctxt}}^{\mathrm{val}},\; 
K=V=\boldsymbol{Z}^{\mathrm{val}}
\right) 
\in \mathbb{R}^{M \times 1 \times d}.
\]
This operation, repeated over $L_{\mathcal{M}_{c}}$ layers, compresses the multi-channel information into a single "covariates-aware" latent representation, effectively selecting the most relevant features from the covariates for the given context.

\item \textbf{Latent Reshaping.} 
To prepare the aggregated representation for global sequence processing, the tensor is reshaped to treat the $M$ latent tokens as a sequence:
\[
\boldsymbol{Z}^{\mathrm{agg}} \in \mathbb{R}^{M \times 1 \times d} 
\quad \xrightarrow{\text{reshape}} \quad 
\boldsymbol{Z}^{\mathrm{agg}} \in \mathbb{R}^{1 \times M \times d}.
\]

\item \textbf{Global Latent Refinement.} 
Finally, $L_{\mathcal{M}_{s}}$ self-attention blocks are applied to the sequence of tokens. This allows the model to capture global dependencies across the aggregated latent space:
\[
\boldsymbol{Z}^{\mathrm{final}} 
= 
\mathrm{Transformer}_L(\boldsymbol{Z}^{\mathrm{agg}}) 
\in \mathbb{R}^{1 \times M \times d}.
\]
The resulting $\boldsymbol{Z}^{\mathrm{final}}$ is the final time series representation, integrating both the local channel information and the global context necessary for the downstream task.
\end{enumerate}


\subsection{The Temporal Context Query Module $\mathcal{C}$}
\label{sec:queryer-details}

\begin{figure}[h!]
    \centering
    \includegraphics[width=0.60\linewidth]{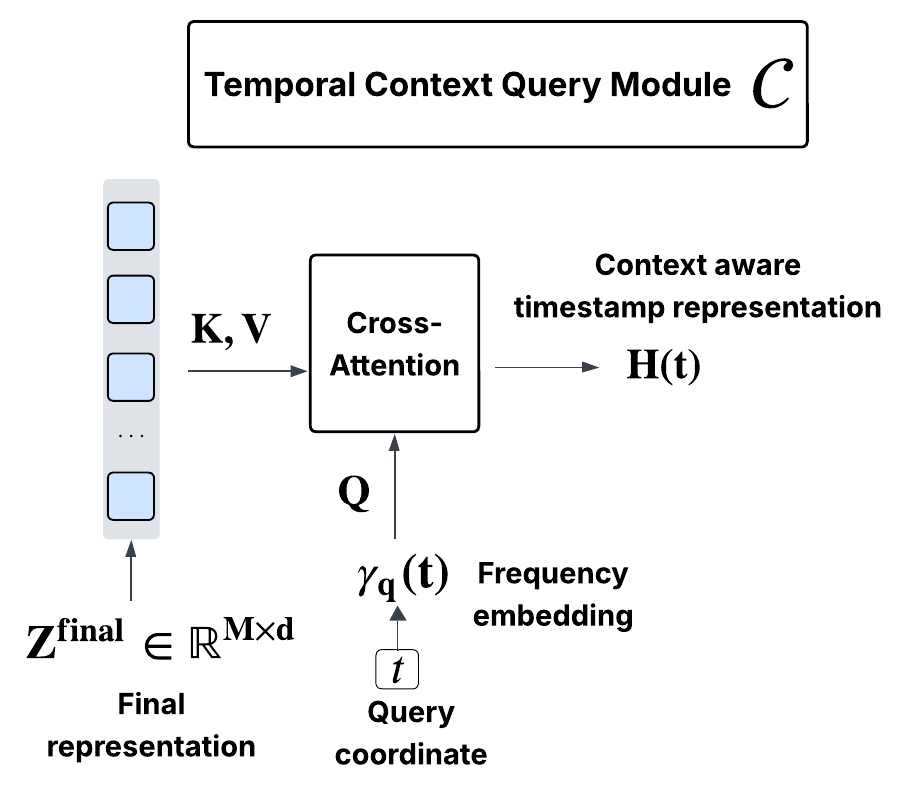}
    \caption{Overview of the Temporal Context Query Module $\mathcal{C}$.
    }
    \label{fig:queryer}
\end{figure}

\paragraph{Module $\mathcal{C}$ Overview.}

The Temporal Context Query Module $\mathcal{C}$ maps the final latent representation 
$\boldsymbol{Z}^{\mathrm{final}} \in \mathbb{R}^{M \times d}$ 
and a query coordinate $t$ to a time-series-aware representation 
$H(t) \in \mathbb{R}^{d}$. 
This module enables continuous-time querying of the encoded context by bridging the gap between the discrete latent tokens and the continuous time domain.

Shown in \Cref{fig:queryer}, the module operates as follows:

\begin{enumerate}[(i)]

\item \textbf{Frequency Encoding.}  
A target timestamp $t \in \mathbb{R}$ is mapped into a higher-dimensional frequency embedding $\gamma_q(t)$ \citep{mildenhall2021nerf}. This encoding uses sinusoidal functions at multiple scales to capture both coarse and fine-grained temporal patterns:
\[
t 
\xrightarrow{\text{Frequency Encoding}} 
\gamma_q(t) \in \mathbb{R}^{d}.
\]

\item \textbf{Contextual Querying via Cross-Attention.}  
The frequency embedding $\gamma_q(t)$ serves as the query ($Q$) in a cross-attention mechanism. It attends to the final time series representation $\boldsymbol{Z}^{\mathrm{final}}$, which provides the keys ($K$) and values ($V$):
\[
H(t)
=
\mathrm{CrossAttn}\!\left(
Q=\gamma_q(t),\;
K=V=\boldsymbol{Z}^{\mathrm{final}}
\right).
\]
This operation extracts a localized, time-series-aware representation from the global latent context, specifically conditioned on the query coordinate $t$.

\item \textbf{Representation Output.}  
The resulting vector $H(t)$ constitutes the "time-series-aware timestamp representation". It integrates the global context stored in the latent tokens with the specific temporal information of the query, serving as the primary input for the downstream in-context regressor.

\end{enumerate}



\subsection{The In-Context Learning Regressor Module $\mathcal{R}$}
\label{sec:regressor-details}

\paragraph{Module $\mathcal{R}$ Overview.}
The In-Context Learning Regressor $\mathcal{R}$ is the final component of the architecture. It treats both forecasting and imputation as in-context regression tasks, where the model learns to map target representations to values by conditioning on observed "input-output" pairs \citep{brown2020language,garg2022what,vanoswald2023transformers}. $\mathcal{R}$ leverages a specific token construction mechanism to align context-aware embeddings with raw covariates.

\begin{figure} 
    \centering
    \begin{subfigure}[b]{0.43\textwidth}
        \centering
        \includegraphics[width=\linewidth]{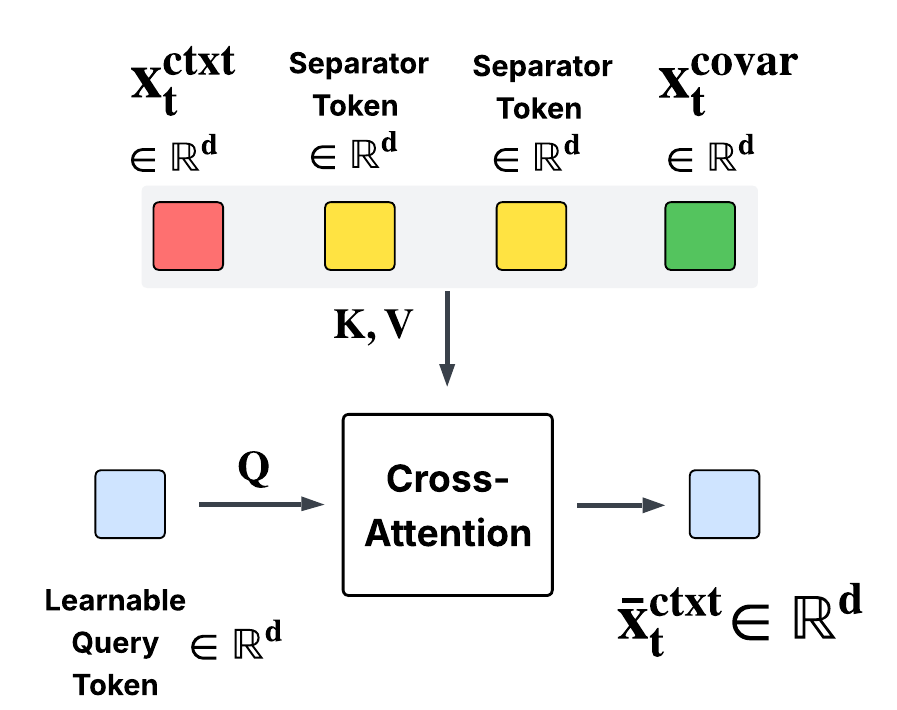}
        \caption{Build context tokens $\mathbf{\bar x^{ctxt}_{t}}$ for $t \in \mathcal{T}^{cxt}$.}
        \label{fig:build-x_ctxt}
    \end{subfigure}
    \hfill
    \begin{subfigure}[b]{0.43\textwidth}
        \centering
        \includegraphics[width=\linewidth]{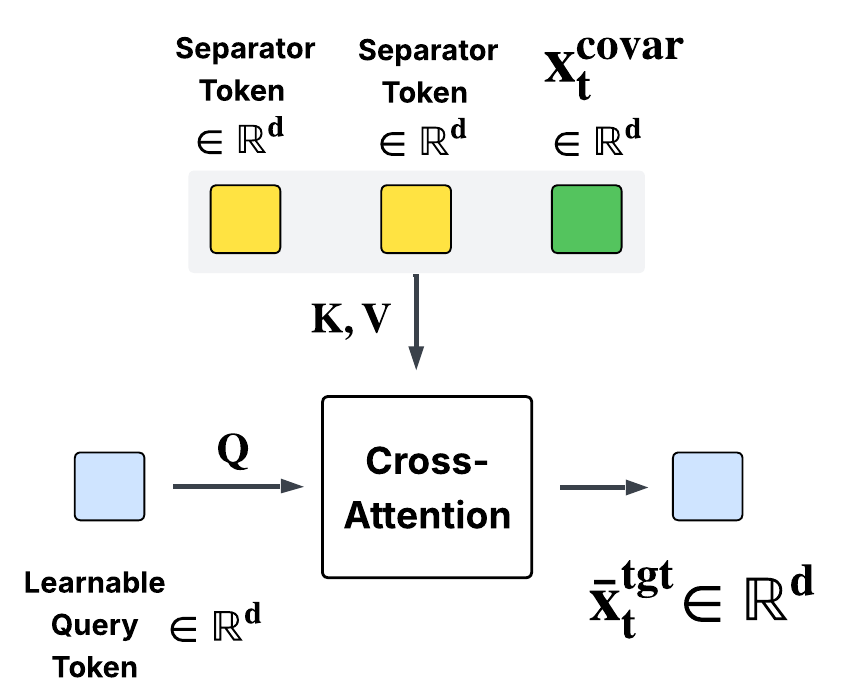}
        \caption{Build target tokens $\mathbf{\bar x^{tgt}_{t}}$ for $t \in \mathcal{T}^{tgt}$.}
        \label{fig:build-x_tgt}
    \end{subfigure}
    \caption{Cross-Attention Mechanism for Context and Target Token Construction.}
    \label{fig:build-tokens-icl}
\end{figure}

\paragraph{Input Projection and Token Construction via Cross-Attention.}
As depicted in \cref{fig:build-tokens-icl}, prior to token construction, all raw inputs (including the observed values $\mathbf{x}_t^{\text{ctxt}}$ and covariates $\mathbf{x}_t^{\text{covar}}$) are linearly projected into a common $d$-dimensional latent space $\mathbb{R}^d$.
A Cross-Attention mechanism then fuses these projected representations with a learnable query token $Q\in\mathbb{R}^d$.

Notably, the covariate input $\mathbf{x}_t^{\text{covar}}$ is strictly optional and can consist of zero, one, or multiple distinct covariates. The Cross-Attention mechanism  accommodates this variability: since attention operates over sets, varying the number of covariates simply changes the sequence length of the Keys and Values, requiring no architectural modifications.
This allows the regressor $\mathcal{R}$, and thus \texttt{TS-ICL}, to operate on covariate grids $\mathcal{T}^{\mathrm{covar}}_{c}$ unaligned with the context grid $\mathcal{T}^{\mathrm{ctxt}}$.

\begin{itemize}
    \item \textbf{Context Tokens ($\bar{\mathbf{x}}_t^{\text{ctxt}}$):} For $t \in \mathcal{T}^{\text{ctxt}}$, the model groups the projected observed value $\mathbf{x}_t^{\text{ctxt}}$ with a set of learned separator tokens and the projected covariates $\mathbf{x}_t^{\text{covar}}$ (if any). These act as Keys ($K$) and Values ($V$) for the learnable Query token $Q$, resulting in a unified context representation $\bar{\mathbf{x}}_t^{\text{ctxt}} \in \mathbb{R}^d$.
    \item \textbf{Target Tokens ($\bar{\mathbf{x}}_t^{\text{tgt}}$):} For $t \in \mathcal{T}^{\text{tgt}}$, the ground truth value is unknown. The target token $\bar{\mathbf{x}}_t^{\text{tgt}} \in \mathbb{R}^d$ is similarly constructed by applying Cross-Attention between the learnable Query $Q$ and the available information: the covariate embeddings $\mathbf{x}_t^{\text{covar}}$ and the separator tokens.
\end{itemize}

\begin{figure}
    \centering
    \includegraphics[width=0.99\linewidth]{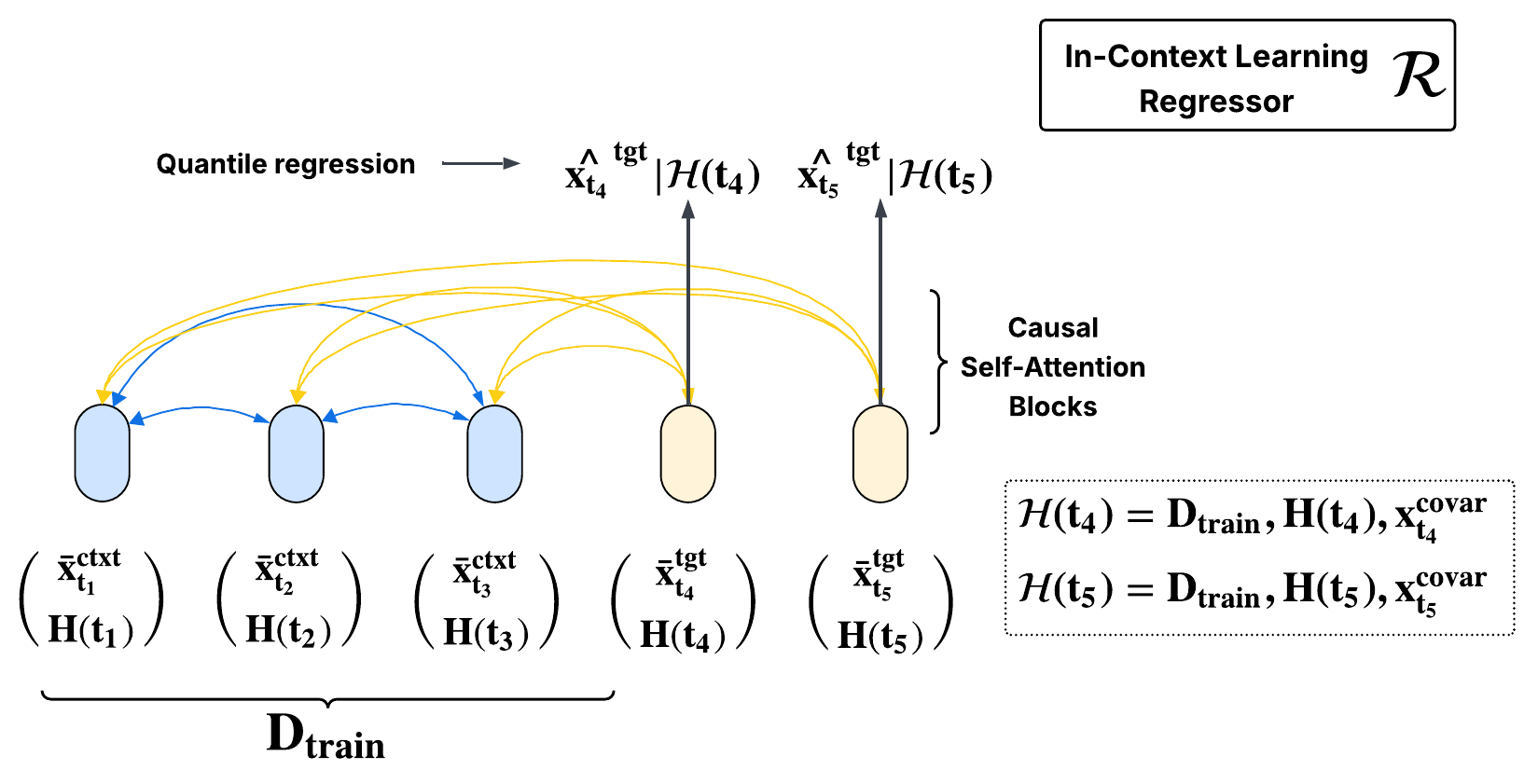}
    \caption{Overview of the In-Context Learning Regressor Module $\mathcal{R}$. Forecasting task shown for illustration.}
    \label{fig:regressor}
\end{figure}

\paragraph{In-Context Input Sequence.}
The regressor processes a sequence $\mathbf{S}$ organized to facilitate relational learning (\cref{fig:regressor}). The sequence consists of paired context tokens followed by target queries:
$$
\mathbf{S} = \big[\, \underbrace{(\bar{\mathbf{x}}_{t_1}^{\text{ctxt}}, \mathbf{H}(t_1)), \dots, (\bar{\mathbf{x}}_{t_n}^{\text{ctxt}}, \mathbf{H}(t_n))}_{\mathcal{D}_{\mathrm{train}}},\, (\bar{\mathbf{x}}_{t_{n+1}}^{\text{tgt}}, \mathbf{H}(t_{n+1})), \dots \,\big]
$$
where $\mathbf{H}(t)$ represents the context-aware temporal embedding.
All $(\bar{\mathbf{x}}_{t}^{\text{ctxt}}, \mathbf{H}(t)$ pairs are summed to form the final regressor input sequence in the $d-$dimensional latent space.

\paragraph{Causal In-Context Regression.}
The sequence $\mathbf{S}$ is processed by $L$ layers of causal self-attention blocks. The causal mask is critical as it ensures a specific information flow:
\begin{itemize}
    \item Each target token $(\bar{\mathbf{x}}_{t_j}^{\text{tgt}}, \mathbf{H}(t_j))$ attends to all previous pairs in $\mathcal{D}_{\mathrm{train}}$ to infer the underlying mapping $\mathbf{H}(t) \to \mathbf{x}(t)$.
    \item The attention mechanism allows the model to dynamically weigh past observations based on their similarity to the current target query in the representation space, without attending to future target values.
\end{itemize}

\paragraph{Quantile Prediction and Loss.}
To capture predictive uncertainty, for each target timestamp $t_j \in \mathcal{T}^{\text{tgt}}$, the model outputs 99 quantiles $\hat{\boldsymbol{q}}(t_j) = (\hat{q}_{\alpha_k}(t_j))_{k=1}^{99}$ via a linear projection of the final hidden states. The model is trained by minimizing a Smooth Pinball Loss:
$$
\mathcal{L} = \sum_{t_j \in \mathcal{T}^{\mathrm{tgt}}} \sum_{k=1}^{99} \left[ \alpha_k e_{j,k} + \beta \log\Big(1 + \exp(-e_{j,k}/\beta)\Big) \right]
$$
where $e_{j,k} = x(t_j) - \hat{q}_{\alpha_k}(t_j)$, $\alpha_k \in (0,1)$ is the quantile level, and $\beta > 0$ is a smoothing parameter. In practice, we set $\beta = 0.01$. During training, gradients are only backpropagated through the target predictions; the context set $\mathcal{D}_{\mathrm{train}}$ is treated strictly as conditioning data and does not contribute to the loss.
\clearpage

\clearpage
\section{Training Details}
\label{training-details}

This appendix provides detailed information about the training procedure of \texttt{TS-ICL}.
\Cref{app:training_prior} covers the data aspect, from pretraining corpus to prior generation, whereas \Cref{appendix:hyperparameters} describes the set of hyperparameters used to instantiate and train \texttt{TS-ICL}.

\subsection{Training Prior} \label{app:training_prior}

\subsubsection{Univariate Time Series Datasets.}
\label{subsec:univar-real-ts}

The univariate pretraining datasets of \texttt{TS-ICL} originate from three main sources, namely:
\begin{enumerate*}[label=(\roman*)]
\item \texttt{LOTSA} \citep{woo2024unified};
\item \texttt{Chronos} training data \citep{Chronosv1} and
\item \texttt{TempoPFN} synthetic data \citep{moroshan2025tempopfn}.
The latter includes in particular the synthetic generators from \texttt{ForecastPFN} \citep{dooley2023forecastpfn}, \texttt{Chronos} \citep{Chronosv1} and \texttt{CauKer} \citep{xie2025cauker}.
\end{enumerate*}
Overall, the pretraining corpus comprises 40 datasets, listed in \cref{tab:pretraining-dataset} with their key features.

\begin{table}[h!]
\centering
\caption{
All 40 univariate time series datasets used to pretrain \texttt{TS-ICL} and their key properties.
The weight column reports the down- or upsampling coefficient applied dynamically at each epoch.
$^{\dagger}$: offline downsampling from the original dataset.
}
\resizebox{\textwidth}{!}{
\begin{tabular}{lllrrrrr}
\toprule
\multirow{2}{*}{\textbf{Dataset}} & 
\textbf{Release} &
\multirow{2}{*}{\textbf{Domain}} & 
\multirow{2}{*}{\textbf{Freq}} & 
\textbf{Num.} & 
\textbf{Num.} & 
\textbf{Max} & \multirow{2}{*}{\textbf{Weight}}
\\
& \textbf{Platform} & & & \textbf{Series} & \textbf{Variates} & \textbf{Length} & \\
\midrule
Australian Electricity & \href{https://huggingface.co/datasets/autogluon/chronos_datasets/tree/main}{Chronos} & Energy & 30T & 5 & 1 & 232,272  & 220 \\
BDG-2 Bull             & \href{https://huggingface.co/datasets/Salesforce/lotsa_data}{LOTSA} & Energy & H  & 41  & 1 & 12,280    & 25  \\
BDG-2 Fox              & \href{https://huggingface.co/datasets/Salesforce/lotsa_data}{LOTSA} & Energy & H  & 135 & 1 & 12,280    & 5   \\
BDG-2 Panther          & \href{https://huggingface.co/datasets/Salesforce/lotsa_data}{LOTSA} & Energy & H  & 105 & 1 & 6,132     & 2.5 \\
BuildingsBench900k     & \href{https://huggingface.co/datasets/Salesforce/lotsa_data}{LOTSA} & Energy & H  & 100,000$^{\dagger}$ & 1 & 8,759 & 0.02048 \\
Residential Load Power & \href{https://huggingface.co/datasets/Salesforce/lotsa_data}{LOTSA} & Energy & 1T & 271 & 3 & 614,880   & 1.2 \\
Residential PV Power   & \href{https://huggingface.co/datasets/Salesforce/lotsa_data}{LOTSA} & Energy & 1T & 233 & 3 & 614,880   & 1.5 \\
Wind Farms H           & \href{https://huggingface.co/datasets/autogluon/chronos_datasets/tree/main}{Chronos} & Energy & H & 337 & 1 & 6,148 & 4 \\
Wind Farms D           & \href{https://huggingface.co/datasets/autogluon/chronos_datasets/tree/main}{Chronos} & Energy & D & 337 & 1 & 366 & 2   \\
\midrule

China Air Quality & \href{https://huggingface.co/datasets/Salesforce/lotsa_data}{LOTSA} & Climate & H  & 437   & 6 & 397,335 & 0.3 \\
CMIP6 2000        & \href{https://huggingface.co/datasets/Salesforce/lotsa_data}{LOTSA} & Climate & 6H & 8,192 & 22$^{\dagger}$ & 7,300 & 0.057 \\
ERA5 1989         & \href{https://huggingface.co/datasets/Salesforce/lotsa_data}{LOTSA} & Climate & H  & 8,192 & 15$^{\dagger}$ & 8,736 & 0.085 \\
ERA5 1990         & \href{https://huggingface.co/datasets/Salesforce/lotsa_data}{LOTSA} & Climate & H  & 8,192 & 15$^{\dagger}$ & 8,736 & 0.085 \\
ERA5 1991         & \href{https://huggingface.co/datasets/Salesforce/lotsa_data}{LOTSA} & Climate & H  & 8,192 & 15$^{\dagger}$ & 8,736 & 0.085 \\
Spanish Weather   & \href{https://www.kaggle.com/datasets/nicholasjhana/energy-consumption-generation-prices-and-weather}{Kaggle}  & Climate & H & 5 & 3 & 24,544 & 105 \\
Subseasonal       & \href{https://huggingface.co/datasets/Salesforce/lotsa_data}{LOTSA} & Climate & 1D & 862 & 4 & 16,470 & 0.3 \\
Subseasonal Precipitation & \href{https://huggingface.co/datasets/Salesforce/lotsa_data}{LOTSA} & Climate & 1D & 862 & 1 & 11,323 & 1.2  \\
Weatherbench daily & \href{https://huggingface.co/datasets/autogluon/chronos_datasets/tree/main}{Chronos} & Climate & 1D & 10,000$^{\dagger}$ & 1 & 14,610 & 0.1024 \\

\midrule
Mexico City Bikes & \href{https://huggingface.co/datasets/autogluon/chronos_datasets/tree/main}{Chronos} & Traffic & H & 494 & 1 & 104,449 & 2.5 \\
PEMS04 & \href{https://huggingface.co/datasets/Salesforce/lotsa_data}{LOTSA} & Traffic & 5T & 307 & 3 & 16,992 & 1.2 \\
PEMS07 & \href{https://huggingface.co/datasets/Salesforce/lotsa_data}{LOTSA} & Traffic & 5T & 883 & 1 & 28,224 & 1.2 \\
PEMS08 & \href{https://huggingface.co/datasets/Salesforce/lotsa_data}{LOTSA} & Traffic & 5T & 170 & 3 & 17,856 & 2.1 \\
Q-TRAFFIC & \href{https://huggingface.co/datasets/Salesforce/lotsa_data}{LOTSA} & Traffic & 15T & 45,148 & 1 & 5,856 & 0.024 \\
Taxi (30 Min.) & \href{https://huggingface.co/datasets/autogluon/chronos_datasets/tree/main}{Chronos} & Traffic & 30T & 2,428 & 1 & 1,488 & 0.88 \\
Taxi (Hourly) & \href{https://huggingface.co/datasets/autogluon/chronos_datasets/tree/main}{Chronos} & Traffic & H & 2,428 & 1 & 744 & 0.88 \\
Uber TLC (Hourly) & \href{https://huggingface.co/datasets/autogluon/chronos_datasets/tree/main}{Chronos} & Traffic & H & 262 & 1 & 4,344 & 4 \\
\midrule

Alibaba Cluster Trace 2018  & \href{https://huggingface.co/datasets/Salesforce/lotsa_data}{LOTSA} & Cloud & 5T & 58,409 & 2 & 1,728 & 0.009 \\
Wiki Daily & \href{https://huggingface.co/datasets/autogluon/chronos_datasets/tree/main}{Chronos} & Web & D & 100,000 & 1 & 2,741 & 0.00512 \\
Monash M3 Monthly & \href{https://huggingface.co/datasets/Salesforce/lotsa_data}{LOTSA} & Econ./Fin. & M & 1,428 & 1 & 126 & 0.72 \\
NN5 Weekly & \href{https://huggingface.co/datasets/Salesforce/lotsa_data}{LOTSA} & Econ./Fin. & W & 111 & 1 & 105 & 5 \\
Project Tycho  & \href{https://huggingface.co/datasets/Salesforce/lotsa_data}{LOTSA} & Health & W & 1,258 & 1 & 3,854 & 0.21 \\
\midrule

Anomaly     & \href{https://github.com/magnusross/TempoPFN}{TempoPFN} & Synthetic & - & 5,000     & 1 & 10,000 & 0.0256 \\
ForecastPFN & \href{https://github.com/magnusross/TempoPFN}{TempoPFN} & Synthetic & - & 5,000     & 1 & 10,000 & 1 \\
GP          & \href{https://github.com/magnusross/TempoPFN}{TempoPFN} & Synthetic & - & 5,000     & 1 & 10,000 & 0.4096 \\
Kernel Synth 1M & \href{https://huggingface.co/datasets/autogluon/chronos_datasets/tree/main}{Chronos} & Synthetic & - & 1,000,000 & 1 & 1,024 & 0.001024 \\
Sawtooth    & \href{https://github.com/magnusross/TempoPFN}{TempoPFN} & Synthetic & - & 5,000 & 1 & 10,000 & 0.0512 \\
Sinewave    & \href{https://github.com/magnusross/TempoPFN}{TempoPFN} & Synthetic & - & 5,000 & 1 & 10,000 & 0.1024 \\
Spikes      & \href{https://github.com/magnusross/TempoPFN}{TempoPFN} & Synthetic & - & 5,000 & 1 & 10,000 & 0.0256 \\
Step        & \href{https://github.com/magnusross/TempoPFN}{TempoPFN} & Synthetic & - & 5,000 & 1 & 10,000 & 0.0512 \\
OU          & \href{https://github.com/magnusross/TempoPFN}{TempoPFN} & Synthetic & - & 5,000 & 1 & 10,000 & 0.4096 \\
\bottomrule
\end{tabular}
}
{\footnotesize
}
\label{tab:pretraining-dataset}
\end{table}

\cref{tab:pretraining-dataset} highlights that our selection 
\begin{enumerate*}[label=(\roman*)]
\item spans \emph{multiple domains}, including energy, nature/climate, transport, cloud, health/economics;
\item covers a broad range of \emph{frequencies}, from minutely- to weekly- and monthly-sampled time series;
\item includes series of \emph{varying context length}, from 126 timesteps to 614k.
\end{enumerate*}
In total, this forms a corpus of about 2M series, strictly non-overlapping with the different benchmarks on which \texttt{TS-ICL} is evaluated (\texttt{fm-imputation}, \texttt{fev-bench} and \texttt{TIME}).

\paragraph{Sampling strategy.}
We adopt a simple three-step stratified sampling strategy to encourage training on maximally diverse patterns.
\begin{enumerate*}[(i)]
\item Very large hourly datasets are downsampled offline, thereby reducing memory footprint:
we use 100k random samples from \emph{BuildingsBench900k}, 10k random samples from \emph{Weatherbench daily} and 15 (resp., 22) representative channels out of 45 (resp. 53) for the \emph{ERA5} (resp., \emph{CMIP6}) datasets.
\item Similarly to \texttt{Moirai} \citep{woo2024unified}, a random subsample or upsample is drawn from each dataset at each training epoch, to avoid biases towards hourly datasets in the energy and climate domains.
The sampling coefficients are reported in \cref{tab:pretraining-dataset}.
\item For each sample, we select a single window drawn at random from the available series length.
\end{enumerate*}

\subsubsection{Covariates Problem Generators.}
\label{subsec:dag-explanations}

In this section, we provide the technical specifications for the synthetic target--covariate(s) generation process described in Section~\ref{sec:data-prior}.

\paragraph{Graph Construction Logic} The generator constructs a fixed Directed Acyclic Graph (DAG) for each time series batch. The construction follows a topological ordering to ensure acyclicity:

\begin{enumerate}[(i)]
    \item \textbf{Node Initialization.} We start with $R$ root nodes, where $R$ is randomly drawn from a geometric distribution (favoring smaller values), and capped at a maximum of $R_{\max}=3$. Each root corresponds to one of the base univariate time series in the pretraining corpus in \cref{tab:pretraining-dataset} (e.g., signals from real-world datasets or \texttt{TempoPFN}).
    \item \textbf{Graph Size.} The total number of generated series $C$ (channels) corresponds to the number of time series that will define the final covariates--target problem. Specifically, the constructed problem will consist of $C-1$ covariate time series and 1 target time series. The value of $C$ is sampled from a shifted exponential distribution to favor smaller, more manageable graphs while allowing for high complexity:
    \[
    C = \min\left( C_{\max}, \lfloor \text{Exp}(\lambda) \rfloor + C_{\min} \right),
    \]
    where we set $C_{\min}=2$, $C_{\max}=20$, and $\lambda=0.8$ by default. The total number of nodes in the graph is $V = 2C + R$.
    \item \textbf{Edge Sampling.} For each non-root node $i \in \{R, \dots, V-1\}$, the number of parents $k_i$ is sampled from a geometric distribution $k_i \sim \text{Geometric}(p)$, clipped to the number of available ancestors. Parents are then sampled uniformly without replacement from the set of all preceding nodes $\{0, \dots, i-1\}$.
    \item \textbf{Operator Assignment.} Each non-root node is assigned a Structural Causal Model (SCM) sampled from the Structural Causal Model registry.
\end{enumerate}

\begin{figure}[t]
\centering

\begin{subfigure}{0.49\textwidth}
    \centering
    \includegraphics[width=\linewidth]{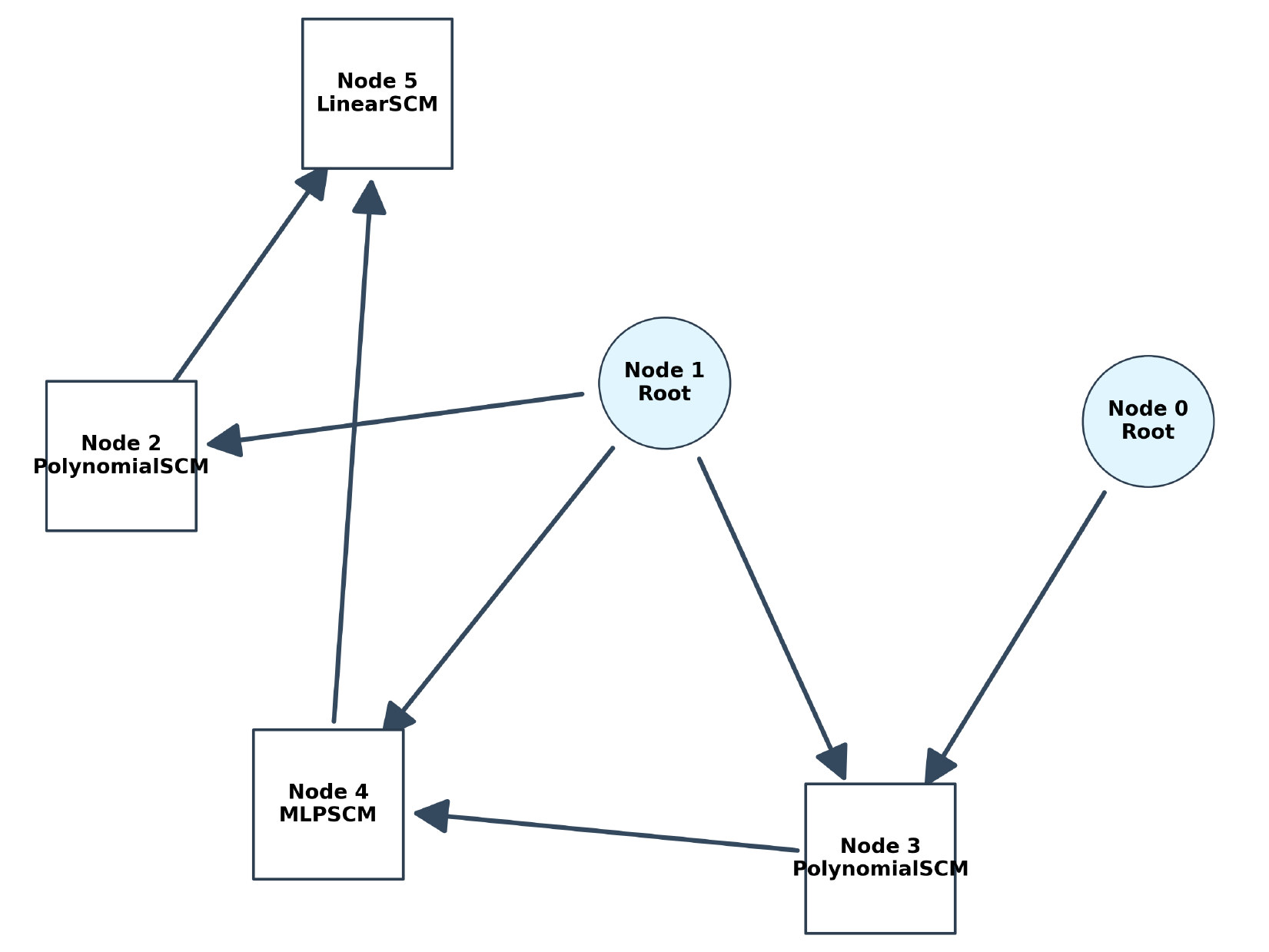}
    \caption{Fully dependent structure (all nodes are connected).}
    \label{fig:dag_dep}
\end{subfigure}
\hfill
\begin{subfigure}{0.49\textwidth}
    \centering
    \includegraphics[width=\linewidth]{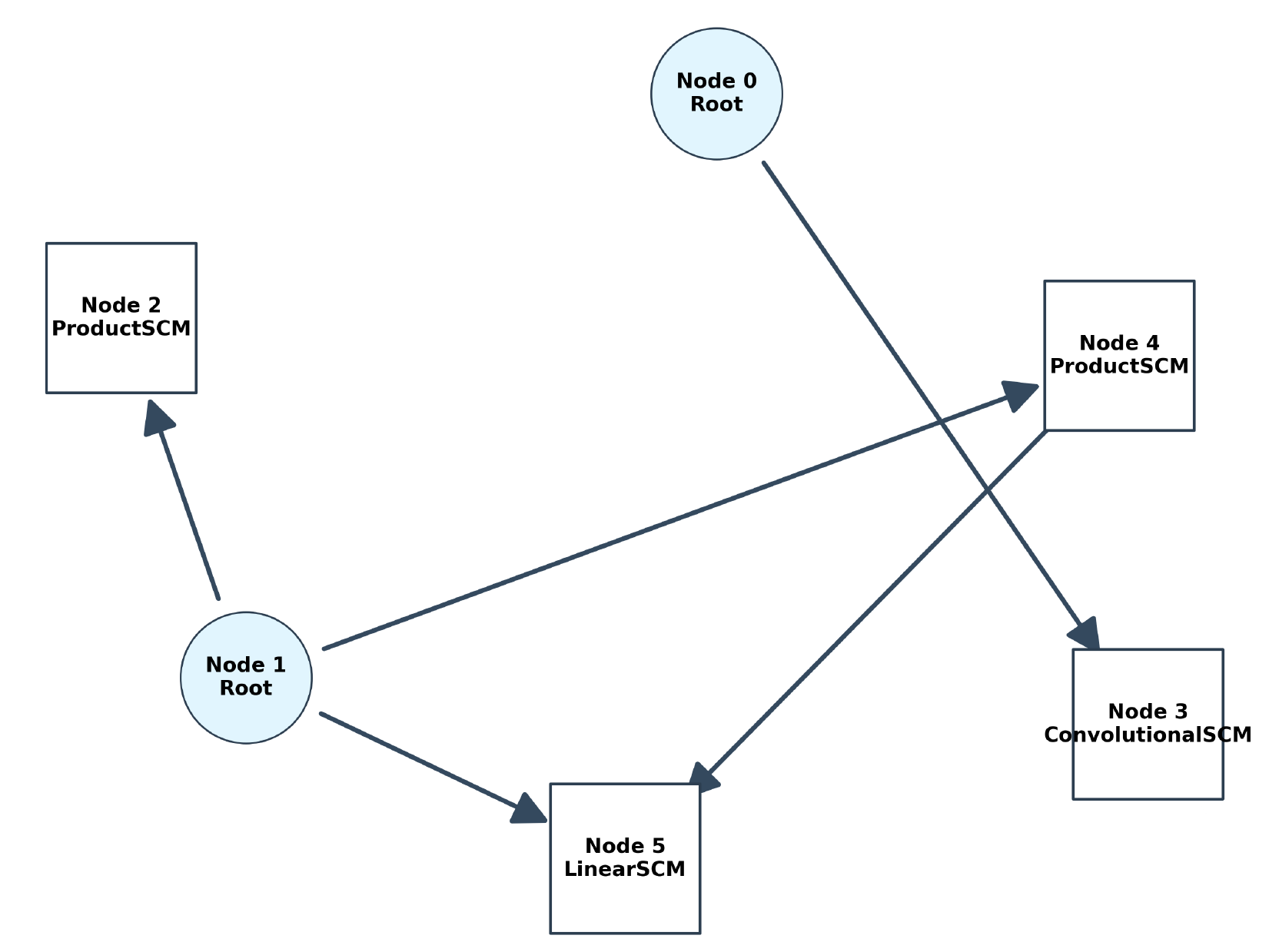}
    \caption{Partially independent structure (not all nodes are connected).}
    \label{fig:dag_indep}
\end{subfigure}

\caption{Examples of randomly generated DAG structures with six nodes. The left graph enforces strong dependencies between nodes, while the right graph allows for partial independence, leading to more diverse causal structures.}
\label{fig:dag_examples}
\end{figure}

This DAG structure allows for the generation of both dependent and independent child nodes, which is essential for constructing realistic target--covariate time series relationships. This design also ensures that covariates are not always informative, thereby encouraging models to learn to ignore irrelevant inputs when appropriate.
As an illustration, \cref{fig:dag_examples}
displays two randomly generated graphs with 6 nodes each, including 2 channels (for one target and one covariate).

\begin{figure} 
\centering

\begin{subfigure}{0.49\textwidth}
\includegraphics[width=\linewidth]{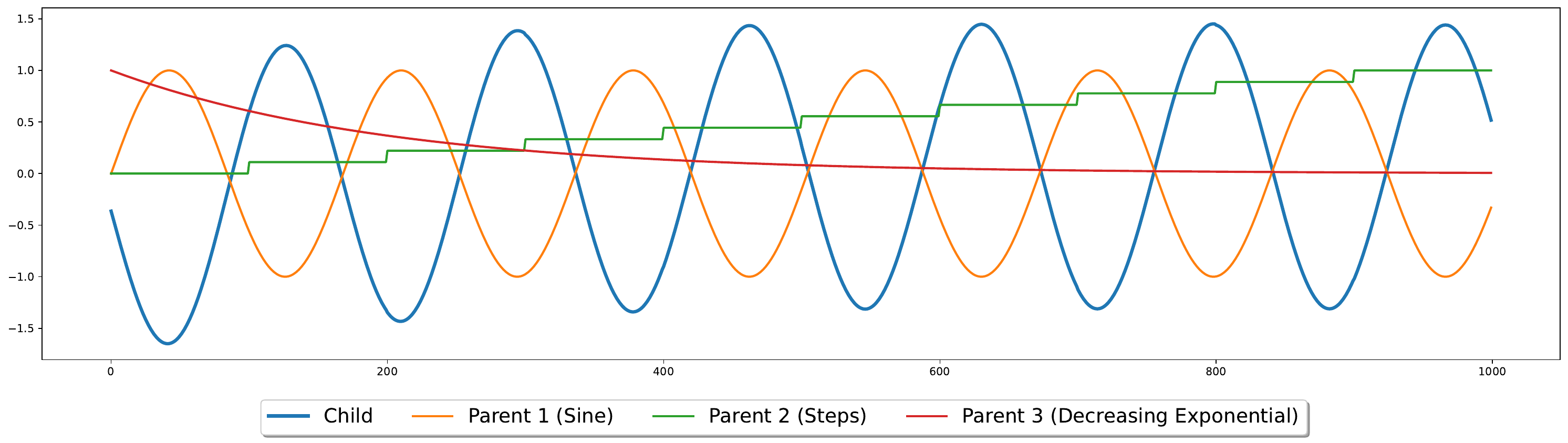}
\caption{Linear}
\end{subfigure}
\hfill
\begin{subfigure}{0.49\textwidth}
\includegraphics[width=\linewidth]{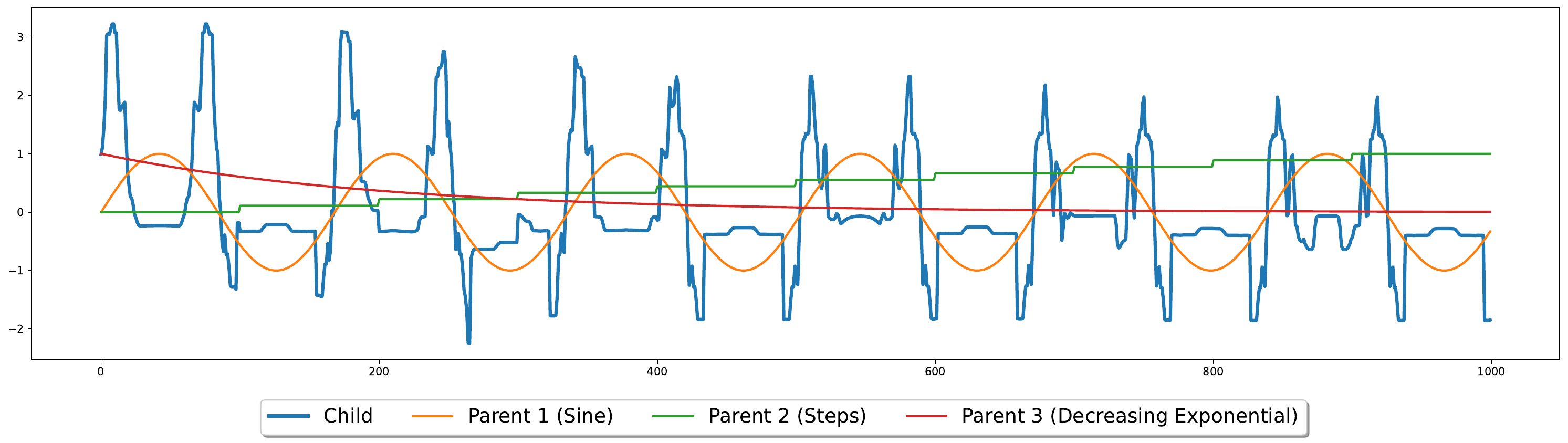}
\caption{MLP}
\end{subfigure}

\begin{subfigure}{0.49\textwidth}
\includegraphics[width=\linewidth]{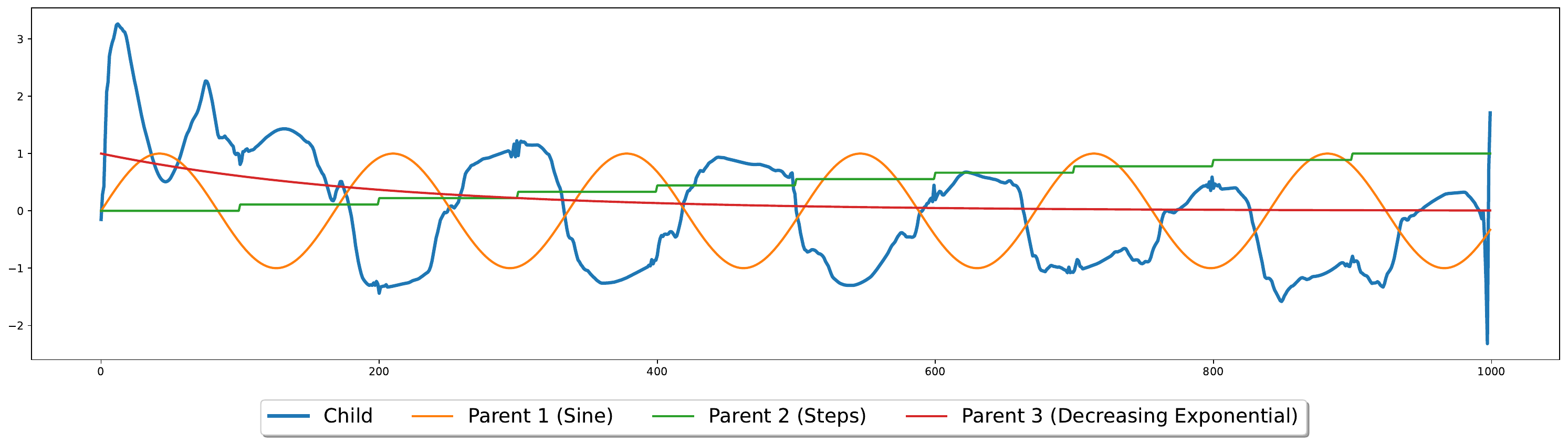}
\caption{Convolutional}
\end{subfigure}
\hfill
\begin{subfigure}{0.49\textwidth}
\includegraphics[width=\linewidth]{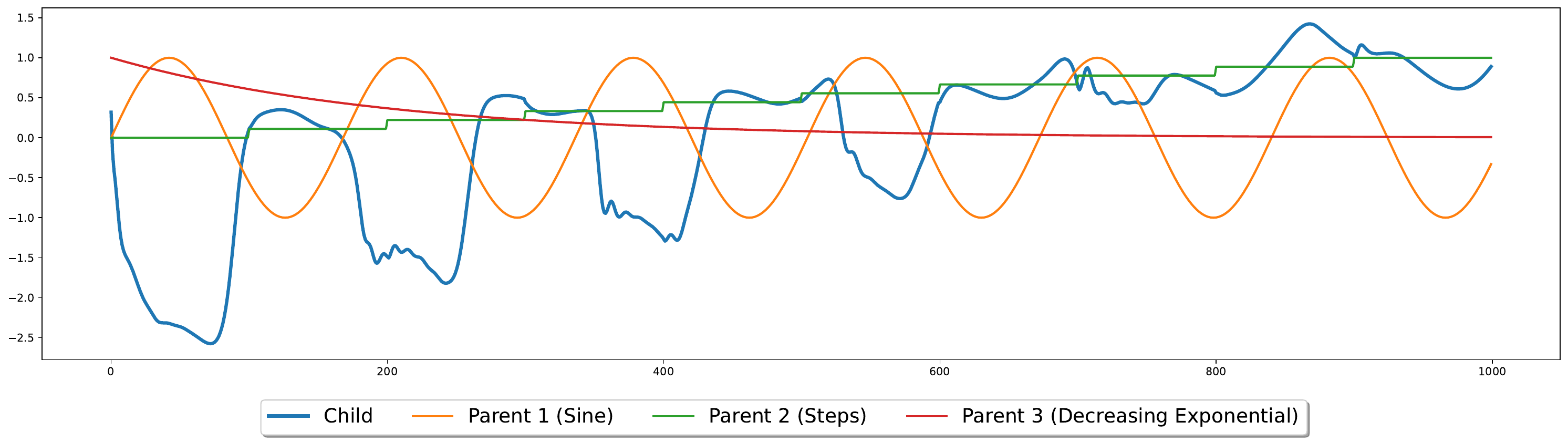}
\caption{RNN}
\end{subfigure}

\begin{subfigure}{0.49\textwidth}
\includegraphics[width=\linewidth]{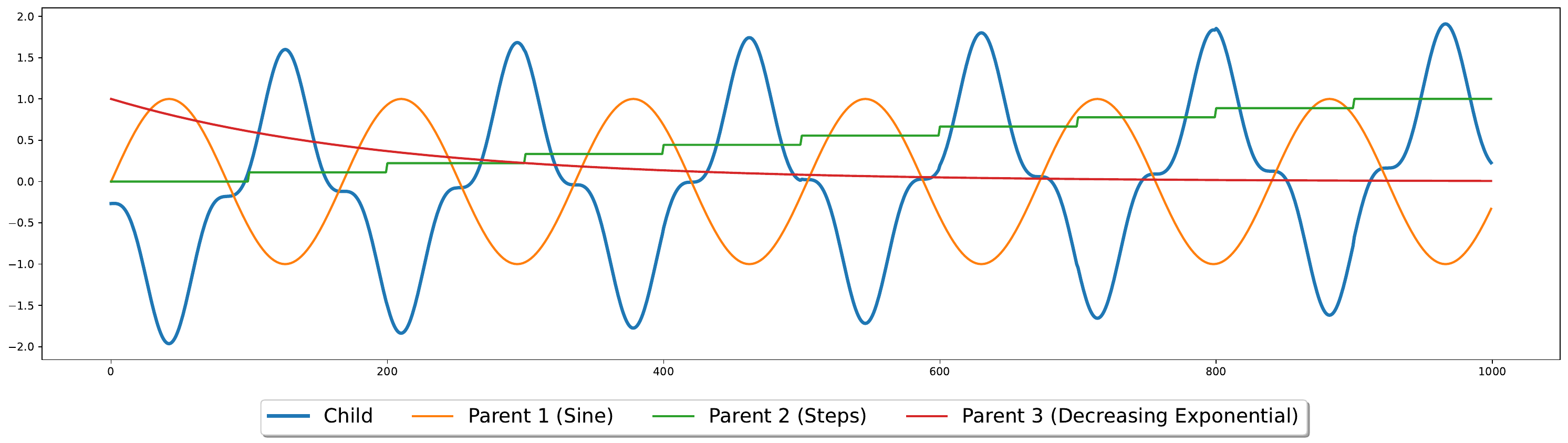}
\caption{Polynomial}
\end{subfigure}
\hfill
\begin{subfigure}{0.49\textwidth}
\includegraphics[width=\linewidth]{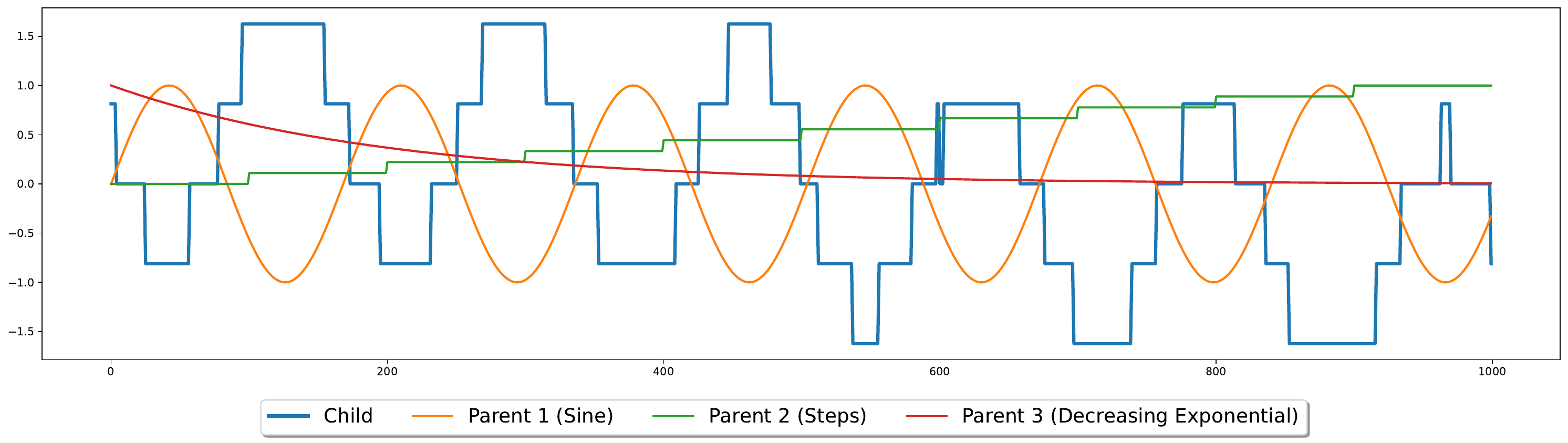}
\caption{Discretize}
\end{subfigure}

\begin{subfigure}{0.49\textwidth}
\includegraphics[width=\linewidth]{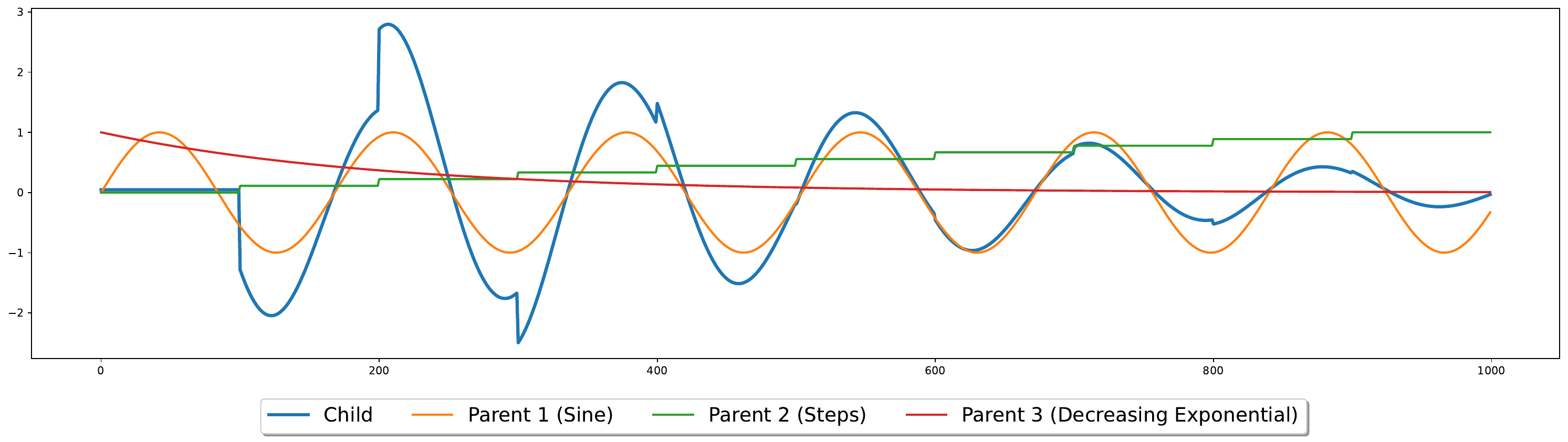}
\caption{Product}
\end{subfigure}

\caption{Examples of time series generated by different Structural Causal Models (SCMs) from the registry. Each plot shows the transformation of three root signals (sinusoidal, linear trend, and exponential decay) into a child time series through the corresponding operator.}
\label{fig:scm_examples}
\end{figure}

\paragraph{Structural Causal Model (SCM) Registry.} To ensure a wide variety of functional relationships, we implement a set of diverse SCMs inspired by \citep{qu2026tabiclv2}. Let $\mathbf{X}{pa(i)}$ denote the collection of parent time series for node $i$. The child node $Y_i$ is generated as $Y_i = \text{Normalize}(f(\mathbf{X}{pa(i)}))$, where $f \in \textit{Registry}$. The registry includes:

\begin{itemize}
    \item \textbf{LinearSCM.} A simple linear combination $Y = \mathbf{W}\mathbf{X} + b$.
    \item \textbf{MLPSCM.} A multi-layer perceptron with random depth (2--10 layers) and hidden dimensions (8--128). Activations are randomly selected for each layer (ReLU, Tanh, ELU, etc.). We use a sparsity-inducing "block-wise dropout" initialization to create specific feature-group dependencies.
    \item \textbf{ConvolutionalSCM.} Models local temporal dependencies using 1D convolutions with random kernel sizes (3--8) and random channel depths.
    \item \textbf{RNNSCM.} Captures deep temporal dependencies using a GRU architecture. These are strictly causal, ensuring the value at time $t$ only depends on $t' \leq t$.
    \item \textbf{PolynomialSCM.} Each input is raised to a random power $d \in \{1, 2, 3, 4\}$ before being linearly combined, inducing symmetric non-linearities.
    \item \textbf{DiscretizeSCM.} Simulates quantization effects by mapping a linear mixture of inputs into a fixed number of discrete bins ($2$ to $15$).
    \item \textbf{ProductSCM.} Computes the element-wise product of all parent signals, representing multiplicative interactions.
\end{itemize}
For visual illustrations of the types of dependencies induced by each SCM, we refer the reader to \cref{fig:scm_examples}.

\paragraph{Normalization and Stability.} To prevent numerical instability and exploding values across deep DAGs, every node's output is z-normalized.
This ensures that every generated time series $Y_i$ maintains a mean of 0 and a standard deviation of 1 before it is used as an input for further child nodes.

\begin{figure}[p]
\centering
\begin{subfigure}{0.48\textwidth}
\includegraphics[width=\linewidth]{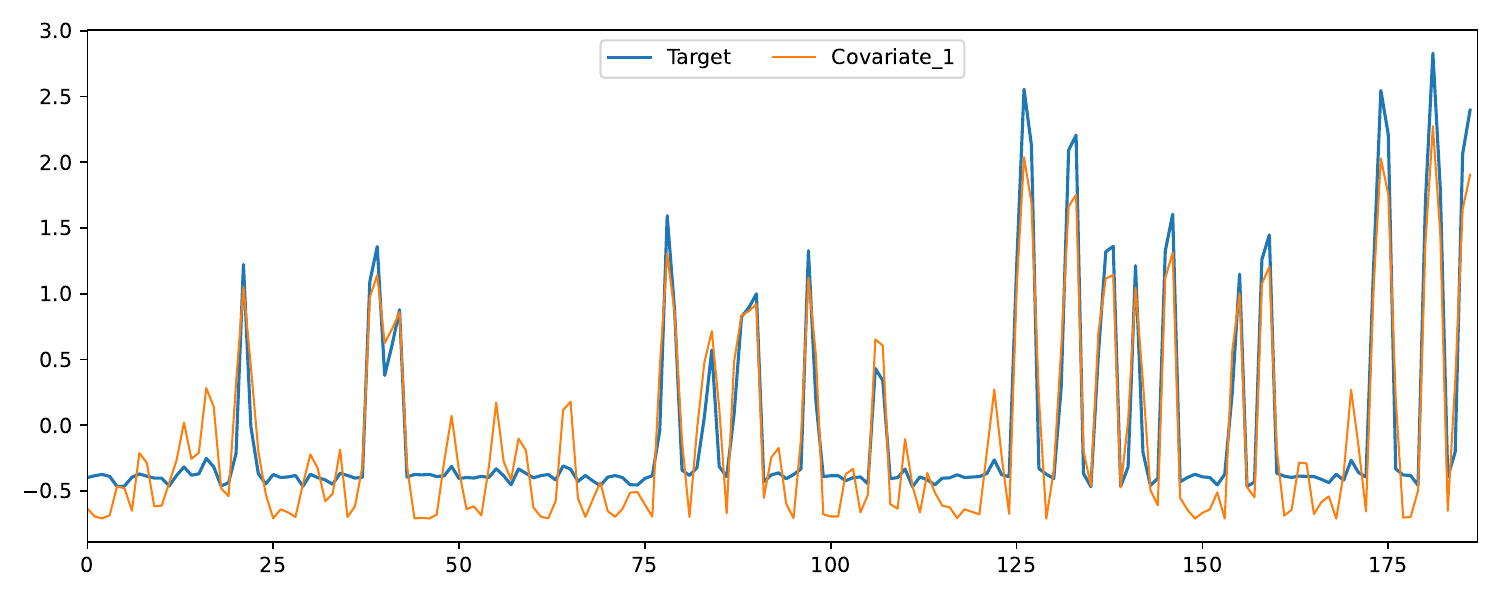}
\caption{}
\end{subfigure}
\hfill
\begin{subfigure}{0.48\textwidth}
\includegraphics[width=\linewidth]{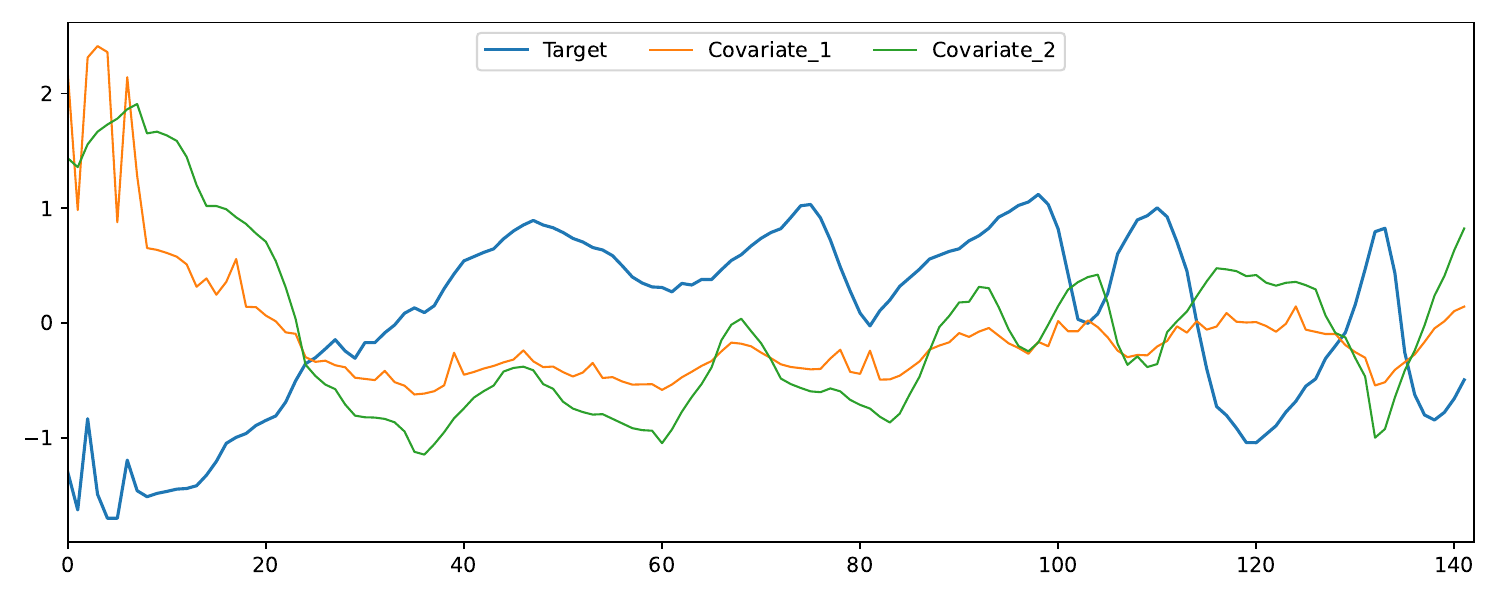}
\caption{}
\end{subfigure}

\begin{subfigure}{0.48\textwidth}
\includegraphics[width=\linewidth]{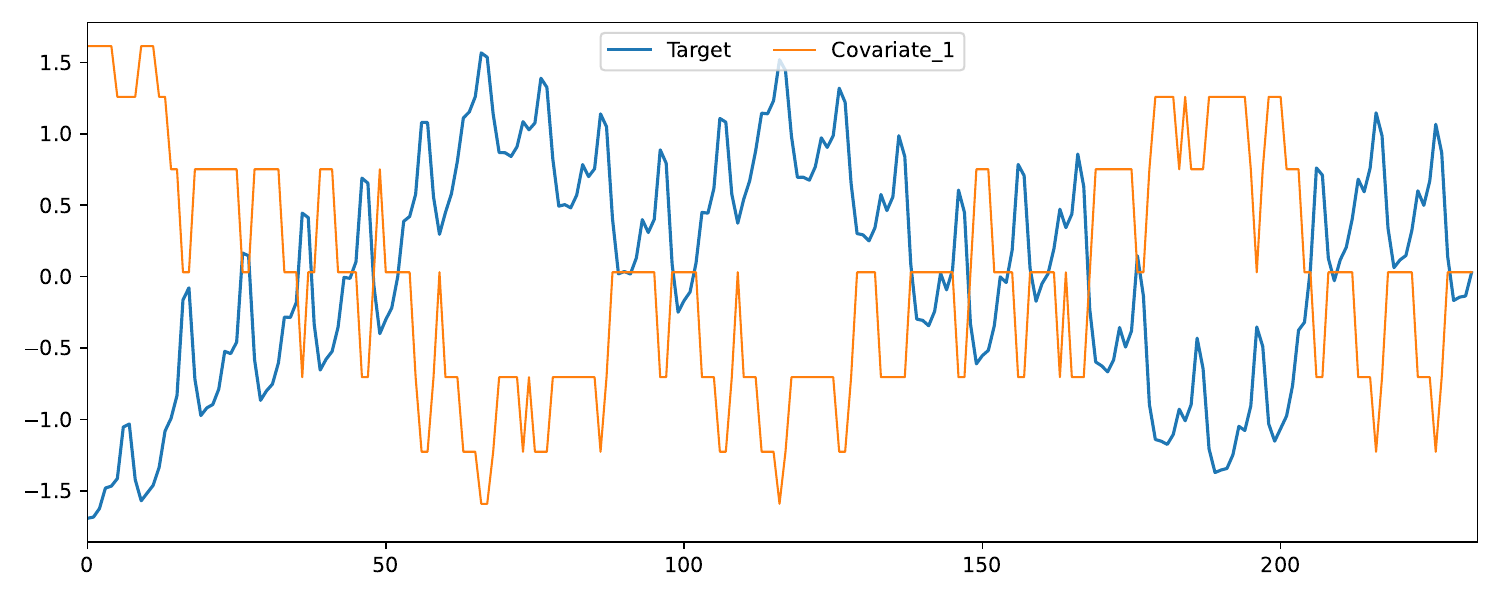}
\caption{}
\end{subfigure}
\hfill
\begin{subfigure}{0.48\textwidth}
\includegraphics[width=\linewidth]{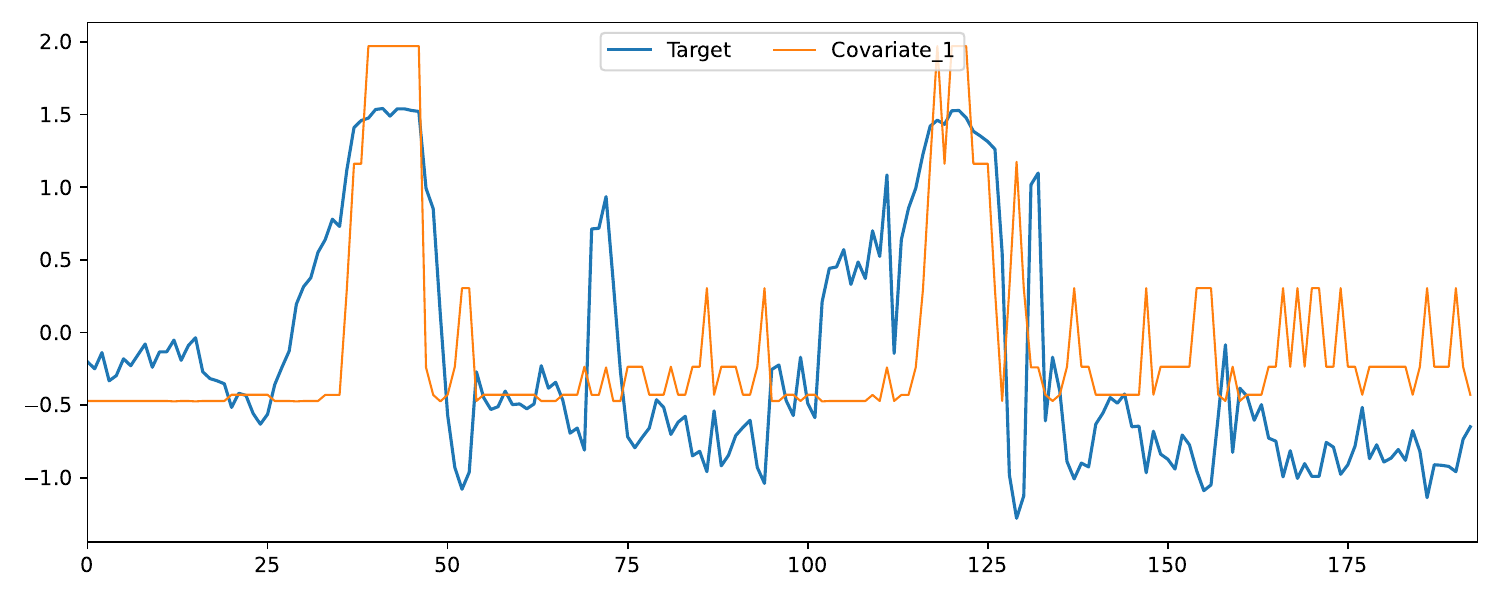}
\caption{}
\end{subfigure}

\begin{subfigure}{0.48\textwidth}
\includegraphics[width=\linewidth]{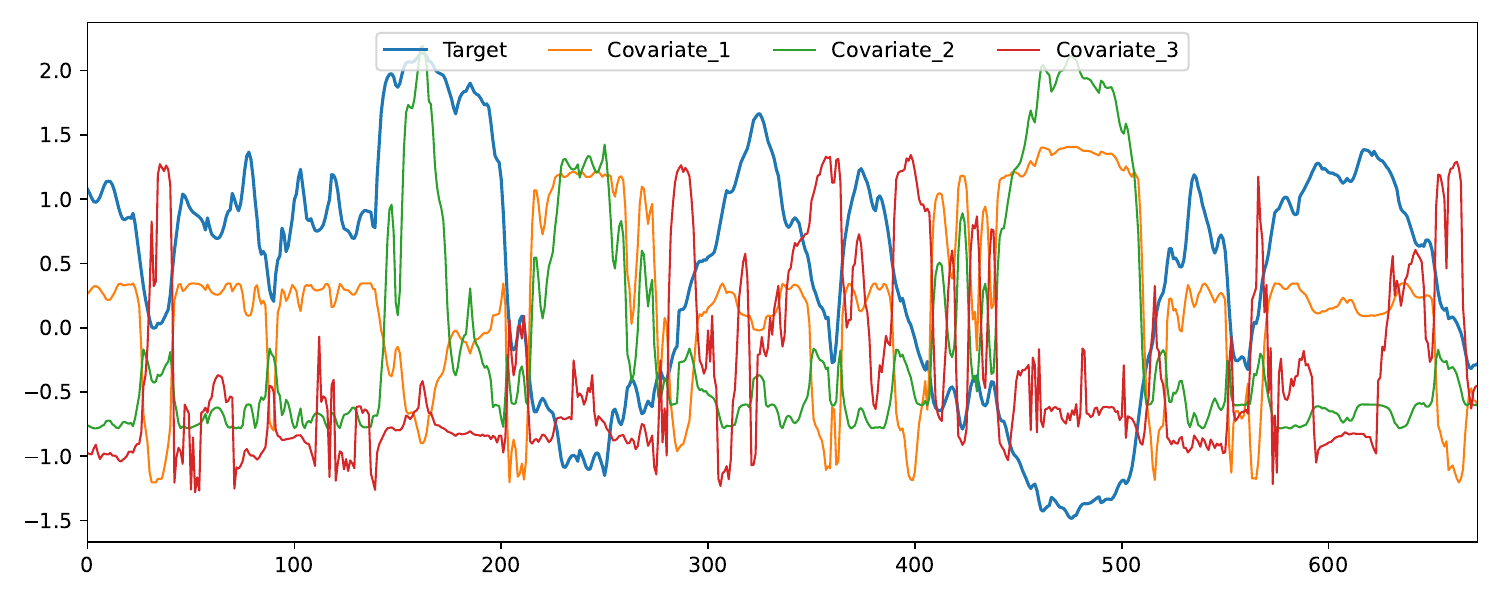}
\caption{}
\end{subfigure}
\hfill
\begin{subfigure}{0.48\textwidth}
\includegraphics[width=\linewidth]{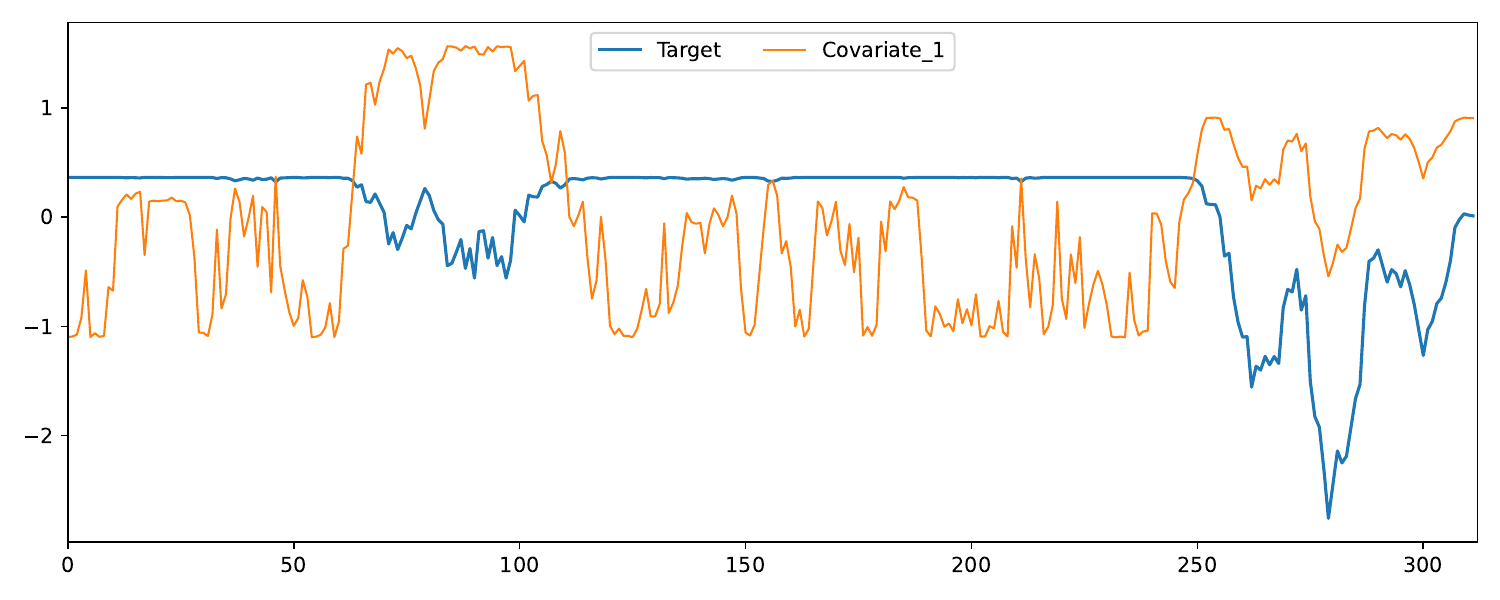}
\caption{}
\end{subfigure}

\begin{subfigure}{0.49\textwidth}
\includegraphics[width=\linewidth]{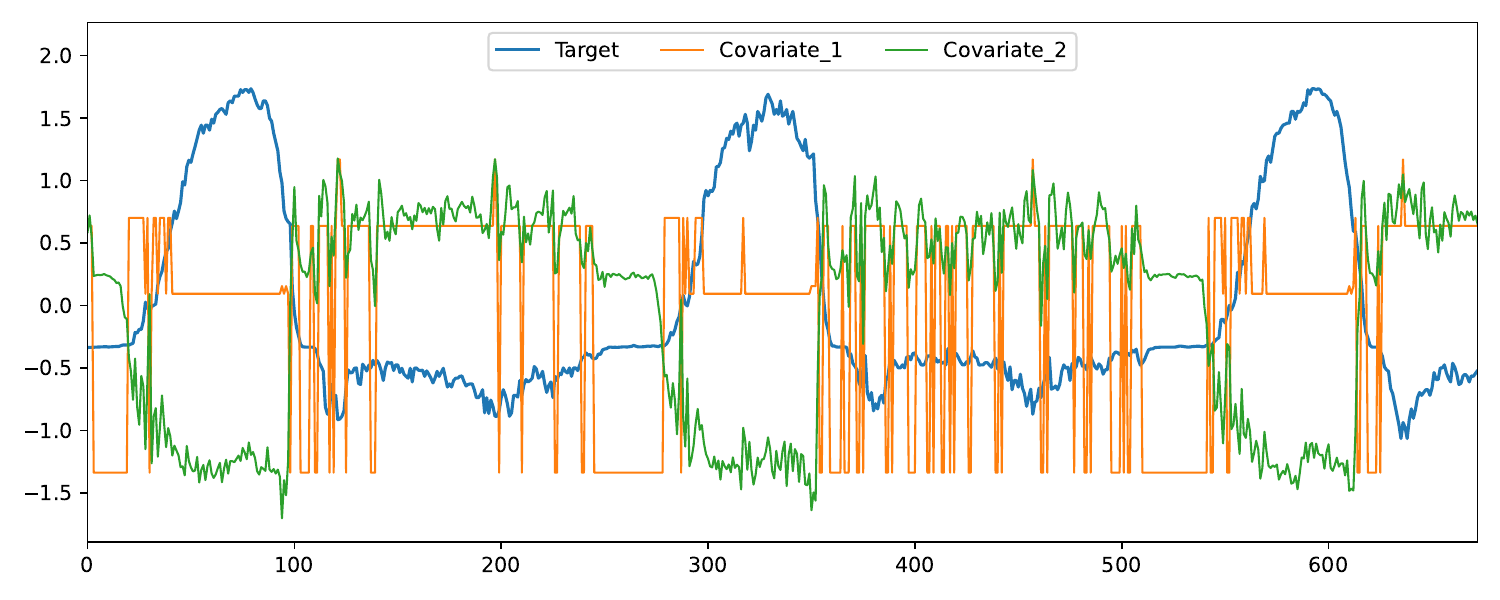}
\caption{}
\end{subfigure}
\hfill
\begin{subfigure}{0.49\textwidth}
\includegraphics[width=\linewidth]{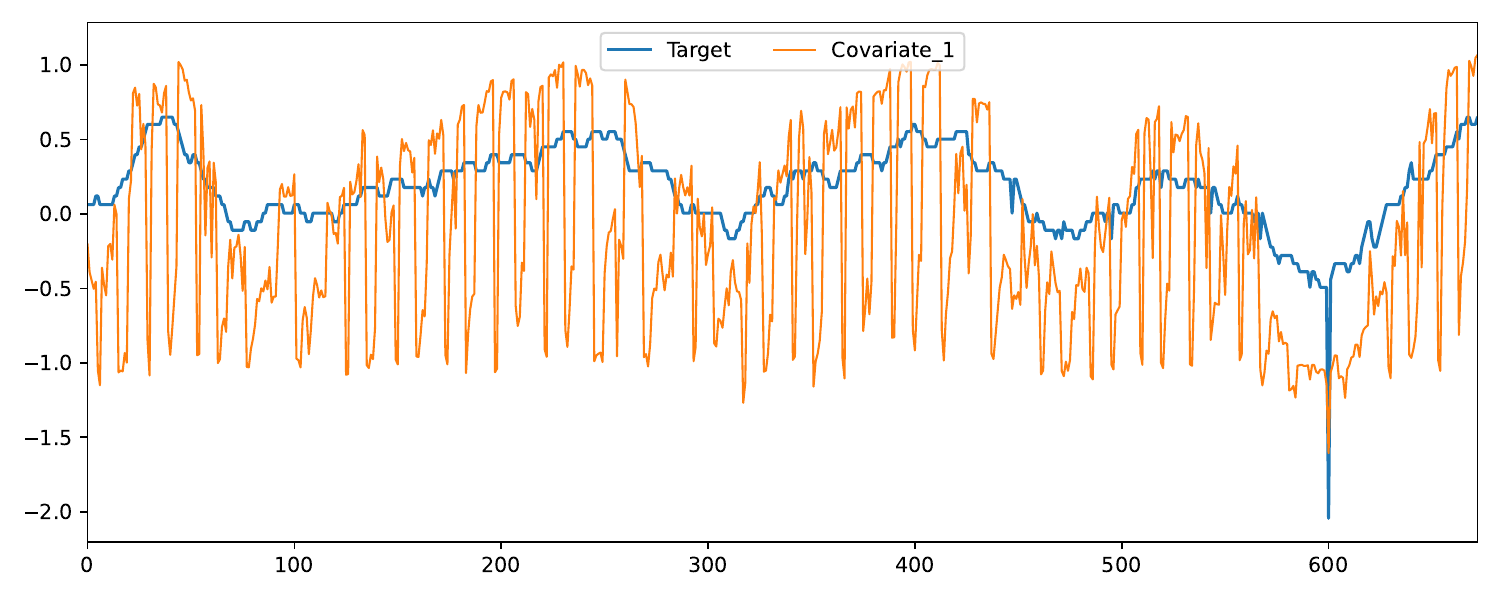}
\caption{}
\end{subfigure}

\begin{subfigure}{0.48\textwidth}
\includegraphics[width=\linewidth]{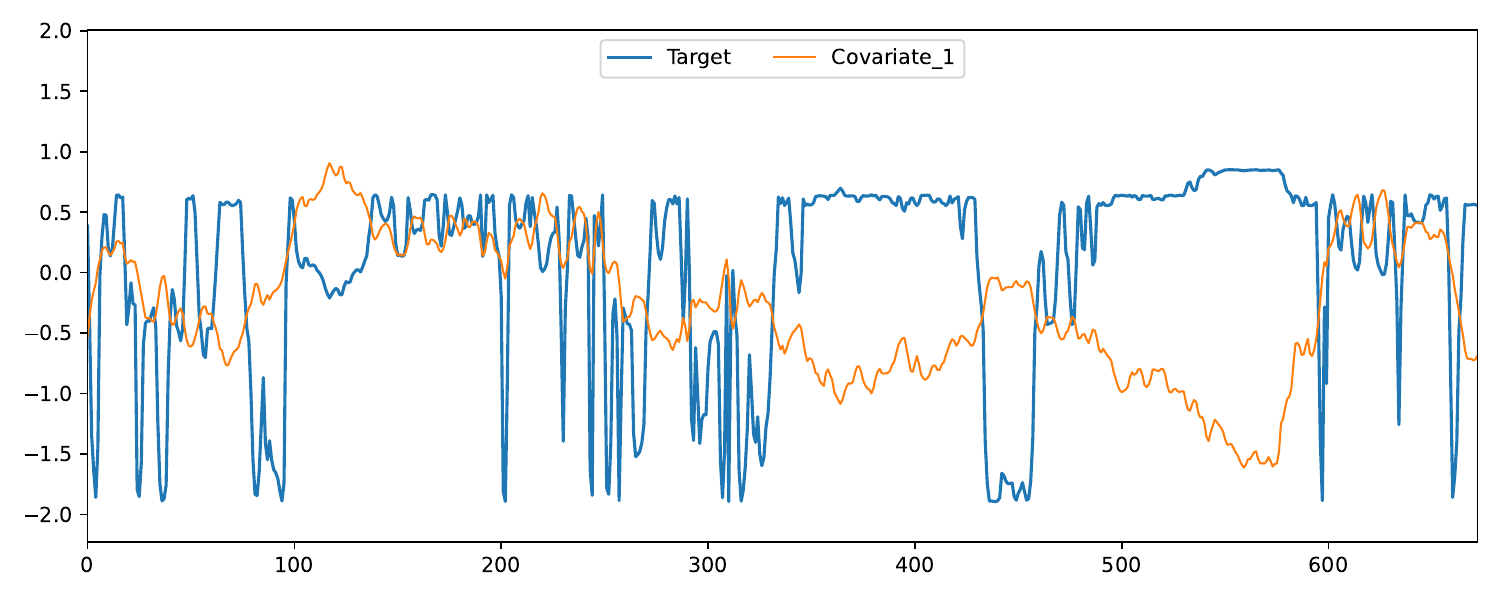}
\caption{}
\end{subfigure}
\hfill
\begin{subfigure}{0.48\textwidth}
\includegraphics[width=\linewidth]{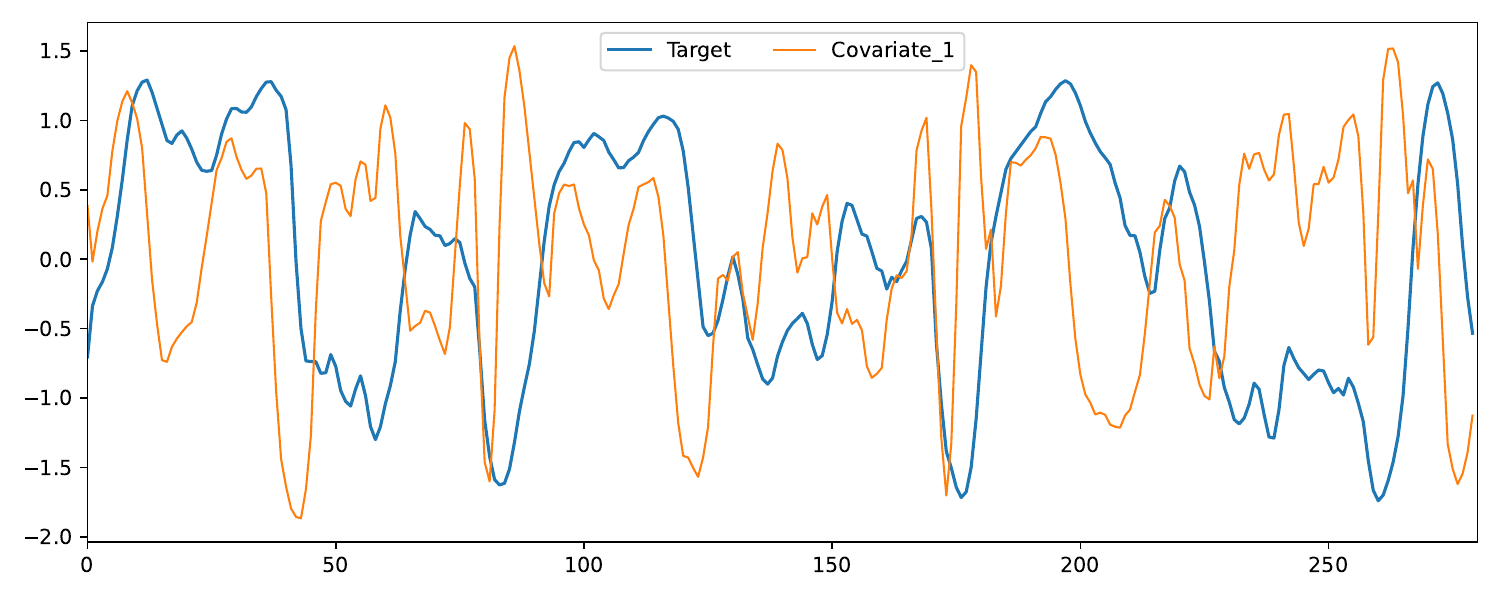}
\caption{}
\end{subfigure}

\caption{Examples of different covariate-target time series sampled from the data prior.}
\label{fig:covar_target_examples}
\end{figure}

\paragraph{Problem Formulation (Target and Covariates).} Once the DAG is computed, we finalize the target--covariate problem by:
\begin{enumerate}[(i)]
\item Sampling a subset of $C$ nodes from the graph to be "visible" to the model.
\item Randomly designating one of these nodes as the Target ($y$).
\item Designating the remaining $C-1$ nodes as Covariates ($\mathbf{x}$).
\end{enumerate}
To encourage learning of informative covariate-target relationships, we first seek to fulfill steps (ii) and (iii) by looking for connected components of the DAG.
We thus sample the $C$ channels from the subset of nodes that have all, or at least one, root node as common ancestor.
\begin{itemize}
\item This approach ensures that covariates are not merely "noise" but share a common underlying causal structure with the target, sometimes acting as direct causes, sometimes as effects, and sometimes as siblings sharing a latent root cause.
\item If this subset is empty or contains less than $C$ nodes, the remaining nodes are drawn uniformly at random within the entire graph, allowing for independent or unrelated (covariate, target) pairs.
\end{itemize}
Examples of such multivariate problems drawn from the data prior are shown in \cref{fig:covar_target_examples}.

\clearpage

\subsubsection{The whole procedure}
\label{subsec:whole-training-procedure}

We summarize the full data generation pipeline in \cref{alg:data_prior}. The procedure samples heterogeneous training tasks combining real-world time series, synthetic time series and target--covariates time series problems. This enables the model to learn both standard univariate forecasting/imputation tasks and more complex multivariate settings generated via structural causal models.

A Bernoulli switch controlled by probability $\pi$ determines whether a task is sampled from the univariate or covariate (SCM-based) mode.

\begin{algorithm}[h!]
\caption{Data Prior Sampling Procedure for \texttt{TS-ICL}}
\label{alg:data_prior}
\begin{algorithmic}[1]

\Require Collection of base univariate time series $\mathcal{D}$ (real + synthetic)
\Require SCM registry $\Upsilon$
\Require Probability $\pi$ of sampling a univariate task

\State Sample $u \sim \mathcal{U}(0,1)$

\If{$u < \pi$}

    \State \textbf{Univariate mode:}
    \State Sample time series $x \sim \mathcal{D}$
    \State Apply task masking (imputation or forecasting)
    \State \Return $x$

\Else

    \State \textbf{Covariate mode:}

    \State Sample number of task channels:
    \State \hspace{0.5cm} $C \sim \text{Exp}(\lambda)$, clipped to $[C_{\min}, C_{\max}]$

    \State Sample number of root signals:
    \State \hspace{0.5cm} $R \sim \text{Geometric}(q)$, clipped to $R_{\max}$

    \State Define latent DAG size:
    \State \hspace{0.5cm} $V = 2C + R$

    \State Sample root time series:
    \State \hspace{0.5cm} $S_1, \dots, S_R \sim \mathcal{D}$

    \State Initialize DAG nodes:
    \State \hspace{0.5cm} $\{X_i\}_{i=1}^R \leftarrow \{S_i\}_{i=1}^R$

    \State \textbf{DAG construction}

    \For{$i = R+1, \dots, V$}
        \State Sample number of parents $k_i \sim \text{Geometric}(p)$
        \State Sample parent set $\mathcal{P}_i \subset \{1,\dots,i-1\}$
        \State Sample SCM $f_i \sim \Upsilon$
        \State Compute:
        \State \hspace{0.5cm} $X_i \leftarrow \text{Normalize}\big(f_i(\mathbf{X}_{\mathcal{P}_i})\big)$
    \EndFor

    \State \textbf{Observation step}

    \State Sample observed set $\mathcal{O} \subset \{1,\dots,V\}$ such that $|\mathcal{O}| = C$
    \State Sample target node $y \sim \mathcal{O}$
    \State Define covariates $\mathbf{x} = \mathcal{O} \setminus \{y\}$

    \State \textbf{Task masking}

    \State \hspace{0.5cm} - imputation: random point/block masking
    \State \hspace{0.5cm} - forecasting: future horizon masking

    \State \Return $(\mathbf{x}, y)$

\EndIf

\end{algorithmic}
\end{algorithm}

\clearpage

\subsection{Architecture Hyperparameters and Implementation Details}
\label{appendix:hyperparameters}

The \texttt{TS-ICL} architecture is governed by a set of hyperparameters controlling model capacity, tokenization granularity, and depth across modules. All latent vectors across the four modules share a common dimensionality $d$, ensuring seamless information flow.

\subsubsection{Architectural Hyperparameters.}


\paragraph{Time Series Encoder $\mathcal{E}$ hyperparameters.}
The encoder serves as the primary feature extractor, compressing multi-channel time series into a fixed-size latent bottleneck - see \cref{fig:encoder}.
\begin{itemize}
\item \textbf{Latent dimension ($d$):} The feature dimensionality used throughout the encoder is set to $d = 256$.
\item \textbf{Temporal Encoding:} Timestamps are mapped using Fourier features in logarithmic scale with base 2, with $L_{\text{fourier}} = 128$ frequencies up to maximum frequency $2^{10}$, followed by a linear projection layer to $\mathbb{R}^d$.
\item \textbf{Number of latent tokens ($M$):} The number of learnable tokens per channel is $M = 32$. This parameter controls the resolution of the latent representation.
\item \textbf{Refinement block:} The refinement stage consists of $L_{\mathcal{E}} = 3$ channel-independent Transformer blocks, each with $n_h^{\mathcal{E}}=8$ self-attention heads of dimension 64.
\end{itemize}


\paragraph{Channel Mixer Module $\mathcal{M}$ hyperparameters.}
The mixer facilitates task-oriented channel mixing by conditioning the target series on covariates - see \cref{fig:mixer}.
\begin{itemize}
\item \textbf{Latent dimension ($d$):} The feature dimensionality used throughout the channel mixer is set to $d = 256$.
\item \textbf{Cross-Channel Attention:} The number of cross-attention layers used to aggregate channel-independent tokens into a covariate-aware representation is $L_{\mathcal{M}}^{\text{cross}} = 3$.
Each layer contains $n_h^{\mathcal{M}}=8$ attention heads of dimension 64.
\item \textbf{Global Latent Refinement:}
After reshaping, the latent sequence is processed by $L_{\mathcal{M}}^{\text{self}} = 3$ self-attention layers to capture dependencies across the aggregated latent space.
Each layer contains $n_h^{\mathcal{M}}=8$ self-attention heads of dimension 64.
\end{itemize}


\paragraph{Temporal Context Query Module $\mathcal{C}$ hyperparameters.}
This module acts as a continuous-time interface between the latent tokens and the regressor - see \cref{fig:queryer}.
\begin{itemize}
\item \textbf{Latent dimension ($d$):} The feature dimensionality used throughout the context query module is set to $d = 256$.
\item \textbf{Frequency Encoding:} Query coordinates $t$ are encoded using high-frequency sinusoidal features (NeRF-style) to preserve fine-grained temporal localizations before being projected to $\mathbb{R}^d$. 
We use three frequency bands each with 128 frequencies and respective maximum frequencies $2^6,\,2^7,\,2^{10}$.
\item \textbf{Query mechanism:} A single $8\times64$ cross-attention layer is used to extract $H(t)$ from $\boldsymbol{Z}^{\mathrm{final}}$.
This layer operates in parallel on the three frequency bands.
After concatenating, the context-aware time representation $H(t)$ has dimension $3\times d=768$.
\end{itemize}


\paragraph{In-Context Learning Regressor $\mathcal{R}$ hyperparameters.}
The regressor performs the final predictive task using a causal Transformer architecture - see Figures \ref{fig:build-tokens-icl} \& \ref{fig:regressor}.
\begin{itemize}
\item \textbf{Latent dimension ($d$):} The feature dimensionality used throughout the in-context learning regressor is set to $d = 512$.
\item \textbf{Input Value Projection:}
Before entering the cross-attention block for token construction, all available observed values $x_t$ are projected to $\mathbb{R}^d$ via a linear layer.
Similarly, if available, covariates $X_t$ are projected to $\mathbb{R}^d$ via another linear layer, shared by all covariates.
\item \textbf{ICL Transformer:}
The causal sequence processing is performed by $L_{\mathcal{R}} = 12$ Transformer layers, each with $n_h^{\mathcal{R}}=8$ heads of dimension 64.
\item \textbf{Quantile Head:} A final linear layer maps the $d$-dimensional hidden states of target tokens to a 99-dimensional vector representing the equidistant predicted quantiles.
\end{itemize}

\subsubsection{Training hyperparameters and task-specific strategies.} \label{app:training_strategies}

This section provides additional details about the training strategy.

\paragraph{Input scaling.}
Following common practice in time series forecasting, raw time series are preprocessed by an instance-normalization layer, scaling each sample to zero mean and unit variance.
We then apply a pointwise $\sinh^{-1}$ transform to the standardized inputs, to stabilize training against outlier values \citep{ansari2024chronos2}.

\paragraph{Task mixing and zero-shot robustness.}
We employ a task mixing probability $\pi = 0.8$:
20\% of the training batches are drawn from the univariate pretraining corpus in \cref{tab:pretraining-dataset}, whereas the remaining 80\% are further processed by the SCM prior in \cref{alg:data_prior} to create target--covariates tasks (with $C-1$ covariates, $C\geq2$) or new univariate tasks ($C=1$).
This mixture ensures the model remains a robust univariate predictor while learning to leverage exogenous signals.
\begin{itemize}
\item \textbf{Covariate Complexity Sampling:} For covariate tasks, the number of channels $C$ is (i) either $C=1$ (univariate task) with probability 0.2 (ii) or sampled from a truncated exponential distribution:
\[
P(K=k) \propto e^{-\lambda k}, \quad k \in \{2, \dots, 20\}.
\]
We set $\lambda = 0.5$ to favor simpler tasks with few covariates while maintaining significant exposure to high-dimensional inputs (up to 20 covariates).
\end{itemize}

\paragraph{Specialized Checkpoints.}
While the architecture $\mathcal{R}$ is identical for all tasks, we provide two specialized checkpoints. The \textit{Forecasting} checkpoint is trained strictly with right causal masking, whereas the \textit{Imputation} checkpoint is trained to reconstruct missing values using both preceding and succeeding context.

\paragraph{Imputation Training.}
We enforce task diversity while training the imputation checkpoint with a two-step procedure.
\begin{enumerate}
\item \textbf{Window sampling:}
a per-batch context length $T$ sampled from multiple regimes to handle varying context lengths:
\[
T \sim
\begin{cases}
\mathcal{U}(128,\,336) & \text{with probability } p_1=0.15 \\
\mathcal{U}(512,\,1024) & \text{with probability } p_2=0.6 \\
\mathcal{U}(1300,\,1400) & \text{with probability } p_3=0.05 \\
\mathcal{U}(2000,\,2100) & \text{with probability } p_4=0.1 \\
\mathcal{U}(4000,\,4096) & \text{with probability } p_5=0.1
\end{cases},
\]
where $p_1, \dots, p_5$ are dataset balancing probabilities.
The maximum supported window length for imputation is $T=4096$.

\item \textbf{Masking Procedure:}
once a context window has been sampled, we then apply a random masking strategy.
A fraction $\rho$ of the observations are held out as target queries $\mathcal{T}^{\text{tgt}}$. The remaining points form $\mathcal{D}_{\text{train}}$.
\begin{itemize}
\item $\rho$ is either a pointwise missingness rate, sampled randomly from $\{0.05,\,0.06,\dots,0.95\}$;
\item or corresponds to up to four missing blocks of random length between 12 and 168 (adjusted depending on available context length) and at most 50\% pointwise missing values.
\end{itemize}
\end{enumerate}

\paragraph{Forecasting training.}
A similar procedure is applied to the forecasting checkoint;
\begin{enumerate}
\item \textbf{Window sampling:} the lookback and horizon pairs are sampled jointly:


\[
(L,\,H) \sim
\begin{cases}
(\mathcal{U}(50,\,200),\;\mathcal{U}(8,\,20)) & \text{with probability } p_1=0.2 \\
(\mathcal{U}(200,\,672),\;\mathcal{U}(18,\,36)) & \text{with probability } p_2=0.15 \\
(\mathcal{U}(1000,\,1100),\;\mathcal{U}(48,\,336)) & \text{with probability } p_3=0.5 \\
(\mathcal{U}(2000,\,2100),\;\mathcal{U}(48,\,336)) & \text{with probability } p_4=0.1 \\
(\mathcal{U}(3062,\,4096),\;\mathcal{U}(48,\,672)) & \text{with probability } p_5=0.05
\end{cases}.
\]

The maximum supported configuration is:
\[
T_{\mathrm{look-back}} \leq 4096, \quad T_{\mathrm{horizon}} \leq 672.
\]

\item \textbf{Masking Procedure:}
We enforce robustness to irregularly sampled time series by removing a fraction $\rho$ of the observations in the lookback window, with probability 0.15.
We draw $\rho$ from the set $\{0.1,\,0.25,\,0.5,\,0.75\}$.
\end{enumerate}


\paragraph{Quantile head.}
The model predicts a set of 99 quantiles $\{\alpha_k\}_{k=1}^{99}$, uniformly spaced in $(0,1)$, allowing for full density estimation.
We set the smoothing coefficient of the pinball loss to $\beta = 0.01$ to ensure differentiability.


\subsubsection{Optimization Hyperparameters.}
\label{app:optimization_hyperparams}

For both imputation and forecasting checkpoints, the optimization is configured as follows:
\begin{itemize}
\item \textbf{Optimizer:} \textit{Muon} \citep{jordan2024muon} for 2D parameters and AdamW for 1D parameters (embeddings, scales, biases).
\item \textbf{Learning rate:} Max $lr = 4e-4$.
\item \textbf{Scheduler:} Cosine decay down to $5e-4$.
\item \textbf{Hardware:} 4$\times$ Nvidia H100 GPUs (92GB).
\item \textbf{Batch size:} the global batch size is $B = 256$, obtained through a mini-batch size of 32 and two steps of gradient accumulation.
\item \textbf{Training Budget:} The imputation checkpoint is trained for $500k$ optimization steps, representing approximately 5 training days.
The forecasting checkpoint is trained for $650k$ optimization steps, representing approximately 9 training days.
\end{itemize}

\clearpage
\section{Ablation Studies}
\label{appendix:ablations}

In this section, we provide a comprehensive analysis of the architectural choices and scaling properties of \texttt{TS-ICL}. All ablation models were trained for a fixed budget of $100\text{k}$ steps on a H100 GPU to ensure fair comparison. Evaluation is performed on a subset of 11 datasets (44 tasks) of \texttt{fm-impute-bench}, namely \texttt{fm-impute-mini}, commonly used for ablations \cite{TSFMImputationBenchmark} (see \cref{tab:fm-impute-mini-details}).

\begin{table}[h!]
\centering
\caption{\texttt{fm-impute-mini} subset for zero-shot imputation ablation studies.}
\scalebox{0.75}{
\begin{tabular}{lcccccc}
\toprule
\multirow{2}{*}{\textbf{Dataset}} & \multirow{2}{*}{\textbf{Domain}} & \multirow{2}{*}{\textbf{Freq}} & \textbf{Num.} & \textbf{Series} & \textbf{Num.} & \textbf{Window} \\
& & & \textbf{Series} & \textbf{Length} & \textbf{Test Windows} & \textbf{Size} \\
\midrule
{BDG2-Bear}         & Energy    & 1H    & 91  & 17544  & 7522  & 672 \\
{BDG2-Rat}          & Energy    & 1H    & 280 & 17544  & 24915 & 672 \\
{Covid19 Energy}    & Energy    & 1H    & 1   & 31912  & 195   & 672 \\
{GFC12 Load}        & Energy    & 1H    & 20  & 39414  & 4960  & 672 \\
{Hog}               & Energy    & 1H    & 24  & 17544  & 2310  & 672 \\
{Jena Weather 10T}  & Climate   & 10min & 21  & 52704  & 1428  & 4032 \\
{Jena Weather 1H}   & Climate   & 1H    & 21  & 8784   & 1344  & 672 \\
{Oikolab Weather}   & Climate   & 1H    & 8   & 100057 & 5288  & 672 \\
{PDB}               & Energy    & 1H    & 1   & 17520  & 96    & 672 \\
{Pedestrian Counts} & Transport & 1H    & 66  & 96400  & 7733  & 672 \\
{Weather}           & Climate   & 1H    & 11  & 35064  & 2398  & 672 \\
\bottomrule
\end{tabular}
}
\label{tab:fm-impute-mini-details}
\end{table}

\subsection{Architecture Scaling}
\label{appendix:scaling}

We evaluate the impact of model capacity by varying
\begin{enumerate*}[(i)]
\item the number of attention heads $n_h=n_h^{\mathcal{E}}=n_h^{\mathcal{M}}=n_h^{\mathcal{C}}$ in the three Transformers of the encoder;
\item the number $\mathcal{L}_{\mathcal{R}}$ of self-attention layers in the in-context regressor $\mathcal{R}$;
\item the number $n_h^{\mathcal{R}}$ of self-attention heads in $\mathcal{R}$.
\end{enumerate*}
We define three model configurations: \textit{Small}, \textit{Medium}, and \textit{Large}, with respectively \textbf{8.5M}, \textbf{12M} and \textbf{27M} parameters.


\begin{table}[h]
\centering
\caption{Model Capacity Ablation. 
Average CRPS over 44 univariate imputation tasks from the \texttt{fm-impute-mini} benchmark (lower is better).}
\label{tab:scaling-results}
\begin{tabular}{lcccccc}
\toprule
\textbf{Model} & $n_h$ & $\mathcal{L}_{\mathcal{R}}$ & $n_h^{\mathcal{R}}$ & \textbf{Params} & \textbf{\texttt{fm-impute-mini}} \\
\midrule
\texttt{Small}  & $4$ & $8$  & $4$ & $\sim$8.5M  & 0.201 \\
\texttt{Medium} & $8$ & $12$ & $4$ & $\sim$12M  & 0.197 \\
\texttt{Large}  & $8$ & $12$ & $8$ & $\sim$27M & \textbf{0.194} \\
\bottomrule
\end{tabular}
\end{table}

\paragraph{Results.}
Table~\ref{tab:scaling-results} shows that increasing model capacity consistently improves imputation accuracy, with the \textit{Large} configuration obtaining the best average CRPS. 
In particular, moving from \textit{Small} to \textit{Large} reduces the average CRPS from $0.201$ to $0.194$, suggesting that additional attention capacity in both the encoder and the in-context regressor improves the model's ability to exploit contextual information across tasks.
The gains, however, are relatively smooth and exhibit diminishing returns: the \textit{Medium} model already closes a substantial fraction of the gap to the \textit{Large} model, while using less than half the number of parameters. 
Moreover, the \textit{Small} model remains competitive despite its lower capacity. 
Overall, this suggests a favorable accuracy--efficiency trade-off, with smaller variants remaining attractive under computational constraints.

\subsection{Component Ablations: Synergy between Encoder and Regressor}
\label{appendix:component-ablations}

To justify the hybrid structure of \texttt{TS-ICL}, we compare the full architecture against two baseline configurations:
\begin{itemize}
\item \textbf{Encoder-Only.}
We remove the In-Context Regressor $\mathcal{R}$. The context-aware representation $H(t)$ from module $\mathcal{C}$ is passed directly through a dense MLP network - with $5\times256$ hidden layers - to project onto the target values.
\item \textbf{Regressor-Only (Pure ICL).}
We remove the Encoder $\mathcal{E}$ and the Mixer $\mathcal{M}$.
The Regressor $\mathcal{R}$ receives only raw temporal Fourier features \citep{mildenhall2021nerf} as representation $H(t)$, performing pure in-context regression without the benefit of the refined latent context.
\end{itemize}

\begin{table}[h]
\centering
\caption{\textbf{Model Capacity Ablation.}
Average CRPS over 44 univariate imputation tasks from the \texttt{fm-impute-mini} benchmark (lower is better).}
\label{tab:component-results}
\begin{tabular}{lccc}
\toprule
\textbf{Configuration} & \textbf{Architecture Change} & \textbf{\texttt{fm-impute-mini}} \\
\midrule
Encoder-Only & $H(t) \to \text{MLP Head}$ & 0.297 \\
Regressor-Only & No $\mathcal{E}$, Raw Fourier Features & 0.204 \\
\textbf{Full TS-ICL} & \textbf{Encoder $\mathcal{E}$ + Regressor $\mathcal{R}$} & \textbf{0.194}  \\
\bottomrule
\end{tabular}
\end{table}

\paragraph{Results.}
Table~\ref{tab:component-results} highlights the complementarity between the encoder and the in-context regressor. 
The \textit{Encoder-Only} variant performs substantially worse, indicating that the latent context representation alone is not sufficient without an expressive regression mechanism. 
Conversely, the \textit{Regressor-Only} baseline is much stronger, but still underperforms the full model, showing that raw Fourier features provide a competitive ICL baseline but lack the refined context produced by the encoder. 
The full \texttt{TS-ICL} architecture achieves the best CRPS, suggesting that the encoder and regressor act synergistically: the encoder builds informative context-aware representations, while $\mathcal{R}$ effectively uses them for in-context prediction.

\subsection{Covariate Management Strategies}
\label{appendix:covar-ablations}

We investigate the optimal placement of the covariate mixing mechanism. Since \texttt{TS-ICL} allows for multi-stage conditioning, we compare three strategies for handling exogenous signals:
\begin{itemize}
    \item \textbf{Early Mixing (Encoder Only):} Covariates are processed in the Encoder $\mathcal{E}$ and mixed in the Mixer $\mathcal{M}$. The Regressor $\mathcal{R}$ only receives $H(t)$ and context-target pairs of the main series, with no cross-attention on covariates during token construction.
    \item \textbf{Late Mixing (Regressor Only):} The Encoder $\mathcal{E}$ is univariate (no covariates). All covariate information is provided directly to the Regressor $\mathcal{R}$ via the Cross-Attention mechanism during input token construction.
    \item \textbf{Dual Mixing (Full Architecture):} Covariates are leveraged both in the global representation (Encoder/Mixer) and for local token conditioning (Regressor).
\end{itemize}

\begin{table}[h]
\centering
\caption{Covariate Mixing Strategy.
Average CRPS over 6 covariate-aware imputation tasks from \texttt{fm-impute-covars} (lower is better).}
\label{tab:covar-results}
\begin{tabular}{lccc}
\toprule
\textbf{Mixing Strategy} & \textbf{Mixing Stage} & \texttt{fm-impute-covars}\\
\midrule
Early Mixing & $\mathcal{E} + \mathcal{M}$ only & 0.123 \\
\textbf{Late Mixing} & $\mathcal{R}$ only & \textbf{0.085} \\
\textbf{Dual Mixing} & $\mathcal{E} + \mathcal{M}$ and $\mathcal{R}$ & \textbf{0.085} \\
\bottomrule
\end{tabular}
\end{table}

\paragraph{Results.}
The results in \cref{tab:covar-results} show that late covariate mixing is already sufficient to obtain strong performance, with \textit{Late Mixing} matching the \textit{Dual Mixing} variant. 
This suggests that injecting covariate information directly in the regressor $\mathcal{R}$ provides effective pointwise conditioning at prediction time. 
In contrast, \textit{Early Mixing} alone performs worse, indicating that global covariate information in the encoder is not sufficient without late-stage conditioning. 
We nevertheless retain the dual strategy, as early mixing may provide useful global context about covariate structure in harder settings, while late mixing supplies local, pointwise covariate information to the regressor.

\clearpage
\section{Evaluation Metrics}
\label{sec:scores-metrics}

This section provides formal definitions for the evaluation metrics employed across the various experimental benchmarks presented in \cref{sec:experiements} and Appendices~\ref{extended-imputation-expes}, \ref{extended-forecasting-expes}. These metrics are designed to provide a comprehensive assessment of model performance, covering (i) computational efficiency during inference, (ii) the calibration and quality of the predicted probability distributions, and (iii / iv) the accuracy of point predictions via scale-independent error measures.

\subsection{Metrics Definition}

\paragraph{(i) Inference efficiency score definition.}
The computational efficiency is measured by the median inference time per window (as in \citep{shchur2025fev}), expressed in milliseconds (ms). The calculation follows a two-step aggregation: first, the mean inference time is computed for each individual dataset to account for domain-specific variations; second, the median of these mean values is taken. For a collection of $D$ datasets, where each dataset $d$ contains $M_d$ windows with individual inference times $\Delta t_{m,d}$, the efficiency score is defined as:
\begin{equation*}
\mu_d = \frac{1}{M_d} \sum_{m=1}^{M_d} \Delta t_{m,d},
\end{equation*}
\begin{equation*}
\mathrm{Efficiency} = \mathrm{median}(\{\mu_1, \ldots, \mu_D\}).
\end{equation*}
This procedure ensures that the final metric is representative of typical performance while remaining robust to outliers across different data distributions.

\paragraph{(ii) Weighted Quantile Loss (WQL) and Continuous Ranked Probability Score (CRPS) definitions.}
To evaluate the quality of the predicted distribution, the Continuous Ranked Probability Score (CRPS) is employed, which measures the compatibility between the predicted cumulative distribution function $\hat F$ and the observed ground truth $x$. The CRPS can be expressed in its integral form as:
\begin{equation*}
\mathrm{CRPS}(\hat F, x) = \int_{0}^{1} 2 \cdot \mathrm{QL}_{\alpha}(\hat F^{-1}(\alpha), x) \, \mathrm{d}\alpha,
\end{equation*}
where $\mathrm{QL}_{\alpha}(q, x)$ represents the quantile loss (or pinball loss) at level $\alpha$:
\begin{equation*}
\mathrm{QL}_{\alpha}(q, x) = (\alpha - \mathbb{I}_{\{x < q\}}) (x - q).
\end{equation*}

To ensure computational tractability and provide a normalized metric for cross-dataset comparison, a normalized discrete approximation of the CRPS is utilized, known as the Weighted Quantile Loss (WQL) \citep{koenker2001quantile, gneiting2007probabilistic}. For a set of $K$ discrete quantiles $\{\alpha_1, \ldots, \alpha_K\}$, the WQL is calculated as:
\begin{equation*}
\mathrm{WQL} = \frac{1}{K} \sum_{j=1}^{K} \mathrm{WQL}_{\alpha_j},
\end{equation*}
where each individual $\mathrm{WQL}_{\alpha}$ is normalized by the absolute scale of the targets:
\begin{equation*}
\mathrm{WQL}_{\alpha} = 
\frac{2 \sum_{i,t \in \mathcal{T}^{\mathrm{tgt}}} \mathrm{QL}_{\alpha}(q^{(\alpha)}_{i,t}, x_{i,t})}
{\sum_{i,t \in \mathcal{T}^{\mathrm{tgt}}} |x_{i,t}|}.
\end{equation*}

In the evaluation, $K=9$ equidistant quantiles $\alpha \in \{0.1, 0.2, \ldots, 0.9\}$ are used, following standard pratice, e.g. \citep{shchur2025fev,qiao2026sTIME}.
This formulation allows the WQL to serve as a robust, scale-invariant proxy for the CRPS, capturing the accuracy of the entire predicted distribution.

\textit{Important note.} For deterministic baselines such as \texttt{Linear}, \texttt{Seasonal}, or \texttt{LOCF}, the same point prediction is used across all quantile levels when computing the WQL.

\paragraph{(iii) Normalized Mean Absolute Error (NMAE) definition.}
To assess point prediction accuracy while accounting for differing scales across series, the Normalized Mean Absolute Error (NMAE) is used (as in \citep{PatchTST}). This metric rescales the standard Mean Absolute Error (MAE) by the standard deviation of the observations in the context set.

For a series $i$ and a target horizon $\mathcal{T}^{\mathrm{tgt}}$, let $x_{i,t}$ be the ground truth and $\hat{x}_{i,t}$ the predicted median (quantile $0.5$ for \texttt{TS-ICL}). The NMAE for series $i$ is defined as:
\begin{equation*}
\mathrm{NMAE}_i = \frac{\frac{1}{|\mathcal{T}^{\mathrm{tgt}}|} \sum_{t \in \mathcal{T}^{\mathrm{tgt}}} |x_{i,t} - \hat{x}_{i,t}|}
{\sigma_{i, \mathrm{ctxt}}},
\end{equation*}
where $\sigma_{i, \mathrm{ctxt}}$ is the standard deviation of the series $i$ calculated over the observed context set $\mathcal{T}^{\mathrm{ctxt}}$:
\begin{equation*}
\sigma_{i, \mathrm{ctxt}} = \sqrt{\frac{1}{|\mathcal{T}^{\mathrm{ctxt}}|} \sum_{t \in \mathcal{T}^{\mathrm{ctxt}}} (x_{i,t} - \bar{x}_{i, \mathrm{ctxt}})^2}.
\end{equation*}

This normalization provides a scale-independent measure of the error relative to the inherent volatility of the series. The global NMAE is obtained by averaging across all $N$ series:
\begin{equation*}
\mathrm{NMAE} = \frac{1}{N} \sum_{i=1}^{N} \mathrm{NMAE}_i.
\end{equation*}



\paragraph{(iv) Mean Absolute Scaled Error (MASE) definition.}
To evaluate point prediction accuracy across datasets with varying scales, the Mean Absolute Scaled Error (MASE) is used \citep{hyndman2018forecasting}. MASE normalizes the Mean Absolute Error (MAE) of the model by the mean absolute error of a seasonal naïve baseline.

For a series $i$, let $x_{i,t}$ be the ground truth and $\hat{x}_{i,t}$ the predicted median. The MASE for series $i$ is defined as:
\begin{equation*}
\mathrm{MASE}_i = \frac{1}{|\mathcal{T}^{\mathrm{tgt}}|} \sum_{t \in \mathcal{T}^{\mathrm{tgt}}} \frac{|x_{i,t} - \hat{x}_{i,t}|}{a_i},
\end{equation*}
where $a_i$ is the seasonal normalization factor calculated over the target set $\mathcal{T}^{\mathrm{tgt}}$:
\begin{equation*}
a_i = \frac{1}{|\mathcal{T}^{\mathrm{tgt}}| - s} \sum_{t \in \mathcal{T}^{\mathrm{tgt}}, t > s} |x_{i,t} - x_{i,t-s}|,
\end{equation*}
and $s$ is the seasonal periodicity. This normalization ensures that the metric is scale-independent by comparing the model's error to the typical seasonal variations within the same target horizon.

The global metric is obtained by averaging across all $N$ series:
\begin{equation*}
\mathrm{MASE} = \frac{1}{N} \sum_{i=1}^{N} \mathrm{MASE}_i.
\end{equation*}



\clearpage
\section{Extended Imputation Experiments}
\label{extended-imputation-expes}

This section provides broader insights into \texttt{TS-ICL} imputation performances.
A detailed description of the \texttt{fm-impute-bench} benchmark used in the main text (\cref{sec:imputation-expe}) is given in \cref{sec:tmlr-benchmark-imputation}, together with complementary results and qualitative visualizations.
\cref{sec:time-benchmark-imputation} further extends the evaluation to a second benchmark, \texttt{TIME} \cite{qiao2026sTIME}, which we adapt to the univariate imputation setting.


\subsection{\texttt{Fm-impute-bench} Benchmark}
\label{sec:tmlr-benchmark-imputation}

\subsubsection{Datasets and Baselines}

\paragraph{Univariate inference datasets.} \cref{tab:tmlr-bench-datasets} details the \textit{univariate datasets} used for the zero-shot imputation experiments in \cref{sec:imputation-expe}. These datasets cover a diverse range of domains, including energy, transport, and climate science, with sampling frequencies varying from 5 minutes to 1 hour (specifically 5, 10, 15, 30, and 60 minutes). The imputation tasks are performed on four-week windows.
When considering the four distinct missingness scenarios, this benchmark represents a large-scale evaluation involving approximately 1.3 million windows to be imputed.

\begin{table}[h!]
\centering
\caption{All datasets used for zero-shot imputation in the \texttt{fm-impute-bench} benchmark under the \textit{univariate setting.}}
\resizebox{\textwidth}{!}{
\begin{tabular}{lccccccr}
\toprule
\multirow{2}{*}{\textbf{Dataset}} & 
\textbf{Release} & 
\multirow{2}{*}{\textbf{Domain}} & 
\multirow{2}{*}{\textbf{Freq}} & 
\textbf{Num.} & 
\textbf{Series} & \textbf{Num.} & \textbf{Window}
\\
&\textbf{Platform}&& & \textbf{Series} & \textbf{Length} & \textbf{Test Windows} & \textbf{Size} \\
\midrule
{BDG2-Bear}         & \href{https://huggingface.co/datasets/Salesforce/lotsa_data}{LOTSA} 
& Energy    & 1H    & 91  & 17544  & 7522  & 672 \\
{BDG2-Rat}          & \href{https://huggingface.co/datasets/Salesforce/lotsa_data}{LOTSA} 
& Energy    & 1H    & 280 & 17544  & 24915 & 672 \\
{Borealis}          & \href{https://huggingface.co/datasets/Salesforce/lotsa_data}{LOTSA} 
& Energy    & 1H    & 15  & 7447   & 77    & 672 \\
{Covid19 Energy}    & \href{https://huggingface.co/datasets/Salesforce/lotsa_data}{LOTSA} 
& Energy    & 1H    & 1   & 31912  & 195   & 672 \\
{GFC12 Load}        & \href{https://huggingface.co/datasets/Salesforce/lotsa_data}{LOTSA} 
& Energy    & 1H    & 20  & 39414  & 4960  & 672 \\
{Hog}               & \href{https://huggingface.co/datasets/Salesforce/lotsa_data}{LOTSA} 
& Energy    & 1H    & 24  & 17544  & 2310  & 672 \\
{Ideal}             & \href{https://huggingface.co/datasets/Salesforce/lotsa_data}{LOTSA} 
& Energy    & 1H    & 217 & 16167  & 156   & 672 \\
{PDB}               & \href{https://huggingface.co/datasets/Salesforce/lotsa_data}{LOTSA} 
& Energy    & 1H    & 1   & 17520  & 96    & 672 \\
{KDD Cup2022}       & \href{https://huggingface.co/datasets/Salesforce/lotsa_data}{LOTSA} 
& Energy    & 10min & 134 & 35279  & 2546  & 4032 \\
{ERA5 geopotential} & \href{https://huggingface.co/datasets/Salesforce/lotsa_data}{LOTSA} 
& Climate   & 1H    & 500 & 8736   & 19000 & 672 \\
{ERA5 humidity}     & \href{https://huggingface.co/datasets/Salesforce/lotsa_data}{LOTSA} 
& Climate   & 1H    & 500 & 8736   & 19000 & 672 \\
{ERA5 temperature}  & \href{https://huggingface.co/datasets/Salesforce/lotsa_data}{LOTSA} 
& Climate   & 1H    & 500 & 8736   & 19000 & 672 \\
{ERA5 wind speed}   & \href{https://huggingface.co/datasets/Salesforce/lotsa_data}{LOTSA} 
& Climate   & 1H    & 500 & 8736   & 19000 & 672 \\
{Oikolab Weather}   & \href{https://huggingface.co/datasets/Salesforce/lotsa_data}{LOTSA} 
& Climate   & 1H    & 8   & 100057 & 5288  & 672 \\
{Pedestrian Counts} & \href{https://huggingface.co/datasets/Salesforce/lotsa_data}{LOTSA} 
& Transport & 1H    & 66  & 96400  & 7733  & 672 \\
{Traffic}           & \href{https://huggingface.co/datasets/Salesforce/lotsa_data}{LOTSA} 
& Transport & 1H    & 861 & 17544  & 83479 & 672 \\
{PEMS BAY}          & \href{https://huggingface.co/datasets/Salesforce/lotsa_data}{LOTSA} 
& Transport & 5min  & 325 & 52128  & 2275  & 8064 \\
{PEMS 03}           & \href{https://huggingface.co/datasets/Salesforce/lotsa_data}{LOTSA} 
& Transport & 5min  & 358 & 26208  & 358   & 8064 \\
{SHMETRO}           & \href{https://huggingface.co/datasets/Salesforce/lotsa_data}{LOTSA} 
& Transport & 15min & 576 & 8809   & 576   & 2688 \\
\midrule
{ETT1-15T}      & \href{https://huggingface.co/datasets/Salesforce/GiftEval/tree/main}{GIFT-eval} 
& Energy    & 15min & 7    & 69680  & 1050  & 2688 \\
{ETT1-1H}       & \href{https://huggingface.co/datasets/Salesforce/GiftEval/tree/main}{GIFT-eval} 
& Energy    & 1H    & 7    & 17420  & 1092  & 672 \\
{ETT2-15T}      & \href{https://huggingface.co/datasets/Salesforce/GiftEval/tree/main}{GIFT-eval} 
& Energy    & 15min & 7    & 69680  & 1050  & 2688 \\
{ETT2-1H}       & \href{https://huggingface.co/datasets/Salesforce/GiftEval/tree/main}{GIFT-eval} 
& Energy    & 1H    & 7    & 17420  & 1092  & 672 \\
{Solar-1H}              & \href{https://huggingface.co/datasets/Salesforce/GiftEval/tree/main}{GIFT-eval} 
& Energy  & 1H & 137  & 8760   & 8768  & 672 \\
{Jena Weather 10T} & \href{https://huggingface.co/datasets/Salesforce/GiftEval/tree/main}{GIFT-eval} 
& Climate   & 10min & 21   & 52704  & 1428  & 4032 \\
{Jena Weather 1H}  & \href{https://huggingface.co/datasets/Salesforce/GiftEval/tree/main}{GIFT-eval} 
& Climate   & 1H    & 21   & 8784   & 1344  & 672 \\
{Loop Seattle 5T}  & \href{https://huggingface.co/datasets/Salesforce/GiftEval/tree/main}{GIFT-eval} 
& Transport & 5min  & 323  & 105120 & 21964 & 8064 \\
{Loop Seattle 1H}  & \href{https://huggingface.co/datasets/Salesforce/GiftEval/tree/main}{GIFT-eval} 
& Transport & 1H    & 323  & 8760   & 20672 & 672 \\
{MDense}        & \href{https://huggingface.co/datasets/Salesforce/GiftEval/tree/main}{GIFT-eval} 
& Transport & 1H    & 30   & 17520  & 4710  & 672 \\
\midrule
{Enedis LDM Small}          & \href{https://zenodo.org/records/15232742}{Zenodo} 
& Energy  & 30min & 500 & 17424 & 20500 & 1344 \\
{London Smart Meters Small} & \href{https://huggingface.co/datasets/autogluon/chronos_datasets/tree/main/monash_london_smart_meters}{Chronos} 
& Energy  & 30min & 500 & 22000 & 25779 & 1344 \\
{Spanish Energy}            & \href{https://www.kaggle.com/datasets/nicholasjhana/energy-consumption-generation-prices-and-weather}{Kaggle} 
& Energy  & 1H    & 9 & 35064 & 1962  & 672 \\
{Weather}                   & \href{https://github.com/zhouhaoyi/Informer2020}{Informer} & Climate  & 1H    & 11 & 35064 & 2398  & 672 \\
\bottomrule
\end{tabular}
}
\label{tab:tmlr-bench-datasets}
\end{table}

\paragraph{Inference datasets with covariates.} \cref{tab:dataset-covar-tmlr-bench} details the six datasets used to evaluate zero-shot imputation with \textit{exogenous covariates}. Following the protocol in \cref{tab:tmlr-bench-datasets}, four-week windows are generated for these experiments. As described in \citep{TSFMImputationBenchmark}, the \texttt{PV} and \texttt{Wind} datasets map regional renewable energy production in 2021 to solar irradiance and wind speed, respectively. In contrast, \texttt{Load-France} tracks national electricity demand using average temperature as the primary covariate to model consumption patterns. When considering the four distinct missingness scenarios, the covariate benchmark represents an evaluation involving approximately 1k windows to be imputed.

\begin{table}[h!]
\centering
\caption{All datasets used for zero-shot imputation in the \texttt{fm-impute-bench} benchmark under the \textit{known-covariate setting.}}
\resizebox{\textwidth}{!}{
\begin{tabular}{lccccccr}
\toprule
\multirow{2}{*}{\textbf{Dataset}} & 
\textbf{Release} & 
\multirow{2}{*}{\textbf{Domain}} & 
\multirow{2}{*}{\textbf{Freq}} & 
\textbf{Target /} & 
\textbf{Series} & \textbf{Num. Test} & \textbf{Window}
\\
&\textbf{Platform}&& & \textbf{Covariate} & \textbf{Length} & \textbf{Windows} & \textbf{Size} \\
\midrule
{PV-OCC} & \href{https://www.rte-france.com/en/eco2mix/download-indicators}{RTE} / \href{https://meteo.data.gouv.fr/datasets/donnees-climatologiques-de-base-horaires/}{Meteo} & Energy & 1H & 1 / 1 & 8760 & 38 & 672 \\
{PV-PACA} & \href{https://www.rte-france.com/en/eco2mix/download-indicators}{RTE} / \href{https://meteo.data.gouv.fr/datasets/donnees-climatologiques-de-base-horaires/}{Meteo} & Energy & 1H & 1 / 1 & 8760 & 38 & 672 \\
{Wind-HDF} & \href{https://www.rte-france.com/en/eco2mix/download-indicators}{RTE} / \href{https://meteo.data.gouv.fr/datasets/donnees-climatologiques-de-base-horaires/}{Meteo} & Energy & 1H & 1 / 1 & 8760 & 38 & 672 \\
{Wind-GE} & \href{https://www.rte-france.com/en/eco2mix/download-indicators}{RTE} / \href{https://meteo.data.gouv.fr/datasets/donnees-climatologiques-de-base-horaires/}{Meteo} & Energy & 1H & 1 / 1 & 8760 & 38 & 672 \\
{Load-France 21} & \href{https://www.rte-france.com/en/eco2mix/download-indicators}{RTE} / \href{https://data.enedis.fr/explore/dataset/donnees-de-temperature-et-de-pseudo-rayonnement/information/}{Enedis} & Energy & 30min & 1 / 1 & 17520 & 41 & 1344 \\
{Load-France 22} & \href{https://www.rte-france.com/en/eco2mix/download-indicators}{RTE} / \href{https://data.enedis.fr/explore/dataset/donnees-de-temperature-et-de-pseudo-rayonnement/information/}{Enedis} & Energy & 30min & 1 / 1 & 17520 & 41 & 1344 \\
\bottomrule
\end{tabular}
}
\label{tab:dataset-covar-tmlr-bench}
\end{table}

\paragraph{Baselines details.}
A brief description of the baselines used in the benchmark is provided below.
\begin{itemize}
\item \texttt{TabPFNv2.5-TS}~\cite{hoo2024tables2time}
is a time series foundation model that adapts the tabular foundation model \texttt{TabPFN} - a transformer-based model pretrained on synthetic supervised-learning tasks for in-context prediction of unseen tabular datasets ~\cite{hollmann2022tabpfn} - to temporal data.
\texttt{TabPFN-TS} leverages a \texttt{TabPFN} regression backbone and reformulates time-series prediction as an in-context tabular regression problem.
Originally proposed for zero-shot forecasting, \texttt{TabPFN-TS} still naturally applies to imputation:
observed target values form the in-context training set, while missing timestamps are treated as query points, enabling the model to impute gaps using all available non-missing observations.
For consistency with \texttt{TabICLv2-TS}, we use the same feature-generation pipeline as in the \texttt{TabICLv2-TS} package: each timestamp is converted into a tabular row with temporal features, automatically extracted seasonal features, and, when available, exogenous covariates \cite{qu2026tabiclv2}.
The pretrained \texttt{TabPFN} regressor is then queried in zero-shot to obtain point predictions.
Our experiments use the \texttt{TabPFNv2.5} regression checkpoint as the backbone, together with the official implementation:
\url{https://github.com/PriorLabs/tabpfn-time-series}.

\item Similarly to \texttt{TabPFNv2.5-TS}, \texttt{TabICLv2-TS}
is a foundation model for time series analysis adapted from the \texttt{TabICLv2}~\cite{qu2026tabiclv2} tabular foundation model, designed for scalable in-context learning on regression and classification tasks.
We use the package-provided time-series transformation pipeline to construct the tabular representation.
\texttt{TabICLv2} then performs regression by conditioning on the resulting table in a single in-context inference procedure. 
In our experiments, we use the \texttt{TabICLv2} implementation and forecasting utilities from the official release:
\url{https://github.com/soda-inria/tabicl}.

\item \texttt{Linear}
imputes missing values by linearly interpolating between the closest observed neighbors surrounding the gap.
If a gap has no future (resp., past) anchor, it falls fack to \texttt{NOCB}, next-observation-carried-backward (resp., \texttt{LOCF}, last-observation-carried-forward).

\item \texttt{Seasonal Naive}
imputes a missing value at timestamp $t$ by repeating the observation from the previous seasonal period, i.e., the value at $t-S$.
$S$ is pre-defined for each dataset based on its dominant frequency (e.g., daily). If the value at $t-S$ is also missing, the method sequentially searches for an available observation at other seasonal timestamps (e.g., $t+S$, then $t-2S$, etc.).
The method falls back to \texttt{LOCF} in case this search fails.

\item \texttt{LOCF}
imputes a missing value by copying the most recent available past value. 

\item \texttt{SAITS}~\cite{du2023saits}
is a supervised Transformer-based imputation model designed for partially observed multivariate time series.
It uses two diagonally-masked self-attention blocks to capture temporal and cross-variable dependencies, whose outputs are combined through a learned gating mechanism to reconstruct missing values.

\item \texttt{BRITS}~\cite{cao2018brits}
is a supervised recurrent imputation model based on bidirectional RNNs with learned temporal decay.
It processes each window forward and backward to account for irregular gaps, jointly estimating hidden states and missing values while encouraging consistency between directions.

\end{itemize}

\paragraph{Task specific baseline detailed training.}
The two supervised baselines, \texttt{SAITS} and \texttt{BRITS}, are trained on the training split of \texttt{fm-impute-bench} (see \citep{TSFMImputationBenchmark} for more details).
Both implementations are taken from the \texttt{PyPOTS} toolbox~\cite{du2023pypots} and trained on fixed-length windows with a masked-reconstruction objective: a subset of observed entries is randomly masked, and the model reconstructs these values from the remaining observations and the corresponding binary observation mask.
Inputs are z-score normalized per variable, training minimizes MAE on the artificially masked positions, and model selection is performed using validation MSE with early stopping.
We use the default hyperparameter configurations recommended by the original authors, the Adam optimizer~\cite{kingma2015adam}, a batch size of 64, at most 50 training epochs, and early stopping with patience 5.
Since \texttt{SAITS} and \texttt{BRITS} produce pointwise imputations, we adapt them to quantile-based evaluation by replicating each point prediction across all requested quantile levels.

\subsubsection{Extended Results}

\begin{figure}[h!]
    \centering
    \begin{subfigure}[b]{0.495\textwidth}
        \centering
        \includegraphics[width=\linewidth]{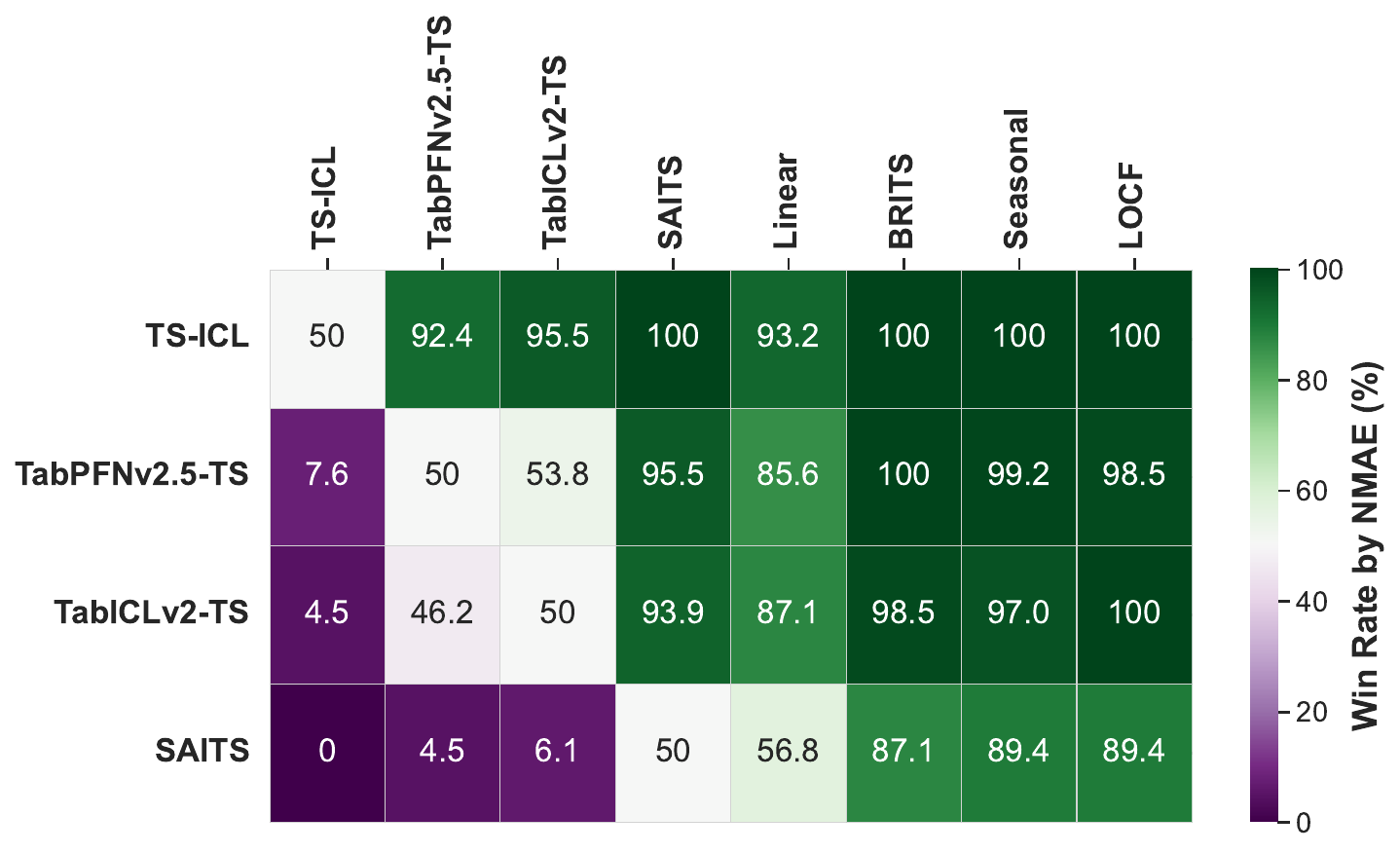}
        \caption{\textit{Univariate} imputation across 132 tasks.}
        \label{fig:Univariate-imputation-annexe-wins-tmlr}
    \end{subfigure}
    \hfill
    \begin{subfigure}[b]{0.495\textwidth}
        \centering
        \includegraphics[width=\linewidth]{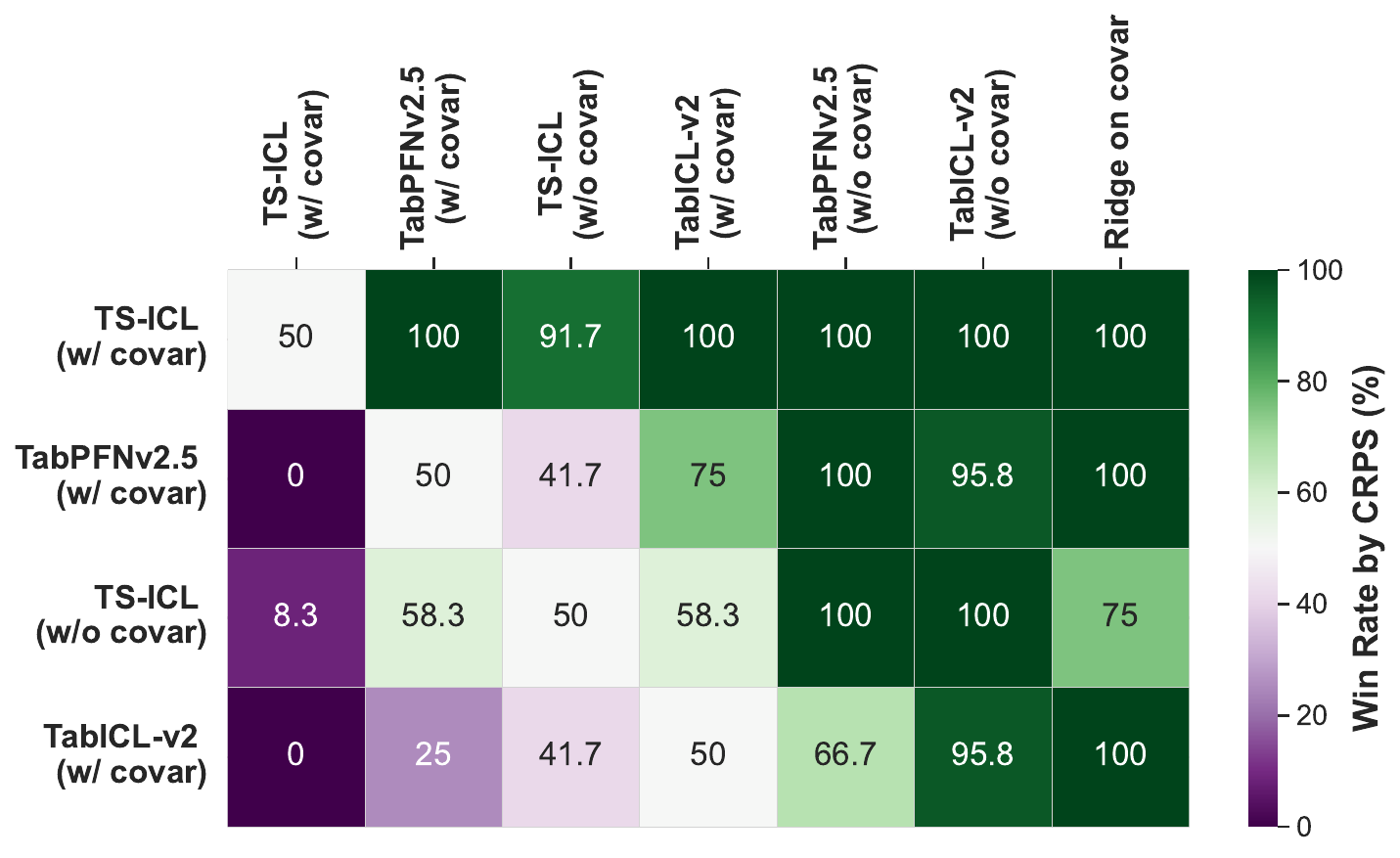}
        \caption{Imputation with \textit{known covariates} across 24 tasks.}
        \label{fig:covariates-imputation-annexe-wins-tmlr}
    \end{subfigure}
    \caption{Pairwise win rates of the top-4 models for imputation on the \texttt{fm-impute} benchmark. Each entry indicates the fraction of tasks where a method outperforms another according to the NMAE.}
    \label{fig:imputations-wins-annexe-tmlr}
\end{figure}

This section extends the empirical evaluation in~\cref{sec:imputation-expe} with a more detailed analysis of imputation performance across all experimental settings. Specifically, we provide:

\begin{itemize}
\item \textbf{Aggregated detailed performance tables:} We report the average NMAE and CRPS (metrics definition in \cref{sec:scores-metrics}) across the 132 \textit{univariate} tasks and 24 \textit{covariate-aware} tasks of \texttt{fm-impute-bench}. These results, detailed in \cref{tab:imputation-detailed-metrics-univariate-TMLR} and \cref{tab:imputation-detailed-metrics-covariates-TMLR}, provide a detailed view of both point-wise and probabilistic performance. It is important to note that metrics are aggregated across tasks using the arithmetic mean, following the evaluation protocol established in \texttt{fm-impute-bench} \citep{TSFMImputationBenchmark}. 

\item \textbf{NMAE pairwise win rates:} To complement the CRPS-based win rate diagrams presented in the main text \cref{fig:imputations-wins-tmlr}, we include the corresponding pairwise win rate visualizations in terms of NMAE for both \textit{univariate} and \textit{known-covariate} experiments in \cref{fig:imputations-wins-annexe-tmlr}. The NMAE-based pairwise comparisons provide additional evidence of the robustness of \texttt{TS-ICL}, showing consistent superiority regardless of the chosen accuracy metric.
\end{itemize}

\begin{table}[h]
\caption{Aggregated imputation metrics on the 132 tasks of the \textit{univariate setting} in \texttt{fm-impute-bench} (mean $\pm$ std). Best in \textbf{bold}.}
\centering
\scalebox{0.62}{
\begin{tabular}{lcccccccc}
\toprule
 & {TSFM} & \multicolumn{2}{c}{Tabular Foundation Models} & \multicolumn{2}{c}{Task Specific Models} & \multicolumn{3}{c}{Local Models} \\
\cmidrule(r){2-2} \cmidrule(r){3-4} \cmidrule(r){5-6} \cmidrule(r){7-9} 
 & \texttt{TS-ICL} & \texttt{TabPFNv2.5-TS} & \texttt{TabICLv2-TS} 
 & \texttt{SAITS} & \texttt{BRITS}
 & \shortstack{\texttt{Linear} \\ \texttt{interp.}} 
 & \shortstack{\texttt{Seasonal} \\ \texttt{Naive}} 
 & \texttt{LOCF} \\
\midrule
NMAE ($\downarrow$) & \textbf{0.243 $\pm$ 0.118} & 0.296 $\pm$ 0.145 & 0.294 $\pm$ 0.136 & 0.386 $\pm$ 0.140 & 0.470 $\pm$ 0.181 & 0.507 $\pm$ 0.287 & 0.580 $\pm$ 0.177 & 0.612 $\pm$ 0.255 \\
CRPS ($\downarrow$) & \textbf{0.255 $\pm$ 0.137} & 0.303 $\pm$ 0.156 & 0.301 $\pm$ 0.148 & 0.503 $\pm$ 0.193 & 0.605 $\pm$ 0.227 & 0.658 $\pm$ 0.377 & 0.750 $\pm$ 0.230 & 0.793 $\pm$ 0.335 \\
\bottomrule
\end{tabular}}
\label{tab:imputation-detailed-metrics-univariate-TMLR}
\end{table}

\begin{table}[h]
\caption{Aggregated imputation performance metrics across the 24 tasks of the \textit{known covariates} setting in \texttt{fm-impute-bench} (mean $\pm$ std). Best in \textbf{bold}.}
\centering
\scalebox{0.72}{
\begin{tabular}{lccccccc}
\toprule
 & \multicolumn{2}{c}{TSFM} & \multicolumn{4}{c}{Tabular Foundation Models} & Local Model \\
\cmidrule(r){2-3} \cmidrule(r){4-7} \cmidrule(r){8-8}
 & \multicolumn{2}{c}{\texttt{TS-ICL}} & \multicolumn{2}{c}{\texttt{TabPFNv2.5-TS}} & \multicolumn{2}{c}{\texttt{TabICLv2-TS}} & \shortstack{\texttt{Ridge on} \\ \texttt{Covar}} \\
\cmidrule(r){2-3} \cmidrule(r){4-5} \cmidrule(r){6-7} \cmidrule(r){8-8}
 & \texttt{w/ covar} & \texttt{w/o covar} & \texttt{w/ covar} & \texttt{w/o covar} & \texttt{w/ covar} & \texttt{w/o covar} & \texttt{w/ covar} \\
\midrule
NMAE ($\downarrow$) & \textbf{0.077 $\pm$ 0.053} & 0.125 $\pm$ 0.113 & 0.121 $\pm$ 0.081 & 0.202 $\pm$ 0.158 & 0.141 $\pm$ 0.099 & 0.206 $\pm$ 0.136 & 0.388 $\pm$ 0.253 \\
CRPS ($\downarrow$) & \textbf{0.074 $\pm$ 0.047} & 0.119 $\pm$ 0.100 & 0.115 $\pm$ 0.074 & 0.196 $\pm$ 0.147 & 0.134 $\pm$ 0.092 & 0.199 $\pm$ 0.125 & 0.471 $\pm$ 0.306 \\
\bottomrule
\end{tabular}}
\label{tab:imputation-detailed-metrics-covariates-TMLR}
\end{table}


\subsubsection{Qualitative Analysis and Visualizations}

\begin{figure}
\centering
\begin{subfigure}[b]{0.87\textwidth}
\centering
\includegraphics[width=\linewidth]{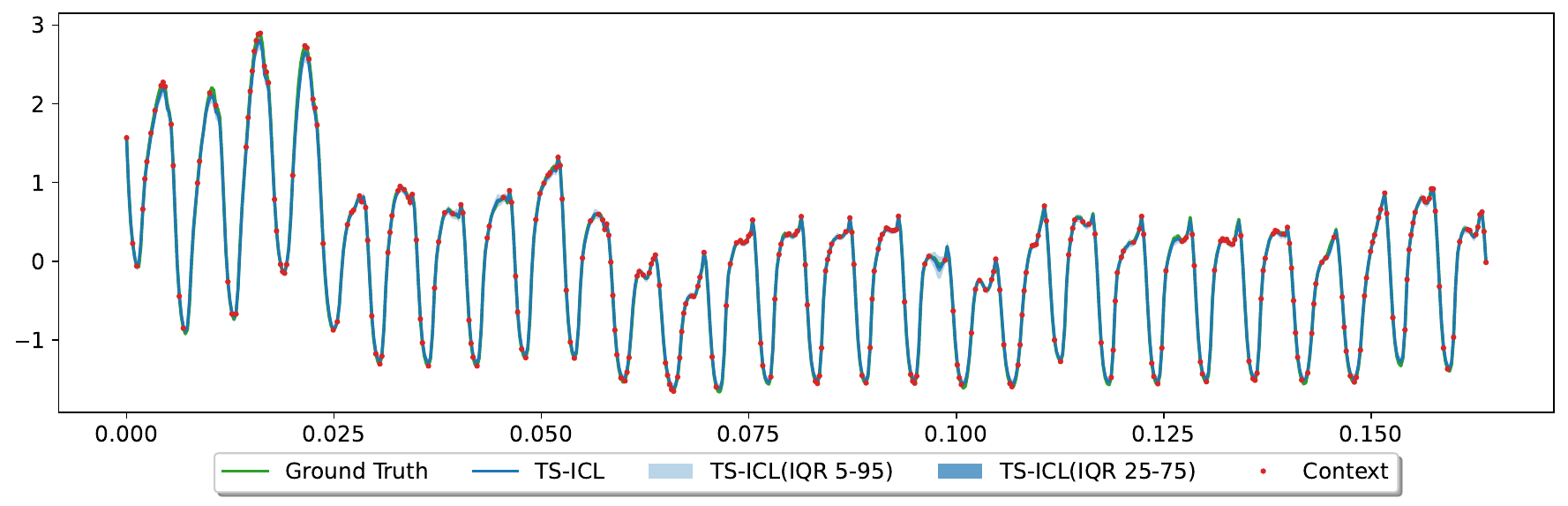}
\caption{\emph{PDB}, 50\% missing values.}
\label{fig:plot_pdb}
\end{subfigure}

\begin{subfigure}[b]{0.87\textwidth}
\centering
\includegraphics[width=\linewidth]{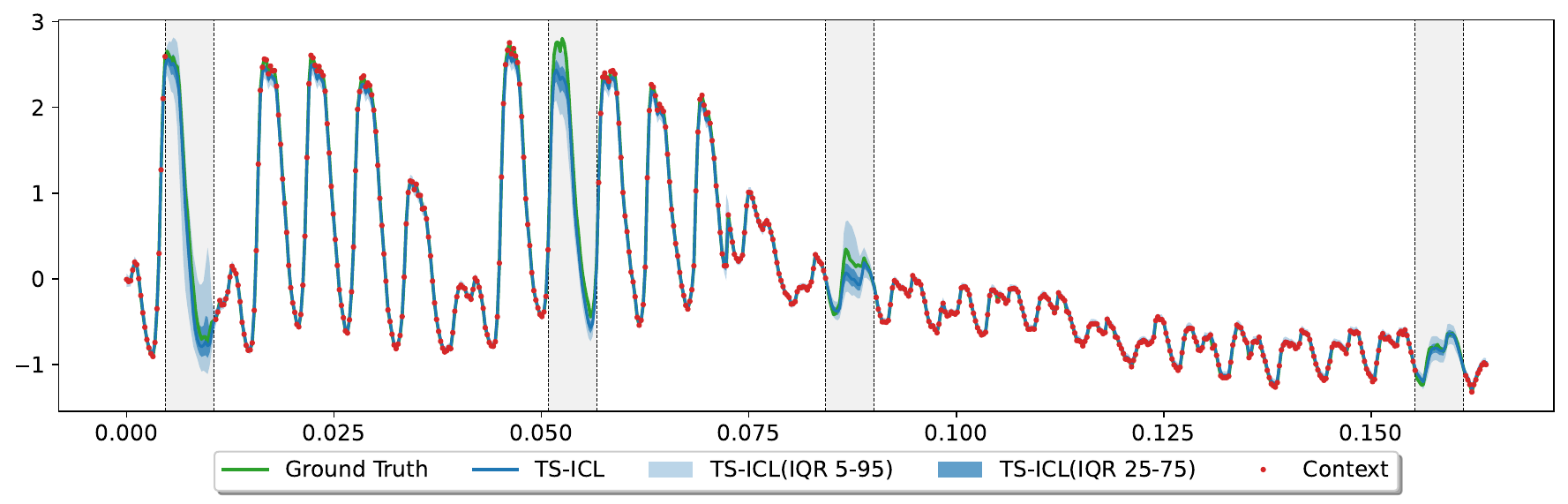}
\caption{\emph{Covid19 Energy}, four one-day missing blocks.}
\label{fig:plot_covid}
\end{subfigure}

\begin{subfigure}[b]{0.87\textwidth}
\centering
\includegraphics[width=\linewidth]{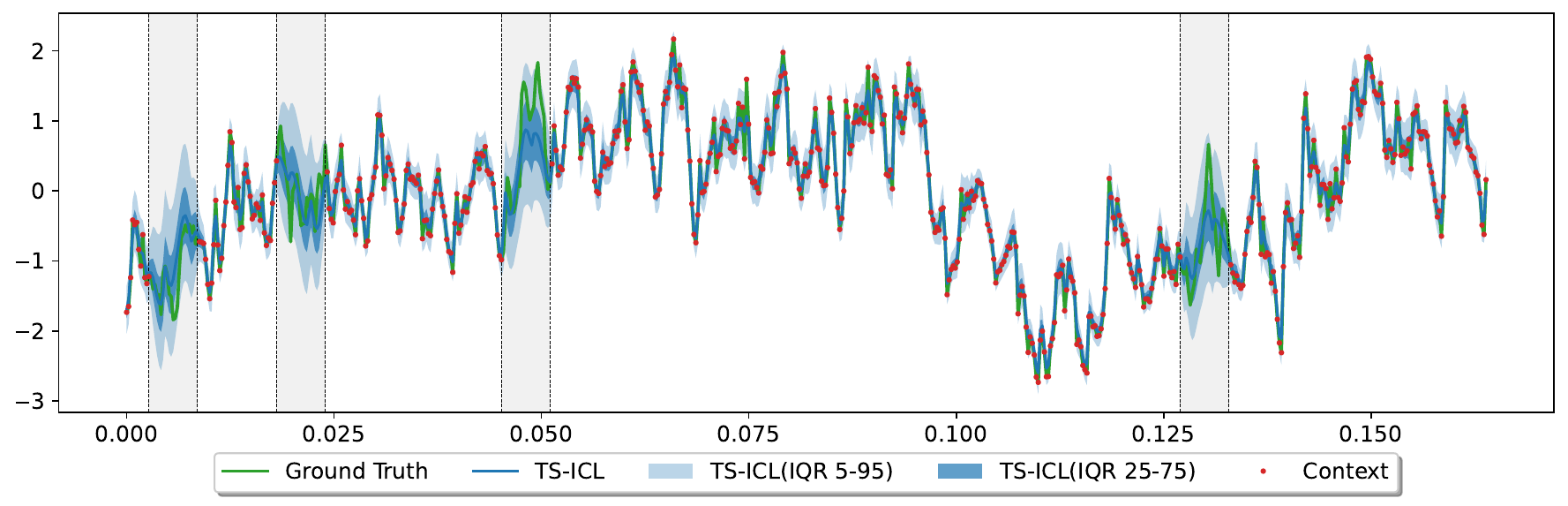}
\caption{\emph{BDG2-Rat}, four one-day missing blocks.}
\label{fig:plot_rat}
\end{subfigure}

\begin{subfigure}[b]{0.87\textwidth}
\centering
\includegraphics[width=\linewidth]{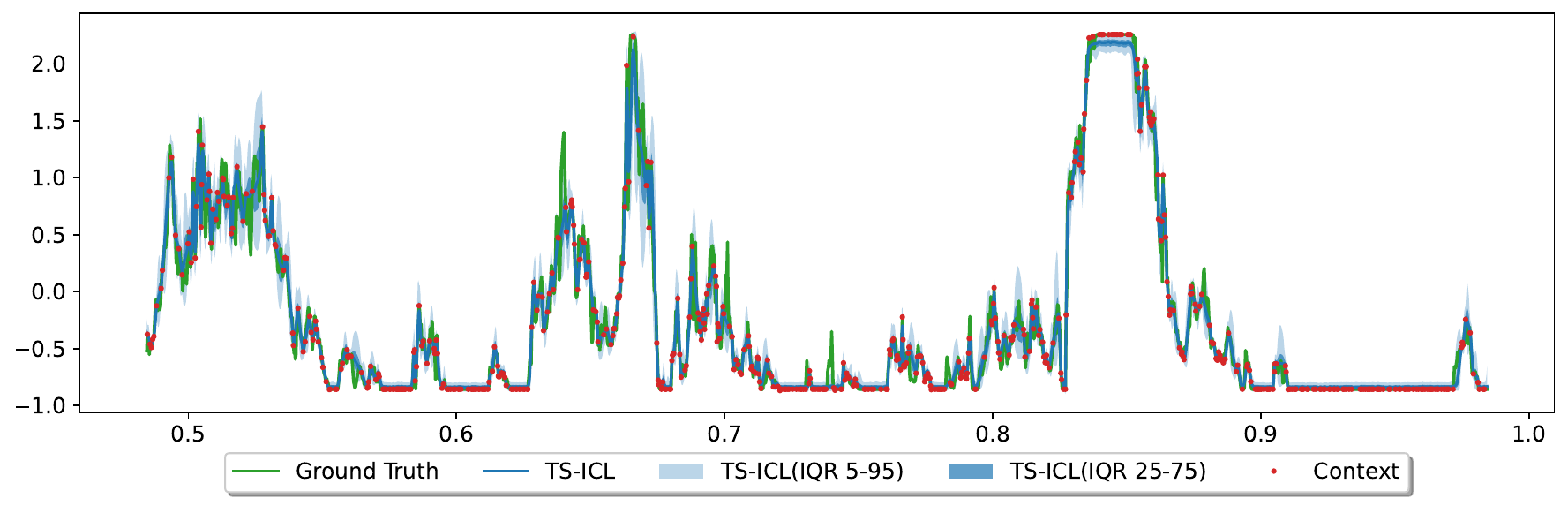}
\caption{\emph{KDD Cup2022}, 70\% missing values.}
\label{fig:plot_kdd}
\end{subfigure}

\begin{subfigure}[b]{0.87\textwidth}
\centering
\includegraphics[width=\linewidth]{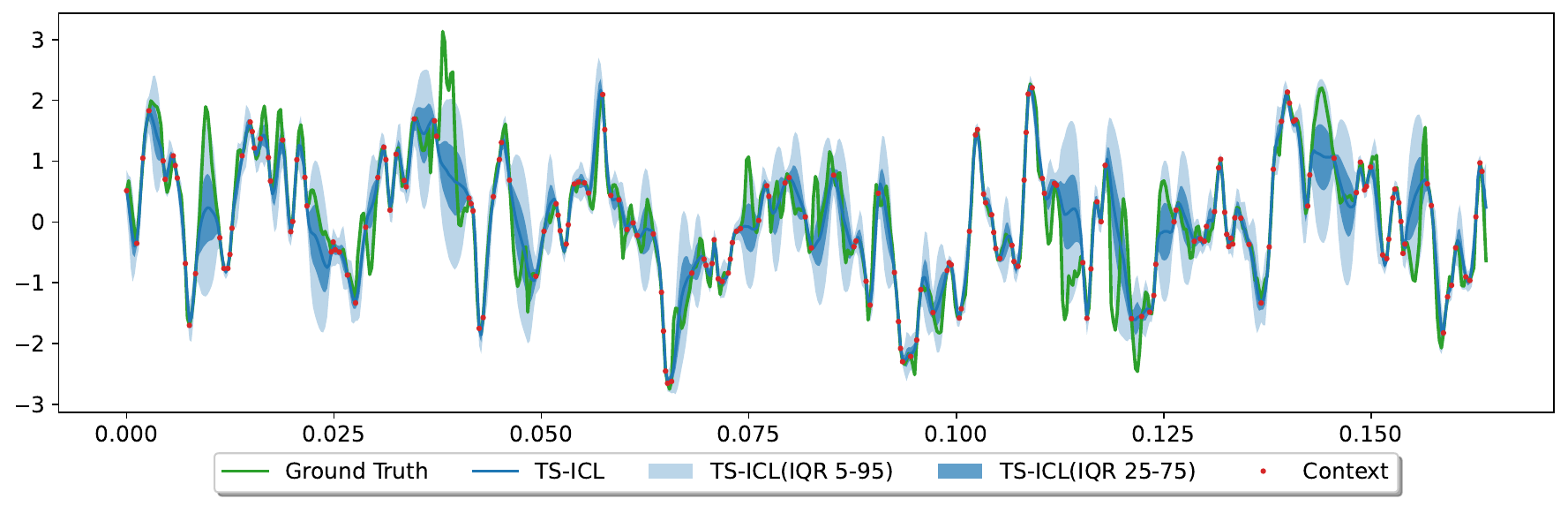}
\caption{\emph{ERA5 wind speed}, 70\% missing values.}
\label{fig:plot_wind}
\end{subfigure}

\caption{
Qualitative assessment of \texttt{TS-ICL} imputations on the \texttt{fm-impute-bench} benchmark.
}
\label{fig:tmlr-plots-1}
\end{figure}

\begin{figure}
\centering
\begin{subfigure}[b]{0.87\textwidth}
\centering
\includegraphics[width=\linewidth]{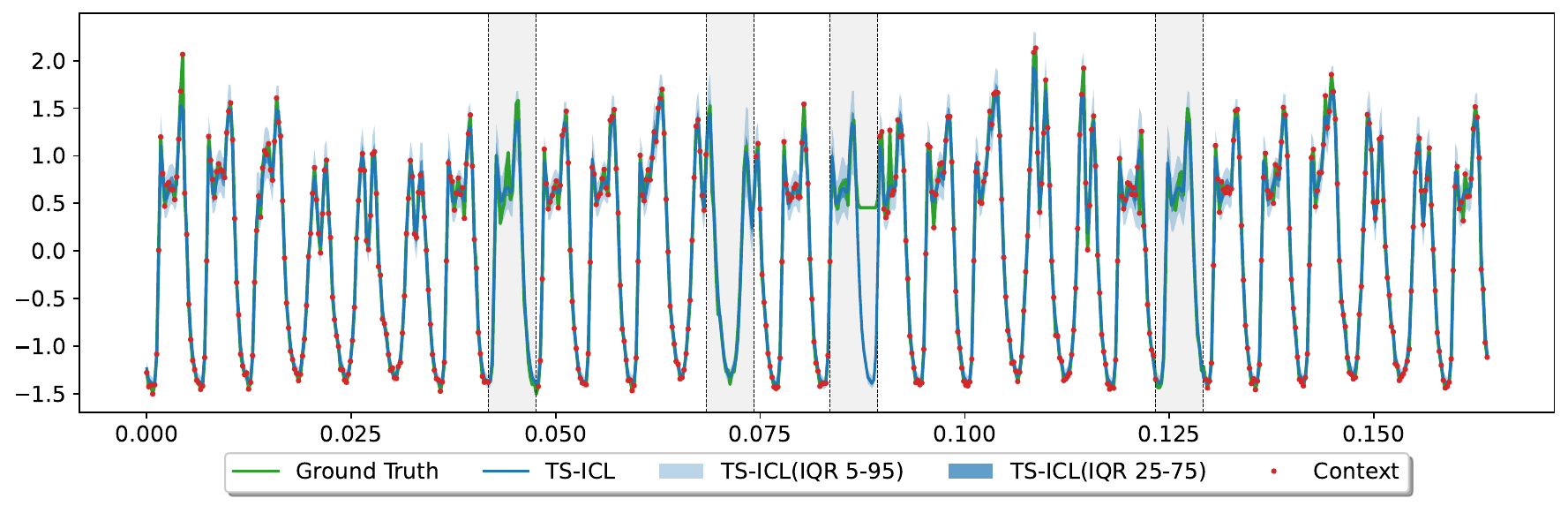}
\caption{\emph{MDense}, four one-day missing blocks.}
\label{fig:plot_mdense}
\end{subfigure}

\begin{subfigure}[b]{0.87\textwidth}
\centering
\includegraphics[width=\linewidth]{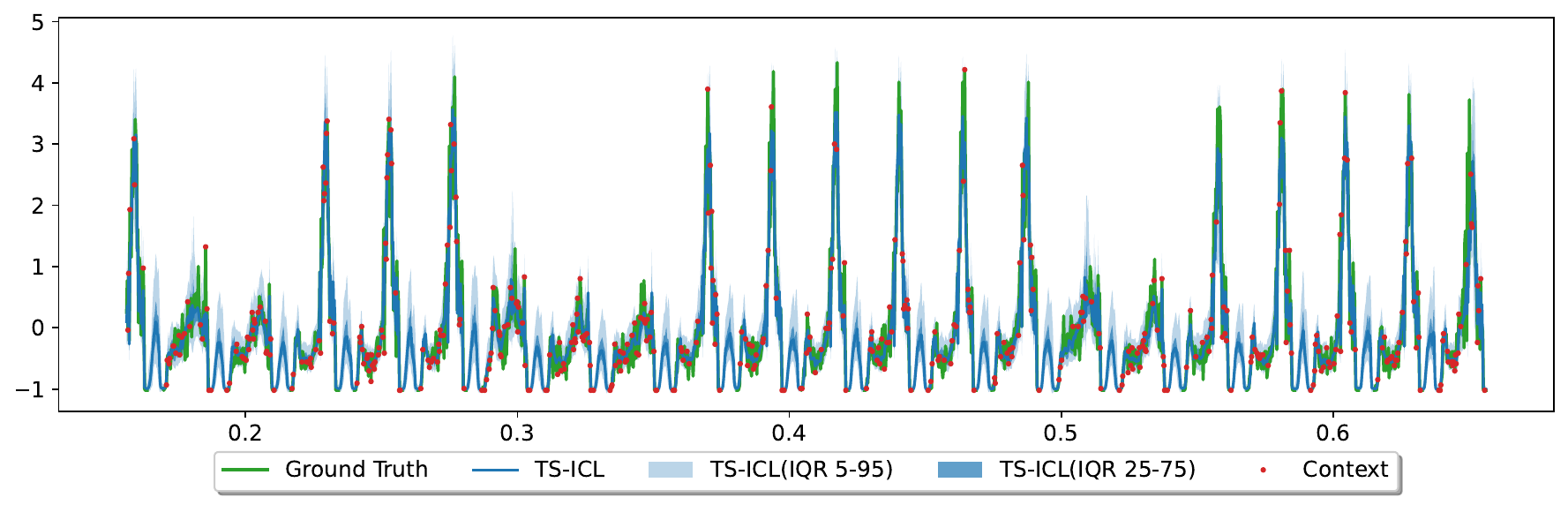}
\caption{\emph{SHMETRO}, 70\% missing values.}
\label{fig:plot_shmetro}
\end{subfigure}

\begin{subfigure}[b]{0.87\textwidth}
\centering
\includegraphics[width=\linewidth]{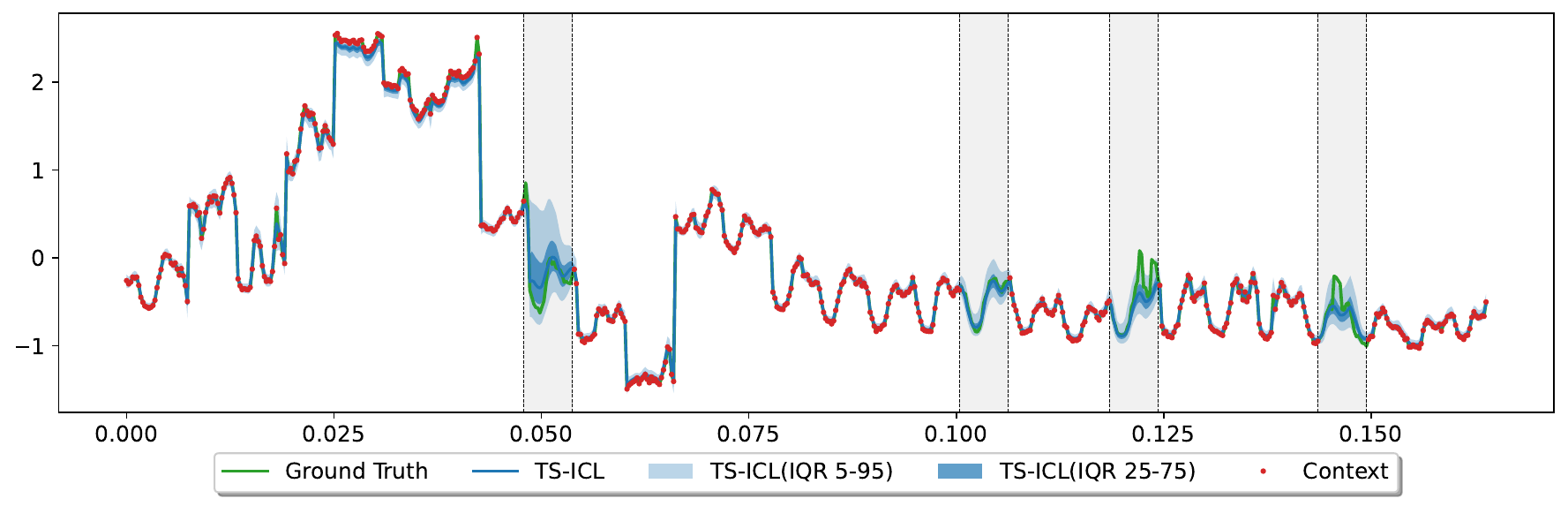}
\caption{\emph{Spanish Energy}, four one-day missing blocks.}
\label{fig:plot_spanishe}
\end{subfigure}

\begin{subfigure}[b]{0.87\textwidth}
\centering
\includegraphics[width=\linewidth]{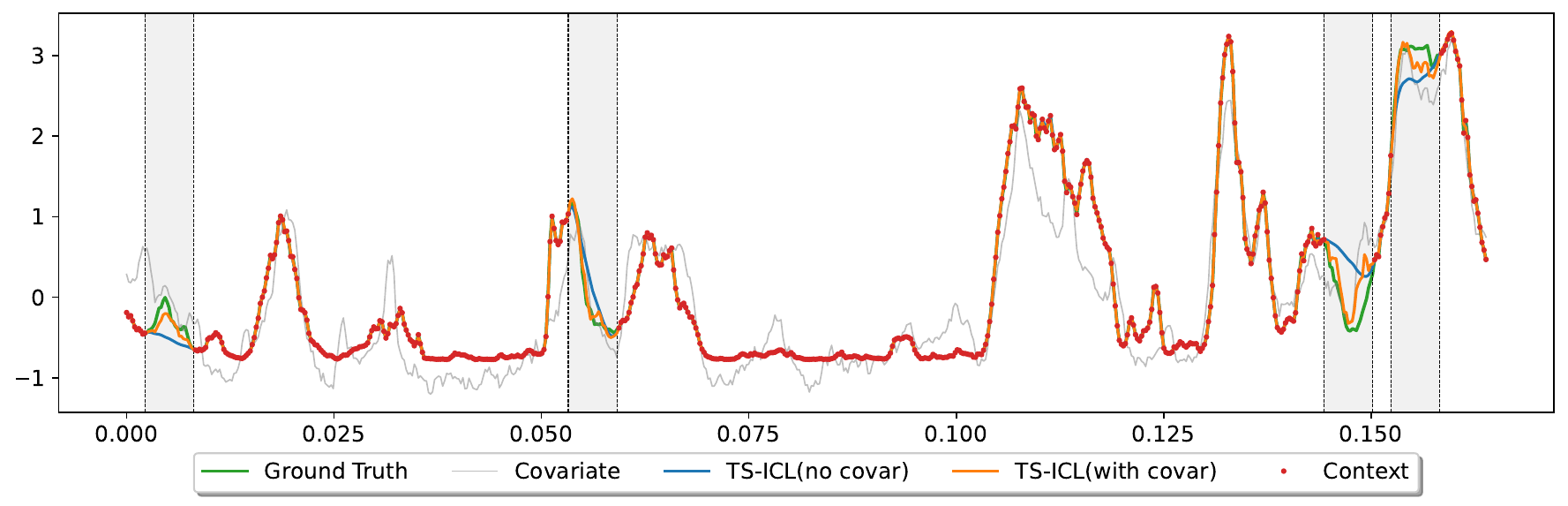}
\caption{\emph{Wind-GE}, four one-day missing blocks.}
\label{fig:plot_covar}
\end{subfigure}

\begin{subfigure}[b]{0.87\textwidth}
\centering
\includegraphics[width=\linewidth]{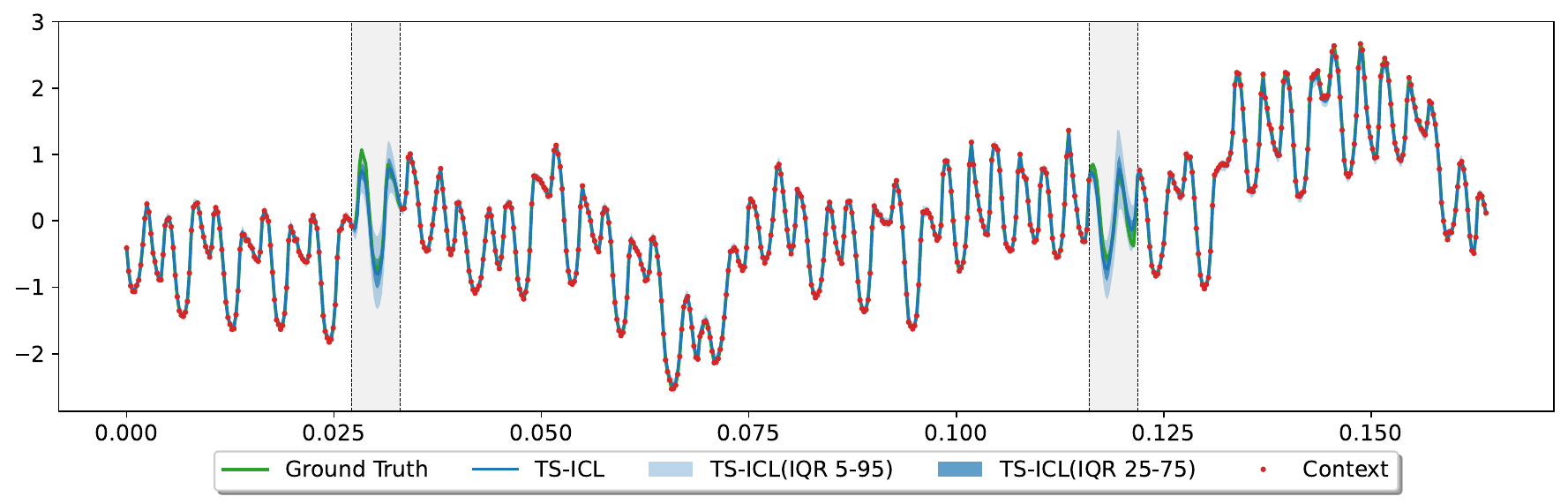}
\caption{\emph{GFC12 Load}, two one-day missing blocks.}
\label{fig:plot_gfc12}
\end{subfigure}
\caption{
Qualitative assessment of \texttt{TS-ICL} imputations on the \texttt{fm-impute-bench} benchmark (continued).
}
\label{fig:tmlr-plots-2}
\end{figure}

This section presents visual examples of \texttt{TS-ICL} imputations for both \textit{univariate} and \textit{known-covariate} settings.
We illustrate in \cref{fig:tmlr-plots-1} and \cref{fig:tmlr-plots-2} the model imputation capabilities across various missingness patterns covering pointwise and blockwise scenarios.

\paragraph{Results.}
Several observations emerge from these plots.
\begin{enumerate}[(i)]
\item With rich context, \texttt{TS-ICL} provides accurate reconstructions of smooth patterns, with tight inter-quantile ranges (\cref{fig:plot_pdb}).
On the contrary, \texttt{TS-ICL} adjusts its uncertainty estimation of sparsely observed yet regular patterns (\cref{fig:plot_shmetro,fig:plot_gfc12}).

\item \texttt{TS-ICL} adapts well to distribution shifts (\cref{fig:plot_covid,fig:plot_spanishe}).

\item \texttt{TS-ICL} tends to provide smooth reconstructions of sparse high-frequency signals, with higher interquantile ranges accomodating their stronger variability (\cref{fig:plot_rat,fig:plot_kdd,fig:plot_wind}).

\item \cref{fig:plot_mdense} (third block) suggests that \texttt{TS-ICL} can serve as a counterfactual estimator, replacing unusual sequences with expected ones under regular conditions.

\item Finally, the ability of \texttt{TS-ICL} to incorporate covariate information at inference is key to produce meaningful imputations in challenging scenarios (\cref{fig:plot_covar}).
\end{enumerate}

\clearpage


\subsection{\texttt{TIME} Benchmark}
\label{sec:time-benchmark-imputation}

In this section, we evaluate the zero-shot imputation capability of \texttt{TS-ICL} on \texttt{TIME} \citep{qiao2026sTIME} a recently introduced benchmark originally designed for TSFM forecasting. We adapt it to cover imputation for the \textit{univariate setting}.
Performance is assessed across diverse missingness patterns, sequence lengths, and application domains (details in \cref{tab:datasets-time-imputation}).

\begin{table}[h!]
\centering
\caption{All datasets used for zero-shot imputation in the \texttt{TIME} benchmark.}
\resizebox{\textwidth}{!}{
\begin{tabular}{ll l ccc r cc cc cc}
\toprule
\multirow{3}{*}{\textbf{Dataset}} & 
\multirow{3}{*}{\makecell[l]{\textbf{Release} \\ \textbf{Platform}}} & 
\multirow{3}{*}{\textbf{Domain}} & 
\multirow{3}{*}{\textbf{Freq}} & 
\multirow{3}{*}{\makecell[c]{\textbf{Num.} \\ \textbf{Series}}} & 
\multirow{3}{*}{\makecell[c]{\textbf{Num.} \\ \textbf{Variate}}} & 
\multirow{3}{*}{\makecell[c]{\textbf{Avg Series} \\ \textbf{Length}}} & 
\multicolumn{2}{c}{\textbf{Short-term}} & 
\multicolumn{2}{c}{\textbf{Med-term}} & 
\multicolumn{2}{c}{\textbf{Long-term}} 
\\
\cmidrule(lr){8-9} \cmidrule(lr){10-11} \cmidrule(lr){12-13}
& & & & & & & \makecell[c]{\textbf{Window} \\ \textbf{Size}} & \makecell[c]{\textbf{Num. Test} \\ \textbf{Windows}} & \makecell[c]{\textbf{Window} \\ \textbf{Size}} & \makecell[c]{\textbf{Num. Test} \\ \textbf{Windows}} & \makecell[c]{\textbf{Window} \\ \textbf{Size}} & \makecell[c]{\textbf{Num. Test} \\ \textbf{Windows}} \\
\midrule
Water Quality-Darwin & \href{https://apps.aims.gov.au/metadata/view/23257155-fa16-4361-ae82-b2a09e4bf9ac}{IMOS} & Nature & 15T & 7 & 6 & 15,229 & 256 & 3,780 & 1024 & 630 & 4096 & 210 \\
Current Velocity & \href{https://catalogue-imos.aodn.org.au/geonetwork/srv/eng/catalog.search\#/metadata/ae86e2f5-eaaf-459e-a405-e654d85adb9c}{IMOS} & Nature & 5T & 1 & 6 & 26,486 & 256 & 720 & 1024 & 90 & 4096 & 30 \\
Current Velocity & \href{https://catalogue-imos.aodn.org.au/geonetwork/srv/eng/catalog.search\#/metadata/ae86e2f5-eaaf-459e-a405-e654d85adb9c}{IMOS} & Nature & 10T & 10 & 6 & 20,669 & 256 & 7,200 & 1024 & 900 & 4096 & 300 \\
Current Velocity & \href{https://catalogue-imos.aodn.org.au/geonetwork/srv/eng/catalog.search\#/metadata/ae86e2f5-eaaf-459e-a405-e654d85adb9c}{IMOS} & Nature & 15T & 5 & 6 & 8,503 & 256 & 3,600 & 1024 & 450 & 4096 & 150 \\
Current Velocity & \href{https://catalogue-imos.aodn.org.au/geonetwork/srv/eng/catalog.search\#/metadata/ae86e2f5-eaaf-459e-a405-e654d85adb9c}{IMOS} & Nature & 20T & 27 & 6 & 6,460 & 256 & 19,440 & 1024 & 2,430 & 4096 & 810 \\
Current Velocity & \href{https://catalogue-imos.aodn.org.au/geonetwork/srv/eng/catalog.search\#/metadata/ae86e2f5-eaaf-459e-a405-e654d85adb9c}{IMOS} & Nature & H & 21 & 6 & 3,502 & 256 & 3,528 & 1024 & 504 & 4096 & 252 \\
CPHL & \href{https://catalogue-imos.aodn.org.au/geonetwork/srv/eng/catalog.search\#/metadata/8964658c-6ee1-4015-9bae-2937dfcc6ab9}{IMOS} & Nature & 15T & 2 & 1 & 10,831 & 256 & 240 & 1024 & 30 & 4096 & 10 \\
CPHL & \href{https://catalogue-imos.aodn.org.au/geonetwork/srv/eng/catalog.search\#/metadata/8964658c-6ee1-4015-9bae-2937dfcc6ab9}{IMOS} & Nature & 30T & 2 & 1 & 14,687 & 256 & 240 & 1024 & 60 & 4096 & 20 \\
CPHL & \href{https://catalogue-imos.aodn.org.au/geonetwork/srv/eng/catalog.search\#/metadata/8964658c-6ee1-4015-9bae-2937dfcc6ab9}{IMOS} & Nature & H & 4 & 1 & 4,971 & 256 & 112 & 1024 & 16 & 4096 & 8 \\
Coastal T-S & \href{https://catalogue-imos.aodn.org.au/geonetwork/srv/eng/catalog.search\#/metadata/7e13b5f3-4a70-4e31-9e95-335efa491c5c}{IMOS} & Nature & 5T & 18 & 3 & 68,604 & 256 & 6,480 & 1024 & 810 & 4096 & 270 \\
Coastal T-S & \href{https://catalogue-imos.aodn.org.au/geonetwork/srv/eng/catalog.search\#/metadata/7e13b5f3-4a70-4e31-9e95-335efa491c5c}{IMOS} & Nature & 15T & 5 & 3 & 20,870 & 256 & 1,800 & 1024 & 225 & 4096 & 75 \\
Coastal T-S & \href{https://catalogue-imos.aodn.org.au/geonetwork/srv/eng/catalog.search\#/metadata/7e13b5f3-4a70-4e31-9e95-335efa491c5c}{IMOS} & Nature & 20T & 1 & 3 & 8,198 & 256 & 360 & 1024 & 45 & 4096 & 15 \\
Coastal T-S & \href{https://catalogue-imos.aodn.org.au/geonetwork/srv/eng/catalog.search\#/metadata/7e13b5f3-4a70-4e31-9e95-335efa491c5c}{IMOS} & Nature & H & 24 & 3 & 5,489 & 256 & 2,016 & 1024 & 288 & 4096 & 144 \\
SG Weather & \href{https://data.gov.sg}{data.gov.sg} & Nature & D & 6 & 4 & 2,953 & 256 & 2,928 & 1024 & 1,272 & 4096 & 648 \\
SG PM 2.5 & \href{https://data.gov.sg/datasets/d_e1058d6974c877257e32048ab128ad83/view\#tag/default/GET/pm25}{data.gov.sg} & Nature & H & 1 & 5 & 38,688 & 256 & 460 & 1024 & 150 & 4096 & 65 \\
NE China Wind & \href{https://github.com/Zhang-zongwei/MFWPN}{Github} & Nature & H & 1 & 4 & 8,764 & 256 & 120 & 1024 & 40 & 4096 & 16 \\
\midrule
Australia Solar & \href{https://www.pvoutput.org}{Pvoutput} & Energy & H & 1 & 3 & 35,064 & 256 & 315 & 1024 & 105 & 4096 & 45 \\
EPF Electricity & \href{https://github.com/jeslago/epftoolbox}{Academic} & Energy & H & 5 & 1 & 52,416 & 256 & 525 & 1024 & 175 & 4096 & 75 \\
OpenElectricity & \href{https://docs.openelectricity.org}{OpenElec} & Energy & 5T & 1 & 10 & 43,488 & 256 & 1,680 & 1024 & 420 & 4096 & 140 \\
EWELD Load & \href{https://pmc.ncbi.nlm.nih.gov/articles/PMC10495315/}{Academic} & Energy & 15T & 1 & 10 & 20,544 & 256 & 560 & 1024 & 140 & 4096 & 20 \\
SG Carpark & \href{https://data.gov.sg/datasets/d_ca933a644e55d34fe21f28b8052fac63/view}{data.gov.sg} & Transport & 15T & 354 & 1 & 14,332 & 256 & 14,868 & 1024 & 2,478 & 4096 & 354 \\
Finland Traffic & \href{https://www.digitraffic.fi/en/}{Digitraffic} & Transport & 15T & 1 & 1 & 35,136 & 256 & 186 & 1024 & 31 & 4096 & 4 \\
\midrule
Port Activity & \href{https://www.kaggle.com/datasets/arunvithyasegar/daily-port-activity-data-and-trade-estimates}{Competition} & Transport & D & 99 & 2 & 2,127 & 256 & 2,376 & & & & \\
Port Activity & \href{https://www.kaggle.com/datasets/arunvithyasegar/daily-port-activity-data-and-trade-estimates}{Competition} & Transport & W & 99 & 2 & 304 & 256 & 792 & & & & \\
ECDC COVID & \href{https://www.ecdc.europa.eu/en/publications-data/download-data-hospital-and-icu-admission-rates-and-current-occupancy-covid-19}{ECDC} & Healthcare & D & 9 & 1 & 1,117 & 256 & 45 & & & & \\
ECDC COVID & \href{https://www.ecdc.europa.eu/en/publications-data/download-data-hospital-and-icu-admission-rates-and-current-occupancy-covid-19}{ECDC} & Healthcare & W & 16 & 1 & 165 & 256 & 64 & & & & \\
Global Influenza & \href{https://www.who.int/tools/flunet}{WHO} & Healthcare & W & 15 & 4 & 205 & 256 & 240 & & & & \\
\midrule
Crypto & \href{https://fred.stlouisfed.org/categories/33913}{FRED} & Finance & D & 1 & 4 & 2,842 & 256 & 36 & & & & \\
US Term Structure & \href{https://fred.stlouisfed.org/categories/33825}{FRED} & Finance & B & 1 & 40 & 9,327 & 256 & 1,400 & & & & \\
Oil Price & \href{https://fred.stlouisfed.org/categories/32217}{FRED} & Finance & B & 1 & 12 & 5,035 & 256 & 420 & & & & \\
\midrule
Job Claims & \href{https://fred.stlouisfed.org/categories/32240}{FRED} & Finance & W & 1 & 2 & 196 & 256 & 8 & & & & \\
Uncertainty-1M & \href{https://fred.stlouisfed.org/categories/33201}{FRED} & Economics & M & 1 & 3 & 780 & 256 & 21 & & & & \\
Housing Inventory & \href{https://fred.stlouisfed.org/categories/97}{FRED} & Economics & M & 1 & 4 & 114 & 256 & 12 & & & & \\
JOLTS & \href{https://fred.stlouisfed.org/categories/32241}{FRED} & Economics & M & 1 & 6 & 297 & 256 & 30 & & & & \\
US Labor & \href{https://fred.stlouisfed.org/categories/12}{FRED} & Economics & M & 1 & 14 & 380 & 256 & 70 & & & & \\
Vehicle Supply & \href{https://fred.stlouisfed.org/categories/32993}{FRED} & Economics & M & 1 & 6 & 391 & 256 & 30 & & & & \\
Auto Production-SF & \href{https://fred.stlouisfed.org/categories/33938}{FRED} & Economics & M & 1 & 1 & 367 & 256 & 5 & & & & \\
Commodity Prod. & \href{https://fred.stlouisfed.org/categories/33062}{FRED} & Economics & M & 32 & 1 & 325 & 256 & 160 & & & & \\
Commodity Import & \href{https://fred.stlouisfed.org/categories/33068}{FRED} & Economics & M & 8 & 1 & 697 & 256 & 40 & & & & \\
WUI-Global & \href{https://fred.stlouisfed.org/categories/33201}{FRED} & Economics & Q & 1 & 15 & 294 & 256 & 75 & & & & \\
Global Price & \href{https://fred.stlouisfed.org/categories/32217}{FRED} & Economics & Q & 1 & 60 & 142 & 256 & 300 & & & & \\
\midrule
Vehicle Sales & \href{https://arxiv.org/pdf/2602.12147}{FRED} & Sales & M & 1 & 10 & 596 & 256 & 50 & & & & \\
Online Retail II & \href{https://archive.ics.uci.edu/dataset/502/online+retail+ii}{Competition} & Sales & D & 1 & 1 & 739 & 256 & 6 & & & & \\
Supply Chain-Cust. & \href{https://www.kaggle.com/datasets/philiphyde1/time-series-supply-chain-dataset}{Competition} & Sales & D & 1 & 36 & 2,007 & 256 & 432 & & & & \\
Supply Chain-Loc. & \href{https://www.kaggle.com/datasets/philiphyde1/time-series-supply-chain-dataset}{Competition} & Sales & D & 1 & 51 & 2,007 & 256 & 612 & & & & \\
\midrule
Azure2019-D & \href{https://github.com/Azure/AzurePublicDataset/blob/master/AzurePublicDatasetV2.md}{Github} & CloudOPS & 5T & 989 & 3 & 8,627 & 256 & 8,901 & & & & \\
Azure2019-I & \href{https://github.com/Azure/AzurePublicDataset/blob/master/AzurePublicDatasetV2.md}{Github} & CloudOPS & 5T & 492 & 3 & 8,630 & 256 & 4,428 & & & & \\
Azure2019-U & \href{https://github.com/Azure/AzurePublicDataset/blob/master/AzurePublicDatasetV2.md}{Github} & CloudOPS & 5T & 78 & 3 & 1,406 & 256 & 1,404 & & & & \\
\midrule
Smart Mfg. & \href{https://www.kaggle.com/datasets/ziya07/smart-manufacturing-iot-cloud-monitoring-dataset}{Competition} & Industry & H & 34 & 5 & 1,666 & 256 & 2,380 & 1024 & 340 & 4096 & 170 \\
MetroPT-3 & \href{https://archive.ics.uci.edu/dataset/791/metropt+3+dataset}{Competition} & Industry & 5T & 1 & 6 & 17,809 & 256 & 216 & 1024 & 36 & 4096 & 18 \\
\bottomrule
\end{tabular}
}
\label{tab:datasets-time-imputation}
\end{table}

\paragraph{Setting.}
We reuse the datasets and windowing protocol originally designed for forecasting, treating each lookback window as a partially observed sequence with synthetically introduced missing values.
The benchmark spans multiple domains (e.g., energy, finance, healthcare, transportation) and covers diverse lengths and frequencies.
Considering short, medium, and long window size configurations yields 98 imputation datasets, each composed of multiple samples to impute.
The per-task window sizes can be found in \cref{tab:datasets-time-imputation}.
To evaluate robustness under different missingness patterns, we define four masking scenarios, namely:
\begin{itemize}
    \item pointwise masking with 50\% missing values;
    \item pointwise masking with 70\% missing values;
    \item a single contiguous missing block of size $1/15$ of the window; and 
    \item two disjoint missing blocks, each of size $1/15$ of the window. 
\end{itemize}
This setup captures both random and structured missingness, reflecting realistic data corruption patterns. Overall, this results in $98 \times 4$ imputation tasks and approximately 440k windows to reconstruct.

\begin{figure}[h!]
\centering
\begin{subfigure}[c]{0.60\textwidth}
\centering
\includegraphics[width=\linewidth]{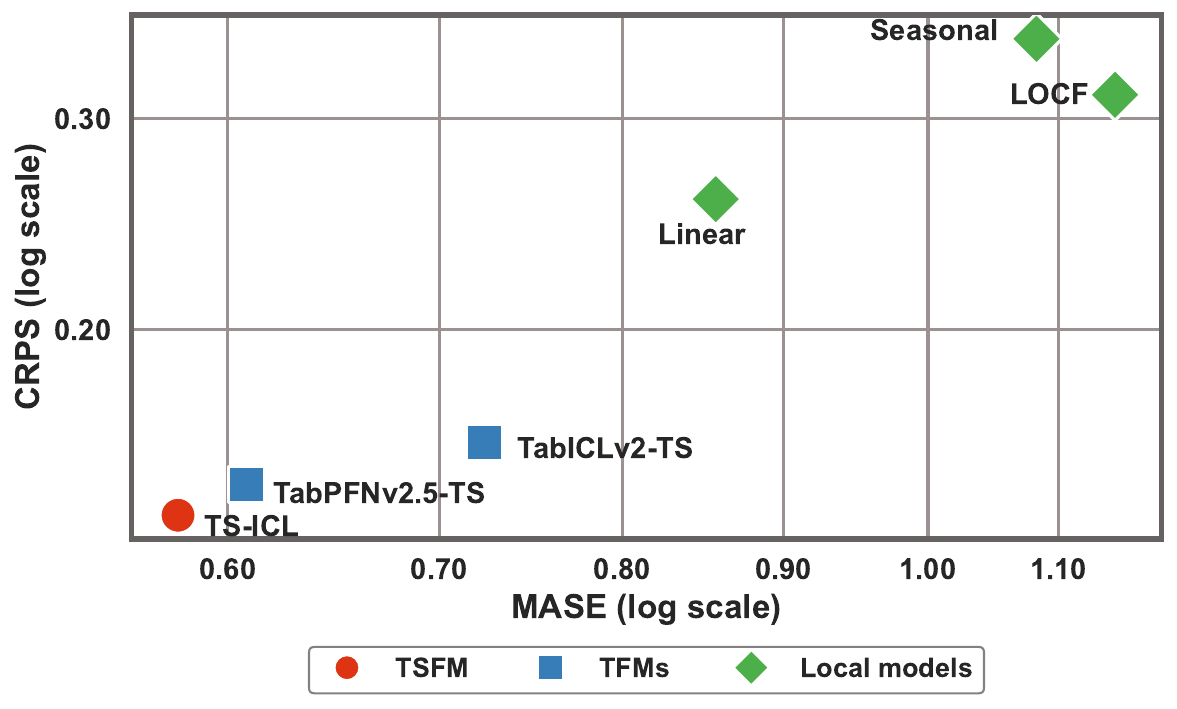}
\caption{Aggregated scores across 392 tasks - \texttt{TIME} benchmark.}
\label{fig:time-imputation-univar}
\end{subfigure}
\hfill
\begin{subfigure}[c]{0.495\textwidth}
\centering
\includegraphics[width=\linewidth]{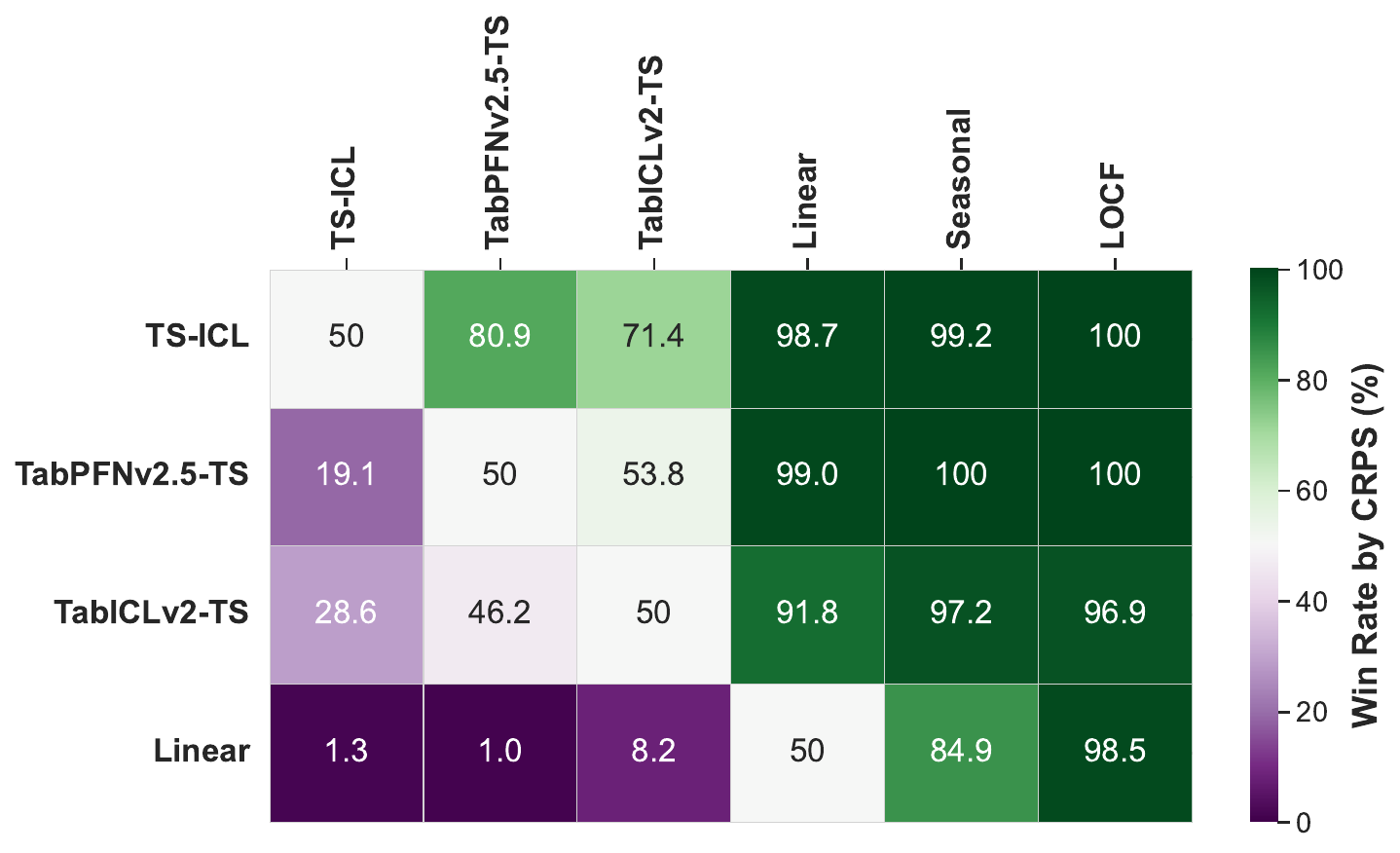}
\caption{Win rates across 392 tasks (CRPS).}
\label{fig:time-imputation-univar-wins_crps}
\end{subfigure}
\begin{subfigure}[c]{0.495\textwidth}
\centering
\includegraphics[width=\linewidth]{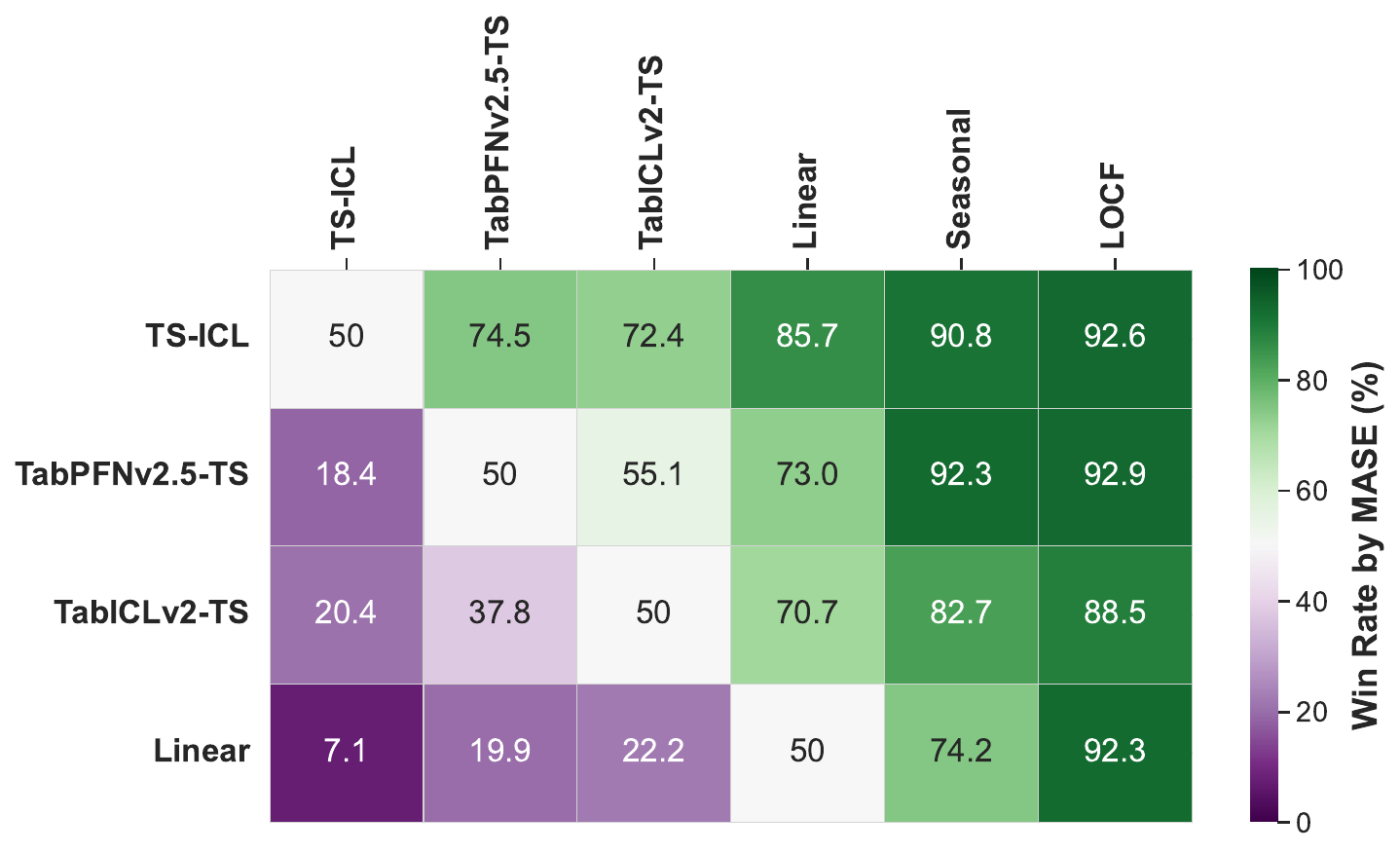}
\caption{Win rates across 392 tasks (MASE).}
\label{fig:time-imputation-univar-wins_mase}
\end{subfigure}

\caption{
Agregated metrics on the \texttt{TIME} univariate time series imputation benchmark.
(a) MASE-CRPS (lower is better).
Each point corresponds to a method, averaged across 392 tasks.
(b-c) Pairwise win rates of the top-4 models.
Each entry indicates the fraction of tasks where a method outperforms another according to the (b) CRPS or (c) the MASE.
}
\label{fig:time-imputation-univar-all}
\end{figure}

\paragraph{Baselines.}
\texttt{TS-ICL} is evaluated against the same tabular foundation model and local baselines as those described in \cref{sec:imputation-expe}.
Final aggregated results (arithmetic mean) are reported in \cref{fig:time-imputation-univar-all} (and \cref{tab:imputation-results}), using both probabilistic (CRPS) and scale-normalized pointwise metrics (MASE; see \cref{sec:scores-metrics} for a definition).


\paragraph{Results.}
Results consistently highlight the strong performance of \texttt{TS-ICL}. As illustrated in \cref{fig:time-imputation-univar}, \texttt{TS-ICL} achieves the lowest overall error, yielding relative improvements of 5.4\% in CRPS and 4.8\% in MASE over \texttt{TabPFNv2.5-TS}, the current state-of-the-art tabular foundation model (TFM). Compared to \texttt{TabICLv2-TS}, these gains extend to 13.0\% (CRPS) and 20.0\% (MASE). Beyond aggregate metrics, \texttt{TS-ICL} demonstrates a dominant pairwise win rate against \texttt{TabPFNv2.5-TS}, outperforming it on 80.9\% of tasks for CRPS and 74.5\% for MASE (\cref{fig:time-imputation-univar-wins_mase}). Notably, \texttt{TS-ICL} achieves this peak accuracy with significant efficiency gains: it maintains an average inference runtime two orders of magnitude faster compared to the most competitive TFMs. This combination positions \texttt{TS-ICL} as a highly scalable solution for large-scale time series imputation.
\begin{table}[h]
\caption{Detailed performance metrics (mean $\pm$ std) for the univariate time series on the \texttt{TIME} imputation benchmark (392 tasks). Best in {\bfseries bold}.}
\centering
\scalebox{0.75}{
\begin{tabular}{lcccccc}
\toprule
 & {TSFM} & \multicolumn{2}{c}{Tabular FMs} & \multicolumn{3}{c}{Local models} \\
\cmidrule(r){2-2} \cmidrule(r){3-4} \cmidrule(r){5-7} 
 & \texttt{TS-ICL} & \texttt{TabPFNv2.5} & \texttt{TabICLv2} & \texttt{Linear} & \texttt{Seasonal} & \texttt{LOCF} \\
\midrule
MASE ($\downarrow$) & \textbf{0.579 $\pm$ 0.323} & 0.608 $\pm$ 0.281 & 0.724 $\pm$ 0.662 & 0.857 $\pm$ 0.539 & 1.082 $\pm$ 0.243 & 1.146 $\pm$ 0.700 \\
CRPS ($\downarrow$) & \textbf{0.140 $\pm$ 0.118} & 0.148 $\pm$ 0.123 & 0.161 $\pm$ 0.134 & 0.257 $\pm$ 0.228 & 0.350 $\pm$ 0.398 & 0.314 $\pm$ 0.256 \\
\bottomrule
\end{tabular}}
\label{tab:imputation-results}
\end{table}


\clearpage

\section{Extended Forecasting Experiments}
\label{extended-forecasting-expes}

This section provides broader insights into \texttt{TS-ICL} forecasting performances.
A detailed description of the \texttt{fev-bench} datasets used in the main benchmark in \cref{sec:forecasting-expe} is given in \cref{sec:fevbench-appendix}, together with complementary results and qualitative visualizations.
\cref{sec:time-forecasting-appendix} further extends the zero-shot evaluation in the \emph{univariate setting} to a second benchmark, \texttt{TIME} \cite{qiao2026sTIME}, across 98 tasks and against 12 foundation models.


\subsection{\texttt{Fev-bench} Benchmark}
\label{sec:fevbench-appendix}

\paragraph{Inference datasets.} \cref{tab:fevbench-full-datasets} details the datasets used for zero-shot forecasting in \cref{sec:forecasting-expe}. As described by \citep{shchur2025fev}, these datasets cover a diverse range of domains, with frequencies ranging from 5 minutes to quarterly data. Each forecasting task has its own prediction horizon. In \cref{sec:forecasting-expe}, we distinguish two scenarios: \textit{univariate zero-shot forecasting} and \textit{zero-shot forecasting with known covariates} when available.

\paragraph{Baseline details.}
We consider the strongest time-series foundation model baselines reported in \texttt{fev-bench} at the time of writing, while excluding models with substantial training-data overlap with the benchmark. 
In particular, we do not include \texttt{Moirai-2.0} and \texttt{TimesFm2.5} among main baselines due to their high reported leakage rates of $28\%$ and $10\%$, respectively, on \texttt{fev-bench}.
All baselines are evaluated in the zero-shot setting, without task-specific fine-tuning. 
We also report the leakage indicator from~\cite{shchur2025fev}, defined as the fraction of model--task pairs for which the model pretraining data overlaps with the benchmark data.

\begin{itemize}
    \item \texttt{Chronos-2}~\cite{ansari2024chronos2} is a 120M-parameter, patch-based encoder-only transformer closely following the T5 encoder design, with alternating time and group attention layers for in-context learning across related series and covariates. 
    It is the only TSFM baseline in \texttt{fev-bench} that natively supports known-future covariates and handles missing values in the look-back window. 
    Its reported leakage rate is $0\%$, making it a clean zero-shot baseline.

    \item \texttt{TiRex}~\cite{auer2025tirex} is a 35M-parameter decoder-only xLSTM model for zero-shot probabilistic forecasting. 
    It predicts quantiles directly and does not use covariates in the \texttt{fev-bench} setup. 
    Its reported leakage rate is $1\%$.

    \item \texttt{TimesFM-2.5}~\cite{TimesFM} is a 200M-parameter patched decoder-only transformer designed for long-context forecasting and direct quantile prediction. 
    It is evaluated as a univariate forecaster in \texttt{fev-bench}. 
    Its reported leakage rate is $10\%$, so its aggregate score should be interpreted with some caution.

    \item \texttt{Toto-1.0}~\cite{TOTO2025} is a 151M-parameter decoder-only transformer optimized for multivariate observability time series. 
    It supports multivariate inputs, but does not use known future covariates in the \texttt{fev-bench} setting. 
    Its reported leakage rate is $8\%$, which is non-negligible but substantially lower than that of \texttt{Moirai-2.0}.

    \item \texttt{Chronos-Bolt}~\cite{Chronosv1} is a 205M-parameter T5 encoder--decoder model and a patch-based variant of Chronos. 
    It chunks the historical context into patches and produces multi-step quantile forecasts. 
    It is evaluated as a univariate forecaster in \texttt{fev-bench}. 
    Its reported leakage rate is $0\%$.
\end{itemize}

\subsubsection{Extended Results}

\begin{figure}[h!]
    \centering
    \begin{subfigure}[b]{0.495\textwidth}
        \centering
        \includegraphics[width=\linewidth]{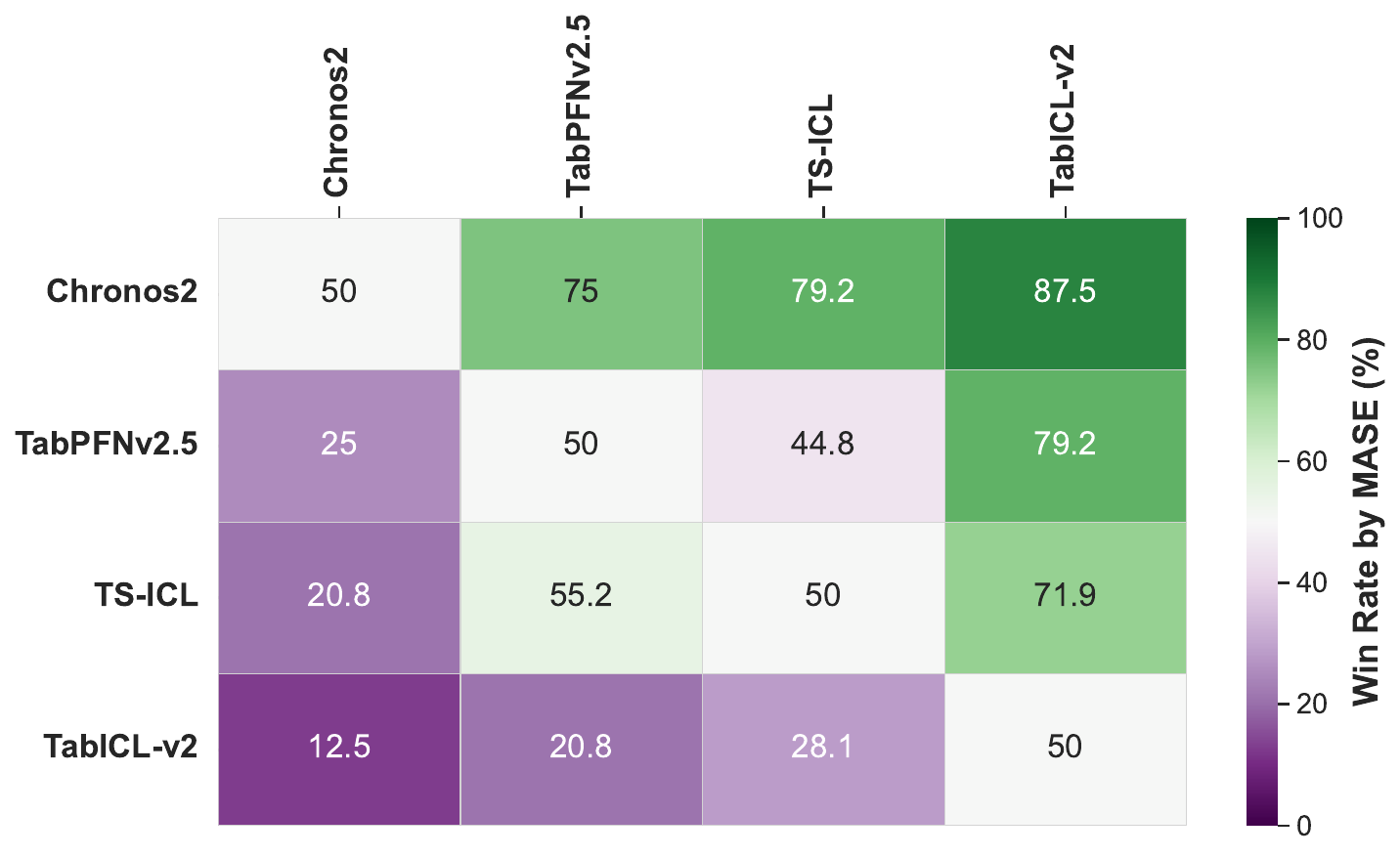}
        \caption{\textit{Univariate} forecasting across 100 tasks.}
        \label{fig:Univariate-forecasting-annexe-fev-wins}
    \end{subfigure}
    \hfill
    \begin{subfigure}[b]{0.495\textwidth}
        \centering
        \includegraphics[width=\linewidth]{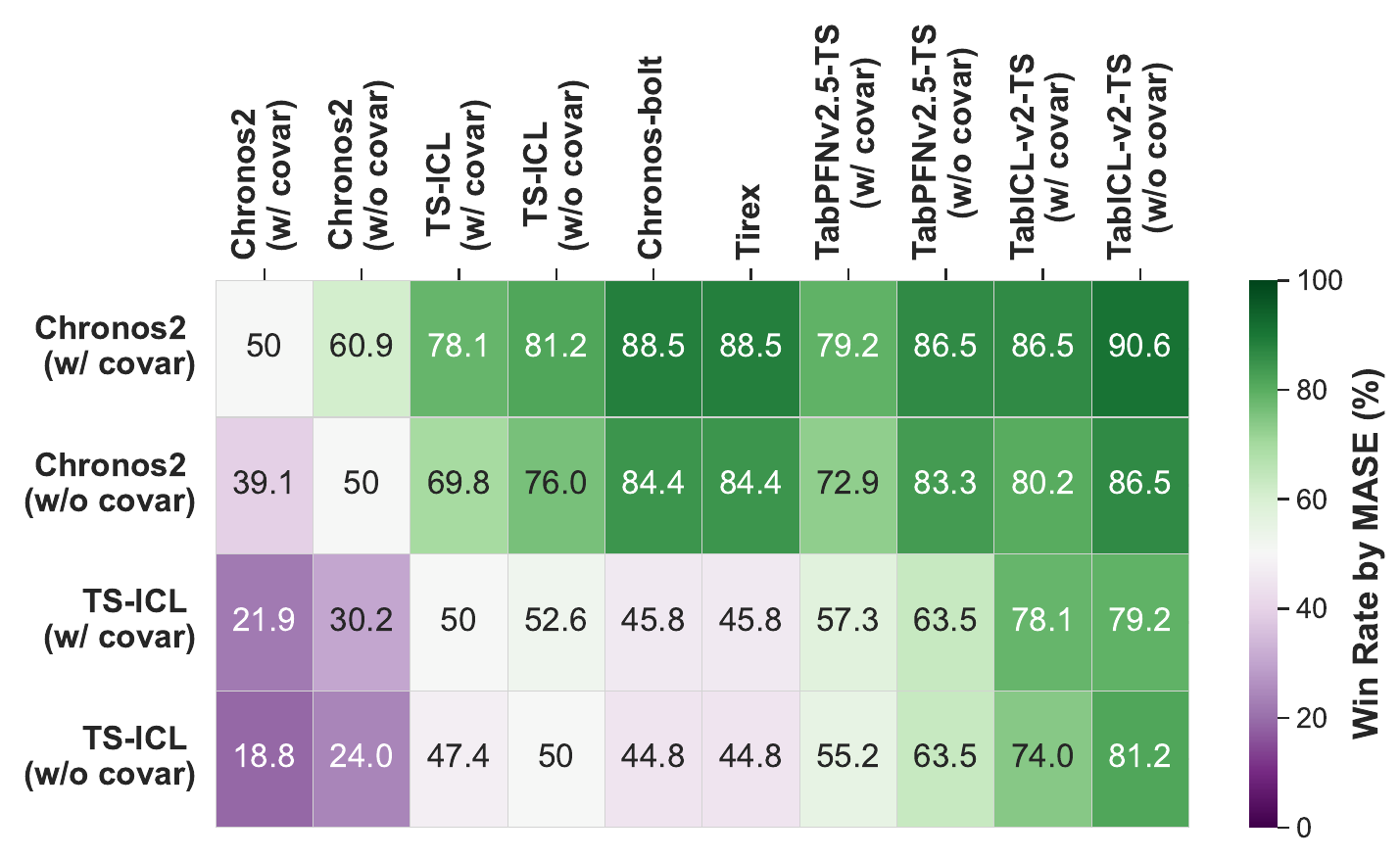}
        \caption{Forecasting with \textit{known covariates} across 30 tasks.}
        \label{fig:covariates-forecasting-annexe-fev-wins}
    \end{subfigure}
    \caption{Pairwise win rates for forecating on the \texttt{fev-bench} benchmark. Each entry indicates the fraction of tasks where a method outperforms another according to the MASE.}
    \label{fig:forecasting-wins-fev-annexe-mase}
\end{figure}

This section extends the empirical evaluation in \cref{sec:forecasting-expe} with a more detailed analysis of forecasting performance across all experimental settings. 
Specifically, we provide:

\begin{itemize}
    \item \textbf{Aggregated detailed performance tables:} We report the average MASE and CRPS (metrics definition in \cref{sec:scores-metrics}) across the \textit{univariate} tasks and \textit{covariates-aware} tasks of \texttt{fm-impute-bench}. These results, detailed in \cref{tab:forecasting-detailed-metrics-univariate-fevbench} and \cref{tab:forecasting-detailed-metrics-covariates-fevbench}, providing a detailed view of both point-wise and probabilistic performance. Note that metrics are aggregated across tasks using the geometric mean, following the evaluation protocol established in \texttt{fev-bench}. 
    \item \textbf{MASE pairwise win rates:} To complement the CRPS-based win rate diagrams presented in the main text (\cref{fig:Univariate-forecasting-fev-wins,fig:covariates-forecasting-fev-wins}), we include the corresponding pairwise win rate visualizations in terms of MASE for both \textit{univariate} and \textit{known-covariate} experiments in \cref{fig:forecasting-wins-fev-annexe-mase}.
\end{itemize}

\begin{table}[h]
\caption{\texttt{Fev-bench} \textit{univariate} forecasting (100 tasks). Performance metrics aggregated (geometric mean $\pm$ geometric std). Best in {\bfseries bold}.}
\centering

\scalebox{0.85}{
\begin{tabular}{lccccc}
\toprule
 & \multicolumn{5}{c}{TSFM} \\
\cmidrule(r){2-6}
 & \texttt{TS-ICL} & \texttt{Chronos-2} & \texttt{Chronos-Bolt} & \texttt{TiRex} & \texttt{Toto} \\
\midrule
MASE ($\downarrow$) & 1.150 $\pm$ 2.175 & \textbf{1.081 $\pm$ 2.201} & 1.158 $\pm$ 2.154 & 1.102 $\pm$ 2.134 & 1.773 $\pm$ 2.148 \\
CRPS ($\downarrow$) & 0.137 $\pm$ 2.995 & \textbf{0.129 $\pm$ 3.009} & 0.140 $\pm$ 3.027 & 0.132 $\pm$ 3.064 & 0.135 $\pm$ 2.958 \\
\bottomrule
\end{tabular}}

\vspace{1em}

\scalebox{0.85}{
\begin{tabular}{lcccc}
\toprule
 & \multicolumn{2}{c}{Tabular Foundation models} & \multicolumn{2}{c}{Local models} \\
\cmidrule(r){2-3} \cmidrule(r){4-5} 
 & \texttt{TabPFNv2.5} & \texttt{TabICLv2} & \texttt{Seasonal} & \texttt{LOCF} \\
\midrule
MASE ($\downarrow$) & 1.218 $\pm$ 2.173 & 1.400 $\pm$ 2.129 & 1.547 $\pm$ 2.062 & 1.839 $\pm$ 2.243 \\
CRPS ($\downarrow$) & 0.141 $\pm$ 3.028 & 0.159 $\pm$ 3.252 & 0.241 $\pm$ 2.954 & 0.260 $\pm$ 2.682 \\
\bottomrule
\end{tabular}}

\label{tab:forecasting-detailed-metrics-univariate-fevbench}
\end{table}

\begin{table}[h]
\caption{\texttt{Fev-bench} forecasting on 100 tasks (30 \textit{covariate-aware})  forecasting (100 tasks). Performance metrics aggregated (geometric mean $\pm$ geometric std). Best in \textbf{bold}.}
\centering
\scalebox{0.85}{
\begin{tabular}{lcccc}
\toprule
 & \multicolumn{4}{c}{TSFM} \\
\cmidrule(r){2-5}
 & \multicolumn{2}{c}{\texttt{TS-ICL}} & \multicolumn{2}{c}{\texttt{Chronos-2}} \\
\cmidrule(r){2-3} \cmidrule(r){4-5}
 & \textit{w/ covar} & \textit{w/o covar} & \textit{w/ covar} & \textit{w/o covar} \\
\midrule
MASE ($\downarrow$) & 1.117 $\pm$ 2.232 & 1.150 $\pm$ 2.175 & \textbf{1.034 $\pm$ 2.250} & 1.081 $\pm$ 2.202 \\
CRPS ($\downarrow$) & 0.131 $\pm$ 3.000 & 0.137 $\pm$ 2.995 & \textbf{0.123 $\pm$ 3.062} & 0.129 $\pm$ 3.018 \\
\midrule \midrule
 & \multicolumn{2}{c}{Tabular Foundation Models} & \multicolumn{2}{c}{Univariate FMs} \\
\cmidrule(r){2-3} \cmidrule(r){4-5}
 & \texttt{TabPFNv2.5-TS} & \texttt{TabICLv2-TS} & \texttt{Chronos-Bolt} & \texttt{TiRex} \\
 & \small{(w/ covar)} & \small{(w/ covar)} & \small{(univar.)} & \small{(univar.)} \\
\midrule
MASE ($\downarrow$) & 1.162 $\pm$ 2.235 & 1.342 $\pm$ 2.194 & 1.158 $\pm$ 2.154 & 1.102 $\pm$ 2.134 \\
CRPS ($\downarrow$) & 0.134 $\pm$ 3.060 & 0.153 $\pm$ 3.279 & 0.140 $\pm$ 3.027 & 0.132 $\pm$ 3.064 \\
\bottomrule
\end{tabular}}
\label{tab:forecasting-detailed-metrics-covariates-fevbench}
\end{table}



\subsubsection{Qualitative Analysis and Visualizations}

This section presents visual examples of \texttt{TS-ICL} imputations for both \textit{univariate} and \textit{known-covariates} settings.
We illustrate model forecasting capabilities across various \textit{fev-bench} tasks.

\begin{figure}[h!]
\centering
\begin{subfigure}[b]{0.87\textwidth}
\centering
\includegraphics[width=\linewidth]{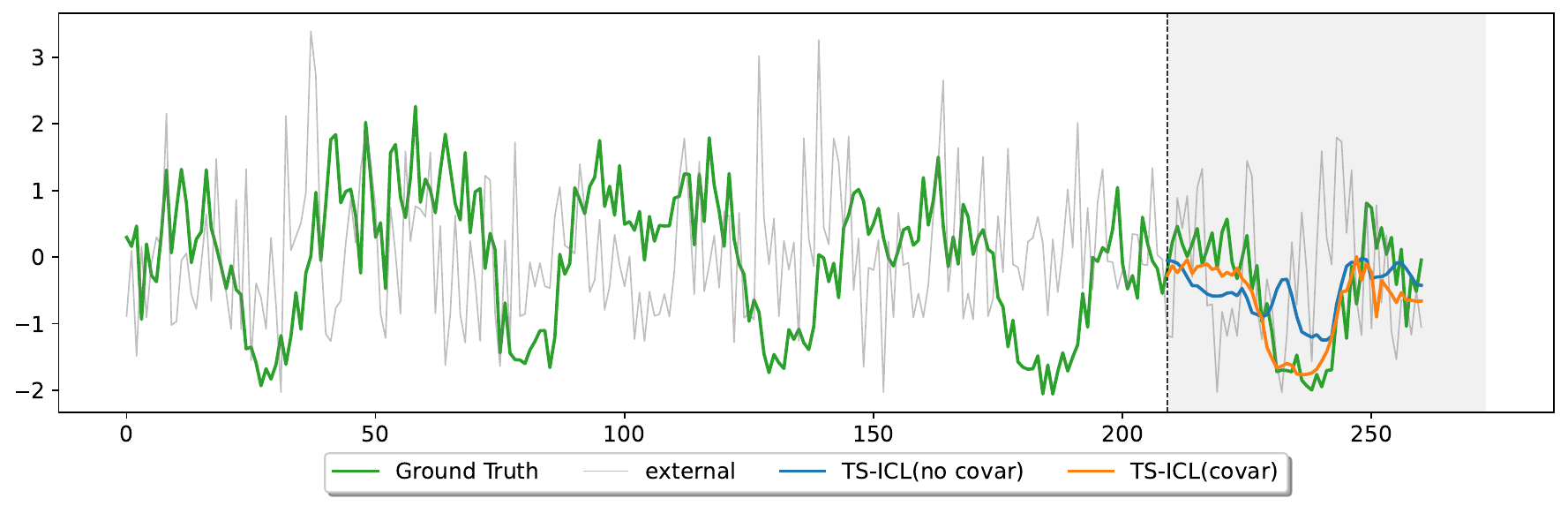}
\caption{\emph{Hermes/W} - $H=52$.}
\label{fig:covar_gfc17}
\end{subfigure}

\begin{subfigure}[b]{0.87\textwidth}
\centering
\includegraphics[width=\linewidth]{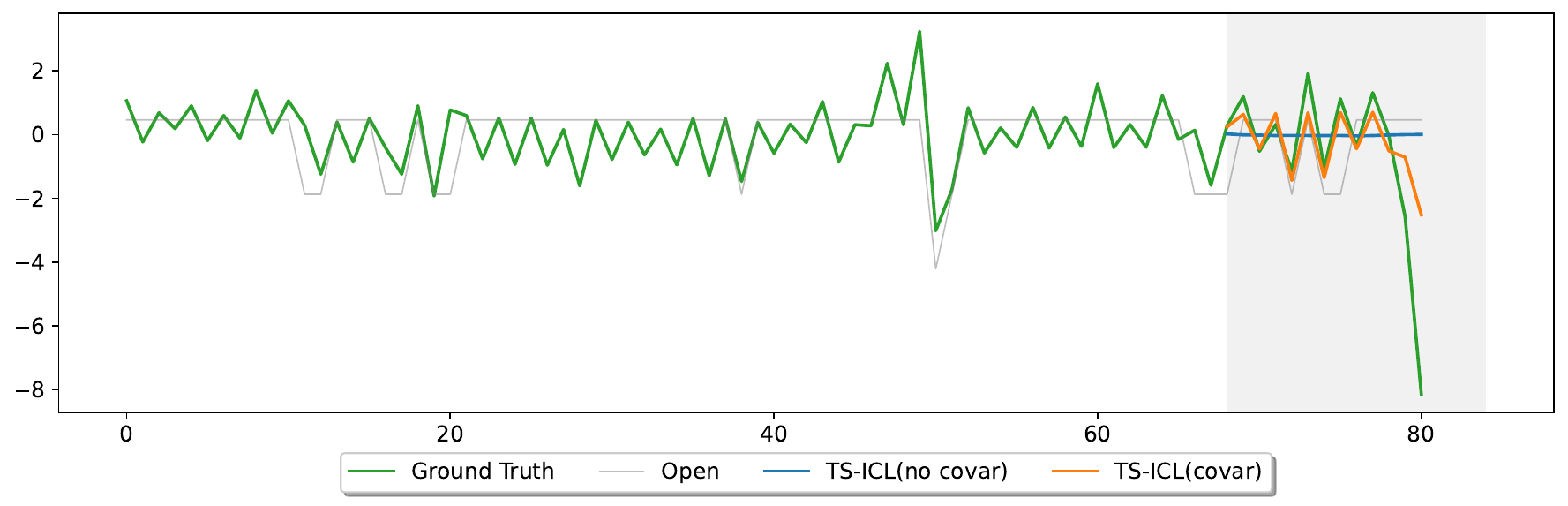}
\caption{\emph{Rossmann/W} - $H=13$.}
\label{fig:covar_rossmann}
\end{subfigure}

\begin{subfigure}[b]{0.87\textwidth}
\centering
\includegraphics[width=\linewidth]{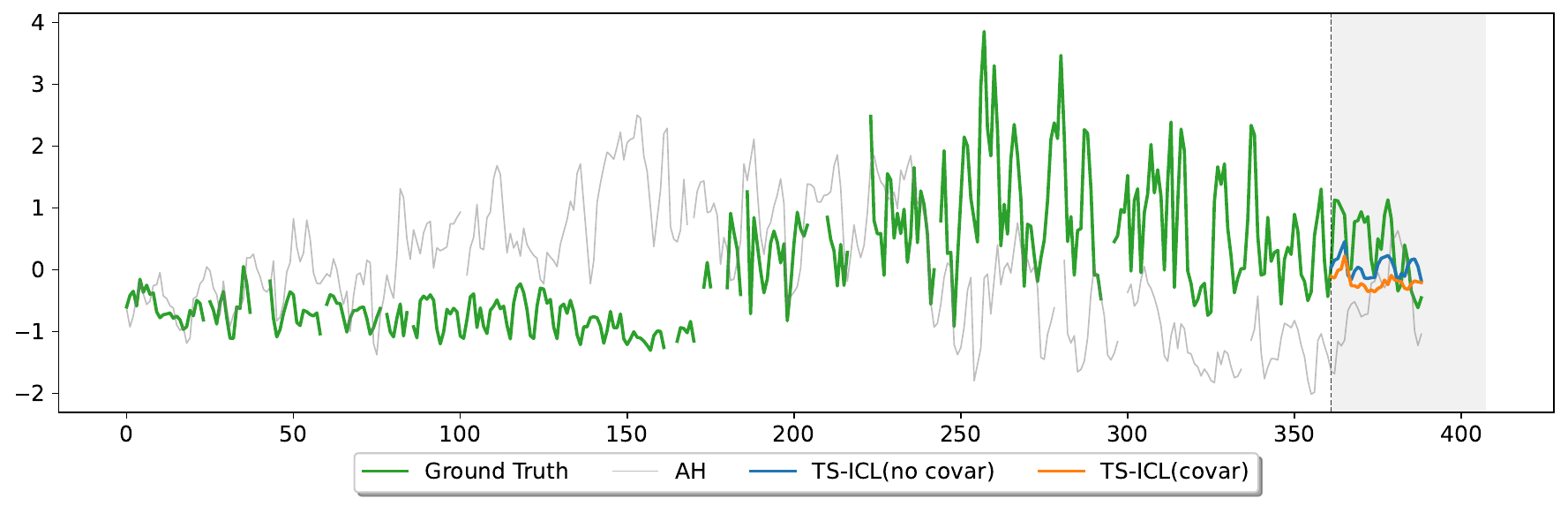}
\caption{\emph{UCI Air Quality/D} - $H=28$.}
\label{fig:covar_uci}
\end{subfigure}

\caption{
Qualitative assessment of \texttt{TS-ICL} forecasts on the \texttt{fev-bench} benchmark, in the \emph{known covariate setting}.
Covariates are shown in light gray.
}
\label{fig:fev-plots-covar}
\end{figure}

\begin{figure}[h!]
\centering
\begin{subfigure}[b]{0.87\textwidth}
\centering
\includegraphics[width=\linewidth]{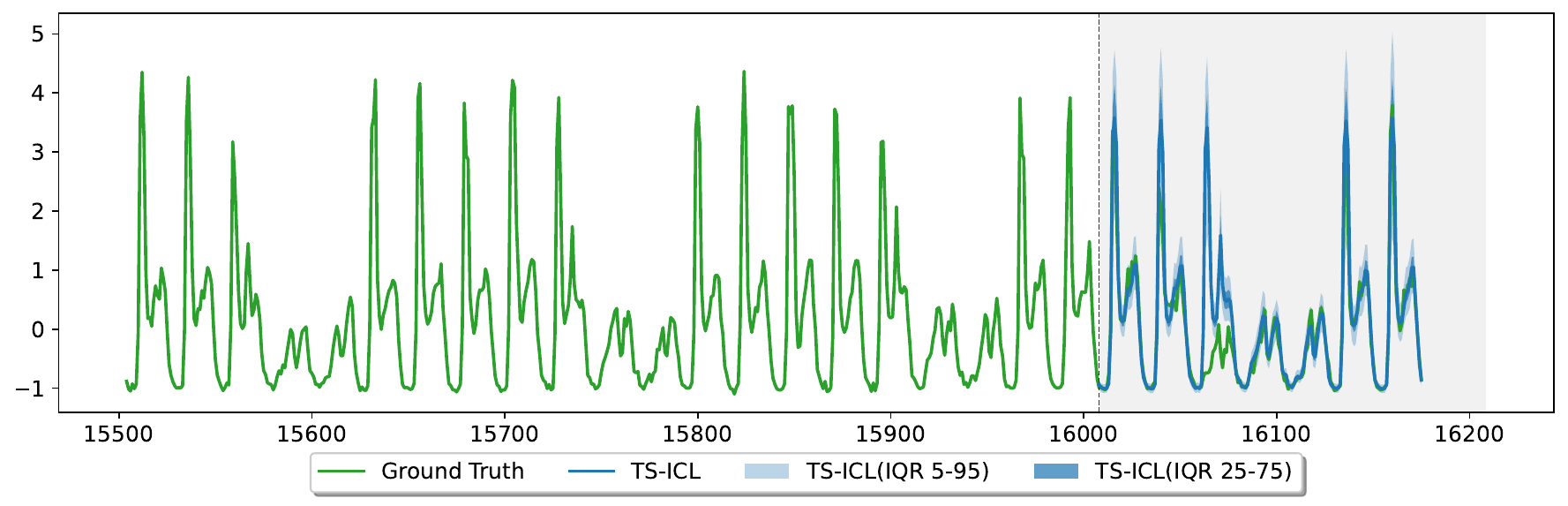}
\caption{\emph{M-DENSE/1H} - $H=168$.}
\label{fig:mdense}
\end{subfigure}

\begin{subfigure}[b]{0.87\textwidth}
\centering
\includegraphics[width=\linewidth]{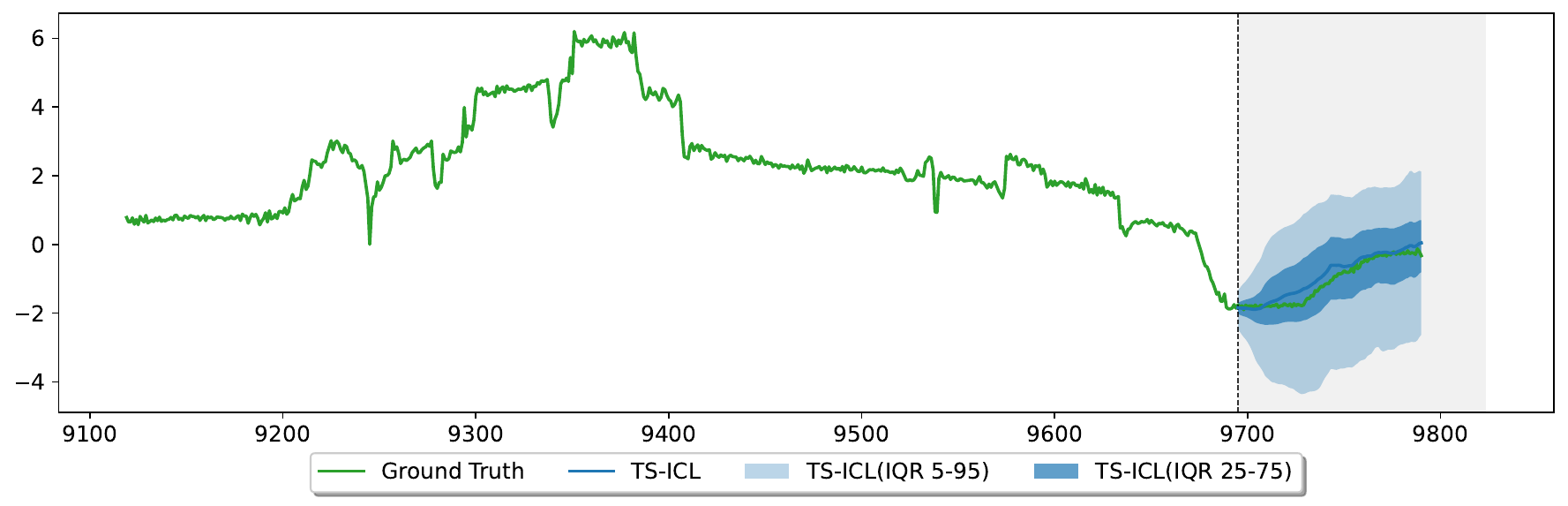}
\caption{\emph{BOOMLET - 1631/30T} - $H=96$.}
\label{fig:boomlet}
\end{subfigure}

\begin{subfigure}[b]{0.87\textwidth}
\centering
\includegraphics[width=\linewidth]{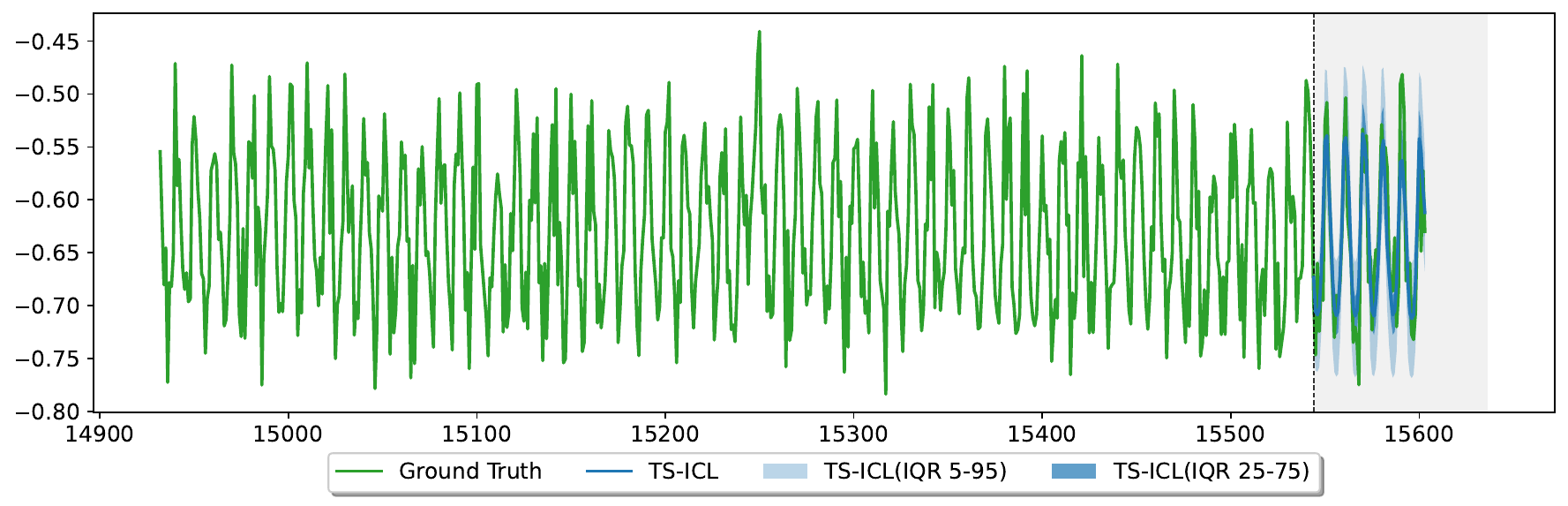}
\caption{\emph{BOOMLET - 1225/1T} - $H=96$.}
\label{fig:boomlet2}
\end{subfigure}

\begin{subfigure}[b]{0.87\textwidth}
\centering
\includegraphics[width=\linewidth]{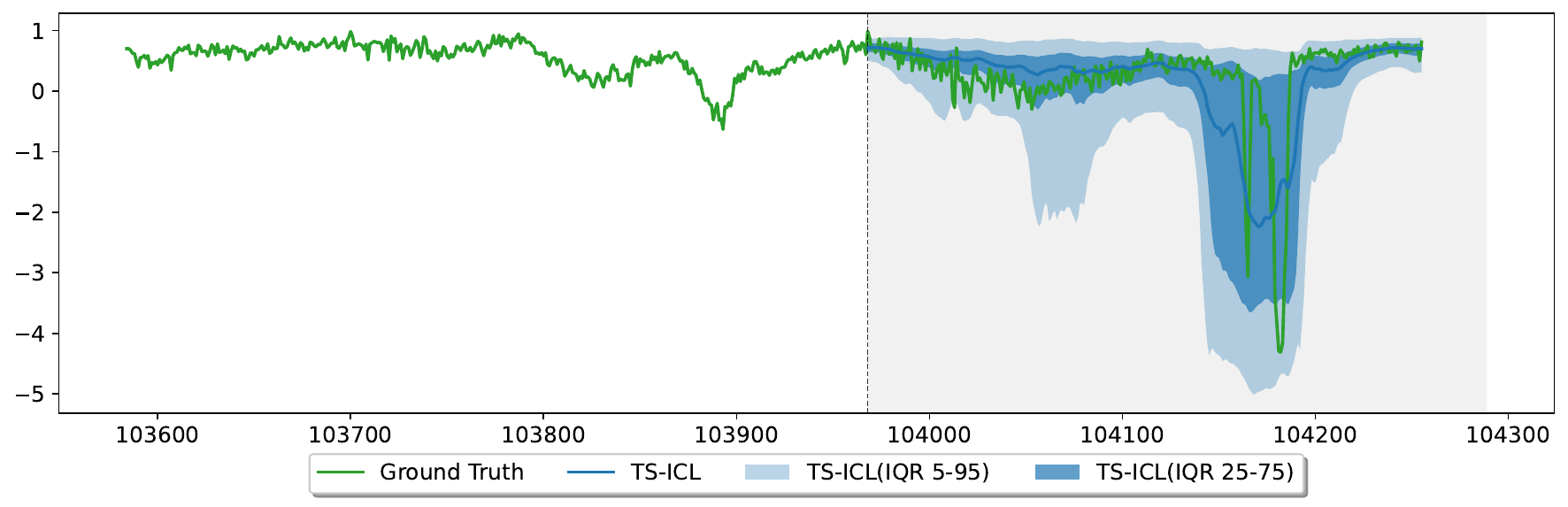}
\caption{\emph{Loop Seattle/5T} - $H=288$.}
\label{fig:seattle}
\end{subfigure}

\begin{subfigure}[b]{0.87\textwidth}
\centering
\includegraphics[width=\linewidth]{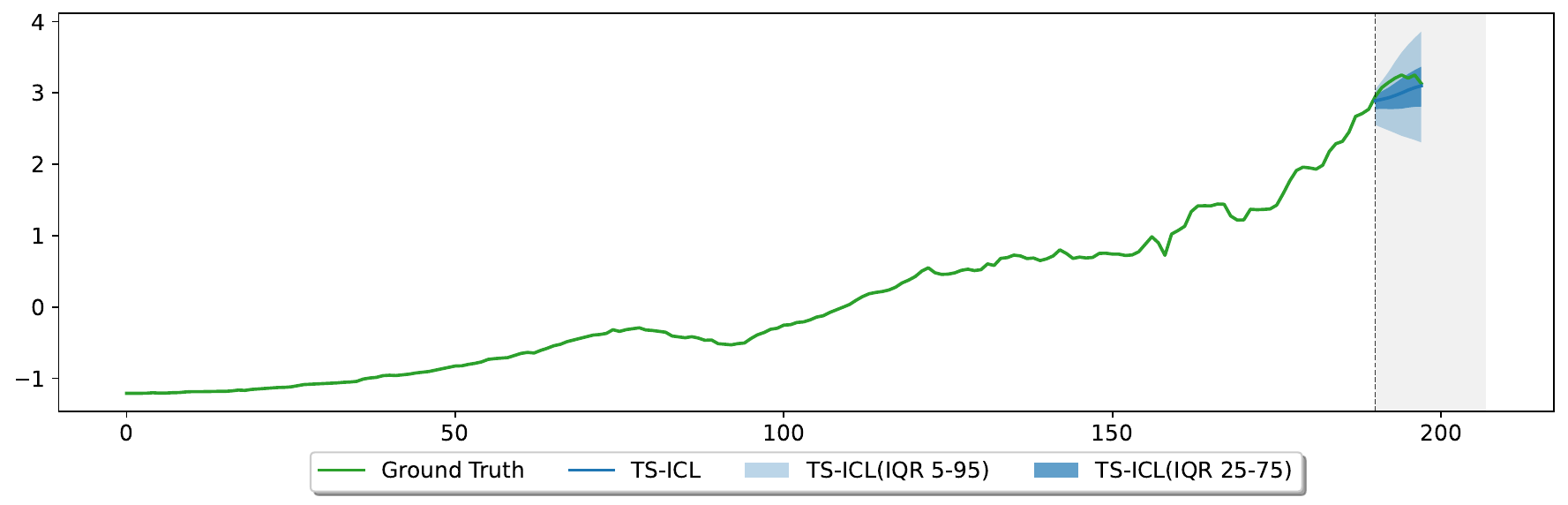}
\caption{\emph{US Consumption/1Q} - $H=8$.}
\label{fig:us_conso}
\end{subfigure}

\caption{
Qualitative assessment of \texttt{TS-ICL} forecasts on the \texttt{fev-bench} benchmark.
}
\label{fig:fev-plots-0}
\end{figure}

\begin{figure}
\centering
\begin{subfigure}[b]{0.87\textwidth}
\centering
\includegraphics[width=\linewidth]{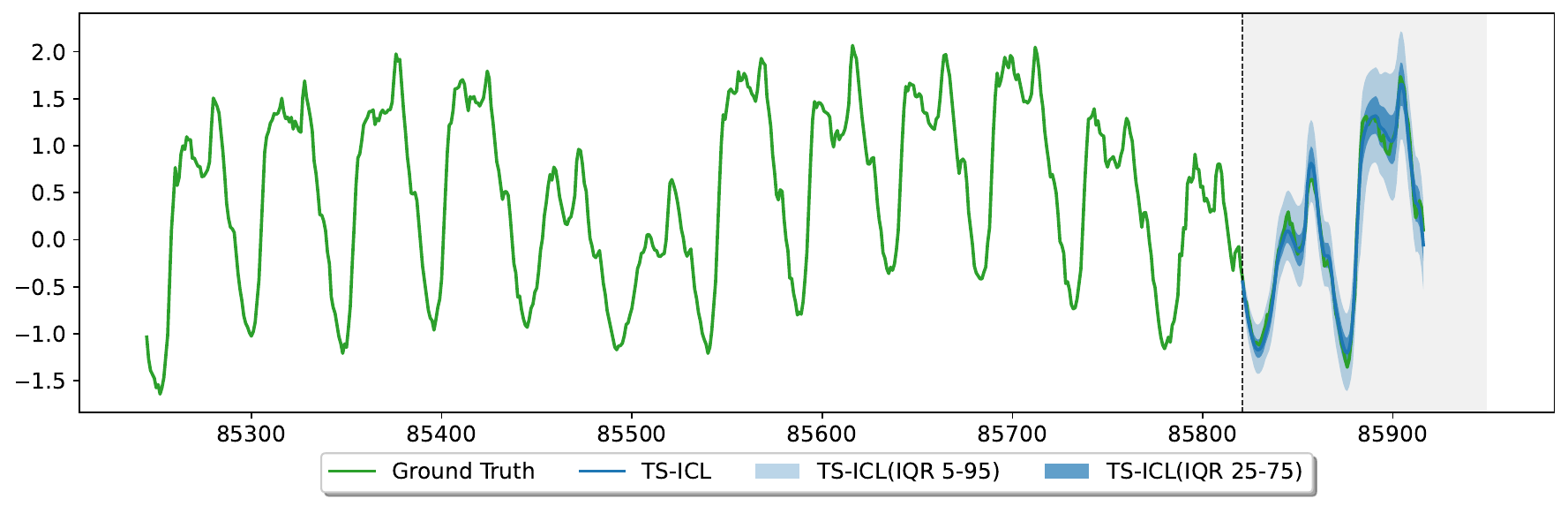}
\caption{\emph{ENTSOE-e Load/30T} - $H=96$.}
\label{fig:plot_entsoe}
\end{subfigure}

\begin{subfigure}[b]{0.87\textwidth}
\centering
\includegraphics[width=\linewidth]{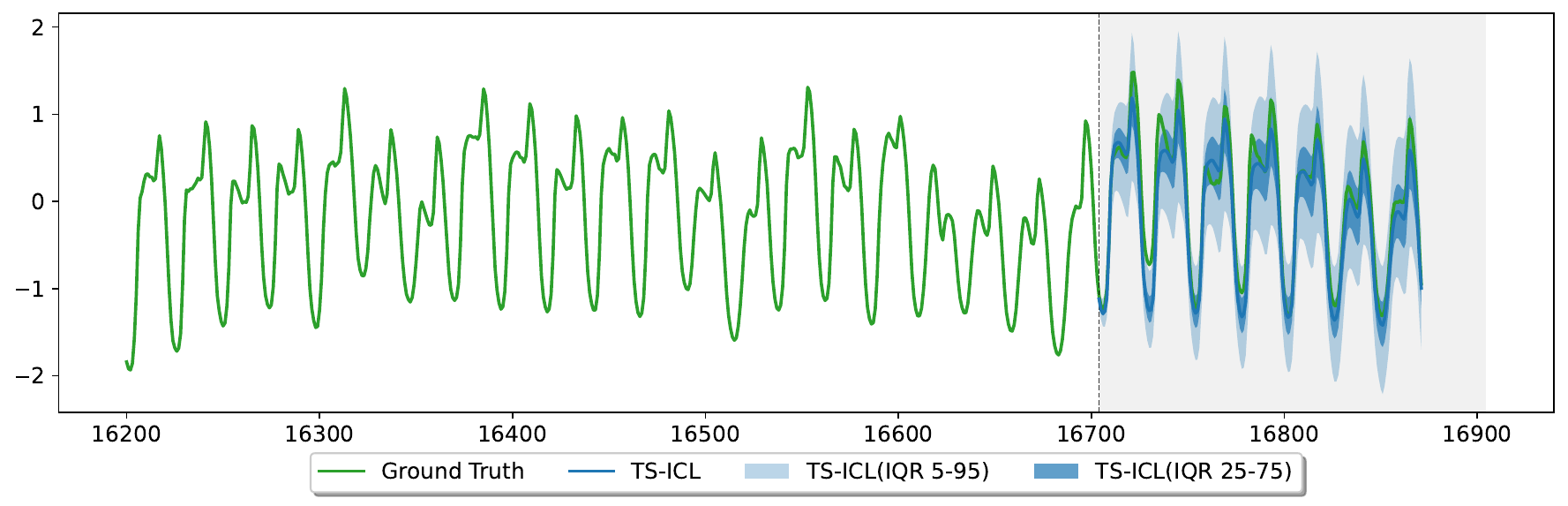}
\caption{\emph{GFC17/1H} - $H=168$.}
\label{fig:plot_gfc17}
\end{subfigure}

\begin{subfigure}[b]{0.87\textwidth}
\centering
\includegraphics[width=\linewidth]{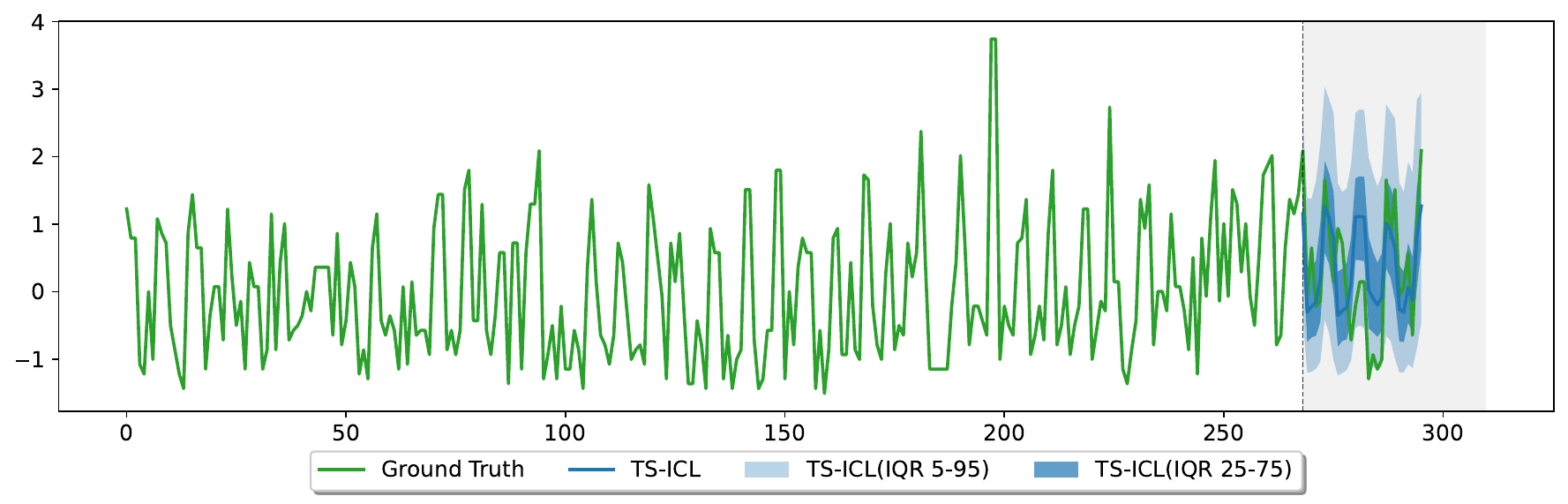}
\caption{\emph{Restaurant/1D} - $H=28$.}
\label{fig:plot_resto}
\end{subfigure}

\begin{subfigure}[b]{0.87\textwidth}
\centering
\includegraphics[width=\linewidth]{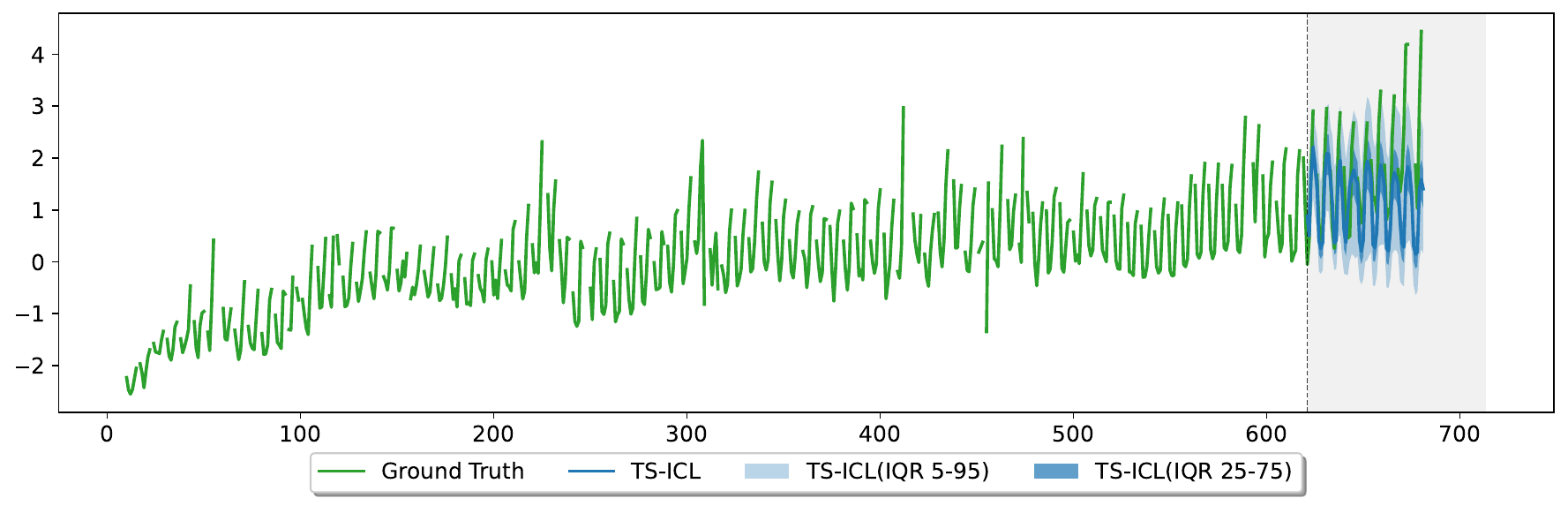}
\caption{\emph{Rohlik Orders/1D} - $H=61$.}
\label{fig:plot_rohlik_ordersD}
\end{subfigure}

\begin{subfigure}[b]{0.87\textwidth}
\centering
\includegraphics[width=\linewidth]{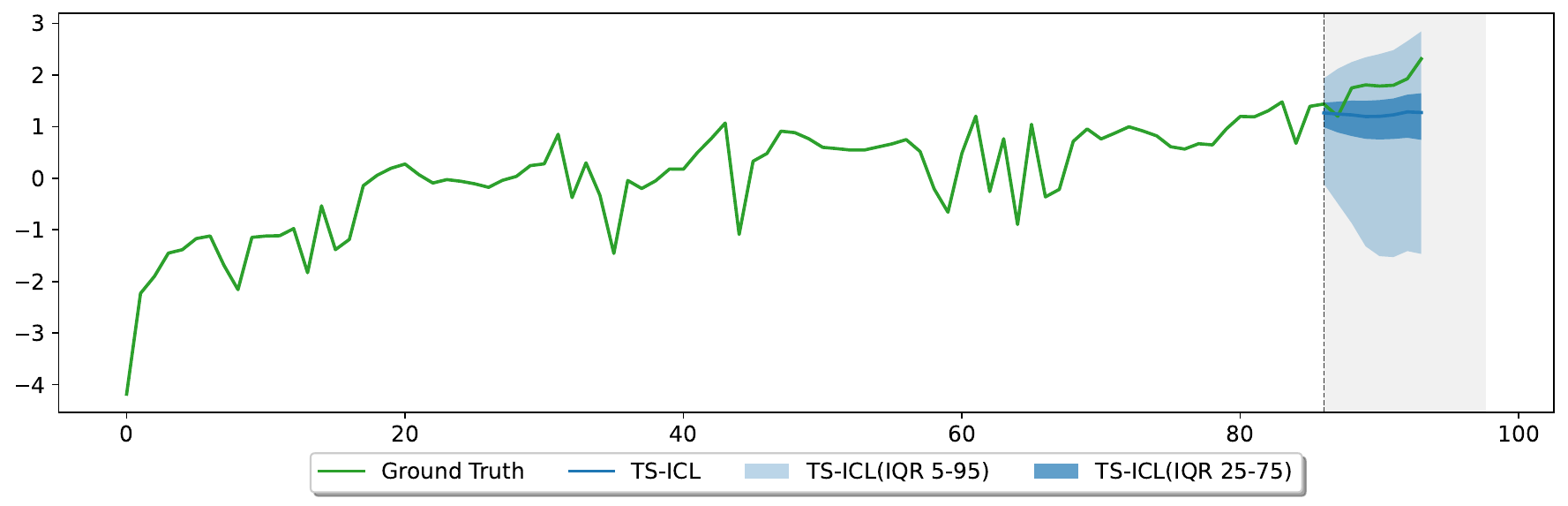}
\caption{\emph{Rohlik Orders/1W} - $H=8$.}
\label{fig:plot_rohlik_ordersW}
\end{subfigure}

\caption{
Qualitative assessment of \texttt{TS-ICL} forecasts on the \texttt{fev-bench} benchmark (continued).
}
\label{fig:fev-plots-1}
\end{figure}

\begin{figure}
\centering
\begin{subfigure}[b]{0.87\textwidth}
\centering
\includegraphics[width=\linewidth]{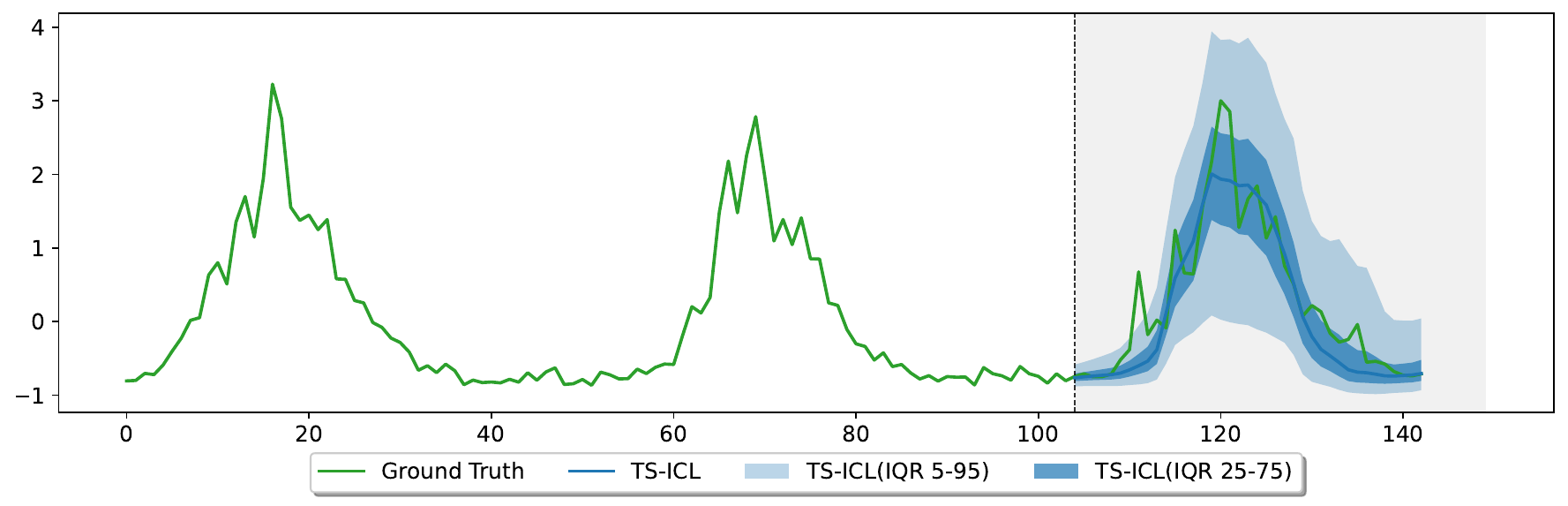}
\caption{\emph{Walmart/1W} - $H=39$.}
\label{fig:plot_walmart}
\end{subfigure}

\begin{subfigure}[b]{0.87\textwidth}
\centering
\includegraphics[width=\linewidth]{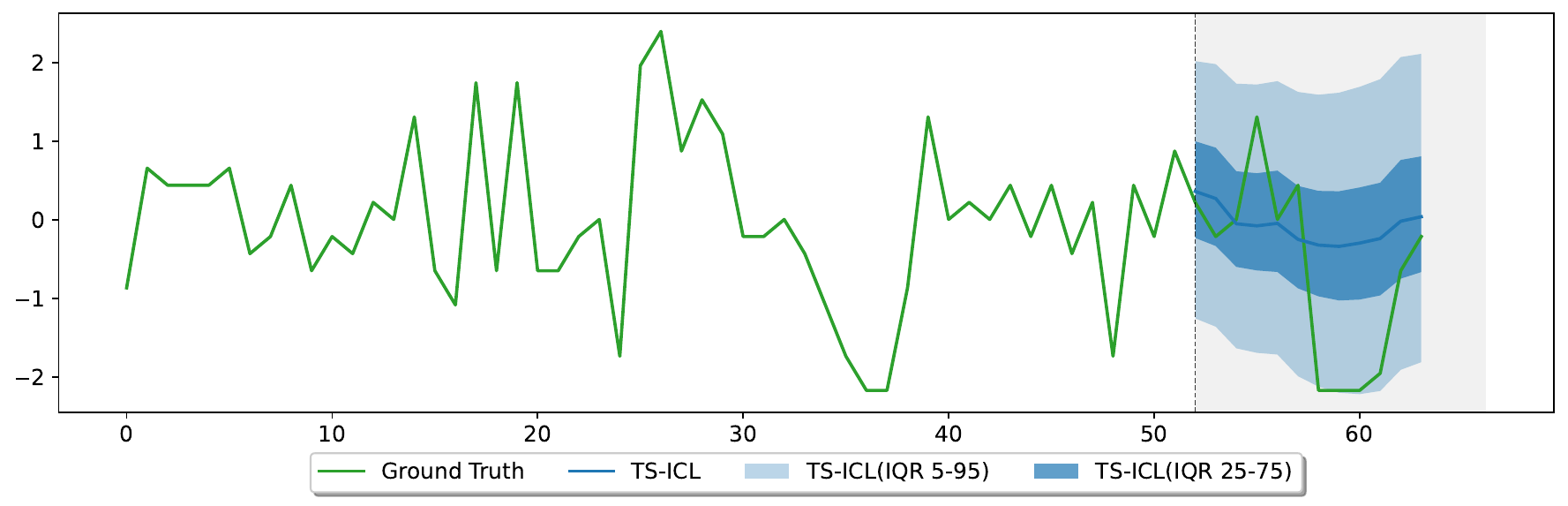}
\caption{\emph{M5/1M} - $H=12$.}
\label{fig:plot_m5_1M}
\end{subfigure}

\begin{subfigure}[b]{0.87\textwidth}
\centering
\includegraphics[width=\linewidth]{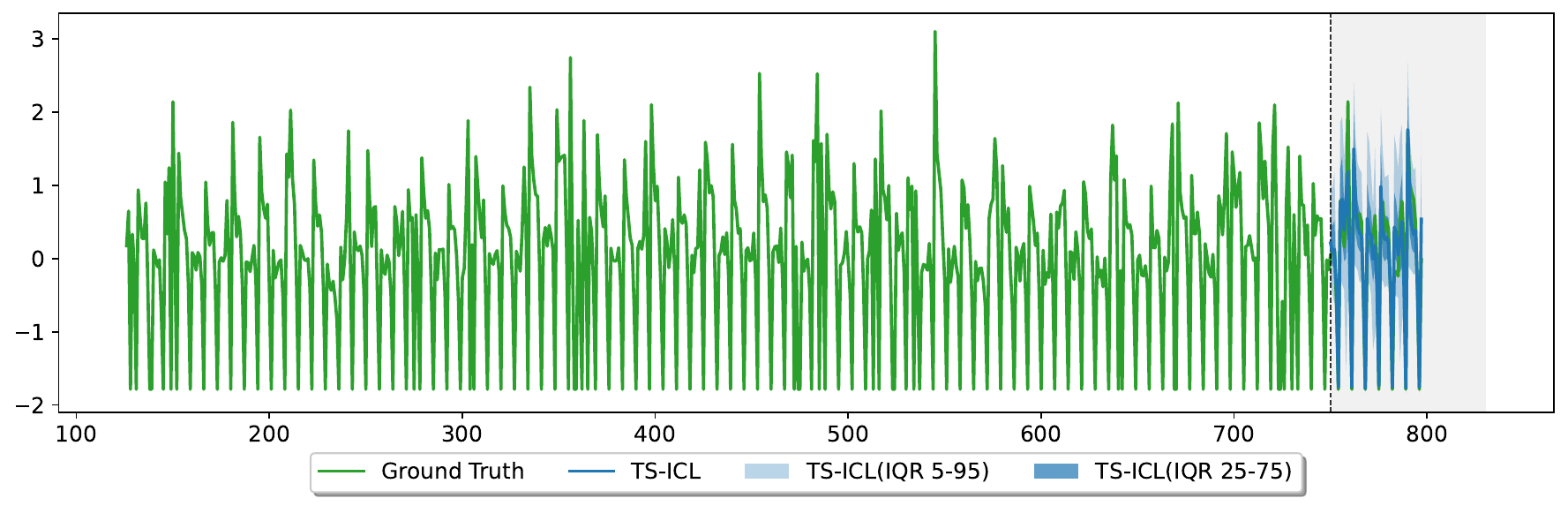}
\caption{\emph{Rossmann/1D} - $H=48$.}
\label{fig:plot_rossmann}
\end{subfigure}

\begin{subfigure}[b]{0.87\textwidth}
\centering
\includegraphics[width=\linewidth]{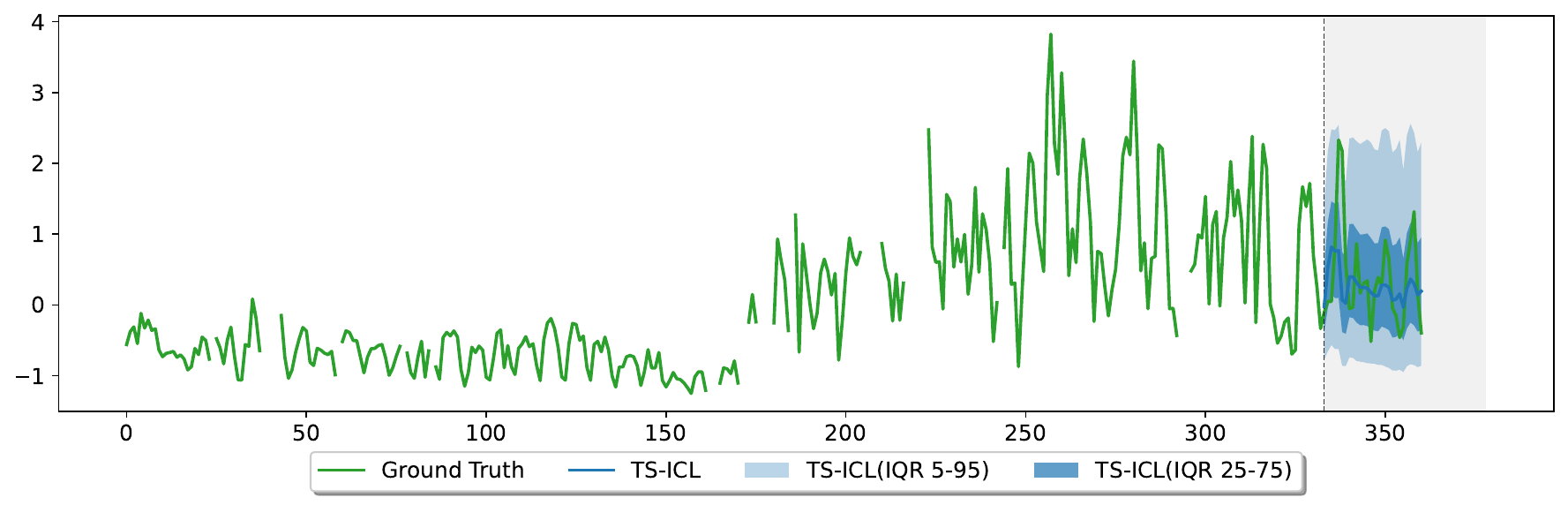}
\caption{\emph{UCI Air Quality/1D} - $H=28$.}
\label{fig:plot_uci_aq}
\end{subfigure}

\begin{subfigure}[b]{0.87\textwidth}
\centering
\includegraphics[width=\linewidth]{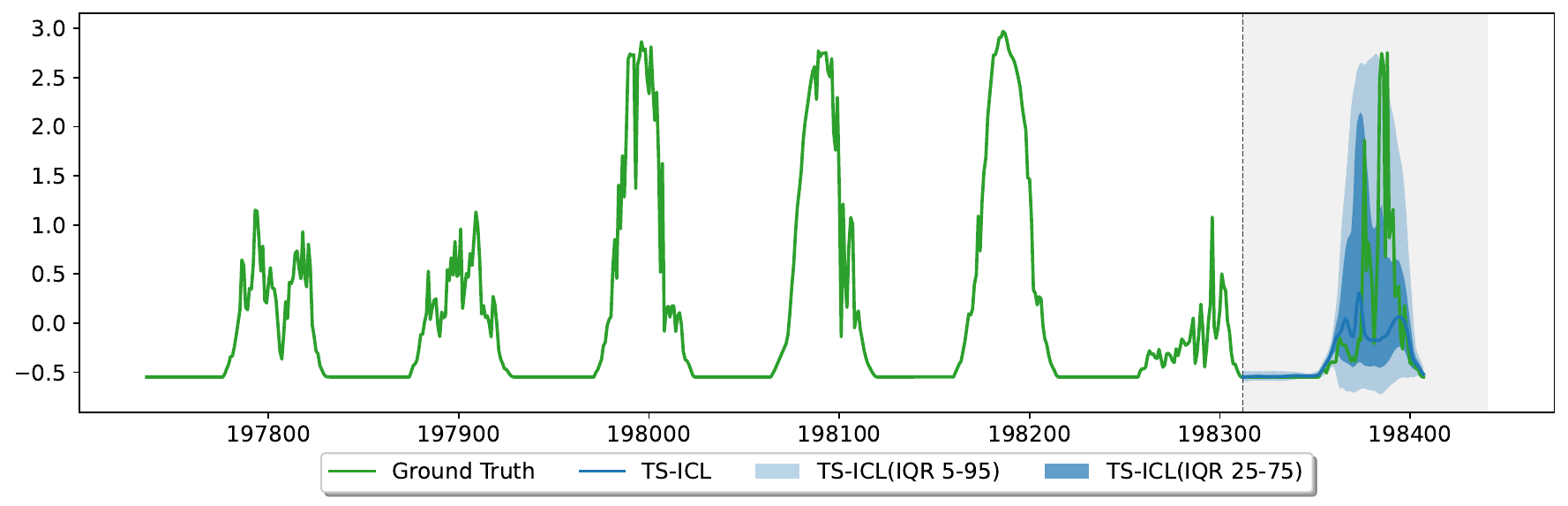}
\caption{\emph{Solar with Weather/15T} - $H=96$.}
\label{fig:plot_solar}
\end{subfigure}

\caption{
Qualitative assessment of \texttt{TS-ICL} forecasts on the \texttt{fev-bench} benchmark (continued).
}
\label{fig:fev-plots-2}
\end{figure}

\paragraph{Results.}
Several observations emerge from the forecasting plots in \cref{fig:fev-plots-covar} (\emph{known covariate setting}) and \cref{fig:fev-plots-0,fig:fev-plots-1,fig:fev-plots-2}(\emph{univariate setting}), where the median forecast is shown together with the corresponding 25-75 and 5-95 inter-quantile ranges.

\begin{enumerate}[(i)]
\item The plots highlight the general ability of \texttt{TS-ICL} to extrapolate from long context windows (with a maximum lookback length of 4096) and regular patterns in heterogeneous sampling rates, domains and seasonalities (e.g. Figures \ref{fig:mdense}, \ref{fig:boomlet}, \ref{fig:plot_entsoe} and \ref{fig:plot_rossmann}).

\item Similarly to the imputation setting, \texttt{TS-ICL} tends to provide smooth forecasts of high-frequency phenomena and adjusts its inter-quantile range accordingly (e.g. Figures \ref{fig:seattle}, \ref{fig:plot_resto} and \ref{fig:plot_solar}).

\item In the \emph{univariate setting}, extrapolating from very short contexts is particularly challenging.
\texttt{TS-ICL} compensates with wider inter-quantile ranges, with mixed success depending on the regularity of the underlying phenomena (Figures \ref{fig:us_conso}, \ref{fig:plot_rohlik_ordersW}, \ref{fig:plot_walmart} and \ref{fig:plot_m5_1M}).

\item In the \emph{known covariate setting}, \texttt{TS-ICL} manages to leverage additional covariate, when the latter informs about the target, while mostly ignoring it otherwise (\cref{fig:fev-plots-covar}).

\item \cref{fig:plot_uci_aq} gives an example of forecasting with missing values, with \texttt{TS-ICL} providing adequate uncertainty estimates.
\end{enumerate}

\begin{table}[h!]
\centering
\caption{Full statistics of \texttt{fev-bench} tasks with data sources. Covariates: P (Past), K (Known), S (Static).}
\scalebox{0.53}{
\begin{tabular}{lll ccc r ccccc r}
\toprule
\multirow{2}{*}{\textbf{Task / Dataset}} & \multirow{2}{*}{\textbf{Source}} & \multirow{2}{*}{\textbf{Domain}} & \multirow{2}{*}{\textbf{Freq}} & \multirow{2}{*}{\textbf{Horizons}} & \multirow{2}{*}{\makecell[c]{\textbf{Num.} \\ \textbf{Series}}} & \multirow{2}{*}{\makecell[c]{\textbf{Num.} \\ \textbf{Target}}} & \multirow{2}{*}{\makecell[c]{\textbf{Median} \\ \textbf{Length}}} & \multicolumn{3}{c}{\textbf{Covariates}} & \multirow{2}{*}{\makecell[c]{\textbf{Num. Test} \\ \textbf{Windows}}} \\
\cmidrule(lr){9-11}
& & & & & & & & \textbf{P} & \textbf{K} & \textbf{S} & \\
\midrule
BizITObs - L2C & \href{https://huggingface.co/datasets/Salesforce/lotsa_data}{LOTSA} & cloud & 5T & 288 & 1 & 7 & 31,968 & 0 & 0 & 0 & 140 \\
BizITObs - L2C & \href{https://huggingface.co/datasets/Salesforce/lotsa_data}{LOTSA} & cloud & H & 24 & 1 & 7 & 2,664 & 0 & 0 & 0 & 140 \\
ETT & \href{https://github.com/zhouhaoyi/ETDataset}{GitHub} & energy & 15T & 96 & 2 & 7 & 69,680 & 0 & 0 & 0 & 280 \\
ETT & \href{https://github.com/zhouhaoyi/ETDataset}{GitHub} & energy & H & 168 & 2 & 7 & 17,420 & 0 & 0 & 0 & 280 \\
ETT & \href{https://github.com/zhouhaoyi/ETDataset}{GitHub} & energy & D & 28 & 2 & 7 & 724 & 0 & 0 & 0 & 280 \\
ETT & \href{https://github.com/zhouhaoyi/ETDataset}{GitHub} & energy & W & 13 & 2 & 7 & 103 & 0 & 0 & 0 & 70 \\
Hierarchical Sales & \href{https://huggingface.co/datasets/Salesforce/lotsa_data}{LOTSA} & retail & D & 28 & 118 & 1 & 1,825 & 0 & 0 & 0 & 1,180 \\
Hierarchical Sales & \href{https://huggingface.co/datasets/Salesforce/lotsa_data}{LOTSA} & retail & W & 13 & 118 & 1 & 260 & 0 & 0 & 0 & 1,180 \\
Hospital & \href{https://huggingface.co/datasets/Salesforce/lotsa_data}{LOTSA} & healthcare & M & 12 & 767 & 1 & 84 & 0 & 0 & 0 & 3,068 \\
Jena Weather & \href{https://www.bgc-jena.mpg.de/wetter/}{MPI Jena} & nature & 10T & 144 & 1 & 21 & 52,704 & 0 & 0 & 0 & 420 \\
Jena Weather & \href{https://www.bgc-jena.mpg.de/wetter/}{MPI Jena} & nature & D & 28 & 1 & 21 & 366 & 0 & 0 & 0 & 231 \\
Jena Weather & \href{https://www.bgc-jena.mpg.de/wetter/}{MPI Jena} & nature & H & 24 & 1 & 21 & 8,784 & 0 & 0 & 0 & 420 \\
Loop Seattle & \href{https://huggingface.co/datasets/Salesforce/lotsa_data}{LOTSA} & mobility & D & 28 & 323 & 1 & 365 & 0 & 0 & 0 & 3,230 \\
Loop Seattle & \href{https://huggingface.co/datasets/Salesforce/lotsa_data}{LOTSA} & mobility & 5T & 288 & 323 & 1 & 105,120 & 0 & 0 & 0 & 3,230 \\
Loop Seattle & \href{https://huggingface.co/datasets/Salesforce/lotsa_data}{LOTSA} & mobility & H & 168 & 323 & 1 & 8,760 & 0 & 0 & 0 & 3,230 \\
M-DENSE & \href{https://huggingface.co/datasets/Salesforce/lotsa_data}{LOTSA} & mobility & D & 28 & 30 & 1 & 730 & 0 & 0 & 0 & 300 \\
M-DENSE & \href{https://huggingface.co/datasets/Salesforce/lotsa_data}{LOTSA} & mobility & H & 168 & 30 & 1 & 17,520 & 0 & 0 & 0 & 300 \\
SZ Taxi & \href{https://huggingface.co/datasets/Salesforce/lotsa_data}{LOTSA} & mobility & 15T & 96 & 156 & 1 & 2,976 & 0 & 0 & 0 & 1,560 \\
SZ Taxi & \href{https://huggingface.co/datasets/Salesforce/lotsa_data}{LOTSA} & mobility & H & 168 & 156 & 1 & 744 & 0 & 0 & 0 & 312 \\
Solar & \href{https://huggingface.co/datasets/Salesforce/lotsa_data}{LOTSA} & energy & W & 13 & 137 & 1 & 52 & 0 & 0 & 0 & 137 \\
Solar & \href{https://huggingface.co/datasets/Salesforce/lotsa_data}{LOTSA} & energy & D & 28 & 137 & 1 & 365 & 0 & 0 & 0 & 1,370 \\
\midrule
Australian Tourism & \href{https://robjhyndman.com/publications/hierarchical-tourism/}{Monash} & econ & Q & 8 & 89 & 1 & 36 & 0 & 0 & 0 & 178 \\
FRED-MD - CEE & \href{https://research.stlouisfed.org/econ/mccracken/fred-databases/}{Fed} & econ & M & 12 & 1 & 3 & 798 & 4 & 0 & 0 & 60 \\
FRED-MD - Macro & \href{https://research.stlouisfed.org/econ/mccracken/fred-databases/}{Fed} & econ & M & 12 & 1 & 51 & 798 & 0 & 0 & 0 & 1,020 \\
FRED-QD - CEE & \href{https://research.stlouisfed.org/econ/mccracken/fred-databases/}{Fed} & econ & Q & 8 & 1 & 3 & 266 & 4 & 0 & 0 & 60 \\
FRED-QD - Macro & \href{https://research.stlouisfed.org/econ/mccracken/fred-databases/}{Fed} & econ & Q & 8 & 1 & 51 & 266 & 0 & 0 & 0 & 1,020 \\
GVAR & \href{https://www.mohaddes.org/gvar}{Mohaddes} & econ & Q & 8 & 33 & 6 & 178 & 3 & 0 & 0 & 1,980 \\
US Consumption & \href{https://otexts.com/fpp3/}{FPP3} & econ & M & 12 & 31 & 1 & 792 & 0 & 0 & 0 & 310 \\
US Consumption & \href{https://otexts.com/fpp3/}{FPP3} & econ & Q & 8 & 31 & 1 & 262 & 0 & 0 & 0 & 310 \\
US Consumption & \href{https://otexts.com/fpp3/}{FPP3} & econ & Y & 5 & 31 & 1 & 64 & 0 & 0 & 0 & 310 \\
World CO2 Emissions & \href{https://data.worldbank.org/}{WorldBank} & econ & Y & 5 & 191 & 1 & 60 & 0 & 0 & 0 & 1,719 \\
World Life Expectancy & \href{https://data.worldbank.org/}{WorldBank} & econ & Y & 5 & 237 & 1 & 74 & 0 & 0 & 0 & 2,370 \\
World Tourism & \href{https://data.worldbank.org/}{WorldBank} & econ & Y & 5 & 178 & 1 & 21 & 0 & 0 & 0 & 356 \\
\midrule
ENTSO-e Load & \href{https://www.entsoe.eu/data/power-stats/}{ENTSO-E} & energy & 15T & 96 & 6 & 1 & 175,292 & 0 & 3 & 0 & 120 \\
ENTSO-e Load & \href{https://www.entsoe.eu/data/power-stats/}{ENTSO-E} & energy & 30T & 96 & 6 & 1 & 87,645 & 0 & 3 & 0 & 120 \\
ENTSO-e Load & \href{https://www.entsoe.eu/data/power-stats/}{ENTSO-E} & energy & H & 168 & 6 & 1 & 43,822 & 0 & 3 & 0 & 120 \\
EPF-BE & \href{https://github.com/mrazavi64/EPF-Standard-Benchmark}{GitHub} & energy & H & 24 & 1 & 1 & 52,416 & 0 & 2 & 0 & 20 \\
EPF-DE & \href{https://github.com/mrazavi64/EPF-Standard-Benchmark}{GitHub} & energy & H & 24 & 1 & 1 & 52,416 & 0 & 2 & 0 & 20 \\
EPF-FR & \href{https://github.com/mrazavi64/EPF-Standard-Benchmark}{GitHub} & energy & H & 24 & 1 & 1 & 52,416 & 0 & 2 & 0 & 20 \\
EPF-NP & \href{https://github.com/mrazavi64/EPF-Standard-Benchmark}{GitHub} & energy & H & 24 & 1 & 1 & 52,416 & 0 & 2 & 0 & 20 \\
EPF-PJM & \href{https://github.com/mrazavi64/EPF-Standard-Benchmark}{GitHub} & energy & H & 24 & 1 & 1 & 52,416 & 0 & 2 & 0 & 20 \\
ERCOT & \href{https://www.ercot.com/griddata}{ERCOT} & energy & D & 28 & 8 & 1 & 6,452 & 0 & 0 & 0 & 160 \\
ERCOT & \href{https://www.ercot.com/griddata}{ERCOT} & energy & H & 168 & 8 & 1 & 154,872 & 0 & 0 & 0 & 160 \\
ERCOT & \href{https://www.ercot.com/griddata}{ERCOT} & energy & M & 12 & 8 & 1 & 211 & 0 & 0 & 0 & 120 \\
ERCOT & \href{https://www.ercot.com/griddata}{ERCOT} & energy & W & 13 & 8 & 1 & 921 & 0 & 0 & 0 & 160 \\
GFC12 & \href{https://huggingface.co/datasets/Salesforce/lotsa_data}{LOTSA} & energy & H & 168 & 11 & 1 & 39,414 & 0 & 1 & 0 & 110 \\
GFC14 & \href{https://huggingface.co/datasets/Salesforce/lotsa_data}{LOTSA} & energy & H & 168 & 1 & 1 & 17,520 & 0 & 1 & 0 & 20 \\
GFC17 & \href{https://huggingface.co/datasets/Salesforce/lotsa_data}{LOTSA} & energy & H & 168 & 8 & 1 & 17,544 & 0 & 1 & 0 & 160 \\
Solar with Weather & \href{https://www.kaggle.com/datasets/the-guardian/solar-generation-and-weather-data}{Kaggle} & energy & 15T & 96 & 1 & 1 & 198,600 & 2 & 7 & 0 & 20 \\
Solar with Weather & \href{https://www.kaggle.com/datasets/the-guardian/solar-generation-and-weather-data}{Kaggle} & energy & H & 24 & 1 & 1 & 49,648 & 2 & 7 & 0 & 20 \\
\midrule
BOOMLET - 1062 & \href{https://huggingface.co/datasets/Datadog/BOOM}{BOOM} & cloud & 5T & 288 & 1 & 21 & 16,384 & 0 & 0 & 0 & 420 \\
BOOMLET - 1209 & \href{https://huggingface.co/datasets/Datadog/BOOM}{BOOM} & cloud & 5T & 288 & 1 & 53 & 16,384 & 0 & 0 & 0 & 1,060 \\
BOOMLET - 1225 & \href{https://huggingface.co/datasets/Datadog/BOOM}{BOOM} & cloud & T & 60 & 1 & 49 & 16,384 & 0 & 0 & 0 & 980 \\
BOOMLET - 1230 & \href{https://huggingface.co/datasets/Datadog/BOOM}{BOOM} & cloud & 5T & 288 & 1 & 23 & 16,384 & 0 & 0 & 0 & 460 \\
BOOMLET - 1282 & \href{https://huggingface.co/datasets/Datadog/BOOM}{BOOM} & cloud & T & 60 & 1 & 35 & 16,384 & 0 & 0 & 0 & 700 \\
BOOMLET - 1487 & \href{https://huggingface.co/datasets/Datadog/BOOM}{BOOM} & cloud & 5T & 288 & 1 & 54 & 16,384 & 0 & 0 & 0 & 1,080 \\
BOOMLET - 1631 & \href{https://huggingface.co/datasets/Datadog/BOOM}{BOOM} & cloud & 30T & 96 & 1 & 40 & 10,463 & 0 & 0 & 0 & 800 \\
BOOMLET - 1676 & \href{https://huggingface.co/datasets/Datadog/BOOM}{BOOM} & cloud & 30T & 96 & 1 & 100 & 10,463 & 0 & 0 & 0 & 2,000 \\
BOOMLET - 1855 & \href{https://huggingface.co/datasets/Datadog/BOOM}{BOOM} & cloud & H & 24 & 1 & 52 & 5,231 & 0 & 0 & 0 & 1,040 \\
BOOMLET - 1975 & \href{https://huggingface.co/datasets/Datadog/BOOM}{BOOM} & cloud & H & 24 & 1 & 75 & 5,231 & 0 & 0 & 0 & 1,500 \\
BOOMLET - 2187 & \href{https://huggingface.co/datasets/Datadog/BOOM}{BOOM} & cloud & H & 24 & 1 & 100 & 5,231 & 0 & 0 & 0 & 2,000 \\
BOOMLET - 285 & \href{https://huggingface.co/datasets/Datadog/BOOM}{BOOM} & cloud & T & 60 & 1 & 75 & 16,384 & 0 & 0 & 0 & 1,500 \\
BOOMLET - 619 & \href{https://huggingface.co/datasets/Datadog/BOOM}{BOOM} & cloud & T & 60 & 1 & 52 & 16,384 & 0 & 0 & 0 & 1,040 \\
BOOMLET - 772 & \href{https://huggingface.co/datasets/Datadog/BOOM}{BOOM} & cloud & T & 60 & 1 & 67 & 16,384 & 0 & 0 & 0 & 1,340 \\
BOOMLET - 963 & \href{https://huggingface.co/datasets/Datadog/BOOM}{BOOM} & cloud & T & 60 & 1 & 28 & 16,384 & 0 & 0 & 0 & 560 \\
\midrule
Favorita Store Sales & \href{https://www.kaggle.com/c/favorita-grocery-sales-forecasting}{Kaggle} & retail & M & 12 & 1,579 & 1 & 54 & 1 & 1 & 6 & 3,158 \\
Favorita Store Sales & \href{https://www.kaggle.com/c/favorita-grocery-sales-forecasting}{Kaggle} & retail & W & 13 & 1,579 & 1 & 240 & 1 & 1 & 6 & 15,790 \\
Favorita Store Sales & \href{https://www.kaggle.com/c/favorita-grocery-sales-forecasting}{Kaggle} & retail & D & 28 & 1,579 & 1 & 1,688 & 1 & 2 & 6 & 15,790 \\
Favorita Transactions & \href{https://www.kaggle.com/c/favorita-grocery-sales-forecasting}{Kaggle} & retail & M & 12 & 51 & 1 & 54 & 1 & 0 & 5 & 102 \\
Favorita Transactions & \href{https://www.kaggle.com/c/favorita-grocery-sales-forecasting}{Kaggle} & retail & W & 13 & 51 & 1 & 240 & 1 & 0 & 5 & 510 \\
Favorita Transactions & \href{https://www.kaggle.com/c/favorita-grocery-sales-forecasting}{Kaggle} & retail & D & 28 & 51 & 1 & 1,688 & 1 & 1 & 5 & 510 \\
KDD Cup 2022 & \href{https://www.kaggle.com/c/wind-power-forecasting}{Kaggle} & energy & D & 14 & 134 & 1 & 243 & 9 & 0 & 0 & 1,340 \\
KDD Cup 2022 & \href{https://www.kaggle.com/c/wind-power-forecasting}{Kaggle} & energy & 10T & 288 & 134 & 1 & 35,279 & 9 & 0 & 0 & 1,340 \\
KDD Cup 2022 & \href{https://www.kaggle.com/c/wind-power-forecasting}{Kaggle} & energy & 30T & 96 & 134 & 1 & 11,758 & 9 & 0 & 0 & 1,340 \\
M5 & \href{https://www.kaggle.com/c/m5-forecasting-accuracy}{Kaggle} & retail & M & 12 & 30,490 & 1 & 58 & 0 & 8 & 5 & 30,490 \\
M5 & \href{https://www.kaggle.com/c/m5-forecasting-accuracy}{Kaggle} & retail & W & 13 & 30,490 & 1 & 257 & 0 & 8 & 5 & 30,490 \\
M5 & \href{https://www.kaggle.com/c/m5-forecasting-accuracy}{Kaggle} & retail & D & 28 & 30,490 & 1 & 1,810 & 0 & 8 & 5 & 30,490 \\
Restaurant & \href{https://www.kaggle.com/competitions/recruit-restaurant-visitor-forecasting}{Kaggle} & retail & D & 28 & 817 & 1 & 296 & 0 & 0 & 4 & 6,536 \\
Rohlik Orders & \href{https://www.kaggle.com/competitions/rohlik-business-case-forecasting}{Kaggle} & retail & W & 8 & 7 & 1 & 170 & 9 & 4 & 0 & 35 \\
Rohlik Orders & \href{https://www.kaggle.com/competitions/rohlik-business-case-forecasting}{Kaggle} & retail & D & 61 & 7 & 1 & 1,197 & 9 & 4 & 0 & 35 \\
Rohlik Sales & \href{https://www.kaggle.com/competitions/rohlik-business-case-forecasting}{Kaggle} & retail & W & 8 & 5,243 & 1 & 150 & 1 & 13 & 7 & 5,243 \\
Rohlik Sales & \href{https://www.kaggle.com/competitions/rohlik-business-case-forecasting}{Kaggle} & retail & D & 14 & 5,390 & 1 & 1,046 & 1 & 13 & 7 & 5,390 \\
Rossmann & \href{https://www.kaggle.com/c/rossmann-store-sales}{Kaggle} & retail & W & 13 & 1,115 & 1 & 133 & 1 & 4 & 10 & 8,920 \\
Rossmann & \href{https://www.kaggle.com/c/rossmann-store-sales}{Kaggle} & retail & D & 48 & 1,115 & 1 & 942 & 1 & 5 & 10 & 11,150 \\
Walmart & \href{https://www.kaggle.com/c/walmart-recruiting-store-sales-forecasting}{Kaggle} & retail & W & 39 & 2,936 & 1 & 143 & 0 & 10 & 4 & 2,936 \\
\midrule
ECDC ILI & \href{https://www.ecdc.europa.eu/en/surveillance-atlas-infectious-diseases}{ECDC} & healthcare & W & 13 & 25 & 1 & 201 & 0 & 0 & 0 & 250 \\
Hermes & \href{https://huggingface.co/datasets/Salesforce/lotsa_data}{LOTSA} & retail & W & 52 & 10,000 & 1 & 261 & 0 & 1 & 2 & 10,000 \\
Hospital Admissions & \href{https://coronavirus.data.gov.uk/}{Gov.UK} & healthcare & D & 28 & 8 & 1 & 1,731 & 0 & 0 & 0 & 160 \\
Hospital Admissions & \href{https://coronavirus.data.gov.uk/}{Gov.UK} & healthcare & W & 13 & 8 & 1 & 246 & 0 & 0 & 0 & 128 \\
Redset & \href{https://github.com/huawei-noah/Redset}{GitHub} & cloud & 5T & 288 & 118 & 1 & 25,920 & 0 & 0 & 1 & 1,180 \\
Redset & \href{https://github.com/huawei-noah/Redset}{GitHub} & cloud & 15T & 96 & 126 & 1 & 8,640 & 0 & 0 & 1 & 1,260 \\
Redset & \href{https://github.com/huawei-noah/Redset}{GitHub} & cloud & H & 24 & 138 & 1 & 2,160 & 0 & 0 & 1 & 1,380 \\
UCI Air Quality & \href{https://archive.ics.uci.edu/ml/datasets/Air+Quality}{UCI} & nature & H & 168 & 1 & 4 & 9,357 & 0 & 3 & 0 & 80 \\
UCI Air Quality & \href{https://archive.ics.uci.edu/ml/datasets/Air+Quality}{UCI} & nature & D & 28 & 1 & 4 & 389 & 0 & 3 & 0 & 44 \\
UK COVID - Nation - Cumul. & \href{https://coronavirus.data.gov.uk/}{Gov.UK} & healthcare & D & 28 & 4 & 3 & 729 & 5 & 0 & 0 & 240 \\
UK COVID - Nation - Cumul. & \href{https://coronavirus.data.gov.uk/}{Gov.UK} & healthcare & W & 8 & 4 & 3 & 105 & 5 & 0 & 0 & 48 \\
UK COVID - Nation - New & \href{https://coronavirus.data.gov.uk/}{Gov.UK} & healthcare & D & 28 & 4 & 3 & 729 & 5 & 0 & 0 & 240 \\
UK COVID - Nation - New & \href{https://coronavirus.data.gov.uk/}{Gov.UK} & healthcare & W & 8 & 4 & 3 & 105 & 5 & 0 & 0 & 48 \\
UK COVID - UTLA - Cumul. & \href{https://coronavirus.data.gov.uk/}{Gov.UK} & healthcare & W & 13 & 214 & 1 & 104 & 0 & 0 & 0 & 1,070 \\
UK COVID - UTLA - New & \href{https://coronavirus.data.gov.uk/}{Gov.UK} & healthcare & D & 28 & 214 & 1 & 721 & 0 & 0 & 0 & 2,140 \\
\bottomrule
\end{tabular}
}
\label{tab:fevbench-full-datasets}
\end{table}

\clearpage

\subsection{\texttt{TIME} Benchmark}
\label{sec:time-forecasting-appendix}

In this section, we evaluate the zero-shot forecasting capabilities of \texttt{TS-ICL} on the \texttt{TIME} benchmark \citep{qiao2026sTIME}, covering univariate settings. This benchmark is particularly valuable as it provides a rigorous framework for zero-shot evaluation, ensuring a total absence of data leakage across all compared foundation models. Performance is assessed across a wide range of missingness patterns, sequence lengths, and application domains (see \cref{tab:dataset-time-forecast} for details).

\begin{table}[h!]
\centering
\caption{
    Individual statistics of forecasting tasks across all datasets. Freq denotes the sampling frequency.
}
\resizebox{\textwidth}{!}{
\begin{tabular}{ll l ccc r cc cc cc}
\toprule
\multirow{3}{*}{\textbf{Dataset}} & 
\multirow{3}{*}{\makecell[l]{\textbf{Release} \\ \textbf{Platform}}} & 
\multirow{3}{*}{\textbf{Domain}} & 
\multirow{3}{*}{\textbf{Freq}} & 
\multirow{3}{*}{\makecell[c]{\textbf{Num.} \\ \textbf{Series}}} & 
\multirow{3}{*}{\makecell[c]{\textbf{Num.} \\ \textbf{Variate}}} & 
\multirow{3}{*}{\makecell[c]{\textbf{Avg Series} \\ \textbf{Length}}} & 
\multicolumn{2}{c}{\textbf{Short-term}} & 
\multicolumn{2}{c}{\textbf{Med-term}} & 
\multicolumn{2}{c}{\textbf{Long-term}} 
\\
\cmidrule(lr){8-9} \cmidrule(lr){10-11} \cmidrule(lr){12-13}
& & & & & & & \textbf{Horizon} & \makecell[c]{\textbf{Num. Test} \\ \textbf{Windows}} & \textbf{Horizon} & \makecell[c]{\textbf{Num. Test} \\ \textbf{Windows}} & \textbf{Horizon} & \makecell[c]{\textbf{Num. Test} \\ \textbf{Windows}} \\
\midrule
Water Quality-Darwin & \href{https://apps.aims.gov.au/metadata/view/23257155-fa16-4361-ae82-b2a09e4bf9ac}{IMOS} & Nature & 15T & 7 & 6 & 15,229 & 16 \scriptsize(4H) & 3,780 & 96 \scriptsize(D) & 630 & 288 \scriptsize(3D) & 210 \\
Current Velocity & \href{https://catalogue-imos.aodn.org.au/geonetwork/srv/eng/catalog.search\#metadata/ae86e2f5-eaaf-459e-a405-e654d85adb9c}{IMOS} & Nature & 5T & 1 & 6 & 26,486 & 36 \scriptsize(3H) & 720 & 288 \scriptsize(D) & 90 & 864 \scriptsize(3D) & 30 \\
Current Velocity & \href{https://catalogue-imos.aodn.org.au/geonetwork/srv/eng/catalog.search\#metadata/ae86e2f5-eaaf-459e-a405-e654d85adb9c}{IMOS} & Nature & 10T & 10 & 6 & 20,669 & 18 \scriptsize(3H) & 7,200 & 144 \scriptsize(D) & 900 & 432 \scriptsize(3D) & 300 \\
Current Velocity & \href{https://catalogue-imos.aodn.org.au/geonetwork/srv/eng/catalog.search\#metadata/ae86e2f5-eaaf-459e-a405-e654d85adb9c}{IMOS} & Nature & 15T & 5 & 6 & 8,503 & 12 \scriptsize(3H) & 3,600 & 96 \scriptsize(D) & 450 & 288 \scriptsize(3D) & 150 \\
Current Velocity & \href{https://catalogue-imos.aodn.org.au/geonetwork/srv/eng/catalog.search\#metadata/ae86e2f5-eaaf-459e-a405-e654d85adb9c}{IMOS} & Nature & 20T & 27 & 6 & 6,460 & 9 \scriptsize(3H) & 19,440 & 72 \scriptsize(D) & 2,430 & 216 \scriptsize(3D) & 810 \\
Current Velocity & \href{https://catalogue-imos.aodn.org.au/geonetwork/srv/eng/catalog.search\#metadata/ae86e2f5-eaaf-459e-a405-e654d85adb9c}{IMOS} & Nature & H & 21 & 6 & 3,502 & 24 \scriptsize(D) & 3,528 & 168 \scriptsize(W) & 504 & 336 \scriptsize(2W) & 252 \\
CPHL & \href{https://catalogue-imos.aodn.org.au/geonetwork/srv/eng/catalog.search\#metadata/8964658c-6ee1-4015-9bae-2937dfcc6ab9}{IMOS} & Nature & 15T & 2 & 1 & 10,831 & 12 \scriptsize(3H) & 240 & 96 \scriptsize(D) & 30 & 288 \scriptsize(3D) & 10 \\
CPHL & \href{https://catalogue-imos.aodn.org.au/geonetwork/srv/eng/catalog.search\#metadata/8964658c-6ee1-4015-9bae-2937dfcc6ab9}{IMOS} & Nature & 30T & 2 & 1 & 14,687 & 12 \scriptsize(3H) & 240 & 48 \scriptsize(D) & 60 & 144 \scriptsize(3D) & 20 \\
CPHL & \href{https://catalogue-imos.aodn.org.au/geonetwork/srv/eng/catalog.search\#metadata/8964658c-6ee1-4015-9bae-2937dfcc6ab9}{IMOS} & Nature & H & 4 & 1 & 4,971 & 24 \scriptsize(D) & 112 & 168 \scriptsize(W) & 16 & 336 \scriptsize(2W) & 8 \\
Coastal T-S & \href{https://catalogue-imos.aodn.org.au/geonetwork/srv/eng/catalog.search\#metadata/7e13b5f3-4a70-4e31-9e95-335efa491c5c}{IMOS} & Nature & 5T & 18 & 3 & 68,604 & 36 \scriptsize(3H) & 6,480 & 288 \scriptsize(D) & 810 & 864 \scriptsize(3D) & 270 \\
Coastal T-S & \href{https://catalogue-imos.aodn.org.au/geonetwork/srv/eng/catalog.search\#metadata/7e13b5f3-4a70-4e31-9e95-335efa491c5c}{IMOS} & Nature & 15T & 5 & 3 & 20,870 & 12 \scriptsize(3H) & 1,800 & 96 \scriptsize(D) & 225 & 288 \scriptsize(3D) & 75 \\
Coastal T-S & \href{https://catalogue-imos.aodn.org.au/geonetwork/srv/eng/catalog.search\#metadata/7e13b5f3-4a70-4e31-9e95-335efa491c5c}{IMOS} & Nature & 20T & 1 & 3 & 8,198 & 9 \scriptsize(3H) & 360 & 72 \scriptsize(D) & 45 & 216 \scriptsize(3D) & 15 \\
Coastal T-S & \href{https://catalogue-imos.aodn.org.au/geonetwork/srv/eng/catalog.search\#metadata/7e13b5f3-4a70-4e31-9e95-335efa491c5c}{IMOS} & Nature & H & 24 & 3 & 5,489 & 24 \scriptsize(D) & 2,016 & 168 \scriptsize(W) & 288 & 336 \scriptsize(2W) & 144 \\
SG Weather & \href{https://data.gov.sg}{data.gov.sg} & Nature & D & 6 & 4 & 2,953 & 3 \scriptsize(3D) & 2,928 & 7 \scriptsize(W) & 1,272 & 14 \scriptsize(2W) & 648 \\
SG PM 2.5 & \href{https://data.gov.sg/datasets/d_e1058d6974c877257e32048ab128ad83/view\#tag/default/GET/pm25}{data.gov.sg} & Nature & H & 1 & 5 & 38,688 & 24 \scriptsize(D) & 460 & 72 \scriptsize(3D) & 150 & 168 \scriptsize(W) & 65 \\
NE China Wind & \href{https://github.com/Zhang-zongwei/MFWPN}{Github} & Nature & H & 1 & 4 & 8,764 & 24 \scriptsize(D) & 120 & 72 \scriptsize(3D) & 40 & 168 \scriptsize(W) & 16 \\
\midrule
Australia Solar & \href{https://www.pvoutput.org}{Pvoutput} & Energy & H & 1 & 3 & 35,064 & 24 \scriptsize(D) & 315 & 72 \scriptsize(3D) & 105 & 168 \scriptsize(W) & 45 \\
EPF Electricity & \href{https://github.com/jeslago/epftoolbox}{Academic} & Energy & H & 5 & 1 & 52,416 & 24 \scriptsize(D) & 525 & 72 \scriptsize(3D) & 175 & 168 \scriptsize(W) & 75 \\
OpenElectricity & \href{https://docs.openelectricity.org}{OpenElec} & Energy & 5T & 1 & 10 & 43,488 & 24 \scriptsize(2H) & 1,680 & 96 \scriptsize(8H) & 420 & 288 \scriptsize(D) & 140 \\
EWELD Load & \href{https://pmc.ncbi.nlm.nih.gov/articles/PMC10495315/}{Academic} & Energy & 15T & 1 & 10 & 20,544 & 24 \scriptsize(6H) & 560 & 96 \scriptsize(D) & 140 & 672 \scriptsize(W) & 20 \\
SG Carpark & \href{https://data.gov.sg/datasets/d_ca933a644e55d34fe21f28b8052fac63/view}{data.gov.sg} & Transport & 15T & 354 & 1 & 14,332 & 16 \scriptsize(4H) & 14,868 & 96 \scriptsize(D) & 2,478 & 672 \scriptsize(W) & 354 \\
Finland Traffic & \href{https://www.digitraffic.fi/en/}{Digitraffic} & Transport & 15T & 1 & 1 & 35,136 & 16 \scriptsize(4H) & 186 & 96 \scriptsize(D) & 31 & 672 \scriptsize(W) & 4 \\
\midrule
Port Activity & \href{https://www.kaggle.com/datasets/arunvithyasegar/daily-port-activity-data-and-trade-estimates}{Competition} & Transport & D & 99 & 2 & 2,127 & 30 \scriptsize(M) & 2,376 & & & & \\
Port Activity & \href{https://www.kaggle.com/datasets/arunvithyasegar/daily-port-activity-data-and-trade-estimates}{Competition} & Transport & W & 99 & 2 & 304 & 13 \scriptsize(Q) & 792 & & & & \\
ECDC COVID & \href{https://www.ecdc.europa.eu/en/publications-data/download-data-hospital-and-icu-admission-rates-and-current-occupancy-covid-19}{ECDC} & Healthcare & D & 9 & 1 & 1,117 & 30 \scriptsize(30D) & 45 & & & & \\
ECDC COVID & \href{https://www.ecdc.europa.eu/en/publications-data/download-data-hospital-and-icu-admission-rates-and-current-occupancy-covid-19}{ECDC} & Healthcare & W & 16 & 1 & 165 & 13 \scriptsize(Q) & 64 & & & & \\
Global Influenza & \href{https://www.who.int/tools/flunet}{WHO} & Healthcare & W & 15 & 4 & 205 & 13 \scriptsize(Q) & 240 & & & & \\
\midrule
Crypto & \href{https://fred.stlouisfed.org/categories/33913}{FRED} & Finance & D & 1 & 4 & 2,842 & 30 \scriptsize(M) & 36 & & & & \\
US Term Structure & \href{https://fred.stlouisfed.org/categories/33825}{FRED} & Finance & B & 1 & 40 & 9,327 & 20 \scriptsize(4W) & 1,400 & & & & \\
Oil Price & \href{https://fred.stlouisfed.org/categories/32217}{FRED} & Finance & B & 1 & 12 & 5,035 & 20 \scriptsize(4W) & 420 & & & & \\
\midrule
Job Claims & \href{https://fred.stlouisfed.org/categories/32240}{FRED} & Finance & W & 1 & 2 & 196 & 13 \scriptsize(Q) & 8 & & & & \\
Uncertainty-1M & \href{https://fred.stlouisfed.org/categories/33201}{FRED} & Economics & M & 1 & 3 & 780 & 6 \scriptsize(6M) & 21 & & & & \\
Housing Inventory & \href{https://fred.stlouisfed.org/categories/97}{FRED} & Economics & M & 1 & 4 & 114 & 12 \scriptsize(A) & 12 & & & & \\
JOLTS & \href{https://fred.stlouisfed.org/categories/32241}{FRED} & Economics & M & 1 & 6 & 297 & 12 \scriptsize(A) & 30 & & & & \\
US Labor & \href{https://fred.stlouisfed.org/categories/12}{FRED} & Economics & M & 1 & 14 & 380 & 12 \scriptsize(A) & 70 & & & & \\
Vehicle Supply & \href{https://fred.stlouisfed.org/categories/32993}{FRED} & Economics & M & 1 & 6 & 391 & 12 \scriptsize(A) & 30 & & & & \\
Auto Production-SF & \href{https://fred.stlouisfed.org/categories/33938}{FRED} & Economics & M & 1 & 1 & 367 & 12 \scriptsize(A) & 5 & & & & \\
Commodity Prod. & \href{https://fred.stlouisfed.org/categories/33062}{FRED} & Economics & M & 32 & 1 & 325 & 12 \scriptsize(A) & 160 & & & & \\
Commodity Import & \href{https://fred.stlouisfed.org/categories/33068}{FRED} & Economics & M & 8 & 1 & 697 & 12 \scriptsize(A) & 40 & & & & \\
WUI-Global & \href{https://fred.stlouisfed.org/categories/33201}{FRED} & Economics & Q & 1 & 15 & 294 & 4 \scriptsize(A) & 75 & & & & \\
Global Price & \href{https://fred.stlouisfed.org/categories/32217}{FRED} & Economics & Q & 1 & 60 & 142 & 4 \scriptsize(A) & 300 & & & & \\
\midrule
Vehicle Sales & \href{https://arxiv.org/pdf/2602.12147}{FRED} & Sales & M & 1 & 10 & 596 & 12 & 50 & & & & \\
Online Retail II & \href{https://archive.ics.uci.edu/dataset/502/online+retail+ii}{Competition} & Sales & D & 1 & 1 & 739 & 30 & 6 & & & & \\
Supply Chain-Cust. & \href{https://www.kaggle.com/datasets/philiphyde1/time-series-supply-chain-dataset}{Competition} & Sales & D & 1 & 36 & 2,007 & 30 & 432 & & & & \\
Supply Chain-Loc. & \href{https://www.kaggle.com/datasets/philiphyde1/time-series-supply-chain-dataset}{Competition} & Sales & D & 1 & 51 & 2,007 & 30 & 612 & & & & \\
\midrule
Azure2019-D & \href{https://github.com/Azure/AzurePublicDataset/blob/master/AzurePublicDatasetV2.md}{Github} & CloudOPS & 5T & 989 & 3 & 8,627 & 288 \scriptsize(D) & 8,901 & & & & \\
Azure2019-I & \href{https://github.com/Azure/AzurePublicDataset/blob/master/AzurePublicDatasetV2.md}{Github} & CloudOPS & 5T & 492 & 3 & 8,630 & 288 \scriptsize(D) & 4,428 & & & & \\
Azure2019-U & \href{https://github.com/Azure/AzurePublicDataset/blob/master/AzurePublicDatasetV2.md}{Github} & CloudOPS & 5T & 78 & 3 & 1,406 & 48 \scriptsize(4H) & 1,404 & & & & \\
\midrule
Smart Mfg. & \href{https://www.kaggle.com/datasets/ziya07/smart-manufacturing-iot-cloud-monitoring-dataset}{Competition} & Industry & H & 34 & 5 & 1,666 & 24 \scriptsize(D) & 2,380 & 168 \scriptsize(W) & 340 & 336 \scriptsize(2W) & 170 \\
MetroPT-3 & \href{https://archive.ics.uci.edu/dataset/791/metropt+3+dataset}{Competition} & Industry & 5T & 1 & 6 & 17,809 & 48 \scriptsize(4H) & 216 & 288 \scriptsize(D) & 36 & 576 \scriptsize(2D) & 18 \\
\bottomrule
\end{tabular}
}
\label{tab:dataset-time-forecast}
\end{table}

\paragraph{Setting.} The evaluation is conducted under a \textit{primarily univariate forecasting} setting, where \texttt{Time} considers short-, medium-, and long-context window sizes, resulting in 98 forecasting tasks and approximately 110k windows to predict. Note, however, that some baselines (namely \texttt{Chronos-2}, \texttt{Toto}, \texttt{Moirai}, and \texttt{VisionTS}) leverage look-back windows from others series at inference time, operating in a multivariate forecasting regime \citep{qiao2026sTIME}.

\begin{figure}[h!]
    \centering
    \includegraphics[width=0.7\linewidth]{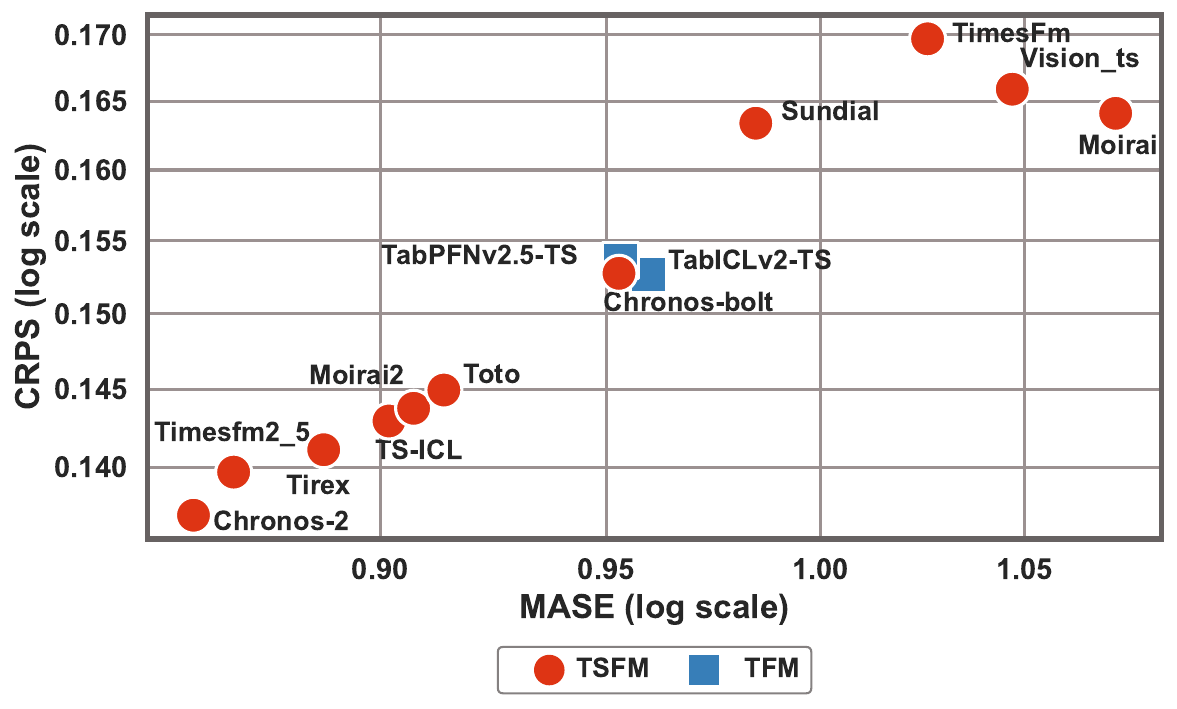}
    \caption{MASE-CRPS time trade-off (lower is better) on the \texttt{TIME} benchmark. The x-axis reports task-averaged MASE for each method, while the y-axis shows task-averaged CRPS for each method. Each point corresponds to a method, averaged across tasks.}
    \label{fig:time-forecasting-score}
\end{figure}

\paragraph{Baselines.} The guaranteed absence of data leakage in the \texttt{TIME} benchmark allows us to expand the comparison to a broader set of state-of-the-art foundation models that were previously excluded in \cref{sec:forecasting-expe}. Final aggregated results are reported in \cref{fig:time-forecasting-score} (and \cref{tab:time-benchmark-results}), using both probabilistic (CRPS) and scale-normalized pointwise metrics (MASE; see \cref{sec:scores-metrics} for a definition). We compare \texttt{TS-ICL} against:
\begin{itemize}
    \item Time Series Foundation Models (TSFMs): \texttt{Chronos-2} \citep{ansari2024chronos2}, \texttt{Timesfm2\_5} \citep{TimesFM}, \texttt{TiRex} \citep{auer2025tirex}, \texttt{Moirai2} \citep{liu2025moirai}, \texttt{Toto} \citep{TOTO2025}, \texttt{Chronos-bolt}, \texttt{Sundial} \citep{liu2025sundial}, \texttt{TimesFm} \citep{TimesFM}, \texttt{Vision\_ts} \citep{chen2024visionts}, and \texttt{Moirai} \citep{woo2024unified}.
    \item Tabular Foundation Models (TFMs): \texttt{TabPFNv2.5-TS} \citep{hollmann2022tabpfn} and \texttt{TabICLv2-TS} \citep{qu2026tabiclv2}.
\end{itemize}

\begin{table}[h] 
\caption{Detailed forecasting performance metrics aggregated (geometric mean $\pm$ geometric std) across the 98 tasks of the \textit{univariate setting} in the \texttt{TIME} benchmark. Best in \textbf{bold}.} 
\centering \scalebox{0.72}{ 
\begin{tabular}{lcccccccc} 
\toprule & \multicolumn{7}{c}{Time Series Foundation Models (TSFMs)} \\ 
\cmidrule(r){2-8} & \texttt{Chronos-2} & \texttt{TimesFM2\_5} & \texttt{TiRex} & \texttt{TS-ICL} & \texttt{Toto} & \texttt{Moirai2} & \texttt{Bolt} \\ 
\midrule 
MASE ($\downarrow$) & \textbf{0.861$\pm$1.621} & 0.869$\pm$1.613 & 0.888$\pm$1.585 & 0.902$\pm$1.597 & 0.907$\pm$1.597 & 0.914$\pm$1.604 & 0.953$\pm$1.629 \\ 
CRPS ($\downarrow$) & \textbf{0.137$\pm$2.885} & 0.140$\pm$2.902 & 0.141$\pm$2.866 & 0.143$\pm$2.807 & 0.144$\pm$2.821 & 0.145$\pm$2.833 & 0.153$\pm$2.688 \\ 
\midrule \midrule
 & \multicolumn{2}{c}{Tabular FMs} & \multicolumn{4}{c}{Other TSFMs} \\ 
\cmidrule(r){2-3} \cmidrule(r){4-7} & \texttt{TabPFNv2.5-TS} & \texttt{TabICLv2-TS} & \texttt{Sundial} & \texttt{TimesFM} & \texttt{Moirai} & \texttt{Vision\_ts} \\ 
\midrule 
MASE ($\downarrow$) & 0.953$\pm$1.601 & 0.960$\pm$1.590 & 0.985$\pm$1.675 & 1.026$\pm$1.593 & 1.073$\pm$1.598 & 1.047$\pm$1.603 \\ 
CRPS ($\downarrow$) & 0.154$\pm$2.852 & 0.153$\pm$2.872 & 0.163$\pm$2.666 & 0.170$\pm$2.840 & 0.164$\pm$2.573 & 0.166$\pm$2.753 \\ 
\bottomrule 
\end{tabular}} 
\label{tab:time-benchmark-results} 
\end{table}

\begin{figure}[h!]
    \centering
    \includegraphics[width=0.9\linewidth]{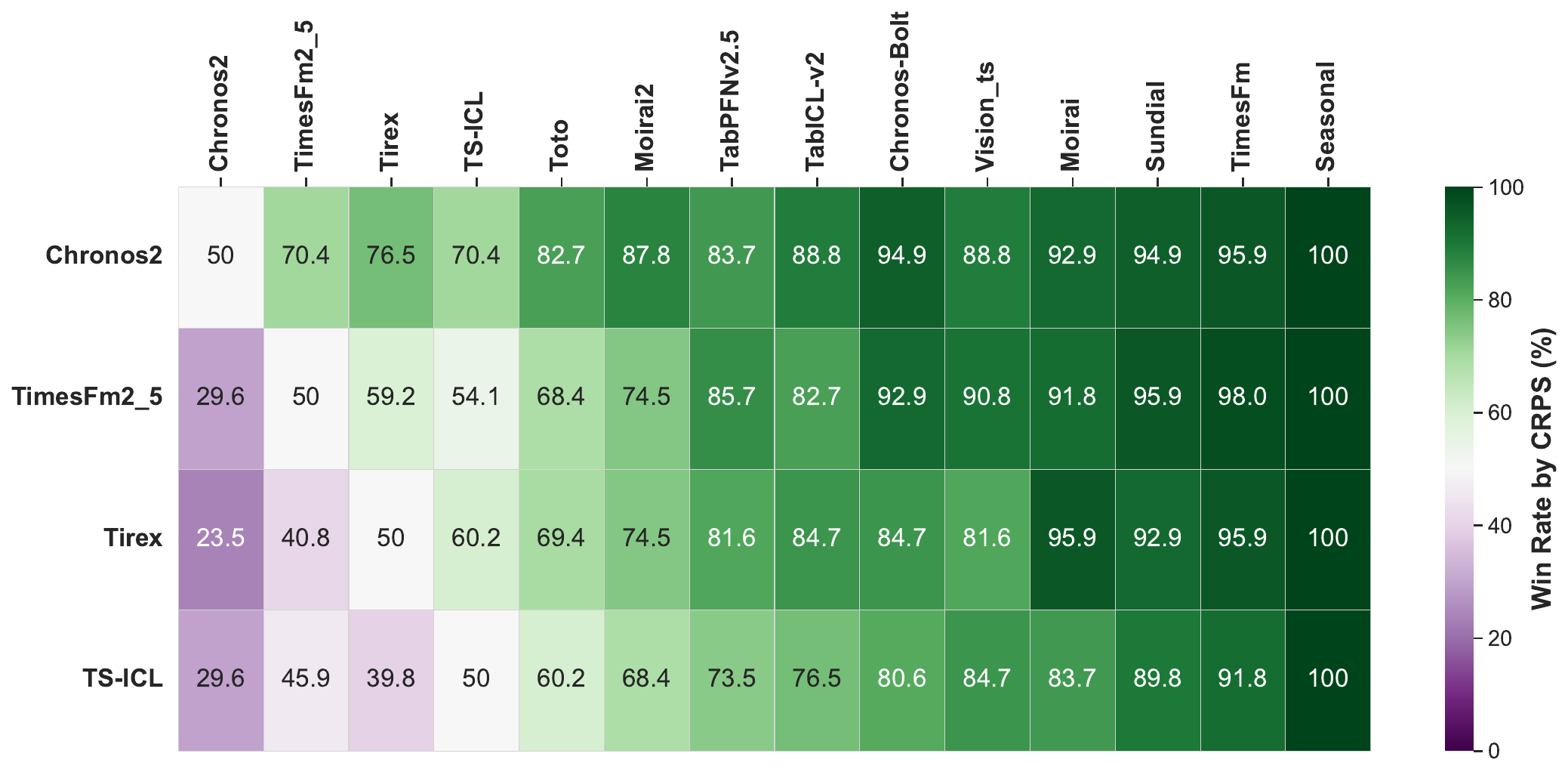}
    \caption{Pairwise win rates for the top-4 models against all other forecasters on the \texttt{TIME} benchmark.
    Each entry indicates the fraction of tasks where a method outperforms another according to the CRPS.}
    \label{fig:time-forecasting-score-wins-crps}
\end{figure}

\begin{figure}[h!]
    \centering
    \includegraphics[width=0.9\linewidth]{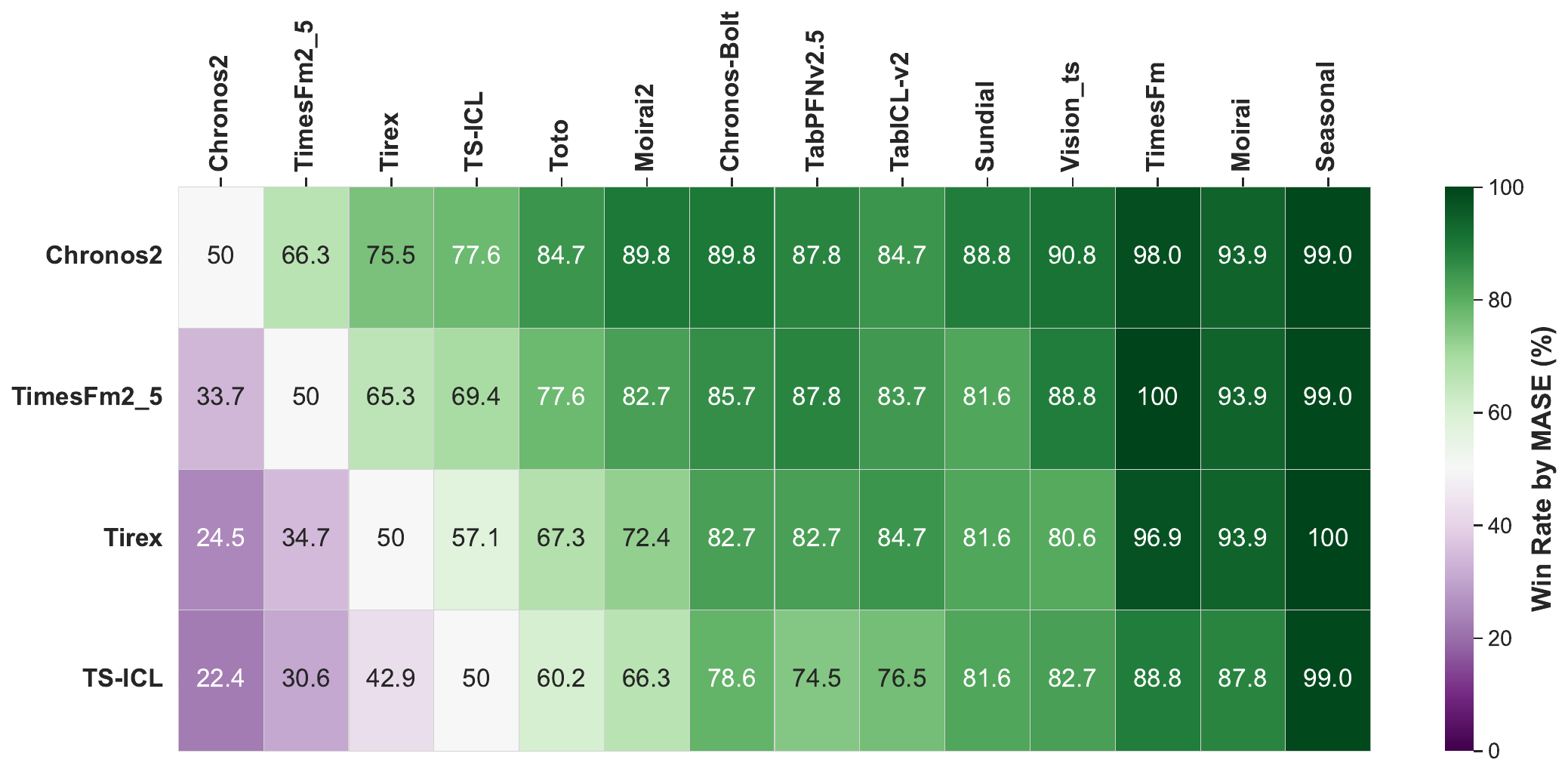}
    \caption{Pairwise win rates for the top-4 models against all other forecasters on the \texttt{TIME} benchmark.
    Each entry indicates the fraction of tasks where a method outperforms another according to the MASE.}
    \label{fig:time-forecasting-score-wins-mase}
\end{figure}

\paragraph{Results.} \texttt{TS-ICL} demonstrates strong competitive performance on the \texttt{Time} benchmark, consistently ranking among the top-tier Time Series Foundation Models (TSFMs). As detailed in \cref{tab:time-benchmark-results}, while \texttt{Chronos-2} maintains its position as the SOTA leader, \texttt{TS-ICL} achieves a highly comparable MASE of 0.902 and a CRPS of 0.143. Notably, the performance gap between \texttt{TS-ICL} model and \texttt{Chronos-2} is minimal, with \texttt{TS-ICL} trailing by only 4.7\% in MASE and 4.3\% in CRPS in relative terms. This narrow margin is further evidenced by the pairwise win rate analysis (\cref{fig:time-forecasting-score-wins-crps,fig:time-forecasting-score-wins-mase}), where \texttt{TS-ICL} successfully outperforms \texttt{Chronos-2} on 29.6\% of tasks for CRPS and 22.4\% for MASE. 

Beyond this head-to-head comparison, \texttt{TS-ICL} shows a clear superiority over the entire category of Tabular Foundation Models (TFMs), significantly outperforming both \texttt{TabPFN} and \texttt{TabICL}. It also surpasses several established TSFMs, including \texttt{TimesFM}, \texttt{Moirai}, and \texttt{Sundial}. By matching the performance of much larger, specialized models within such a tight margin while offering a more efficient architecture, \texttt{TS-ICL} establishes itself as a robust and scalable alternative for high-performance zero-shot forecasting.

\clearpage

\clearpage

\end{document}